\PassOptionsToPackage{table}{xcolor}

\documentclass[11pt]{article}

\usepackage[final]{acl}

\usepackage{times}
\usepackage{latexsym}
\usepackage{inconsolata}

\usepackage[utf8]{inputenc}
\usepackage[T1]{fontenc}
\usepackage{hyperref}
\usepackage{url}
\usepackage{booktabs}
\usepackage{amsmath}
\usepackage{amsfonts}
\usepackage{nicefrac}
\usepackage{microtype}
\usepackage{xcolor}

\usepackage{comment}
\usepackage{algorithm}
\usepackage{algorithmic}
\usepackage{newfloat}
\usepackage{listings}
\usepackage{float}
\usepackage{tabularx}
\usepackage{multirow}
\usepackage{xspace} 
\usepackage{todonotes}
\usepackage{pifont}
\usepackage{etex}
\usepackage{arydshln}
\usepackage{enumitem}
\usepackage{tikz}
\usepackage{colortbl}
\usepackage{siunitx}
\usepackage{graphicx}
\usepackage{subcaption}
\usepackage{adjustbox}
\usepackage{pgf}

\usepackage{tikz}
\usetikzlibrary{calc}
\usetikzlibrary{decorations.pathreplacing}
\usetikzlibrary{fit,backgrounds}
\usetikzlibrary{arrows.meta}
\newcommand{\ds}[1]{%
  \tikz[baseline=(X.base)]\node[draw,circle,inner sep=0.6pt,minimum size=1em] (X) {\scriptsize #1};%
}

\newcommand{\vsep}{\hspace{2pt}}

\definecolor{abovegreen}{RGB}{118,176,83}
\definecolor{belowblue}{RGB}{68,141,192}
\definecolor{heatneutral}{gray}{0.94}

\definecolor{findingsbar}{RGB}{80,119,68}
\definecolor{findingsbg}{RGB}{233,234,230}
\definecolor{findingslabelbg}{RGB}{194,209,180}

\newcommand{\findingbox}[1]{%
  \par\addvspace{5pt}%
  \noindent\begingroup
  \setlength{\fboxsep}{5pt}%
  \hbox to \linewidth{%
    {\color{findingsbar}\vrule width 2.5pt}%
    \colorbox{findingsbg}{%
      \parbox{\dimexpr\linewidth-2.5pt-2\fboxsep\relax}{%
        \normalsize
        {\setlength{\fboxsep}{1pt}%
         \colorbox{findingslabelbg}{%
           \textcolor{findingsbar}{\bfseries\MakeUppercase{Findings}}}}%
        \enspace #1%
      }%
    }%
    \hss%
  }%
  \endgroup\par\addvspace{5pt}%
}

\newcommand{\setbase}[1]{\gdef\base{#1}}
\newcommand{\base}{}

\newcommand{\tcell}[1]{%
  \pgfmathsetmacro{\db}{(#1)-(\base)}%
  \pgfmathsetmacro{\ad}{abs(\db)}%
  \ifdim \db pt < 0pt
    \ifdim \ad pt > 0.50pt {\cellcolor{belowblue!70}#1}%
    \else\ifdim \ad pt > 0.20pt {\cellcolor{belowblue!55}#1}%
    \else\ifdim \ad pt > 0.10pt {\cellcolor{belowblue!40}#1}%
    \else\ifdim \ad pt > 0.01pt {\cellcolor{belowblue!25}#1}%
    \else {\cellcolor{heatneutral}#1}%
    \fi\fi\fi\fi
  \else
    \ifdim \db pt > 0pt
      \ifdim \ad pt > 0.50pt {\cellcolor{abovegreen!70}#1}%
      \else\ifdim \ad pt > 0.20pt {\cellcolor{abovegreen!55}#1}%
      \else\ifdim \ad pt > 0.10pt {\cellcolor{abovegreen!40}#1}%
      \else\ifdim \ad pt > 0.01pt {\cellcolor{abovegreen!25}#1}%
      \else {\cellcolor{heatneutral}#1}%
      \fi\fi\fi\fi
    \else
      {\cellcolor{heatneutral}#1}%
    \fi
  \fi
}

\newcommand{\gcell}[1]{%
  \pgfmathsetmacro{\db}{(#1)-(\base)}%
  \pgfmathsetmacro{\ad}{abs(\db)}%
  \ifdim \db pt < 0pt
    \ifdim \ad pt > 0.50pt {\cellcolor{belowblue!70}#1}%
    \else\ifdim \ad pt > 0.20pt {\cellcolor{belowblue!55}#1}%
    \else\ifdim \ad pt > 0.10pt {\cellcolor{belowblue!40}#1}%
    \else {\cellcolor{belowblue!25}#1}%
    \fi\fi\fi
  \else
    \ifdim \ad pt > 0.50pt {\cellcolor{abovegreen!70}#1}%
    \else\ifdim \ad pt > 0.20pt {\cellcolor{abovegreen!55}#1}%
    \else\ifdim \ad pt > 0.10pt {\cellcolor{abovegreen!40}#1}%
    \else {\cellcolor{abovegreen!25}#1}%
    \fi\fi\fi
  \fi
}

\newcommand{\pcell}[1]{%
  \pgfmathsetmacro{\db}{(\base)-(#1)}%
  \pgfmathsetmacro{\ad}{abs(\db)}%
  \ifdim \db pt < 0pt
    \ifdim \ad pt > 0.50pt {\cellcolor{belowblue!70}#1}%
    \else\ifdim \ad pt > 0.20pt {\cellcolor{belowblue!55}#1}%
    \else\ifdim \ad pt > 0.10pt {\cellcolor{belowblue!40}#1}%
    \else {\cellcolor{belowblue!25}#1}%
    \fi\fi\fi
  \else
    \ifdim \ad pt > 0.50pt {\cellcolor{abovegreen!70}#1}%
    \else\ifdim \ad pt > 0.20pt {\cellcolor{abovegreen!55}#1}%
    \else\ifdim \ad pt > 0.10pt {\cellcolor{abovegreen!40}#1}%
    \else {\cellcolor{abovegreen!25}#1}%
    \fi\fi\fi
  \fi
}

\newcommand{\modelname}{\textit{TileQuant}\xspace}

\title{\modelname: Tile as the Precision-Allocation Unit \\
for LLM Mixed-Precision Post-Training Quantization}

\author{
Hanzhi Zhang\textsuperscript{1},
Qiao Zhang\textsuperscript{2},
Qinglei Cao\textsuperscript{2},
Heng Fan\textsuperscript{1},
Yan Huang\textsuperscript{1},
Kewei Sha\textsuperscript{3},
Yunhe Feng\textsuperscript{1} \\
\textsuperscript{1}LLaVi Lab, Computer Science \& Engineering, University of North Texas \\
\textsuperscript{2}Computer Science, Saint Louis University;
\textsuperscript{3}Data Science, University of North Texas \\
\texttt{\{hanzhi.zhang,heng.fan,yan.huang,kewei.sha,yunhe.feng\}@unt.edu} \\
\texttt{\{qiao.zhang,qinglei.cao\}@slu.edu}
}

\begin{document}

\maketitle

\begin{abstract}
Post-training quantization (PTQ) reduces large language model (LLM) deployment costs, but preserving quality requires accounting for uneven weight sensitivity. Uniform PTQ uses one weight bit-width, importance-aware PTQ adjusts scales or reconstructs weights within that width, and column-wise mixed precision retains selected columns at higher precision. However, sensitivity varies within columns, so whole-column retention also protects less sensitive weights, while runtime outlier detection adds decoding overhead. We therefore introduce \modelname, a training-free framework that assigns precision to hardware-aligned 2D weight tiles. Offline calibration provides input-activation-aware scoring from forward statistics for lower calibration cost and gradient-informed scoring from loss sensitivity for stronger quality preservation. Within each projection's retention budget, the highest-scoring tiles remain at 16-bit and the rest use 8-bit. The resulting map guides reconstruction into compact weight buffers, which the kernel reuses for native integer and 16-bit computation with local accumulation to reduce memory traffic. Experiments show that \modelname reaches $2.66\times$ the original model's generation throughput. Across language modeling, reasoning, long-context retrieval, and question answering, it preserves quality and surpasses quantized and original models on multiple tasks. These results support selecting precision units that capture local sensitivity while enabling efficient hardware execution. The implementation is: \url{https://github.com/HanzhiZhang-Ulrica/TileQuant}
\end{abstract}

\vspace{-10pt}
\section{Introduction}
\vspace{-5pt}

Efficient large language model (LLM) deployment requires reducing storage and inference costs while preserving quality in long-context understanding~\cite{liu2024lost, an2024eval}, retrieval-augmented generation~\cite{lewis2020retrieval, li2024retrieval}, and multi-document reasoning~\cite{bai2024longbench}. Post-training quantization (PTQ) reduces these costs by lowering numerical precision without retraining~\cite{yao2022zeroquant, xiao2023smoothquant, frantar2022gptq}. However, quantization sensitivity varies across regions within each weight matrix, so a matrix-wide precision choice cannot selectively protect the most sensitive weights.

\begin{figure*}[t]
    \centering
    \includegraphics[width=\textwidth]{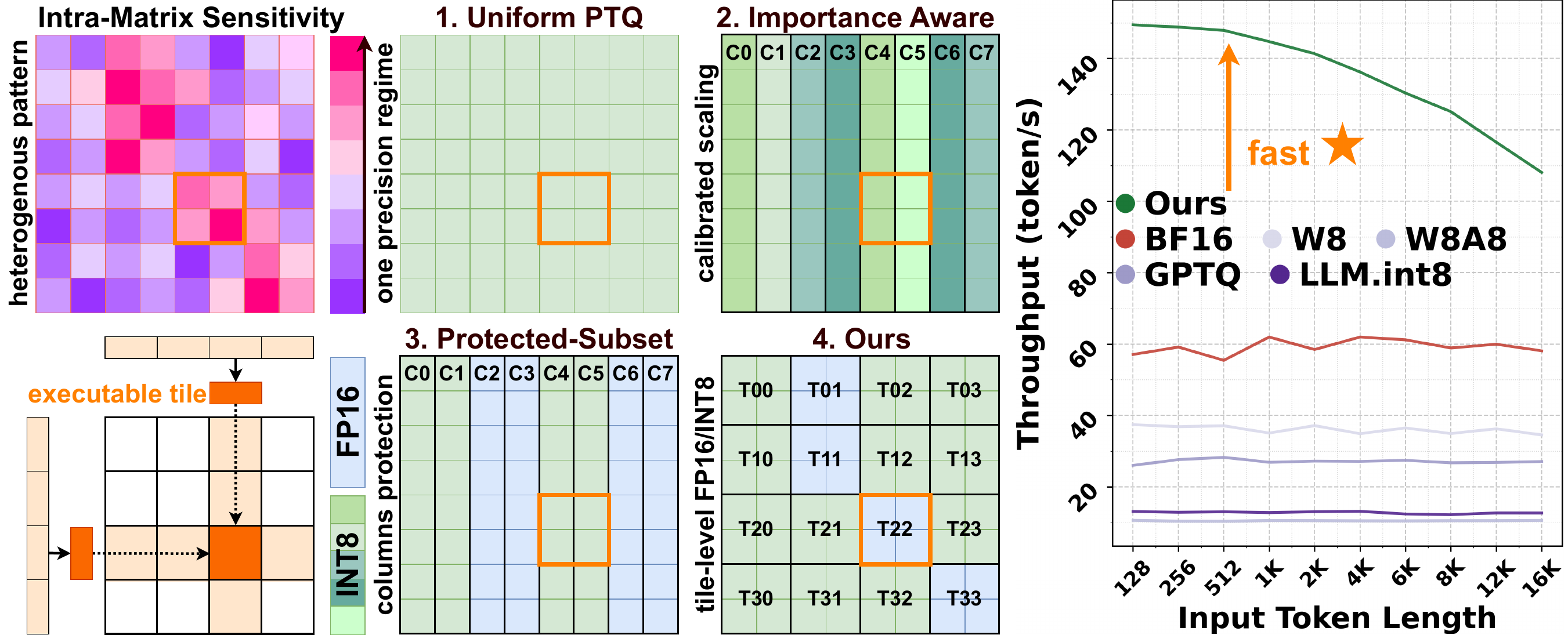}
    \vspace{-20pt}
    \caption{
    Comparison of PTQ precision-allocation units for LLMs.
    Colors denote precision: \textcolor{blue}{blue} indicates 16-bit, and \textcolor{green}{green} indicates 8-bit.
    \emph{Uniform PTQ} uses one weight bit-width.
    \emph{Importance-aware PTQ} adjusts scales or reconstructs weights at the same bit-width, with green shades illustrating calibrated scaling.
    \emph{Protected-subset PTQ} illustrates channel-wise 16-bit retention.
    \emph{\modelname} assigns precision to 2D tiles by local sensitivity.
    }
    \vspace{-15pt}
    \label{fig:intro}
\end{figure*}

Figure~\ref{fig:intro} compares three PTQ designs. \textit{Uniform PTQ}~\cite{yao2022zeroquant} assigns one weight bit-width across a matrix, while \textit{importance-aware PTQ}~\cite{xiao2023smoothquant, lin2024awq, frantar2022gptq} adjusts channel scales or reconstructs weights using calibration data within that width. \textit{Protected-subset PTQ}~\cite{dettmers2022gpt3, lee2024owq} instead retains selected columns at higher precision. However, sensitivity varies even within columns, so whole-column retention also protects less sensitive weights. When selection relies on runtime activation outliers, repeated detection and processing further increase decoding latency~\cite{dettmers2022gpt3}. This motivates our research question: \emph{Which precision-allocation unit captures within-column sensitivity while supporting efficient tile computation?}

Our key insight is to use \textbf{hardware-aligned 2D weight tiles} for local precision allocation. Each tile covers a local region along both matrix dimensions, allowing precision to vary within a column while preserving a regular tile for GPU matrix multiplication~\cite{guide2020cuda, choquette2021nvidia, kerr2017cutlass}. However, turning these choices into efficient inference presents two coupled challenges. First, importance-based selection can retain tiles at different matrix locations, so their layout must combine compact storage with regular access to both precision formats. Second, prefill processes multiple prompt tokens, whereas decode processes one new token per sequence. The same stored layer must therefore exploit weight reuse in prefill and expose sufficient parallel work in decode while limiting intermediate memory traffic.

Building on this insight, we introduce \modelname, a training-free tile-level mixed-precision PTQ framework for LLM linear layers. Two offline calibration routes address different resource and quality needs: input-activation-aware (IAA) scoring combines forward activation statistics with weight magnitudes for lower calibration cost, while gradient-informed (GI) scoring uses loss gradients for stronger quality preservation. Each projection retains its highest-scoring tiles at 16-bit under a prescribed budget and quantizes the rest to 8-bit. The resulting decision map guides load-time reconstruction into separate compact buffers organized for dense tile loading. Prefill and decode reuse these buffers while the kernel adjusts parallel execution to each input shape. It rescales integer products and locally accumulates contributions from both precision paths, writing each completed output tile once to reduce memory traffic.

On Llama-3.2-3B, \modelname reaches $2.66\times$ the generation throughput of the original 16-bit model while preserving quality across language modeling, reasoning, long-context retrieval, and question answering. For example, GI with 20\% tile retention exceeds Original in 26 of the 27 displayed LV-Eval dataset--length settings. Together, these results show that local precision retention supports broad model capabilities alongside faster end-to-end generation.

We summarize our contributions as follows:
\begin{enumerate}[nosep,left=0pt]
    \item \textbf{Tile as the Precision-Allocation Unit.}
    We assign 16-bit or 8-bit precision to hardware-aligned weight tiles, allowing selected regions within a column to retain higher precision.
    \item \textbf{Calibration-Driven Tile Decision Maps.}
    We develop IAA and GI calibration routes that prioritize tiles under per-layer retention budgets, with distinct cost and quality profiles.
    \item \textbf{Executable Mixed-Precision Reconstruction.}
    We convert tile decision maps into executable mixed-precision linear layers through compact metadata and load-time reconstruction, with dense tile computation and local accumulation reducing memory traffic.
\end{enumerate}

\vspace{-5pt}
\section{Related Work}
\vspace{-5pt}

We review uniform PTQ, importance-aware PTQ, and protected-subset PTQ through their choices of weight bit-width, calibration criterion, and high-precision retention. They describe how weights are represented, how quantization error is controlled, and where additional precision is allocated.

\textit{Uniform PTQ} applies a single weight bit-width to all weights within a chosen unit, producing a homogeneous low-precision representation for the target tensor~\cite{gholami2022survey, lang2024comprehensive, zhao2025benchmarking, sharify2024post}. Representative examples include rounding and reconstruction methods~\cite{yao2022zeroquant, chee2025discquant, kogan2025selective, gong2024makes} and scaling- or transformation-based methods~\cite{xiao2023smoothquant, pan2023smoothquant+, wei2022outlier, shao2023omniquant, liu2023qllm, ashkboos2024quarot, liu2024spinquant, sun2024flatquant} that reduce quantization error in weights and activations.
A homogeneous weight format simplifies deployment, while its shared bit-width couples the precision assigned to sensitive and less sensitive regions.

\textit{Importance-aware PTQ} addresses these sensitivity differences by estimating which weights or channels are sensitive to quantization error and adjusting scaling factors or reconstruction objectives accordingly~\cite{li2024tesseraq, zhang2025qronos, arai2025quantization}. Calibration-based methods leverage activation or sensitivity statistics to pinpoint salient weights~\cite{lin2024awq, kim2023squeezellm, kim2024boa, xue2025vlmq}, while reconstruction-based approaches optimize the quantized tensor to better match original outputs~\cite{nagel2020up, li2021brecq, wei2022qdrop, frantar2022gptq, wei2025mppq, yvinec2023gradient}.
These approaches show that quantization quality depends on where error occurs, and they provide useful signals for protecting important model components. However, in fixed-bit formulations, the resulting tensor retains a single weight bit-width within the target unit.

\textit{Protected-subset PTQ} further introduces non-uniform precision by preserving selected substructures in higher precision while quantizing the remaining weights~\cite{dettmers2022gpt3, ashkboos2024quik, dettmers2023spqr, guo2023olive, shang2023pb, yu2024super}. Methods protect groups~\cite{huang2024slim, li2023llm, guan2024aptq, lee2025amq}, channels~\cite{chen2024channel, nrusimha2024mitigating}, or columns and output features~\cite{lee2024owq, zheng2024mixllm, wang2024claq, yuan2023rptq}.
These methods demonstrate that non-uniform precision can recover accuracy under quantization~\cite{li2023llm, saxena2024resq, xie2025fineq, xiao2025task}. Their protected units are usually tied to predefined structural axes or sparse outlier subsets, making them effective for axis-aligned or outlier-driven protection. However, whole-structure retention also protects less sensitive weights, while sparse protection adds indexing and sparse-computation costs. Mixed precision also supports fine-tuning~\cite{zhou2024automixq} and KV-cache compression~\cite{liao2026channel}.

\vspace{-5pt}
\section{Preliminaries}
\label{sec:prelim}
\vspace{-5pt}

We first review quantized linear projections and then introduce their tile-wise decomposition to support local precision allocation.

\vspace{-2pt}
\subsection{PTQ for Transformer Linear Layers}

During LLM inference, a linear layer computes $Y=XW^\top$, where $X\in\mathbb{R}^{N\times K}$, $W\in\mathbb{R}^{M\times K}$, and $Y\in\mathbb{R}^{N\times M}$. Here, $N$ is the number of activation rows, and $K$ and $M$ are the input and output feature dimensions. PTQ reduces storage by replacing 16-bit weights with lower-precision formats.

For a positive quantization scale $\Delta$, the 8-bit integer weights are $\widehat{W}=\operatorname{clip}_{[-127,127]}(\lfloor W/\Delta\rceil)$, with $\Delta\widehat{W}\approx W$. Here, $\lfloor\cdot\rceil$ rounds to the nearest integer, and clipping keeps the result within the quantization range. When activations are also quantized, the projection is approximated as $Y\approx(\Delta_X\Delta_W)(\widehat{X}\widehat{W}^\top)$ for shared operand scales $\Delta_X$ and $\Delta_W$. With tile-wise scales, each tile product is rescaled before accumulation.

\vspace{-2pt}
\subsection{Tile-Wise Linear Projection}

To support local precision allocation, we partition $W$ into a regular grid of fixed-size 2D tiles after any necessary padding. Let $W^{(r,c)}$ denote a tile, where $r=1,\dots,n_{\mathrm{tr}}$ indexes output-feature tiles and $c=1,\dots,n_{\mathrm{tc}}$ indexes input-feature tiles. Their index set is denoted by $\mathcal{T}(W)$.

Since each weight-tile column uses the same input-feature range, we partition $X$ into corresponding tiles $X^{(c)}$. The output-feature tile $Y^{(r)}$ is then computed as
\vspace{-5pt}
\[
\begin{aligned}
Y^{(r)} ={}& \sum_{c=1}^{n_{\mathrm{tc}}} X^{(c)} \big(W^{(r,c)}\big)^\top,
\qquad r = 1, \dots, n_{\mathrm{tr}}.
\end{aligned}
\vspace{-5pt}
\]
Each term is a dense tile product, allowing different weight tiles to use different precisions while contributing to the same output tile.

We define $\mathcal{F}(W)$ and $\mathcal{Q}(W)$ as the tile-index sets assigned 16-bit and 8-bit precision, respectively. They partition $\mathcal{T}(W)$, satisfying $\mathcal{F}(W)\cup\mathcal{Q}(W)=\mathcal{T}(W)$ and $\mathcal{F}(W)\cap\mathcal{Q}(W)=\emptyset$. Section~\ref{sec:method} determines these assignments from calibration scores.

\begin{figure*}[t]
    
    \centering
    \includegraphics[width=\textwidth]{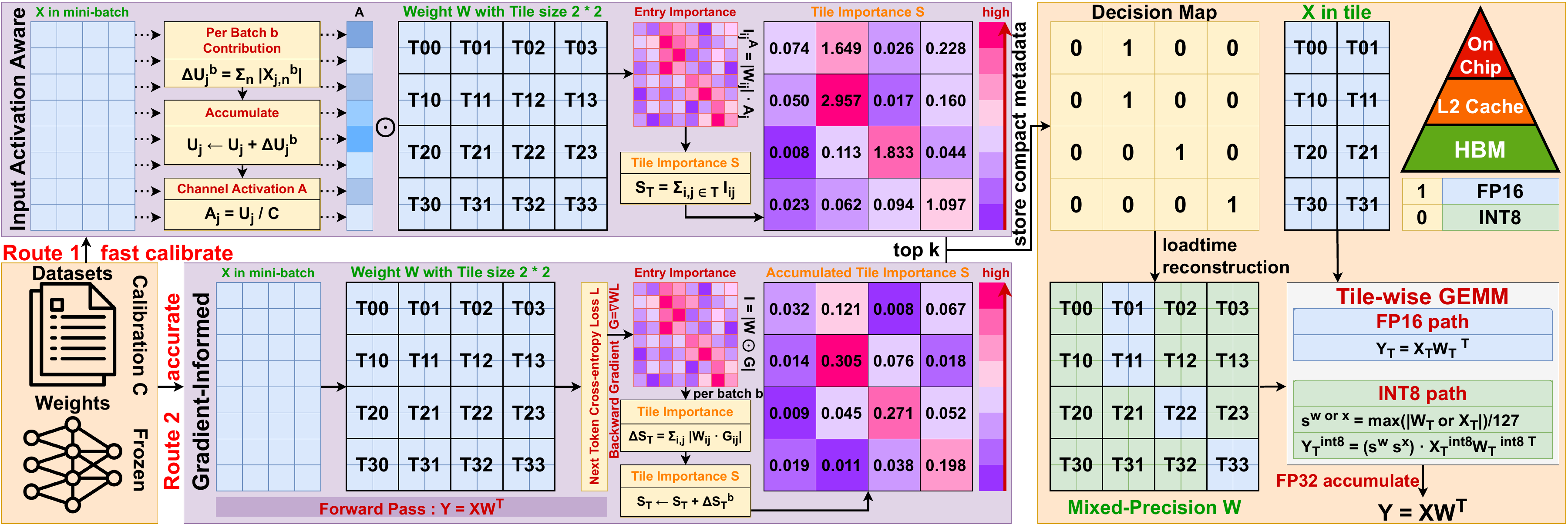}  
    \vspace{-20pt}
    \caption{
    \modelname framework for calibration-driven mixed-precision linear layers.
    The input-activation-aware route (top left) combines forward activation statistics with weight magnitudes, while the gradient-informed route (bottom left) accumulates weight-gradient importance over calibration batches.
    Per-layer top-$k$ selection converts either tile-score map into a 16-bit/8-bit decision map.
    Retained tile IDs and grid metadata guide load-time reconstruction (right).
    The reconstructed layer supplies the selected weights to two arithmetic paths, whose tile products are combined through output-local 32-bit floating-point accumulation.
    }
    \vspace{-15pt}
    \label{fig:framework}
\end{figure*}

\vspace{-5pt}
\section{\modelname}
\label{sec:method}
\vspace{-5pt}

\subsection{\modelname Overview}

Building on the tile-wise projection in Section~\ref{sec:prelim}, \modelname converts attention and feed-forward linear projections into mixed-precision layers. The method must identify local regions worth retaining at 16-bit and organize their computation with the remaining 8-bit weights. Figure~\ref{fig:framework} shows how calibration choices reach the deployed layer.

Offline calibration assigns an importance score to every tile using input-activation-aware (IAA) statistics or gradient-informed (GI) sensitivity. A common rule retains the highest-scoring tiles within each projection's budget and records their locations in a binary decision map (Section~\ref{sec:tile_map}). During loading, this map identifies the regions to retain and those to quantize. The resulting weight buffers and scales supply the mixed-precision kernel (Section~\ref{sec:tile_execution}), so the same offline choices govern arithmetic across inference inputs.

\subsection{Calibration-Driven Tile Map Construction}
\label{sec:tile_map}

Calibration ranks local weight regions for each projection's 16-bit retention budget. With pretrained weights fixed, IAA scores activation-weighted contribution magnitudes, while GI scores first-order loss sensitivity on the same calibration data. These scores provide retention priorities, and the prescribed budget determines how many tiles are selected. Both routes therefore produce score maps for the common selection rule below.

\noindent \textbf{Input-activation-aware route.}
IAA estimates tile importance from contribution magnitudes observed in forward passes. A weight's contribution depends on both its magnitude and the activations multiplied by it. We therefore combine weight magnitudes with input-channel activation statistics.

For each input channel $j$, let $a_j=\mathbb{E}[|x_j|]$ denote the mean absolute activation magnitude gathered from the calibration data. We define the entry-level and aggregated tile-level importance as
\[
\begin{aligned}
I_{ij}^{\mathrm{act}} &= |W_{ij}|\cdot a_j,\quad
S_t^{\mathrm{act}} &= \sum_{(i,j)\in t}|W_{ij}|\cdot a_j,
\end{aligned}
\]
where $t\in\mathcal{T}(W)$ identifies a weight tile. The sum prioritizes regions containing large weights connected to channels with large mean absolute inputs. During calibration, IAA collects these channel means through forward passes and per-channel sums and counts, providing importance scores with a small persistent statistics buffer.

\noindent \textbf{Gradient-informed route.}
For loss-sensitive selection, GI relates weight importance to the model's next-token prediction objective. For calibration batch $b$, a forward pass computes the next-token cross-entropy loss $\mathcal{L}_b$, and backward computation obtains gradients with respect to the target weights. The entry-level importance is $I_{ij,b}^{\mathrm{grad}}=\left|W_{ij}\frac{\partial\mathcal{L}_b}{\partial W_{ij}}\right|$. The batch-level and accumulated tile scores are
\[
\begin{aligned}
S_{t,b}^{\mathrm{grad}} &= \sum_{(i,j)\in t} I_{ij,b}^{\mathrm{grad}},\quad
S_t^{\mathrm{grad}} &= \sum_b S_{t,b}^{\mathrm{grad}}.
\end{aligned}
\]

The weight-gradient product measures first-order sensitivity to a relative weight change under the prediction objective. Taking magnitudes before summation prevents positive and negative terms from canceling within a tile or across calibration inputs. The accumulated score therefore ranks tiles by loss sensitivity over the calibration data. Appendix~\ref{app:gi_interpretation} derives this interpretation, and both routes use the precision-selection rule below.

\noindent \textbf{Per-layer precision selection.}
The resulting scores determine retention priority, and a prescribed ratio controls the retained capacity. Here, each layer denotes one target linear projection. Given a target 16-bit retention ratio $\rho\in(0,1]$, that projection retains
\(
k=\max\!\left(1,\left\lceil\rho\,|\mathcal{T}(W)|\right\rceil\right)
\)
tiles. The $k$ highest-scoring tiles form $\mathcal{F}(W)$, and the others form $\mathcal{Q}(W)$ for 8-bit quantization.

For a fixed tile grid and $\rho$, IAA and GI retain the same number of 16-bit weights, while their scores determine the retained locations. Since scoring precedes budget selection, one completed score map can support several retention ratios. Each resulting decision map records the chosen precision assignments, whose retained indices guide load-time reconstruction.

\vspace{-5pt}
\subsection{Mixed-Precision Reconstruction and Execution}
\label{sec:tile_execution}

The decision map specifies which regions receive each precision. To turn this allocation into storage and execution benefits, the reconstructed layer keeps the selected weight formats in GPU memory and supplies them directly to their arithmetic paths. We first describe how the map becomes a stored layer, then explain how its tile products are computed and combined with a single writeback per output tile.

\noindent \textbf{Load-time reconstruction.}
For each target projection, \modelname stores retained tile IDs and tile-grid metadata. During checkpoint loading, these IDs locate the 16-bit weights, and their complement identifies the weights to quantize to 8-bit. The loader packs the two precision sets into separate contiguous buffers, grouping tiles by output-tile row. This layout places the weights contributing to the same output tile together for traversal by the kernel. Weight scales accompany the integer tiles, and non-target tensors keep their original format.

The scales recover the magnitudes represented by integer weights, and the tile IDs identify their corresponding input-feature tiles. IAA also uses its collected channel statistics to select weight scales, as detailed in Appendix~\ref{app:weight_quantization}. These weights, scales, and indices remain resident across inference calls, so each call begins by preparing activations.

\noindent \textbf{Tile-wise arithmetic.}
Native 8-bit integer general matrix multiplication (INT8 GEMM) requires both operands in integer form. Therefore, the runtime quantizes each input-feature tile $X^{(c)}$ to $\widehat{X}^{(c)}$ using a scale $\Delta_X^{(c)}$ shared across its activation rows. Every 8-bit weight tile in column $c$ reuses this prepared tile, together with its stored weights $\widehat{W}^{(r,c)}$ and scale $\Delta_W^{(r,c)}$. The retained tiles multiply the current 16-bit inputs directly. Thus, the saved map determines weight precision, and input preparation supplies the operands for both paths.

For weight-tile row $r$, let $\mathcal{Q}_r(W)$ and $\mathcal{F}_r(W)$ denote the corresponding column-tile indices in the two sets. Within the current activation-row tile, the mixed-precision projection approximates the original output as
\vspace{-1pt}
\[
\begin{aligned}
Y^{(r)} \approx{}&
\sum_{c\in\mathcal{Q}_r(W)}
\Delta_X^{(c)}\Delta_W^{(r,c)}
\widehat{X}^{(c)}\big(\widehat{W}^{(r,c)}\big)^\top\\
&{}+
\sum_{c\in\mathcal{F}_r(W)}
X^{(c)}\big(W^{(r,c)}\big)^\top.
\end{aligned}
\]
Each integer tile product first accumulates in 32-bit integers and is then converted to floating point for rescaling. The activation and weight scales are applied before summation because they can differ across tiles. This restores all products to the output's numerical units, allowing both precision paths to update a common accumulator.

\noindent \textbf{Output-stationary execution.}
The tile-wise projection sums contributions from multiple weight tiles into each output tile. To reduce memory traffic during this sum, our Triton kernel accumulates successive tile products in registers. Each output tile covers a group of activation rows and the output features associated with one weight-tile row. The kernel loads the corresponding inputs and weights as dense tiles for Tensor Core multiplication~\cite{guide2020cuda, choquette2021nvidia, kerr2017cutlass}. Products from both precision paths enter a 32-bit floating-point (FP32) accumulator, allowing each program to combine its assigned tile contributions within its registers.

Output tiles also provide independent work for parallel execution. When enough tiles are available, one program completes each tile. With fewer output tiles, the kernel distributes the weight tiles contributing to one output tile across multiple programs, exposing additional parallel work along the input dimension. Each program accumulates its assigned products locally, and their partial sums are then combined within the same kernel. Both schedules write each completed output tile once, preserving local accumulation while adapting the available parallel work to the output shape. Appendix~\ref{app:output_scheduling} provides the indexing, scheduling, and partial-sum reduction details.

\noindent \textbf{Prefill and decode.}
The same stored layer must support the different input shapes encountered during generation. Prefill processes multiple prompt-token rows through GEMM, allowing a loaded weight tile to contribute to several rows within an output tile. During decode, each sequence contributes one new activation row, giving general matrix-vector multiplication (GEMV) for a single sequence or a small GEMM for a batch.

\begin{figure*}[t]
    \centering

    \includegraphics[width=0.274\linewidth]{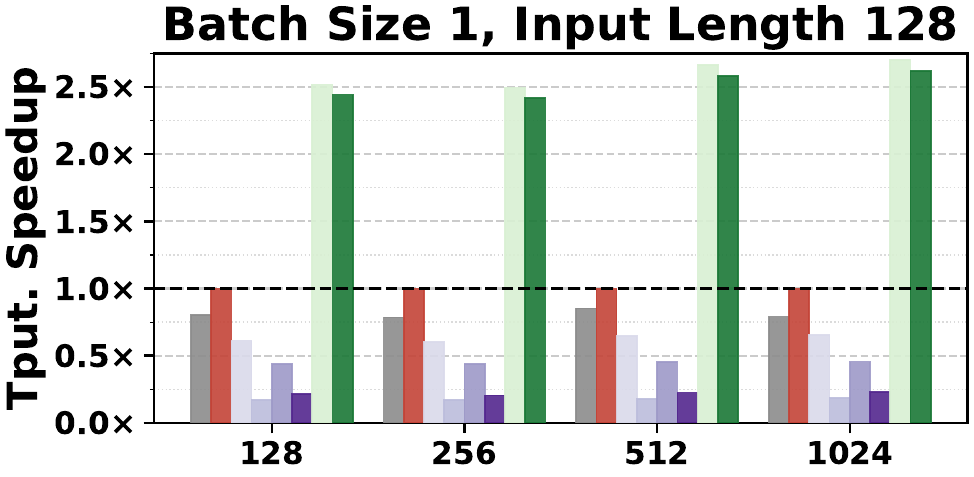}
    \hfill
    \includegraphics[width=0.234\linewidth]{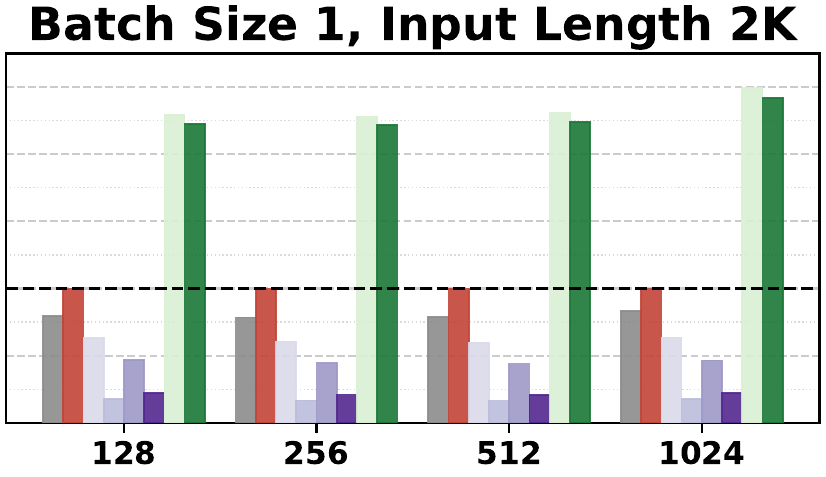}
    \hfill
    \includegraphics[width=0.234\linewidth]{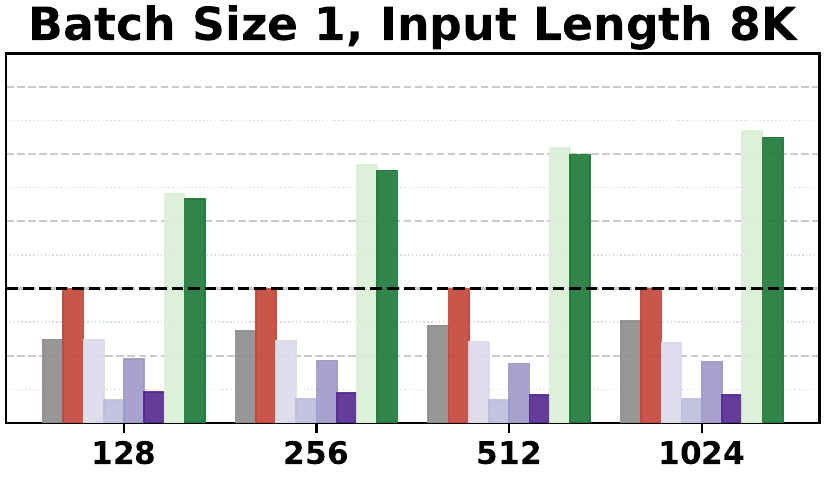}
    \hfill
    \includegraphics[width=0.234\linewidth]{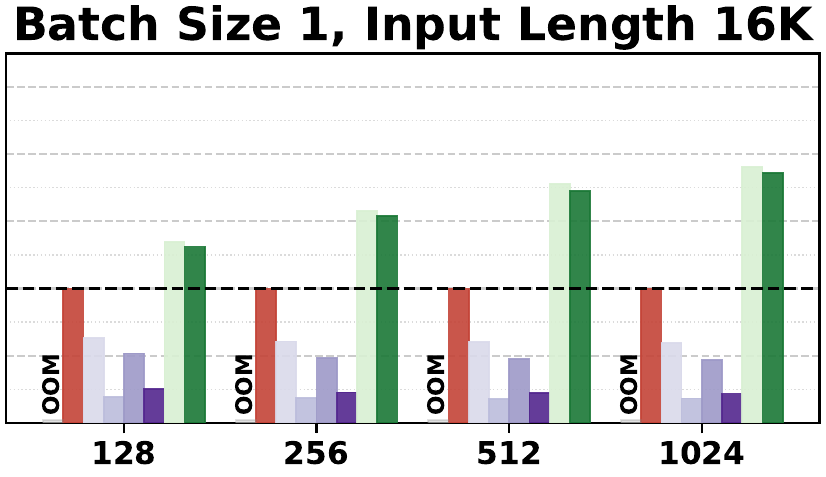}

    \vspace{-3pt}

    \includegraphics[width=0.274\linewidth]{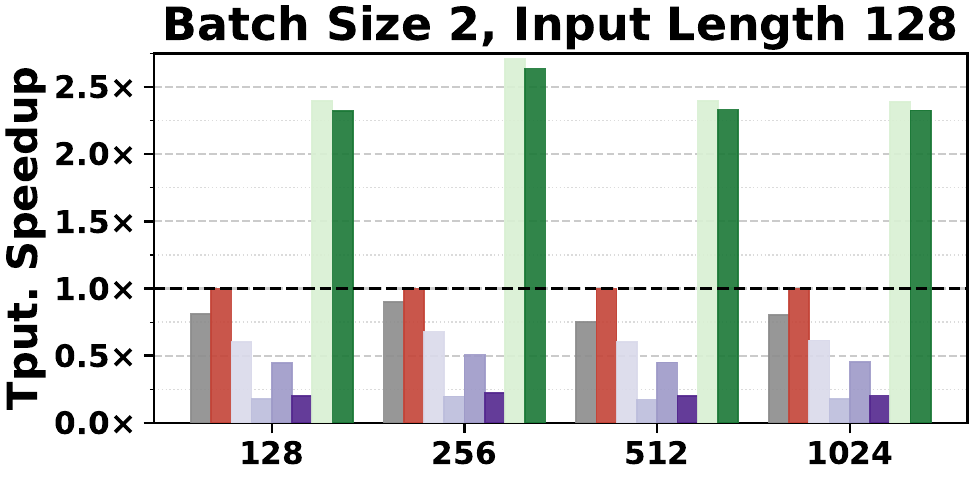}
    \hfill
    \includegraphics[width=0.234\linewidth]{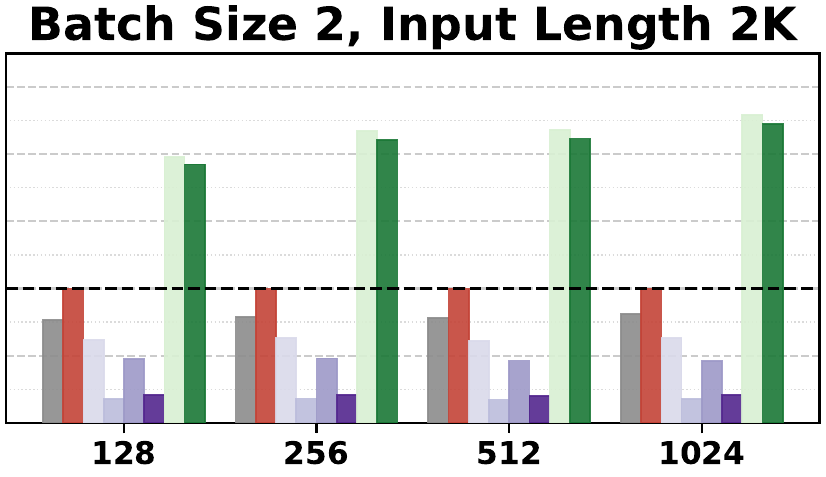}
    \hfill
    \includegraphics[width=0.234\linewidth]{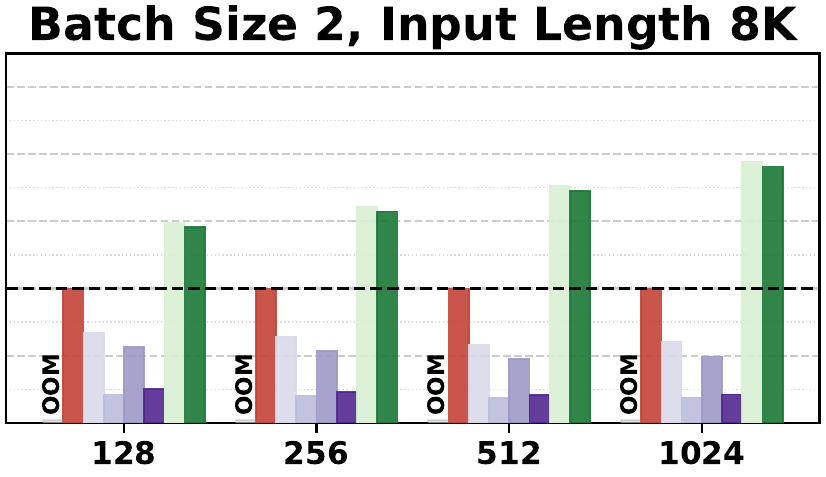}
    \hfill
    \includegraphics[width=0.234\linewidth]{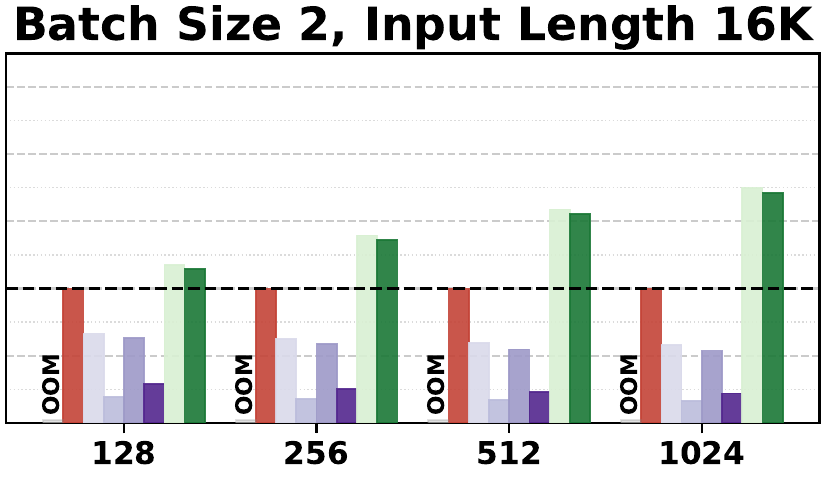}

    \vspace{-3pt}

    \includegraphics[width=0.274\linewidth]{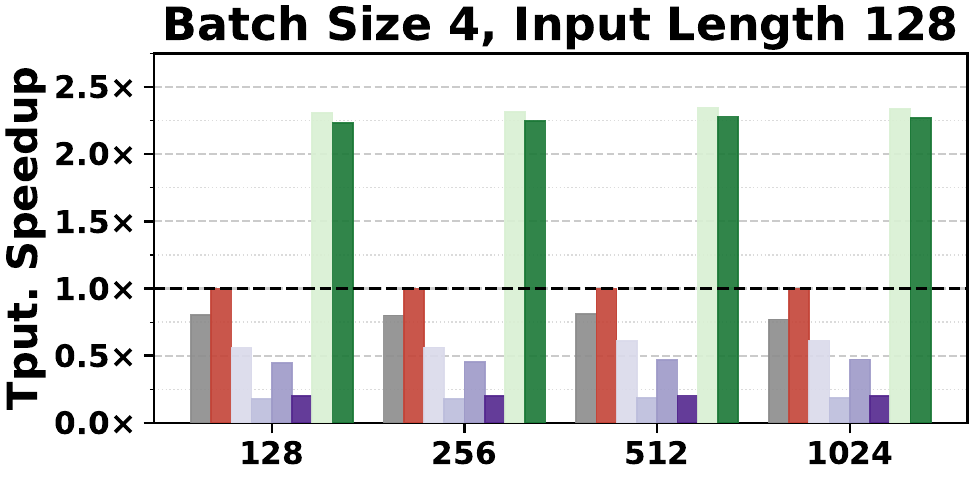}
    \hfill
    \includegraphics[width=0.234\linewidth]{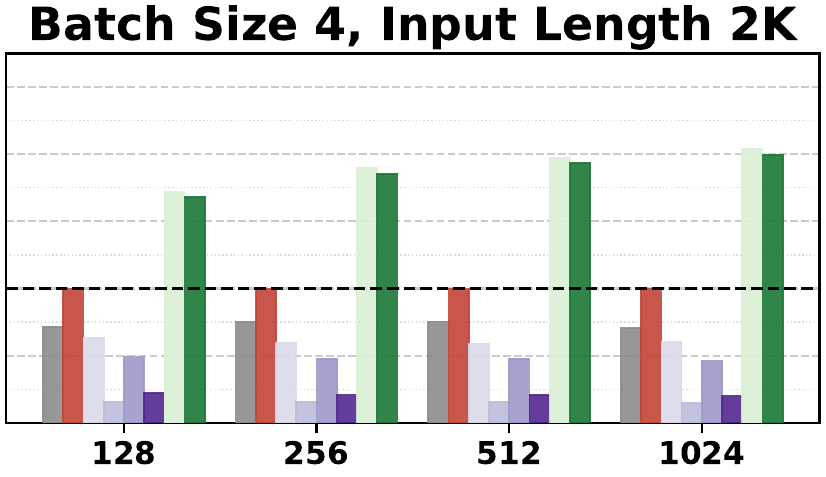}
    \hfill
    \includegraphics[width=0.234\linewidth]{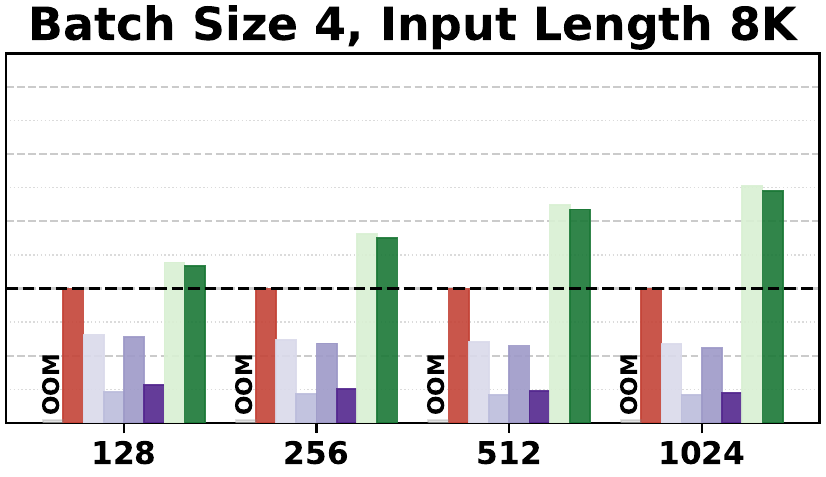}
    \hfill
    \includegraphics[width=0.234\linewidth]{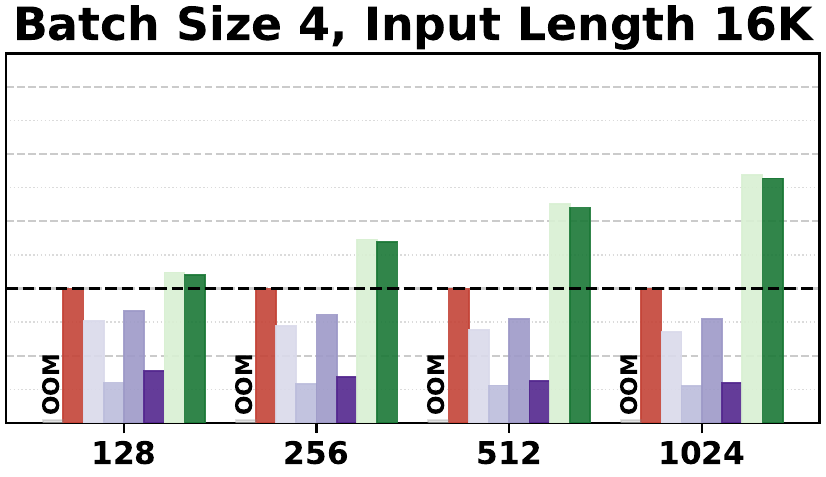}

    \vspace{-3pt}

    \includegraphics[width=0.274\linewidth]{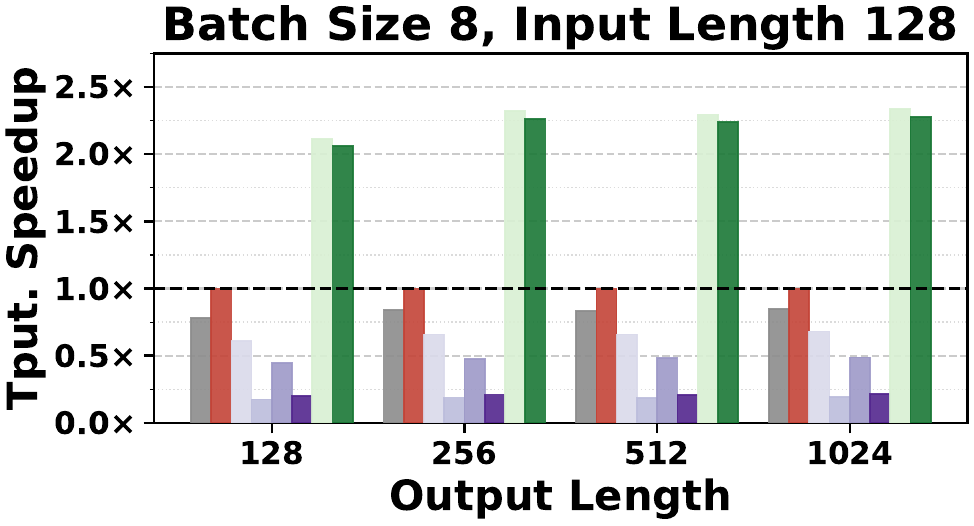}
    \hfill
    \includegraphics[width=0.234\linewidth]{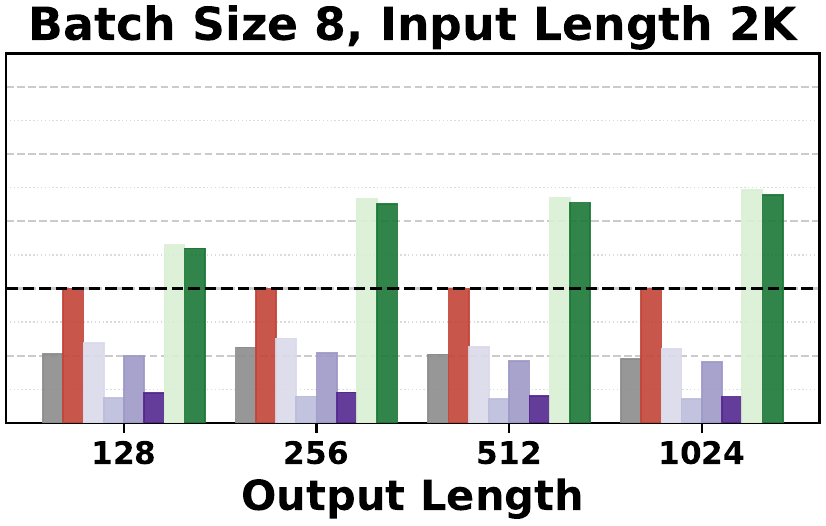}
    \hfill
    \includegraphics[width=0.234\linewidth]{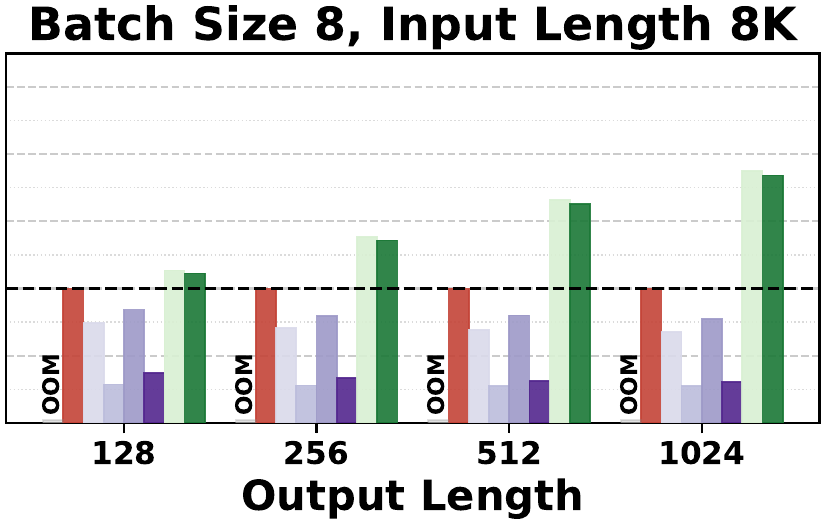}
    \hfill
    \includegraphics[width=0.234\linewidth]{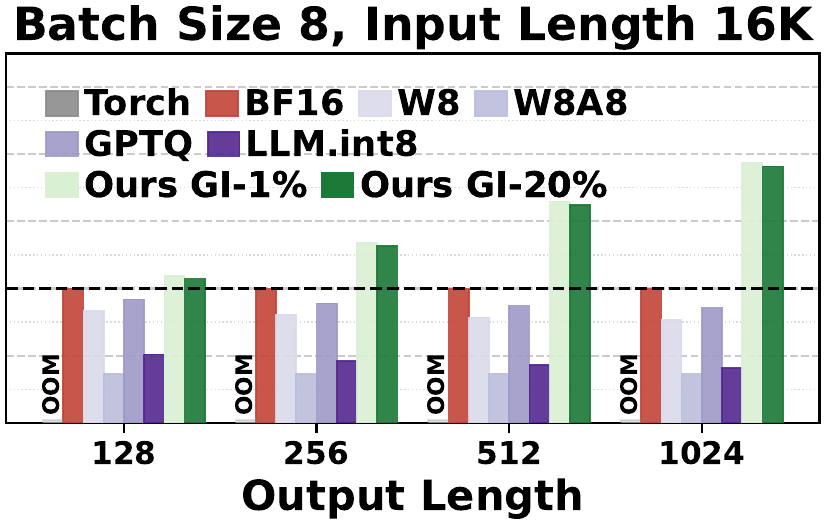}

    \vspace{-8pt}
    \caption{
    Generation-throughput speedup relative to BF16 on Llama-3.2-3B. Rows vary batch size over 1, 2, 4, and 8, and columns vary prompt length over 128, 2K, 8K, and 16K tokens. Each panel varies output length over 128, 256, 512, and 1024 tokens. Throughput includes the complete request and is normalized to BF16 at the same request shape. Values use the median of five measured runs after two warm-up runs. OOM denotes an out-of-memory run.
    }
    \label{fig:generation_throughput_grid}
    \vspace{-18pt}
\end{figure*}

Accordingly, row tiling and launch parameters adapt to the number of activation rows, while both phases reuse the precision map, packed weights, scales, and tile indices. The mapping from weight-tile rows to output features remains unchanged, and the runtime selects the parallel schedule for the current input shape. This organization keeps importance estimation and precision selection outside token generation while reusing the same mixed-precision representation throughout the request.

\vspace{-5pt}
\section{Experiments}
\vspace{-5pt}

\subsection{Experimental Setup}

\begin{table*}[t]

\centering
\scriptsize
\setlength{\tabcolsep}{2.2pt}
\renewcommand{\arraystretch}{0.75}
\caption{
General language-modeling and reasoning results on \textbf{Llama-3.2-3B}. GI and IAA denote gradient-informed and input-activation-aware tile retention, with subscripts indicating the percentage of tiles retained in floating point. $\uparrow$/$\downarrow$ indicate higher/lower is better. Accuracy is reported in percent, while WikiText-2 uses word perplexity. Bold marks the best result per row. Relative to Original (gray), \textcolor{green}{green} indicates equal or better performance and \textcolor{blue}{blue} indicates worse performance.
}
\label{tab:general_llama3b}
\vspace{-10pt}
\begin{adjustbox}{width=\textwidth}
\begin{tabular}{l c c @{\vsep} c c c c @{\vsep} c c @{\vsep} c c c}
\toprule
\multirow{2}{*}{\textbf{Benchmark}} &
\multirow{2}{*}{\textbf{Metric}} &
\multirow{2}{*}{\textbf{Original}} &
\multicolumn{4}{c}{\textbf{Ours}} &
\multicolumn{2}{c}{\textbf{Uniform}} &
\multirow{2}{*}{\rotatebox{25}{\textbf{GPTQ}}} &
\multirow{2}{*}{\rotatebox{25}{\textbf{SmoothQ}}} &
\multirow{2}{*}{\rotatebox{25}{\textbf{LLM.int8}}} \\
\cmidrule(lr){4-7}\cmidrule(lr){8-9}
& & &
\textbf{GI$_1$} & \textbf{IAA$_1$} &
\textbf{GI$_{20}$} & \textbf{IAA$_{20}$} &
\textbf{W8} & \textbf{W8A8} &
& & \\
\midrule

WikiText-2 $\downarrow$ & PPL \setbase{12.487} & \cellcolor{heatneutral}\base & \pcell{12.518} & \pcell{12.537} & {\bfseries {\pcell{12.403}}} & \pcell{12.438} & \pcell{12.453} & \pcell{12.630} & \pcell{12.515} & \pcell{12.609} & \pcell{12.639} \\

WinoGrande $\uparrow$ & Acc. \setbase{71.43} & \cellcolor{heatneutral}\base & \gcell{71.38} & \gcell{70.98} & \gcell{71.46} & \gcell{71.30} & \gcell{70.88} & \gcell{70.64} & \gcell{71.43} & {\bfseries {\gcell{71.59}}} & \gcell{71.11} \\

PIQA $\uparrow$ & Acc. \setbase{77.31} & \cellcolor{heatneutral}\base & \gcell{76.61} & \gcell{77.09} & {\bfseries {\gcell{77.42}}} & \gcell{77.15} & \gcell{76.61} & \gcell{76.55} & \gcell{76.82} & \gcell{76.55} & \gcell{76.82} \\

LAMBADA $\uparrow$ & Acc. \setbase{64.23} & \cellcolor{heatneutral}\base & \gcell{65.62} & \gcell{65.07} & {\bfseries {\gcell{65.78}}} & \gcell{65.49} & \gcell{64.41} & \gcell{63.96} & \gcell{64.43} & \gcell{63.98} & \gcell{64.31} \\

ARC-Challenge $\uparrow$ & Acc. \setbase{52.56} & \cellcolor{heatneutral}\base & \gcell{52.82} & \gcell{52.30} & {\bfseries {\gcell{53.69}}} & \gcell{52.88} & \gcell{52.30} & \gcell{51.79} & \gcell{52.73} & \gcell{51.71} & \gcell{51.88} \\

HellaSwag $\uparrow$ & Acc. \setbase{73.65} & {\bfseries \cellcolor{heatneutral}\base} & \gcell{73.25} & \gcell{73.09} & \gcell{72.95} & \gcell{72.50} & \gcell{73.59} & \gcell{73.53} & \gcell{73.55} & \gcell{73.47} & \gcell{73.48} \\

MMLU $\uparrow$ & Acc. \setbase{60.47} & \cellcolor{heatneutral}\base & \gcell{60.64} & \gcell{60.78} & {\bfseries {\gcell{61.10}}} & \gcell{60.54} & \gcell{60.51} & \gcell{60.80} & \gcell{60.46} & \gcell{60.48} & \gcell{60.89} \\
\bottomrule
\end{tabular}
\end{adjustbox}

\end{table*}

We evaluate efficiency and model quality on Llama-3.2-3B-Instruct~\cite{meta2024llama3.2}. Appendix~\ref{appendix:exp_efficiency} extends the system measurements, while Appendices~\ref{appendix:exp_general} and~\ref{appendix:exp_long} report additional quality results.

\noindent
\textbf{Datasets and Metrics.}
To assess general capabilities, we report WikiText-2 word perplexity~\cite{merity2016pointer} and accuracy on HellaSwag~\cite{zellers2019hellaswag}, MMLU~\cite{hendrycks2020measuring}, ARC-Challenge~\cite{clark2018think}, LAMBADA~\cite{paperno2016lambada}, PIQA~\cite{bisk2020piqa}, and WinoGrande~\cite{sakaguchi2021winogrande}. For long contexts, LongEval~\cite{li2023long} measures line-retrieval exact-match accuracy over 2.7K--43.4K tokens, while LV-Eval~\cite{yuan2024lv} evaluates question answering across 9 datasets. We report LV-Eval at 16K, 32K, and 64K; Appendix~\ref{appendix:exp_long} includes 128K and 256K.

\noindent
\textbf{Configuration and Hardware.}
Experiments use NVIDIA A100 GPUs with 40\,GB memory. \modelname employs $64\times64$ weight tiles, retaining selected tiles at 16-bit and quantizing the rest to INT8. The main tables report 1\% and 20\% retention, while LongEval and Appendix~\ref{appendix:exp} additionally cover 5\% and 10\%.

\noindent
\textbf{Quality Comparisons.}
We compare against the original BF16 model and 5 quantized baselines: Uniform W8, Uniform W8A8, GPTQ~\cite{frantar2022gptq}, LLM.int8~\cite{dettmers2022gpt3}, and SmoothQuant~\cite{xiao2023smoothquant}. Uniform W8 uses channel-wise 8-bit weights with 16-bit activations, whereas W8A8 quantizes activations per token. GPTQ uses 8-bit weights and 16-bit activations, while SmoothQuant uses simulated W8A8 quantization. To ensure consistent quality comparisons, methods share the checkpoint, prompts, decoding settings, and evaluation protocol within each benchmark. Configurations and measurement details are provided in Appendix~\ref{appendix:exp_setup}.

\begin{figure*}[t]
    \vspace{-10pt}
    \centering

    \begin{subfigure}[b]{0.49\linewidth}
        \centering
        \includegraphics[width=\linewidth]{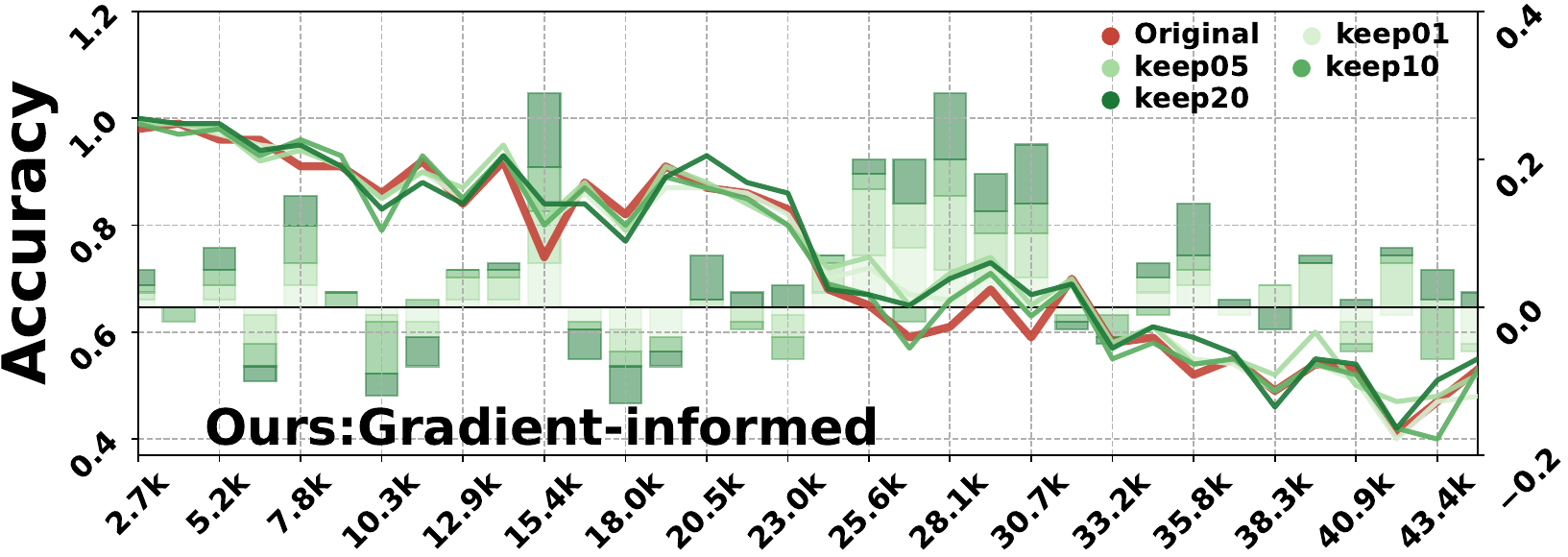}
    \end{subfigure}
    \hfill
    \begin{subfigure}[b]{0.495\linewidth}
        \centering
        \includegraphics[width=\linewidth]{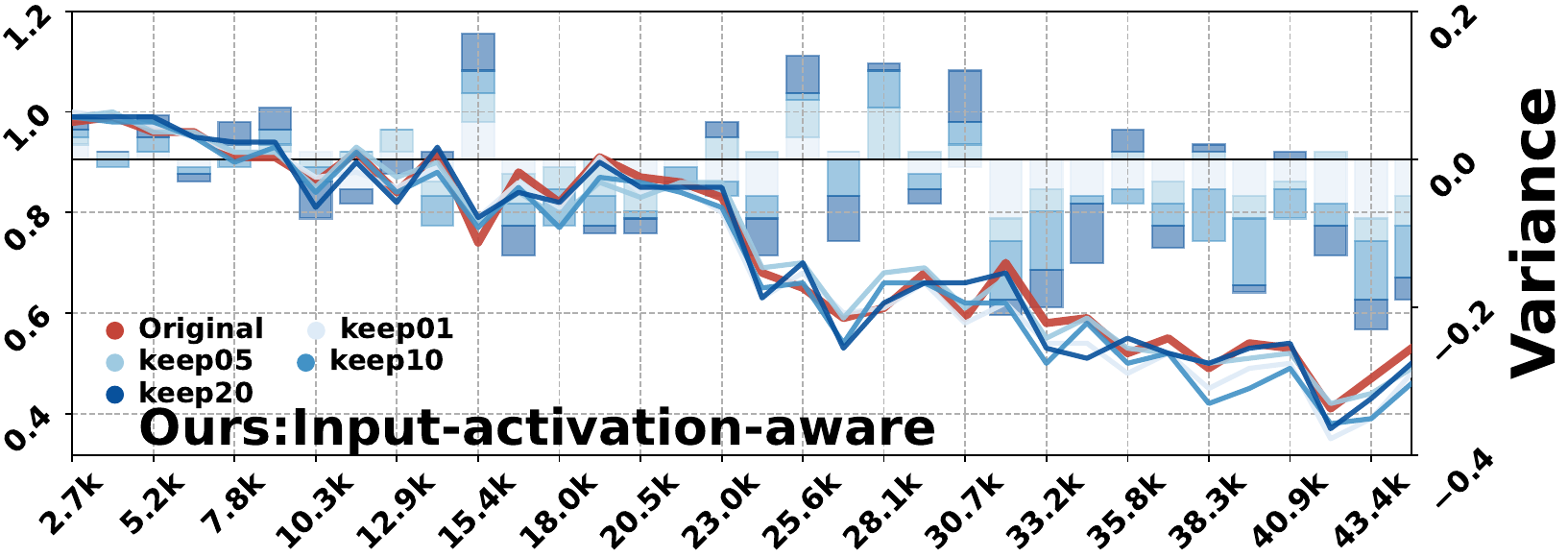}
    \end{subfigure}

    \begin{subfigure}[b]{0.495\linewidth}
        \centering
        \includegraphics[width=\linewidth]{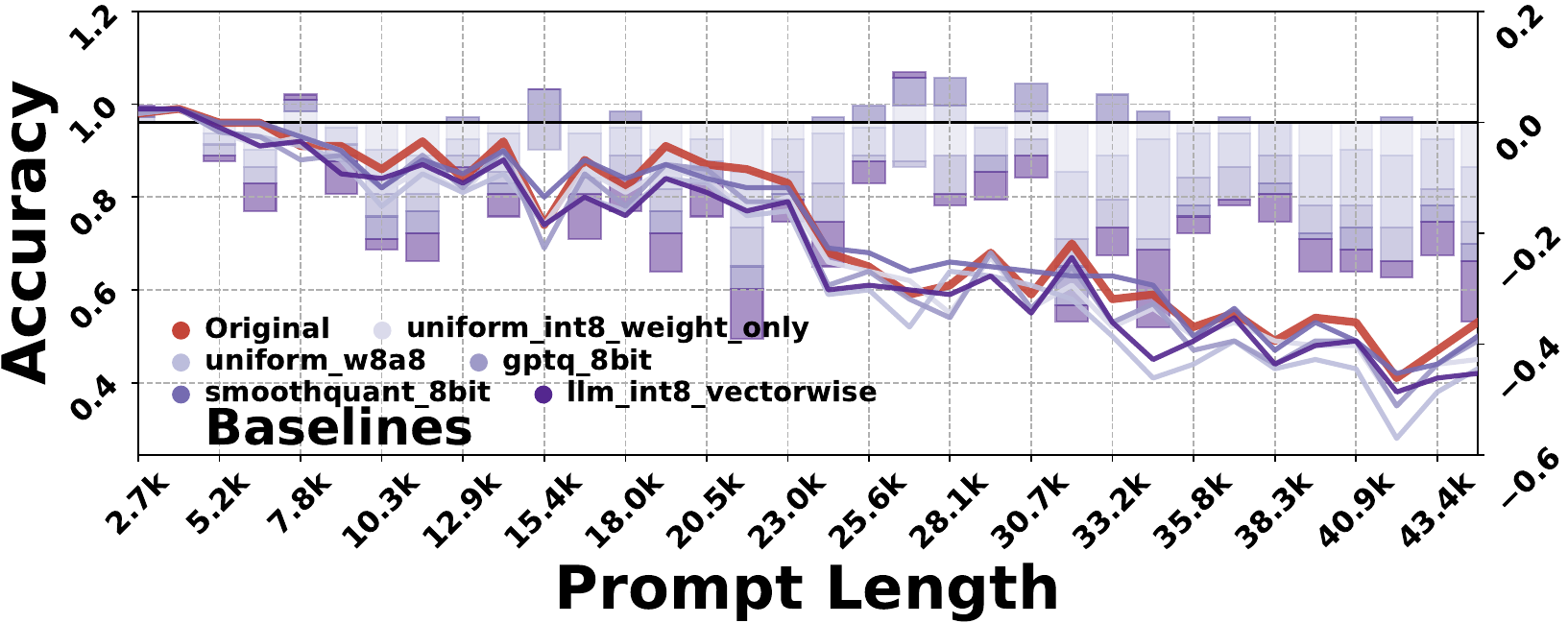}
    \end{subfigure}
    \hfill
    \begin{subfigure}[b]{0.49\linewidth}
        \centering
        \includegraphics[width=\linewidth]{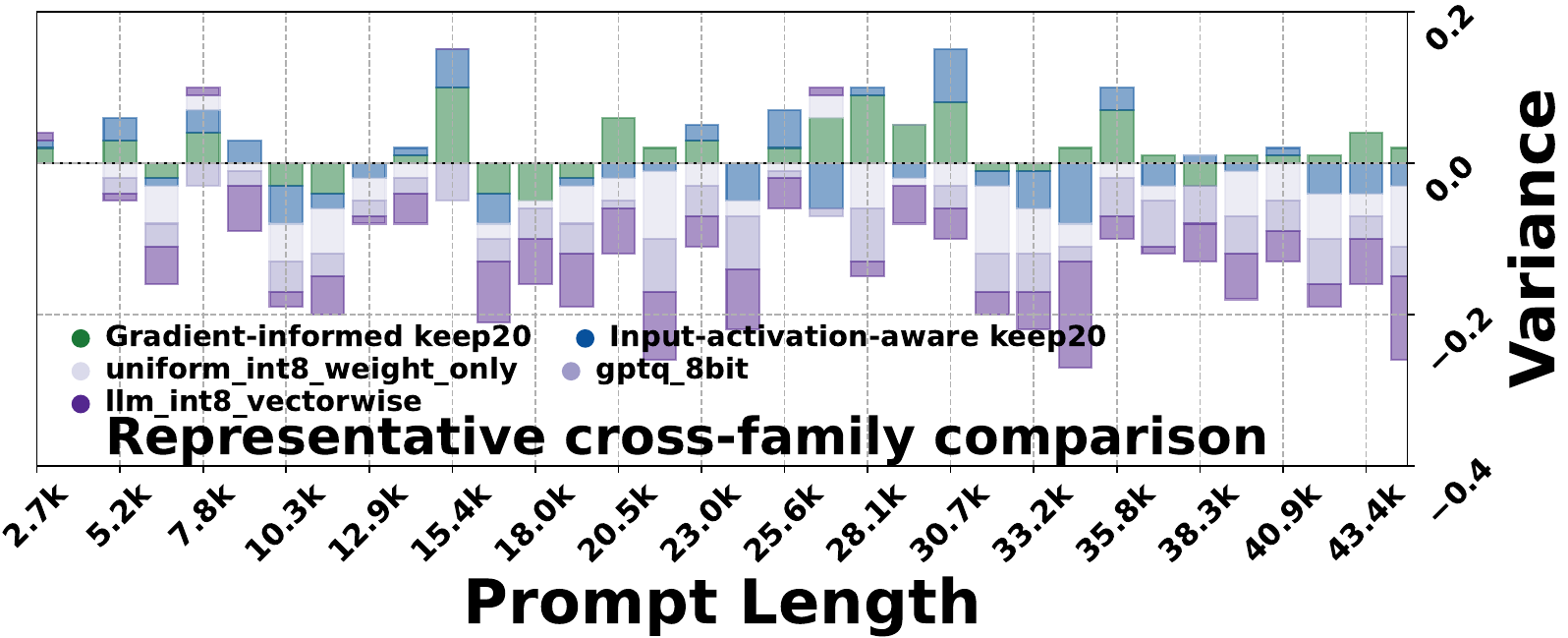}
    \end{subfigure}
    \vspace{-10pt}
\caption{
Line-level retrieval on \textbf{Llama-3.2-3B} across 2.7K--43.4K prompt tokens. Lines (left axis) show exact-match accuracy, while stacked bar segments (right axis) show each configuration's accuracy difference from Original. Top: \modelname with GI maps (left) and IAA maps (right), both retaining 1\%, 5\%, 10\%, or 20\% of tiles in floating point (keep01, keep05, keep10, keep20). Bottom: uniform INT8 weight-only, uniform W8A8, GPTQ, SmoothQuant, and LLM.int8 (left); representative comparison of the strongest \modelname and baseline configurations (right).
}
    \vspace{-15pt}
    \label{fig:longeval_llama3b}
\end{figure*}

\vspace{-5pt}
\subsection{Inference Efficiency}
\label{sec:efficiency}

We measure complete-request generation throughput to evaluate the practical speed benefit of mixed-precision execution. BF16 denotes the original bfloat16 model with scaled dot-product attention (SDPA), while Torch uses Hugging Face half-precision (FP16) generation with eager attention. Both GI$_1$ and GI$_{20}$ exceed the displayed baselines across all 64 settings on Llama-3.2-3B in Figure~\ref{fig:generation_throughput_grid}. With a 128-token prompt and 512 generated tokens at batch size 1, GI$_1$ reaches 154.98 tokens/s, \textbf{$2.66\times$ BF16 and $5.87\times$ GPTQ throughput}. GI$_{20}$ retains 96.7--98.6\% of GI$_1$ throughput across the grid with 20\% of target tiles in BF16. Separate geometry controls report lower prefill and decode latency for tiles than for full-column and compact-strip implementations at matched retention (Appendix~\ref{appendix:geometry_controls}).

The relative benefit varies with the balance between prompt processing and token generation. For long prompts, \textbf{longer continuations strengthen the throughput advantage}: with 16K input tokens at batch size 8, GI$_{20}$ improves from $1.07\times$ to $1.91\times$ BF16 throughput as output length increases from 128 to 1024 tokens. Longer continuations spread prompt-processing cost over more generated tokens, giving repeated decoding a larger role in the complete request. During decoding, \modelname reuses its calibrated precision map and stored tiles in mixed-precision arithmetic.

\vspace{-5pt}
\findingbox{Tile execution speeds up tested generation workloads, preserving nearly all GI throughput from 1\% to 20\% retention.}
\vspace{-5pt}

\begin{table*}[th]

\centering
\scriptsize
\setlength{\tabcolsep}{5.3333pt}
\renewcommand{\arraystretch}{0.75}
\caption{
Long-context question answering results for \textbf{Llama-3.2-3B} on LV-Eval at the 16K, 32K, and 64K length levels. Dataset IDs 1--9 denote CMRC, DuReader, HotpotWikiQA, LIC, LooGLE-CR, LooGLE-MIR, LooGLE-SD, English and Chinese MultiFieldQA, respectively. Cell colors are relative to Original, with \textcolor{green}{green} for higher scores and \textcolor{blue}{blue} for lower scores.
}
\vspace{-8pt}
\label{tab:lveval_llama3b}

\setlength{\aboverulesep}{0.25ex}
\setlength{\belowrulesep}{0.35ex}
\begin{adjustbox}{width=\textwidth}
\begin{tabular}{c c c c c c c c c c c c}
\toprule
\multirow{2}{*}{\textbf{Dataset}} &
\multirow{2}{*}{\textbf{Len}} &
\multirow{2}{*}{\textbf{Original}} &
\multicolumn{4}{c}{\textbf{Ours}} &
\multicolumn{2}{c}{\textbf{Uniform}} &
\multirow{2}{*}{\textbf{GPTQ}} &
\multirow{2}{*}{\textbf{SmoothQ}} &
\multirow{2}{*}{\textbf{LLM.int8}} \\
\cmidrule(lr){4-7}\cmidrule(lr){8-9}
& & &
\textbf{GI$_{1}$} & \textbf{IAA$_{1}$} &
\textbf{GI$_{20}$} & \textbf{IAA$_{20}$} &
\textbf{W8} & \textbf{W8A8} &
& & \\
\midrule

\multirow{3}{*}{\ds{1}}
 & 16k \setbase{21.41} & \cellcolor{heatneutral}\base & \tcell{23.82} & {\bfseries {\tcell{24.21}}} & \tcell{24.06} & \tcell{23.87} & \tcell{21.15} & \tcell{21.39} & \tcell{21.53} & \tcell{20.15} & \tcell{20} \\
 & 32k \setbase{16.86} & \cellcolor{heatneutral}\base & \tcell{18.16} & \tcell{18.23} & {\bfseries {\tcell{18.38}}} & \tcell{18.23} & \tcell{17.95} & \tcell{17.05} & \tcell{17.39} & \tcell{16.06} & \tcell{16.31} \\
 & 64k \setbase{7.5} & \cellcolor{heatneutral}\base & \tcell{8.48} & \tcell{8} & {\bfseries {\tcell{8.59}}} & \tcell{8.04} & \tcell{7.46} & \tcell{6.93} & \tcell{7.58} & \tcell{7.24} & \tcell{6.47} \\
\midrule

\multirow{3}{*}{\ds{2}}
 & 16k \setbase{16.46} & \cellcolor{heatneutral}\base & \tcell{16.45} & \tcell{17.37} & \tcell{16.64} & \tcell{16.5} & \tcell{16.54} & \tcell{17.33} & \tcell{16.07} & \tcell{17.47} & {\bfseries {\tcell{17.88}}} \\
 & 32k \setbase{17.1} & {\bfseries {\cellcolor{heatneutral}\base}} & \tcell{16.48} & \tcell{16.88} & \tcell{16.39} & \tcell{17.03} & \tcell{16.6} & \tcell{16.88} & \tcell{15.83} & \tcell{16.64} & \tcell{16.37} \\
 & 64k \setbase{15.52} & \cellcolor{heatneutral}\base & \tcell{15.4} & \tcell{14.72} & {\bfseries {\tcell{15.76}}} & \tcell{15.47} & \tcell{14.55} & \tcell{14.97} & \tcell{15.04} & \tcell{14.77} & \tcell{14.84} \\
\midrule

\multirow{3}{*}{\ds{3}}
 & 16k \setbase{24.49} & \cellcolor{heatneutral}\base & \tcell{25.29} & \tcell{24.22} & {\bfseries {\tcell{25.56}}} & \tcell{24.37} & \tcell{24.33} & \tcell{20.71} & \tcell{24.74} & \tcell{23.15} & \tcell{23.52} \\
 & 32k \setbase{19.23} & \cellcolor{heatneutral}\base & \tcell{18.43} & \tcell{17.28} & \tcell{19.48} & \tcell{19.15} & \tcell{17.82} & \tcell{19.12} & \tcell{18.35} & {\bfseries {\tcell{20.74}}} & \tcell{17.7} \\
 & 64k \setbase{15.86} & \cellcolor{heatneutral}\base & \tcell{13.61} & {\bfseries {\tcell{17.02}}} & \tcell{16.1} & \tcell{15.96} & \tcell{15.89} & \tcell{14.24} & \tcell{15.08} & \tcell{15.73} & \tcell{15.82} \\
\midrule

\multirow{3}{*}{\ds{4}}
 & 16k \setbase{11.23} & \cellcolor{heatneutral}\base & \tcell{14.08} & \tcell{14.05} & {\bfseries {\tcell{14.21}}} & \tcell{14.09} & \tcell{11.17} & \tcell{11.99} & \tcell{10.76} & \tcell{11.74} & \tcell{12.39} \\
 & 32k \setbase{11.39} & \cellcolor{heatneutral}\base & \tcell{12.68} & \tcell{10.17} & {\bfseries {\tcell{12.83}}} & \tcell{11.34} & \tcell{10.84} & \tcell{11.6} & \tcell{10.72} & \tcell{11.17} & \tcell{12.17} \\
 & 64k \setbase{3.45} & \cellcolor{heatneutral}\base & \tcell{5.22} & {\bfseries {\tcell{5.3}}} & \tcell{5.27} & \tcell{5.24} & \tcell{4.29} & \tcell{4.43} & \tcell{4.04} & \tcell{3.93} & \tcell{4.36} \\
\midrule

\multirow{3}{*}{\ds{5}}
 & 16k \setbase{22.39} & \cellcolor{heatneutral}\base & \tcell{24.12} & \tcell{24.04} & {\bfseries {\tcell{24.37}}} & \tcell{24.11} & \tcell{22.98} & \tcell{24.01} & \tcell{22.37} & \tcell{23.01} & \tcell{22.68} \\
 & 32k \setbase{23.02} & \cellcolor{heatneutral}\base & \tcell{24.32} & {\bfseries {\tcell{24.63}}} & \tcell{24.61} & \tcell{24.41} & \tcell{23.61} & \tcell{23.6} & \tcell{23.12} & \tcell{23.51} & \tcell{22.39} \\
 & 64k \setbase{15.49} & \cellcolor{heatneutral}\base & \tcell{15.6} & \tcell{15.67} & \tcell{15.84} & \tcell{15.7} & \tcell{16.68} & \tcell{16.49} & \tcell{15.54} & {\bfseries {\tcell{17.08}}} & \tcell{15.24} \\
\midrule

\multirow{3}{*}{\ds{6}}
 & 16k \setbase{17.88} & \cellcolor{heatneutral}\base & \tcell{19.77} & \tcell{19.65} & {\bfseries {\tcell{19.97}}} & \tcell{19.71} & \tcell{18.61} & \tcell{19.21} & \tcell{18.62} & \tcell{18.85} & \tcell{18.26} \\
 & 32k \setbase{15.77} & \cellcolor{heatneutral}\base & \tcell{16.13} & \tcell{15.77} & \tcell{16.33} & \tcell{15.83} & \tcell{15.91} & \tcell{16.13} & \tcell{15.3} & {\bfseries {\tcell{16.39}}} & \tcell{16.01} \\
 & 64k \setbase{11.77} & \cellcolor{heatneutral}\base & \tcell{12.48} & \tcell{12.01} & {\bfseries {\tcell{12.66}}} & \tcell{12.07} & \tcell{12.45} & \tcell{11.52} & \tcell{11.63} & \tcell{12.39} & \tcell{11.97} \\
\midrule

\multirow{3}{*}{\ds{7}}
 & 16k \setbase{52.61} & \cellcolor{heatneutral}\base & \tcell{51.86} & {\bfseries {\tcell{53.73}}} & \tcell{52.89} & \tcell{52.67} & \tcell{52.64} & \tcell{51.64} & \tcell{52.84} & \tcell{53.19} & \tcell{52.69} \\
 & 32k \setbase{37.75} & \cellcolor{heatneutral}\base & \tcell{37.72} & \tcell{37.97} & {\bfseries {\tcell{38.07}}} & \tcell{37.85} & \tcell{36.52} & \tcell{36.64} & \tcell{36.98} & \tcell{35.57} & \tcell{38.03} \\
 & 64k \setbase{24.31} & \cellcolor{heatneutral}\base & \tcell{26.22} & \tcell{23.7} & {\bfseries {\tcell{26.59}}} & \tcell{24.24} & \tcell{24.64} & \tcell{24.14} & \tcell{24.69} & \tcell{26.24} & \tcell{25.66} \\
\midrule

\multirow{3}{*}{\ds{8}}
 & 16k \setbase{33.81} & \cellcolor{heatneutral}\base & \tcell{34.28} & {\bfseries {\tcell{36.01}}} & \tcell{34.56} & \tcell{34.34} & \tcell{33.98} & \tcell{34.1} & \tcell{32.69} & \tcell{35.75} & \tcell{34.64} \\
 & 32k \setbase{25.95} & \cellcolor{heatneutral}\base & \tcell{27.51} & \tcell{25.08} & {\bfseries {\tcell{27.83}}} & \tcell{25.85} & \tcell{25.14} & \tcell{26.03} & \tcell{25.84} & \tcell{26.5} & \tcell{25.74} \\
 & 64k \setbase{21.24} & \cellcolor{heatneutral}\base & \tcell{20.22} & \tcell{21.27} & \tcell{21.56} & \tcell{21.38} & \tcell{21.39} & \tcell{20.01} & {\bfseries {\tcell{21.58}}} & \tcell{20.89} & \tcell{21.42} \\
\midrule

\multirow{3}{*}{\ds{9}}
 & 16k \setbase{19.05} & \cellcolor{heatneutral}\base & \tcell{23.38} & \tcell{17.99} & {\bfseries {\tcell{23.59}}} & \tcell{18.96} & \tcell{20.91} & \tcell{19.41} & \tcell{18.84} & \tcell{20.2} & \tcell{22.97} \\
 & 32k \setbase{13.48} & \cellcolor{heatneutral}\base & \tcell{15.95} & \tcell{15.37} & \tcell{16.12} & \tcell{15.42} & \tcell{14.77} & \tcell{15.62} & \tcell{14.19} & {\bfseries {\tcell{18.02}}} & \tcell{14.35} \\
 & 64k \setbase{13.71} & \cellcolor{heatneutral}\base & \tcell{13.85} & \tcell{11.61} & \tcell{14.06} & \tcell{13.67} & \tcell{13.51} & \tcell{13.72} & \tcell{13.95} & {\bfseries {\tcell{14.49}}} & \tcell{12.57} \\
\bottomrule
\end{tabular}
\end{adjustbox}
\vspace{-10pt}
\end{table*}

\vspace{-5pt}
\subsection{General Tasks}
\vspace{-5pt}

Having established generation speedups, we evaluate quality under the same tile-retention settings. GI$_{20}$ leads five of the seven benchmarks in Table~\ref{tab:general_llama3b}. Its LAMBADA accuracy is 65.78\%, compared with 64.43\% for GPTQ and 64.23\% for Original, while WikiText-2 perplexity is 12.403 versus 12.487 for Original. The same loss-guided maps therefore support language modeling and several downstream tasks beyond the calibration inputs.

With only 1\% tile retention, GI$_1$ and IAA$_1$ achieve perplexities of 12.518 and 12.537, respectively, within 0.05 of Original while 99\% of target tiles use 8-bit weights. Appendix~\ref{appendix:exp_general} reports complete retention sweeps, and Appendix~\ref{appendix:selection_controls} compares IAA, magnitude-based, and random retention at equal high-precision capacity.

\vspace{-5pt}
\findingbox{1\% tile retention preserves near-original language-modeling quality; GI improves accuracy on several downstream tasks.}
\vspace{-5pt}

\subsection{Long Context Tasks}

Beyond general benchmarks, we examine whether the same offline-selected tiles preserve evidence retrieval and question answering as inputs grow.

\noindent
\textbf{Long-context Retrieval.}
LongEval requires locating a queried line and returning its content. Figure~\ref{fig:longeval_llama3b} shows that GI generally follows Original closely across the length sweep, with GI$_{20}$ improving accuracy at multiple longer-input settings. At 43.4K tokens, GI$_{20}$ reaches 51\% accuracy, compared with 47\% for Original and 44\% for GPTQ. At matched retention, the GI and IAA comparison further shows how the choice of calibration route affects retrieval quality.

\findingbox{A single calibrated map supports retrieval across lengths. GI$_{20}$ exceeds Original and GPTQ at the longest tested context.}

\noindent
\textbf{Long-context Question Answering.}
Beyond line retrieval, LV-Eval tests question answering from long input contexts across nine datasets, identified in Table~\ref{tab:lveval_llama3b}. We use keyword-recall-based F1, with blacklist-based F1 for CMRC and blacklist-based ROUGE-L for DuReader.

GI$_{20}$ exceeds Original in 26 of the 27 displayed dataset--length settings. On LooGLE-SD at 64K, it scores 26.59, compared with 24.69 for GPTQ and 24.31 for Original. The improvements across datasets and lengths show that a map calibrated once remains useful across long-context tasks.

The forward-only route preserves strong performance with limited retention. On HotpotWikiQA at 64K, IAA$_1$ scores 17.02, compared with 15.86 for Original and 15.89 for Uniform W8. Appendix~\ref{appendix:exp_long} reports more models and length levels.

\findingbox{Both GI and IAA maintain question-answering quality across datasets and input lengths with maps calibrated once.}

\vspace{-5pt}
\section{Conclusion}
\vspace{-5pt}

\modelname uses 2D tiles to connect local precision allocation with executable mixed-precision inference. Calibration scores determine which weight regions retain 16-bit precision, and the resulting maps guide compact storage and dense tile computation. Experiments show faster generation alongside strong language-modeling, reasoning, retrieval, and question-answering quality. These results suggest aligning precision-allocation granularity with hardware-aligned weight tiles for GPU computation. This alignment allows high precision to follow local sensitivity while preserving regular memory access and dense computation.

\clearpage

\section*{Limitations}
\modelname studies training-free mixed-precision inference with 8-bit weights and retained 16-bit tiles. Precision maps are calibrated offline and reused across inference inputs. Input- or task-adaptive allocation would require an additional runtime decision mechanism. Extending the framework to training requires corresponding treatment of gradients, optimizer states, and forward-backward numerical consistency. 

\vspace{-15pt}

\newcommand{\NotationTable}{%
\begin{table*}[!t]
\centering
\caption{Notation used in the main text and Appendix of \modelname. Each calibration batch contains one record.}
\label{tab:notation}
\footnotesize
\setlength{\tabcolsep}{4pt}
\renewcommand{\arraystretch}{0.80}
\setlength{\aboverulesep}{0.20ex}
\setlength{\belowrulesep}{0.25ex}
\begin{adjustbox}{width=\textwidth}
\begin{tabular}{@{}l l l@{}}
\toprule
\textbf{Category} & \textbf{Symbol} & \textbf{Definition} \\
\midrule
\multirow{4}{*}{\textbf{Projection}}
& $X\in\mathbb{R}^{N\times K},\ W\in\mathbb{R}^{M\times K}$ & Input activations and projection weights \\
& $Y=XW^\top$ & Projection output \\
& $N,K,M$ & Activation rows, input features, output features \\
& $x,\ i,j,n$ & Activation row; output-feature, input-feature, row indices \\
\midrule
\multirow{5}{*}{\textbf{Tiling}}
& $r,c;\ n_{\mathrm{tr}},n_{\mathrm{tc}}$ & One-based tile coordinates; tile-row and tile-column counts \\
& $b_M,b_K;\ M',K'$ & Weight-tile dimensions; padded weight dimensions \\
& $\mathcal{T}(W),t;\ \operatorname{id}(r,c)$ & Tile-index set, tile index $t=(r,c)$; zero-based stored ID \\
& $W^{(r,c)}=W_t$ & Weight tile at row $r$ and column $c$ \\
& $X^{(c)},Y^{(r)}$ & Input- and output-feature tiles over the activation rows \\
\midrule
\multirow{9}{*}{\textbf{Calibration}}
& $\theta,\mathcal{P},\mathcal{D}$ & Model parameters, target projections, calibration batches \\
& $b;\ X_b,N_b$ & Calibration batch index; its activations and row count \\
& $z_{b,\tau},T_b,p_\theta$ & Token at position $\tau$, token count, predictive distribution \\
& $\mathcal{L}_b,\ g_{ij,b}$ & Batch loss and its gradient $\partial\mathcal{L}_b/\partial W_{ij}$ \\
& $a_j$ & Mean absolute activation of input feature $j$ \\
& $I_{ij}^{\mathrm{act}},I_{ij,b}^{\mathrm{grad}}$ & IAA entry score; GI entry score for batch $b$ \\
& $S_{t,b}^{\mathrm{grad}}$ & GI score of tile $t$ in batch $b$ \\
& $S_t^{\mathrm{act}},S_t^{\mathrm{grad}}$ & Completed IAA and GI tile scores \\
& $\epsilon,E_{ij}$ & Relative weight change; unit-entry matrix at $(i,j)$ \\
\midrule
\multirow{4}{*}{\textbf{Allocation}}
& $T=|\mathcal{T}(W)|,\ S_t$ & Tile count; completed score of either calibration route \\
& $\rho,k$ & Target 16-bit retention ratio; selected tile count \\
& $\mathcal{F}(W),\mathcal{Q}(W)$ & 16-bit and 8-bit tile-index sets \\
& $\mathcal{F}_r(W),\mathcal{Q}_r(W)$ & Retained and quantized column-tile indices in row $r$ \\
\midrule
\multirow{6}{*}{\textbf{Quantization}}
& $Q(V;\Delta)$ & Signed 8-bit quantizer for array $V$ with positive scale $\Delta$ \\
& $\widehat{X},\widehat{W};\ \widehat{X}^{(c)},\widehat{W}^{(r,c)}$ & Integer operands and their input-feature or weight tiles \\
& $\Delta_X,\Delta_W;\ \Delta_X^{(c)},\Delta_W^{(r,c)}$ & Shared operand scales; per-input-feature and per-weight scales \\
& $m_c$ & Absolute activation maximum over input-feature group $c$ \\
& $\Delta_{0,t},\Delta_{\alpha,t},\alpha$ & Absmax weight scale, candidate scale, candidate multiplier \\
& $E_t(\Delta)$ & Activation-weighted weight-tile reconstruction error \\
\midrule
\multirow{2}{*}{\textbf{Storage}}
& $T_8=|\mathcal{Q}(W)|,T_{16}=|\mathcal{F}(W)|$ & Quantized and retained weight-tile counts \\
& $\rho_{\mathrm{real}},S_{\mathrm{pack}}$ & Realized fraction $T_{16}/T$; packed storage in bytes \\
\midrule
\multirow{6}{*}{\textbf{Execution}}
& $\mathcal{B},b_N,N_{\mathrm{call}}$ & Output-tile activation rows, row-tile size, GEMM-call rows \\
& $X_{\mathcal{B}}^{(c)},\widehat{X}_{\mathcal{B}}^{(c)}$ & 16-bit and integer input tiles restricted to rows $\mathcal{B}$ \\
& $D_{\mathcal{B},r,c},A$ & INT32 tile product; FP32 register accumulator \\
& $p,R_p(r),h_{p,r}$ & Precision ($p\in\{8,16\}$), row pointer, row tile count \\
& $u,S,s$ & Packed-buffer position, split count, zero-based split index \\
& $\delta r$ & Output-tile rows preceding a grouped projection \\
\midrule
\multirow{2}{*}{\textbf{Evaluation}}
& $B,P,O,T_{\mathrm{total}}$ & Request batch size, prompt tokens, output tokens, request time \\
& $\mathcal{R}_1,\mathcal{R}_2;\ J(\mathcal{R}_1,\mathcal{R}_2)$ & Retained sets from two draws; their Jaccard similarity \\
\bottomrule
\end{tabular}
\end{adjustbox}
\end{table*}
}
\begingroup
\let\notationoriginalbibitem\bibitem
\renewcommand{\bibitem}{%
  \ifnum\value{page}>10\relax
    \global\let\bibitem\notationoriginalbibitem
    \NotationTable
  \fi
  \notationoriginalbibitem
}
\bibliography{reference}
\endgroup

\appendix
\setcounter{topnumber}{3}
\setcounter{bottomnumber}{2}
\setcounter{totalnumber}{5}
\setcounter{dbltopnumber}{3}
\renewcommand{\topfraction}{0.92}
\renewcommand{\bottomfraction}{0.95}
\renewcommand{\dbltopfraction}{0.92}
\renewcommand{\textfraction}{0.05}
\clubpenalty=10000
\widowpenalty=10000
\displaywidowpenalty=10000
\renewcommand{\floatpagefraction}{0.80}
\renewcommand{\dblfloatpagefraction}{0.75}
\setlength{\textfloatsep}{8pt plus 2pt minus 2pt}
\setlength{\dbltextfloatsep}{8pt plus 2pt minus 2pt}
\setlength{\floatsep}{8pt plus 2pt minus 2pt}
\setlength{\dblfloatsep}{8pt plus 2pt minus 2pt}
\makeatletter
\setlength{\@fptop}{0pt}
\setlength{\@fpsep}{8pt}
\setlength{\@fpbot}{0pt plus 1fil}
\setlength{\@dblfptop}{0pt}
\setlength{\@dblfpsep}{8pt}
\setlength{\@dblfpbot}{0pt plus 1fil}
\makeatother

\section{Notation and Reading Guide}
\label{app:notation}

Table~\ref{tab:notation} summarizes the notation used for tile-wise precision allocation, calibration, and mixed-precision execution.
Appendix~\ref{app:calibration_procedure} details calibration, and Appendix~\ref{app:score_interpretation} explains the importance scores and per-layer selection.
Appendix~\ref{app:numerical_representation} describes numerical representation and load-time reconstruction, followed by GPU execution in Appendix~\ref{app:gpu_execution}.
Appendix~\ref{appendix:exp} follows the main-text evaluation sequence, beginning with the detailed generation protocol and inference-efficiency measurements in Appendix~\ref{appendix:exp_setup} and Appendix~\ref{appendix:exp_efficiency}. Extended general-task results appear in Appendix~\ref{appendix:exp_general}, followed by long-context retrieval and question answering in Appendix~\ref{appendix:exp_long}. Appendix~\ref{appendix:allocation_analysis} examines calibration and allocation choices through controls.

Within a weight-tile row, $r$ identifies the output-feature tile, while $c$ traverses the corresponding input-feature tiles.
The retained and quantized tile sets therefore determine which 16-bit and 8-bit products contribute to the same output tile.

\section{Calibration Procedure}
\label{app:calibration_procedure}

Calibration converts model observations into one importance map per target weight matrix. This section describes the observations collected by input-activation-aware (IAA) and gradient-informed (GI) scoring. The resulting maps enter the common selection rule in Appendix~\ref{app:retention_rule}, while IAA channel statistics also guide the weight-scale construction in Appendix~\ref{app:weight_quantization}.

\subsection{Calibration Inputs and Target Projections}
\label{app:calibration_inputs}

The calibration builder uses 8,192 document-summary records from the Multi-News training split in dataset order. Each document is placed in a summarization prompt, followed by its reference summary. The prompt and summary are tokenized separately with limits of 4,096 and 256 tokens, respectively. The tokenizer's beginning-of-sequence token is prepended when available, and its end-of-sequence token is appended to the summary. Each calibration batch $b$ contains one concatenated sequence. Both routes therefore process the same records in dataset order.

The target set $\mathcal{P}$ contains the query, key, value, and attention-output projections, together with the gate, up, and down projections in each decoder layer. Calibration runs the pretrained model in evaluation mode with key/value caching disabled. Its parameters remain fixed throughout observation collection. IAA collects the inputs to these projections, while GI differentiates the mean next-token cross-entropy over the concatenated sequence. Thus, the two routes share the model and records while extracting different information for local precision placement.

\subsection{Input-Activation Statistics}
\label{app:iaa_statistics}

IAA needs the mean absolute input magnitude of each channel. For a target projection, let $X_b\in\mathbb{R}^{N_b\times K}$ contain the activation rows observed in calibration batch $b$, and let $\mathcal{D}$ index these batches. Input hooks accumulate channel sums and the number of observed rows, giving
\begin{equation}
 a_j=
 \frac{\sum_{b\in\mathcal{D}}\sum_{n=1}^{N_b}|X_{b,nj}|}
      {\sum_{b\in\mathcal{D}}N_b}.
 \label{eq:app_iaa_mean}
\end{equation}
Weighting by the observed row count makes every activation row contribute equally, including when records have different lengths.

Per-record channel reductions use 32-bit floating point (FP32), and their running totals use 64-bit floating point (FP64). The collected means are stored in FP32 and combined with fixed weights:
\begin{equation}
 S_t^{\mathrm{act}}=\sum_{(i,j)\in t}|W_{ij}|a_j.
 \label{eq:app_iaa_score}
\end{equation}
Each input hook immediately reduces its observations to channel statistics, so the persistent calibration state consists of sums and counts. The final tile scores determine retention priority, and the channel means accompany the IAA decision artifacts for load-time scale selection.

\subsection{Gradient-Informed Score Accumulation}
\label{app:gi_accumulation}

GI identifies weights whose relative changes influence prediction loss. For calibration batch $b$ with sequence $(z_{b,1},\ldots,z_{b,T_b})$, the loss is
\begin{equation}
 \mathcal{L}_b=-\frac{1}{T_b-1}
 \sum_{\tau=1}^{T_b-1}
 \log p_\theta(z_{b,\tau+1}\mid z_{b,1:\tau}).
 \label{eq:app_calibration_loss}
\end{equation}
The shifted prediction targets cover the concatenated prompt and summary. Backpropagation gives $g_{ij,b}=\partial\mathcal{L}_b/\partial W_{ij}$, so the entry score is $I_{ij,b}^{\mathrm{grad}}=|W_{ij}g_{ij,b}|$.

The collector forms the absolute weight-gradient products in FP32, reduces them within each tile, and adds the resulting tile scores across records:
\begin{equation}
\begin{aligned}
 S_{t,b}^{\mathrm{grad}}&=\sum_{(i,j)\in t}I_{ij,b}^{\mathrm{grad}},\\
 S_t^{\mathrm{grad}}&=\sum_{b\in\mathcal{D}}S_{t,b}^{\mathrm{grad}}.
\end{aligned}
 \label{eq:app_gi_score}
\end{equation}
\textbf{Taking magnitudes before aggregation prevents cancellation} between positive and negative weight-gradient terms within a tile or across records. Gradients are cleared before the next record, while accumulated FP32 tile maps remain on the host. The completed maps enter the common retention rule, as summarized in Algorithm~\ref{alg:app_calibration}.

\begin{algorithm*}[t]
\caption{Collecting calibration statistics and tile scores}
\label{alg:app_calibration}
\small
\begin{algorithmic}[1]
\REQUIRE Fixed pretrained model $\theta$, target projections $\mathcal{P}$, ordered one-record batches $\mathcal{D}$, route $\in\{\mathrm{IAA},\mathrm{GI}\}$
\ENSURE One tile-score map per target projection; channel means for IAA
\IF{route is IAA}
    \STATE Initialize a channel-sum vector and row count for each projection in $\mathcal{P}$
    \FORALL{calibration batches $b\in\mathcal{D}$}
        \STATE Run the model with gradient recording disabled
        \STATE Input hooks add channel-wise absolute activation sums and row counts
    \ENDFOR
    \FORALL{projections in $\mathcal{P}$}
        \STATE Compute $a_j$ using Eq.~\eqref{eq:app_iaa_mean}
        \STATE Compute $S_t^{\mathrm{act}}$ using Eq.~\eqref{eq:app_iaa_score}; save channel means
    \ENDFOR
\ELSE
    \STATE Initialize each target projection's tile-score map to zero
    \FORALL{calibration batches $b\in\mathcal{D}$}
        \STATE Clear parameter gradients and evaluate $\mathcal{L}_b$ using Eq.~\eqref{eq:app_calibration_loss}
        \STATE Backpropagate $\mathcal{L}_b$ through the fixed model
        \FORALL{target weights $W$ and tiles $t\in\mathcal{T}(W)$}
            \STATE $S_t^{\mathrm{grad}}\leftarrow S_t^{\mathrm{grad}}+\sum_{(i,j)\in t}|W_{ij}g_{ij,b}|$
        \ENDFOR
    \ENDFOR
    \STATE Clear parameter gradients
\ENDIF
\STATE Save the tile-score maps for the retention rule in Appendix~\ref{app:retention_rule}
\end{algorithmic}
\end{algorithm*}

\section{Score Interpretation and Precision Selection}
\label{app:score_interpretation}

The collected statistics provide two complementary views of local weight importance. We first relate IAA to activation-weighted contributions and GI to first-order loss sensitivity. We then show how a fixed per-layer retention count turns either score into a precision map.

\subsection{Activation-Weighted Magnitudes}
\label{app:iaa_interpretation}

For a fixed weight $W_{ij}$ and an observed activation row $x$, the corresponding multiply contribution is $W_{ij}x_j$. Taking its magnitude and averaging over calibration rows gives $|W_{ij}|\mathbb{E}[|x_j|]$. Therefore, the total expected magnitude of the multiply contributions inside tile $t$ is
\begin{equation}
\begin{aligned}
 \mathbb{E}\!\left[\sum_{(i,j)\in t}|W_{ij}x_j|\right]
 &=\sum_{(i,j)\in t}|W_{ij}|\mathbb{E}[|x_j|]\\
 &=S_t^{\mathrm{act}}.
\end{aligned}
\label{eq:app_iaa_interpretation}
\end{equation}
The identity gives the IAA score a direct interpretation as the sum of expected absolute multiplication contributions within a tile. At a fixed weight magnitude, a larger mean absolute input increases the score. Summing these contributions lets IAA rank local regions using the channel statistics collected during forward computation.

\subsection{First-Order Weight Sensitivity}
\label{app:gi_interpretation}

GI connects local weight importance to the full model's prediction loss. Consider a relative change $W_{ij}\mapsto(1+\epsilon)W_{ij}$, with all other parameters fixed. Let $E_{ij}$ be a matrix with one at $(i,j)$ and zeros elsewhere. At a differentiable calibration state,
\begin{equation}
 \left.\frac{\mathrm d}{\mathrm d\epsilon}
 \mathcal{L}_b(W+\epsilon W_{ij}E_{ij})
 \right|_{\epsilon=0}
 =W_{ij}\frac{\partial\mathcal{L}_b}{\partial W_{ij}}.
 \label{eq:app_relative_sensitivity}
\end{equation}
Thus, $|W_{ij}g_{ij,b}|$ measures the magnitude of the first-order loss response to a relative change of that weight. Summing these magnitudes within a tile and across calibration records yields Eq.~\eqref{eq:app_gi_score}.

The gradient carries the effect of the target projection through downstream model computation. Consequently, GI prioritizes regions that are important under the prediction objective, while IAA prioritizes regions with large activation-weighted contributions. Both interpretations lead to the same capacity-constrained selection below.

\subsection{Per-Layer Retention and Decision Maps}
\label{app:retention_rule}

Here, a layer denotes one target linear projection with weight matrix $W$. Let $T=|\mathcal{T}(W)|$ and let $S_t$ be the completed IAA or GI score for tile $t$. For a retention ratio $\rho\in(0,1]$, the selected count is
\begin{equation}
 k=\max\!\left(1,\left\lceil\rho T\right\rceil\right).
 \label{eq:app_retained_count}
\end{equation}
The count is applied independently to each target projection. Equal-sized tiles make $k$ a direct control over the number of stored 16-bit weight elements on that tile grid. The zero-retention execution control sets $\mathcal{F}(W)=\emptyset$ and assigns every target tile to $\mathcal{Q}(W)$.

At fixed scores and $k$, retaining the highest-scoring tiles solves
\begin{equation}
 \max_{\mathcal{F}(W)\subseteq\mathcal{T}(W),\ |\mathcal{F}(W)|=k}
 \sum_{t\in\mathcal{F}(W)}S_t.
 \label{eq:app_retention_objective}
\end{equation}
To see this, consider a feasible set that retains a lower-scoring tile while excluding a higher-scoring tile. Exchanging the two increases its retained score. Repeating these exchanges produces a top-$k$ set, which attains the maximum. Equal scores permit equally valued selections under this objective.

The implementation identifies the top-$k$ positions through an array-partition operation and saves the concrete selected tile IDs. Their complement defines $\mathcal{Q}(W)$, and the pair $(\mathcal{F}(W),\mathcal{Q}(W))$ determines the layer's binary precision map. Multiple retention ratios can therefore be constructed from the same completed score map. The saved IDs preserve each chosen allocation through loading and inference, using the representation described next.

\section{Numerical Representation and Load-Time Reconstruction}
\label{app:numerical_representation}

The precision map identifies locations, and reconstruction associates those locations with values, scales, and addresses that the kernel can use. We first define tile coordinates and integer quantization, then describe how the decision archive becomes packed weight buffers. Execution of the mixed-precision modules follows Appendix~\ref{app:gpu_execution}.

\subsection{Tile Coordinates and Padding}
\label{app:tile_coordinates}

We retain the convention $X\in\mathbb{R}^{N\times K}$ and $W\in\mathbb{R}^{M\times K}$, with $N$ activation rows and $M$ output features. Let $b_M$ and $b_K$ be the weight-tile dimensions. The deployed configuration uses $b_M=b_K=64$. Zero padding completes the weight grid to
\begin{equation}
 M'=b_M\left\lceil M/b_M\right\rceil,
 \qquad
 K'=b_K\left\lceil K/b_K\right\rceil,
 \label{eq:app_padded_shape}
\end{equation}
with $n_{\mathrm{tr}}=M'/b_M$ and $n_{\mathrm{tc}}=K'/b_K$.

With one-based tile coordinates $t=(r,c)$, the zero-based stored identifier is
\begin{equation}
 \operatorname{id}(r,c)=(r-1)n_{\mathrm{tc}}+(c-1).
 \label{eq:app_tile_identifier}
\end{equation}
Consequently, integer division and remainder recover the output-tile row and input-tile column. Boundary tiles contain zeros outside the original matrix. The kernel masks activation accesses beyond $K$ and output writes beyond $M$, preserving the original linear-layer interface while processing complete weight tiles.

\subsection{Integer Quantization and Weight Scales}
\label{app:weight_quantization}

Each quantized tile needs integer values and a scale that restores their numerical magnitude. For an array $V$ and positive scale $\Delta$, define
\begin{equation}
 Q(V;\Delta)=\operatorname{clip}_{[-127,127]}
 \left(\operatorname{round}_{\mathrm{even}}(V/\Delta)\right),
 \label{eq:app_integer_quantizer}
\end{equation}
where nearest-integer rounding resolves ties to even. The zero point is zero; reconstruction gives $\Delta Q(V;\Delta)$. For a weight tile $W_t=W^{(r,c)}$, the base scale is
\begin{equation}
 \Delta_{0,t}=
 \begin{cases}
 \displaystyle\max_{(i,j)\in t}|W_{ij}|/127,
     & W_t\ne0,\\
 1, & \text{otherwise}.
 \end{cases}
 \label{eq:app_weight_base_scale}
\end{equation}
The GI path uses this absmax scale. The maximum is obtained from the original 16-bit values, and the scale is computed and stored in FP32.

IAA additionally reuses its channel means to select scales for the tiles assigned to 8-bit. For each such tile, it tests $\Delta_{\alpha,t}=\max\{\alpha\Delta_{0,t},10^{-12}\}$ for $\alpha\in\{0.75,1.00,1.25\}$ and minimizes the activation-weighted reconstruction error
\begin{equation}
\begin{aligned}
 E_t(\Delta)={}&\frac{1}{b_Mb_K}\sum_{(i,j)\in t}\max\{a_j,10^{-8}\}\\
 &{}\times\left(\Delta Q(W_t;\Delta)_{ij}-W_{ij}\right)^2.
\end{aligned}
 \label{eq:app_iaa_scale_error}
\end{equation}
Reducing the scale gives finer quantization steps and a narrower representable range. The weighted error therefore accounts for rounding and clipping while emphasizing channels with larger observed inputs. Candidates are evaluated in FP32 in the listed order, with the first retained in a tie.

The chosen scale $\Delta_W^{(r,c)}$ accompanies the signed INT8 tile $\widehat W^{(r,c)}$ throughout execution. Retained tiles preserve their deployment values in bfloat16 (BF16) or half precision (FP16). Thus, calibration determines the protected locations, while scale construction supplies the numerical representation of the remaining weights.

\subsection{Decision Archives and Packed Tile Layout}
\label{app:tile_layout}

The decision archive records each target projection's name, tile-grid shape, and retained IDs. It stores IDs as unsigned 16-bit integers for grids containing at most $2^{16}-1$ tiles and as unsigned 32-bit integers for larger grids. Loading recovers $\mathcal{F}(W)$ from these IDs and constructs $\mathcal{Q}(W)$ from the complementary grid positions.

For GPU execution, the two precision sets occupy separate contiguous tile buffers. Each buffer is sorted by the identifier in Eq.~\eqref{eq:app_tile_identifier}, so tiles belonging to the same output-tile row are contiguous and ordered by input-tile column. Let $R_p(r)$ count the precision-$p$ tiles in rows preceding $r$, for $p\in\{8,16\}$. Row $r$ occupies positions
\begin{equation}
 [R_p(r),R_p(r+1)).
 \label{eq:app_row_range}
\end{equation}
At buffer position $u$, the kernel reads the weight tile and, for the integer path, its scale. The tile ID supplies the corresponding input-feature tile. The kernel can therefore traverse weights contiguously while using their IDs to select activations.

The packed layer stores $T_8=|\mathcal{Q}(W)|$ integer tiles of shape $[T_8,b_M,b_K]$ and $T_{16}=|\mathcal{F}(W)|$ retained tiles of shape $[T_{16},b_M,b_K]$. Integer tiles use INT8; retained tiles and the optional bias of shape $[M]$ use BF16 or FP16. An FP32 scale array of shape $[T_8]$ restores weight magnitudes.

Each precision set has INT32 tile-ID and row-order arrays of length $T_8$ or $T_{16}$. IDs identify the corresponding input-tile columns, while row-order indices record the packing permutations. Two INT32 row-pointer arrays of shape $[n_{\mathrm{tr}}+1]$ delimit each output-tile row's contiguous range.

The packed representation separates numerical payload from indexing. With realized retained fraction $\rho_{\mathrm{real}}=T_{16}/(T_8+T_{16})$, the weight values require $8+8\rho_{\mathrm{real}}$ bits per padded weight element. Including these scales and indexing arrays, the packed tensor storage before bias is
\begin{equation}
\begin{aligned}
 S_{\mathrm{pack}}={}&b_Mb_K(T_8+2T_{16})+4T_8\\
 &+8(T_8+T_{16})+8(n_{\mathrm{tr}}+1)
 \quad\text{bytes}.
\end{aligned}
\label{eq:app_packed_tensor_bytes}
\end{equation}
The terms account for weight values, integer-tile scales, tile IDs plus row-order indices, and the two row-pointer arrays. Activation buffers, outputs, and execution workspace have input-dependent lifetimes and are described separately in Appendix~\ref{app:gpu_execution}.

\subsection{Loading the Mixed-Precision Model}
\label{app:load_time_details}

The loader constructs the deployed representation directly from the pretrained checkpoint and decision archive. It initializes the model structure, checks each target projection's saved grid against its weight shape, and reads checkpoint tensors on the host. As each target weight becomes available, the loader gathers the retained tiles, quantizes the complementary tiles using Appendix~\ref{app:weight_quantization}, builds their row schedules, and installs the resulting mixed-precision module on the GPU.

The original biases and non-target tensors retain their roles in the model. Compatible projection modules are subsequently grouped using the procedure in Appendix~\ref{app:projection_reuse}. After loading, the selected weights, scales, and indices remain resident across requests. Runtime work therefore begins with preparing the current activations and consuming the already constructed tile representation.

\section{Mixed-Precision GPU Execution}
\label{app:gpu_execution}

The packed layer supplies persistent weights and their numerical interpretation. Execution must prepare the current input, evaluate both precision paths, and combine their contributions into the model's output format. We describe this data flow first, then explain how output scheduling, shared projection inputs, and repeated-generation state reuse the same selected model.

\subsection{Activation Preparation}
\label{app:activation_preparation}

The integer path requires 8-bit activations, while retained tiles use the current 16-bit input. The runtime therefore preserves $X$ and prepares an integer copy grouped by the weight tiles' input-feature ranges. For column group $c$, let
\begin{equation}
 m_c=\max_{1\leq n\leq N,\ j\in c}|X_{nj}|.
 \label{eq:app_activation_max}
\end{equation}
Here, $j\in c$ denotes the original input features covered by that tile column. The maximum covers all activation rows of the current projection input, so one scale is shared across its output-tile rows.

For a nonzero group, the fused GPU quantizer computes $m_c/127$ in FP32, rounds the result to the input's 16-bit datatype, and stores that value in FP32 as $\Delta_X^{(c)}$. An all-zero group uses scale one. Division and nearest-integer rounding then produce
\begin{equation}
 \widehat X^{(c)}=Q(X^{(c)};\Delta_X^{(c)}),
 \label{eq:app_activation_quantization}
\end{equation}
using the integer range in Eq.~\eqref{eq:app_integer_quantizer}. The prepared integer values and scales supply every quantized weight tile consuming those input features.

For inputs with at most 128 rows, maximum reduction and quantization are fused. Larger inputs first compute the shared scales and then quantize row tiles. Above 4,096 rows, the runtime applies quantization and mixed GEMM in chunks of at most 4,096 rows using scales computed over the complete input. Each chunk writes directly into its assigned rows of one output tensor. Consequently, row chunking bounds the temporary integer-activation allocation while retaining the same scale definition.

\subsection{Tile Products and Scaled Accumulation}
\label{app:scaled_arithmetic}

Let $\mathcal{B}$ denote the activation rows assigned to an output tile, and let $r$ identify its output-feature tile. The integer path reads $\widehat X_{\mathcal{B}}^{(c)}$ and the stored $\widehat W^{(r,c)}$ as signed INT8 operands for Tensor Core matrix multiplication. Their tile product accumulates in a signed 32-bit integer (INT32) result,
\begin{equation}
 D_{\mathcal{B},r,c}=
 \widehat X_{\mathcal{B}}^{(c)}
 \big(\widehat W^{(r,c)}\big)^\top.
 \label{eq:app_integer_tile_product}
\end{equation}
The kernel converts this result to FP32 and multiplies it by $\Delta_X^{(c)}\Delta_W^{(r,c)}$ before adding it to the output accumulator. Applying the scales per tile allows differently scaled integer products to contribute in the same numerical units.

The retained path instead multiplies $X_{\mathcal{B}}^{(c)}$ by the stored 16-bit weights, with FP32 accumulation. Within each assigned range, the kernel evaluates integer tiles in ascending input-column order, followed by retained tiles in ascending input-column order. Both paths consequently share a local partial sum while preserving their operand formats.

Once all contributions are combined, the result is cast to the deployment datatype. The original bias is added in that datatype when present, and valid output elements are stored. The module restores the input's leading batch and sequence dimensions, completing the linear-layer interface.

\subsection{Output-Tile Scheduling}
\label{app:output_scheduling}

Prefill provides multiple activation rows that can reuse a loaded weight tile, whereas decode provides fewer rows and benefits from additional parallel work across the weight grid. The kernel therefore chooses its activation-row tile size from the current call's shape, while keeping the weight tiles and precision assignments unchanged. Let $N_{\mathrm{call}}$ be the rows passed to one GEMM invocation, including a row chunk when chunking is active. 

Each output tile spans 64 output features. The kernel uses 16 activation rows per tile for $N_{\mathrm{call}}\leq16$ and 32 rows for $16<N_{\mathrm{call}}\leq32$. For $N_{\mathrm{call}}>64$ with $n_{\mathrm{tr}}\geq4n_{\mathrm{tc}}$, it uses 128 rows; all remaining shapes use 64 rows. Every configuration uses four warps and three loop-pipeline stages.

The split count balances the number of parallel programs against the tile work assigned to each. With activation-row tile size $b_N$, the grid contains $n_{\mathrm{tr}}\lceil N_{\mathrm{call}}/b_N\rceil$ output tiles before splitting. The target is four times the GPU's streaming-multiprocessor count. Once the output grid reaches this target, one program handles each tile. Otherwise, the kernel rounds the target-to-output-tile ratio up to a power of two and caps the split count $S$ at 16. It then halves $S$ while $S>1$ and $\lfloor n_{\mathrm{tc}}/S\rfloor<4$, limiting splitting when few input tiles are available.

Each split receives a contiguous range from both precision-specific tile lists. For precision $p$, let $h_{p,r}=R_p(r+1)-R_p(r)$ be its tile count in row $r$. Split $s\in\{0,\ldots,S-1\}$ processes the buffer positions in
\begin{equation}
 \left[
 R_p(r)+\left\lfloor\frac{h_{p,r}s}{S}\right\rfloor,
 R_p(r)+\left\lfloor\frac{h_{p,r}(s+1)}{S}\right\rfloor
 \right).
 \label{eq:app_split_range}
\end{equation}
These ranges cover each stored tile exactly once. Sequential positions provide contiguous weight addresses, and the three-stage loop pipeline schedules tile loads alongside multiplication.

For $S=1$, the program keeps its complete FP32 sum in registers and writes the final output tile. For $S>1$, each program publishes its FP32 partial to a reusable workspace and increments an integer completion counter with acquire-release ordering. The last arriving program loads the partials in ascending split order, combines them inside the same kernel, and writes the completed output once. The fixed combination order makes the accumulation order independent of program completion order for a given schedule.

The workspace has active FP32 partial slices of shape $[S,N_{\mathrm{call}},M]$ and one INT32 counter per output tile. Buffers are cached per device and CUDA stream, and the completing program resets its counter for reuse. Algorithm~\ref{alg:app_mixed_execution} summarizes the arithmetic and completion steps.

\begin{algorithm*}[t]
\caption{Mixed-precision projection with output-tile scheduling}
\label{alg:app_mixed_execution}
\small
\begin{algorithmic}[1]
\REQUIRE Current input $X$; packed weight tiles, scales, IDs, and row pointers; optional bias
\ENSURE Output in the deployment datatype with the original feature dimensions
\STATE Prepare $\widehat X$ and $\{\Delta_X^{(c)}\}_c$ as in Appendix~\ref{app:activation_preparation}
\STATE Select row-tile size and split count $S$; obtain workspace with reusable zeroed counters
\FORALL{output tiles $(\mathcal{B},r)$ and splits $s$ in parallel}
    \STATE Initialize register accumulator $A\leftarrow0$ in FP32
    \FORALL{positions $u$ in the assigned 8-bit range from Eq.~\eqref{eq:app_split_range}}
        \STATE Recover input-tile column $c$ from its stored ID
        \STATE Compute INT32 product $D_{\mathcal{B},r,c}$ using Eq.~\eqref{eq:app_integer_tile_product}
        \STATE $A\leftarrow A+\operatorname{FP32}(D_{\mathcal{B},r,c})\big(\Delta_X^{(c)}\Delta_W^{(r,c)}\big)$
    \ENDFOR
    \FORALL{positions $u$ in the assigned 16-bit range from Eq.~\eqref{eq:app_split_range}}
        \STATE Recover $c$ and accumulate $X_{\mathcal{B}}^{(c)}(W^{(r,c)})^\top$ into $A$
    \ENDFOR
    \IF{$S=1$}
        \STATE Cast $A$, add bias when present, and write valid output elements
    \ELSE
        \STATE Publish $A$ to the workspace slot for $(s,\mathcal{B},r)$
        \STATE Atomically increment the output tile's integer completion counter
        \IF{this program is the last to arrive}
            \STATE Combine FP32 workspace partials in split order $0,\ldots,S-1$
            \STATE Cast the result, add bias when present, and write valid output elements
            \STATE Reset the completion counter to zero
        \ENDIF
    \ENDIF
\ENDFOR
\end{algorithmic}
\end{algorithm*}

\subsection{Shared Projection Inputs}
\label{app:projection_reuse}

Query, key, and value projections consume the same attention input, while gate and up projections consume the same feed-forward input. For members sharing input width and tile geometry, with tile-aligned output widths and compatible device and bias configuration, the loader stacks their packed weight rows along the output dimension. A member preceded by $\delta r$ output-tile rows receives a tile-ID offset of $\delta r\,n_{\mathrm{tc}}$, following Eq.~\eqref{eq:app_tile_identifier}. Its row pointers are offset by the preceding tile count in the corresponding precision buffer. Weight values and scales stay with their original tiles.

When the first projection in a group is invoked, the runtime prepares the shared input once and computes the combined output. Subsequent projections read their feature slices from that result. The group therefore shares activation preparation and one mixed-GEMM invocation while preserving each projection's precision choices and output features. With row chunking, all grouped projections follow the same chunk sequence.

The runtime checks input identity and layout before reusing a grouped output and releases that output after every projection has consumed its slice. Separately invoked projections also reuse prepared integer activations through an inference-only, per-stream cache for compatible inputs of at least 32 rows. Both mechanisms reuse input preparation across projections sharing the same activations.

\subsection{Repeated Generation and Runtime Reuse}
\label{app:generation_reuse}

Successive token steps invoke the same projection operations with new inputs, allowing the runtime to reuse model data and execution state. Weight buffers, scales, and row schedules remain resident. Compiled launcher handles are cached for repeated argument specializations, and device- and stream-local workspaces serve successive split reductions. Reusing these launchers and buffers reduces recurring setup across generation steps.

For fixed request shapes, the runtime prepares CUDA Graphs for prefill and one-token decode. It allocates the key/value cache, input-token storage, position state, and generated-token buffer before replay. Prefill initializes the cache and produces the first greedy token. Each decode replay consumes the current token, updates the cache, selects the next token on the GPU, and advances the GPU-resident position. Attention uses the valid cache prefix indicated by the current position.

Prefill capture is used for requests with at most 16,384 prompt activation rows across the batch. Larger prefills execute eagerly into the same preallocated cache, while the repeated decode body remains captured. This reserves graph-owned activation storage for the phases in which replay is used. Across both paths, the runtime continues to consume the same selected mixed-precision weights, while input values, cache contents, and token positions evolve with generation.

\begin{figure*}[t]
    \centering

    \includegraphics[width=0.274\linewidth]{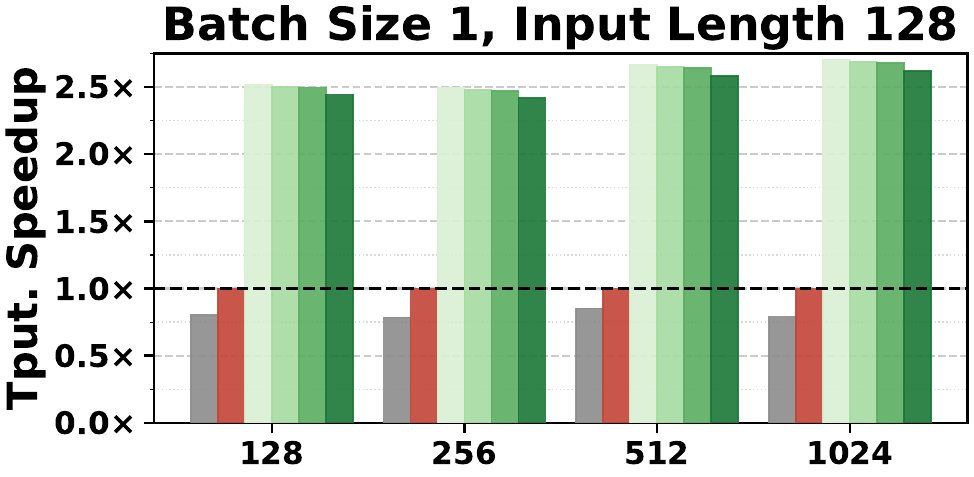}
    \hfill
    \includegraphics[width=0.234\linewidth]{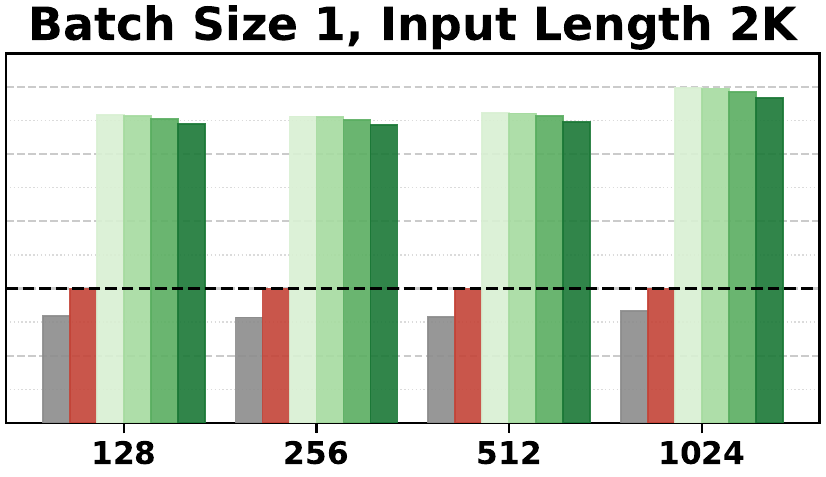}
    \hfill
    \includegraphics[width=0.234\linewidth]{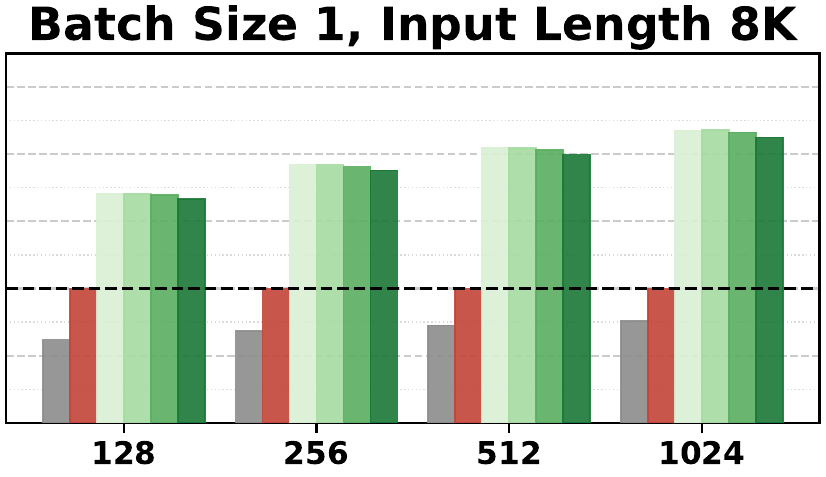}
    \hfill
    \includegraphics[width=0.234\linewidth]{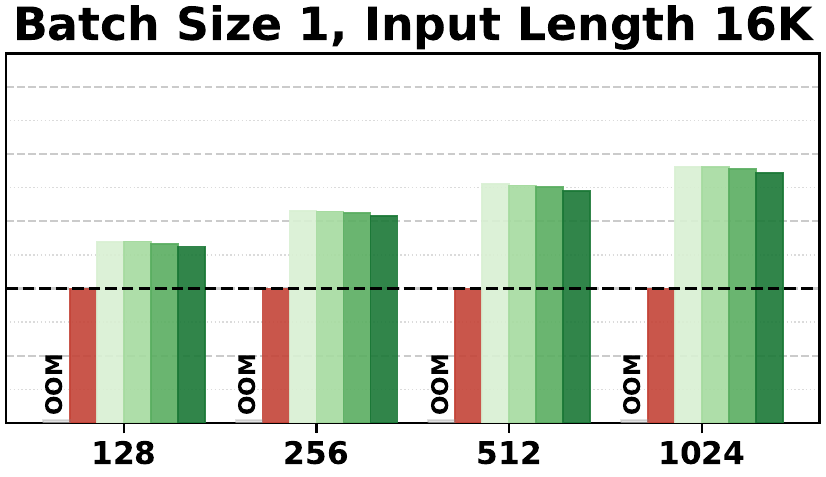}

    \includegraphics[width=0.274\linewidth]{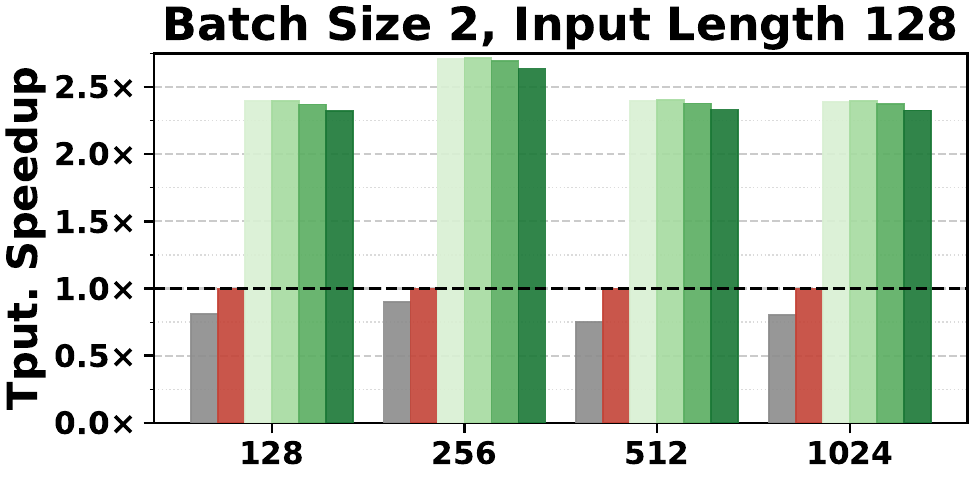}
    \hfill
    \includegraphics[width=0.234\linewidth]{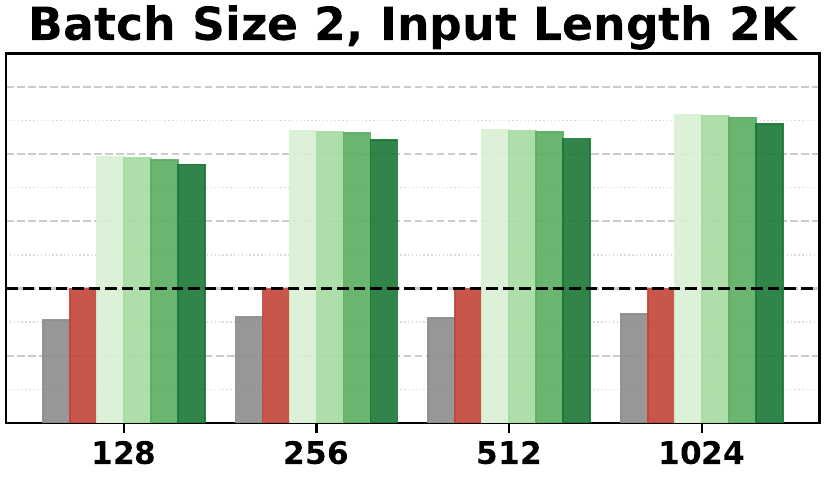}
    \hfill
    \includegraphics[width=0.234\linewidth]{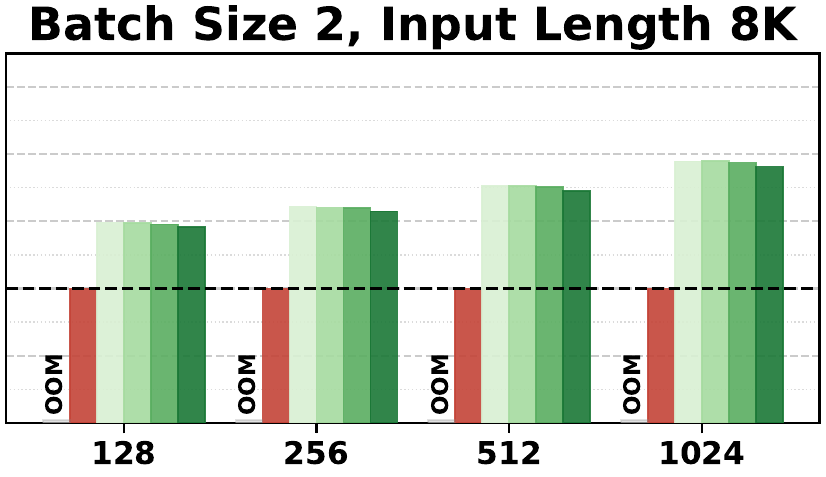}
    \hfill
    \includegraphics[width=0.234\linewidth]{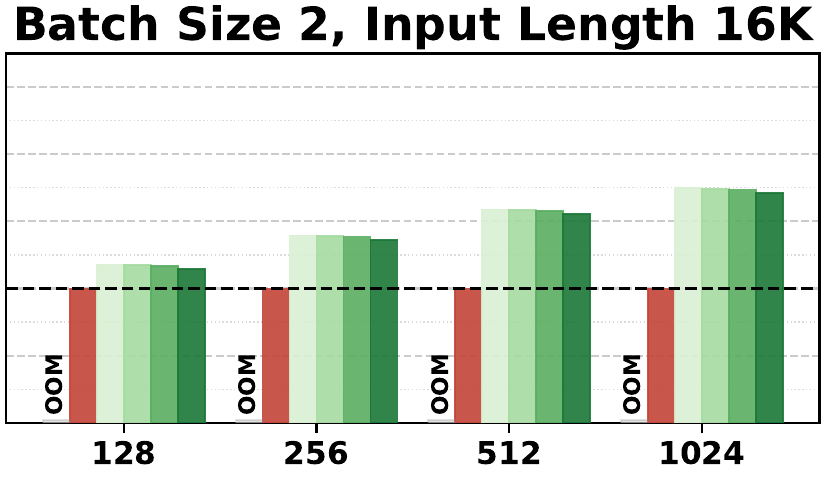}

    \includegraphics[width=0.274\linewidth]{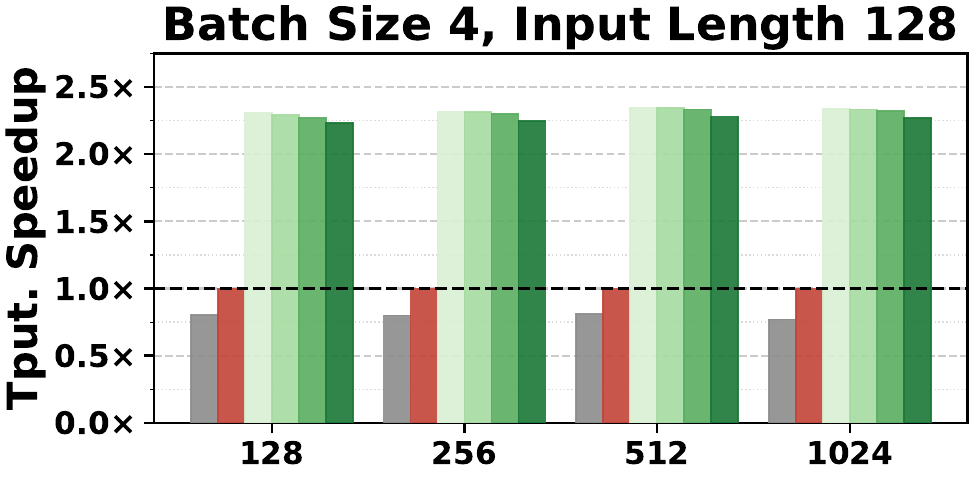}
    \hfill
    \includegraphics[width=0.234\linewidth]{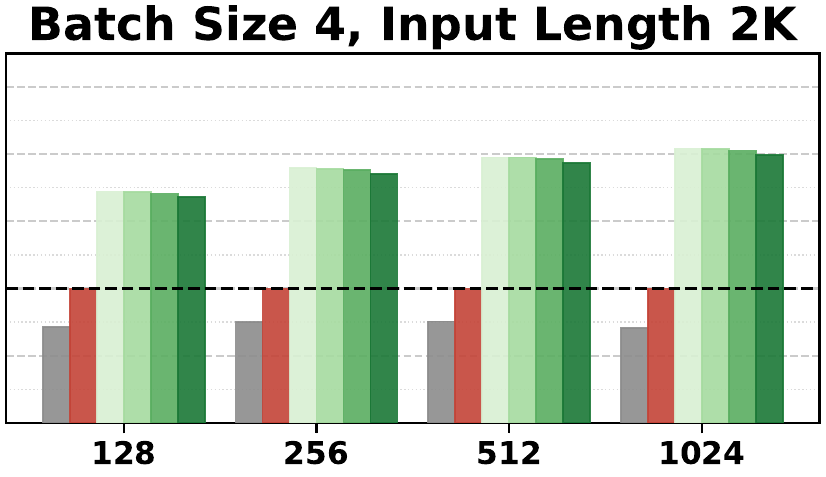}
    \hfill
    \includegraphics[width=0.234\linewidth]{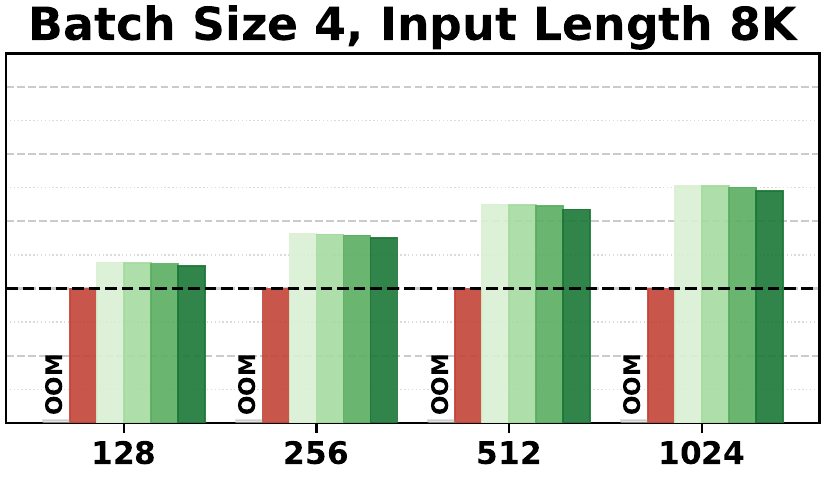}
    \hfill
    \includegraphics[width=0.234\linewidth]{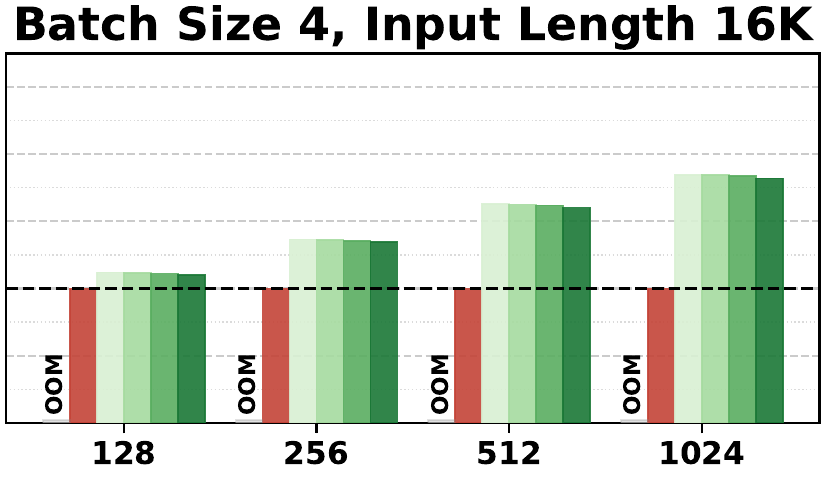}

    \includegraphics[width=0.274\linewidth]{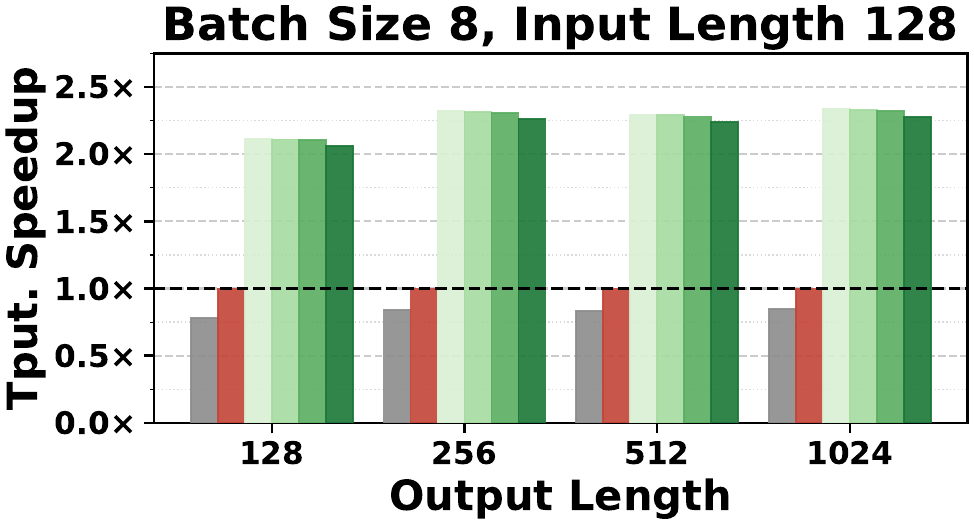}
    \hfill
    \includegraphics[width=0.234\linewidth]{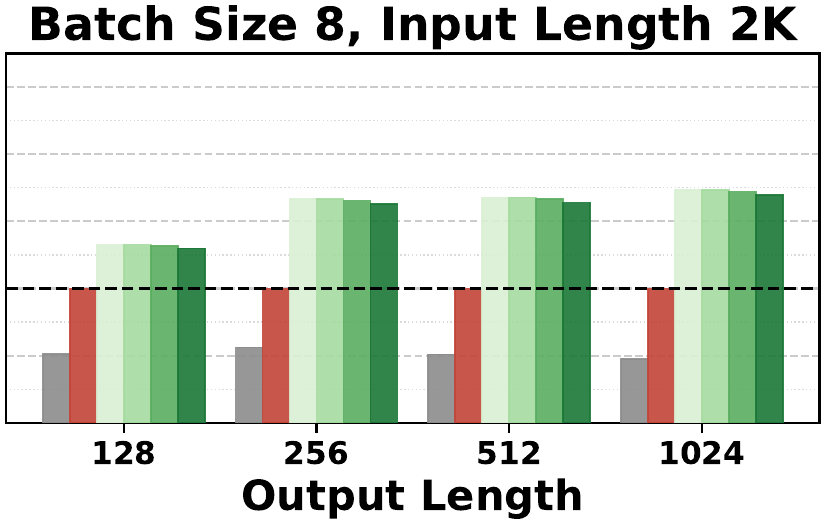}
    \hfill
    \includegraphics[width=0.234\linewidth]{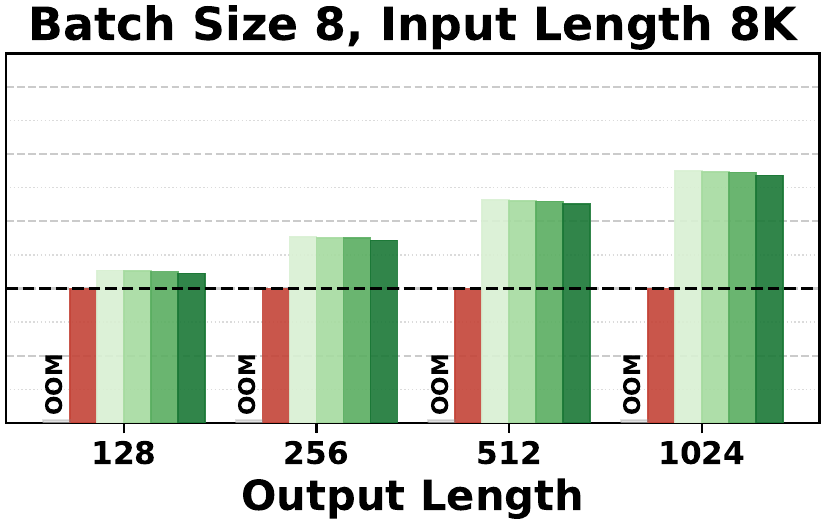}
    \hfill
    \includegraphics[width=0.234\linewidth]{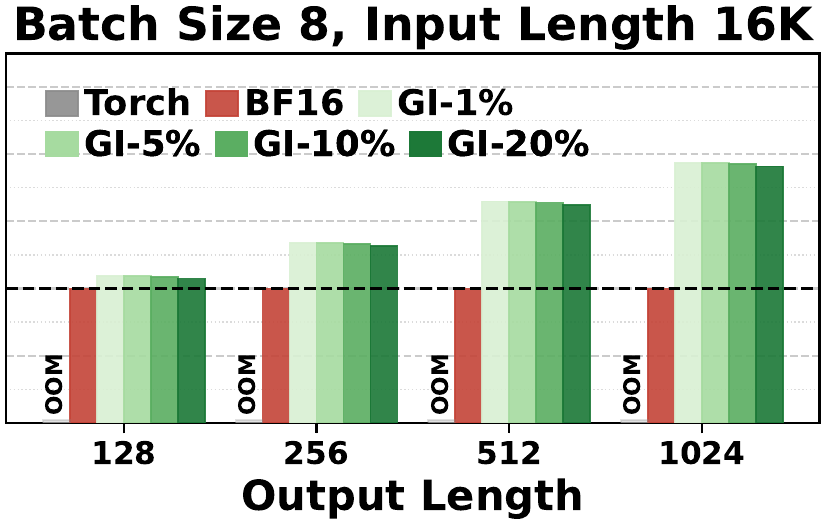}

    \caption{
    Generation-throughput speedup relative to BF16 on Llama-3.2-3B for \textbf{gradient-informed (GI)} \modelname variants.
    Rows vary batch size over 1, 2, 4, and 8; columns vary prompt length over 128, 2K, 8K, and 16K tokens.
    Each panel varies output length and compares GI retention ratios; OOM denotes an out-of-memory run.
    }
    \label{fig:generation_throughput_gi}
    
\end{figure*}

\begin{figure*}[t]
    \centering

    \includegraphics[width=0.274\linewidth]{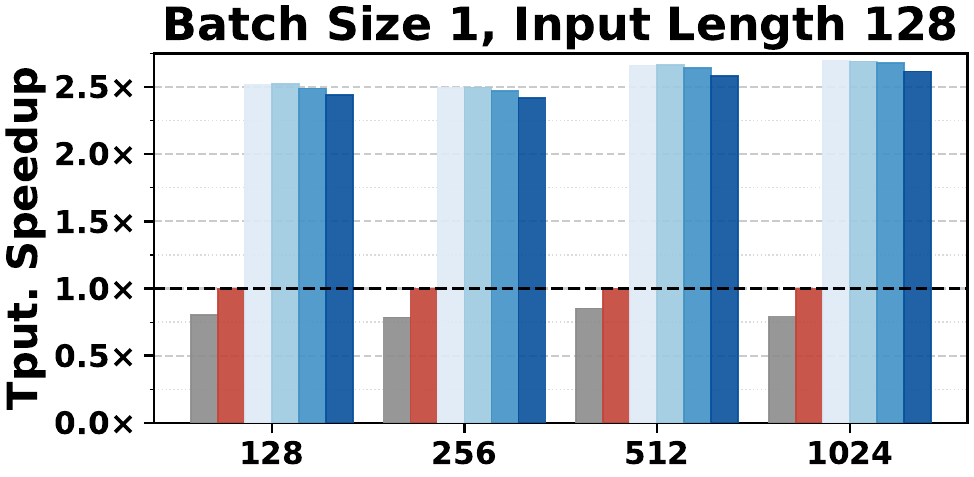}
    \hfill
    \includegraphics[width=0.234\linewidth]{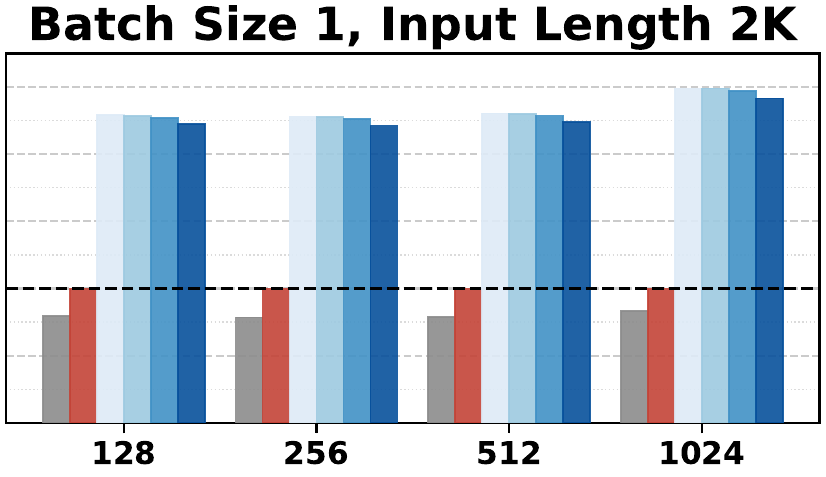}
    \hfill
    \includegraphics[width=0.234\linewidth]{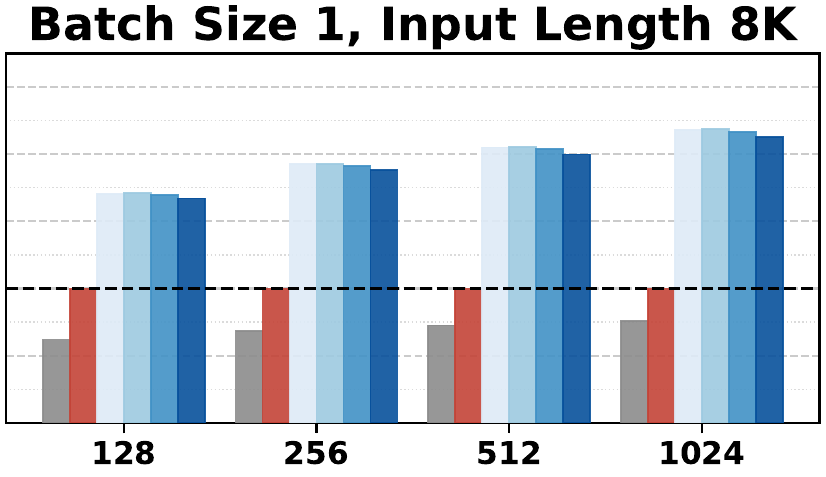}
    \hfill
    \includegraphics[width=0.234\linewidth]{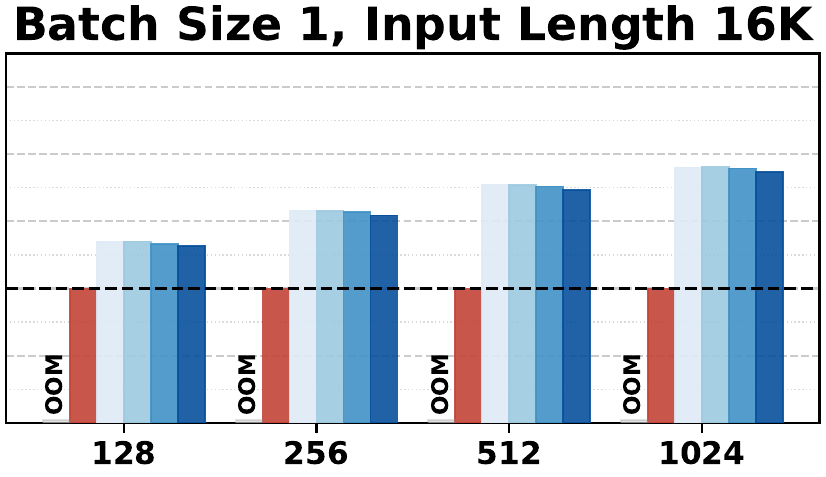}

    \includegraphics[width=0.274\linewidth]{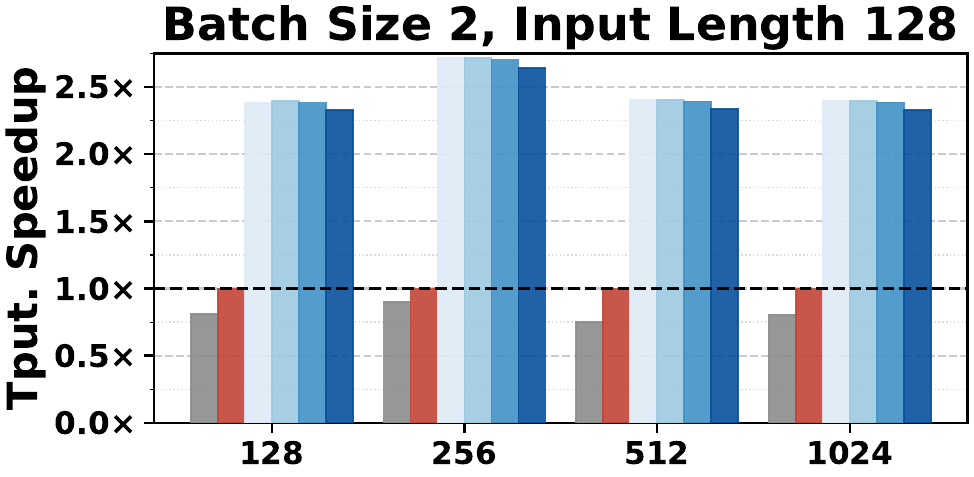}
    \hfill
    \includegraphics[width=0.234\linewidth]{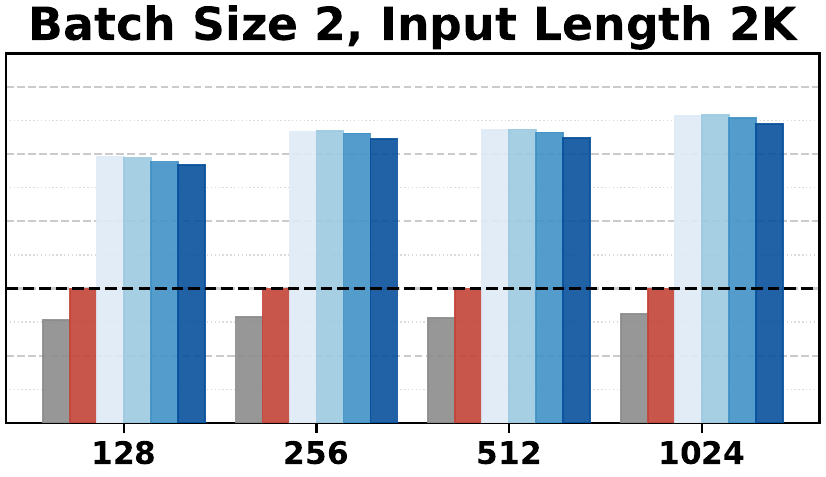}
    \hfill
    \includegraphics[width=0.234\linewidth]{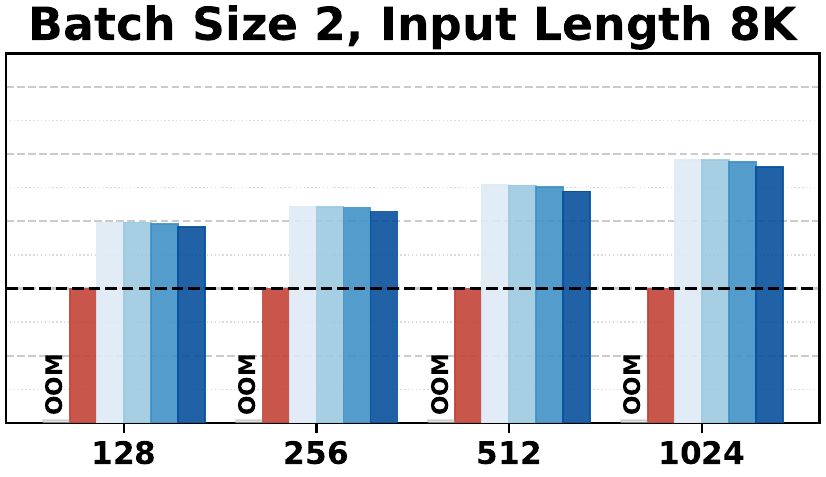}
    \hfill
    \includegraphics[width=0.234\linewidth]{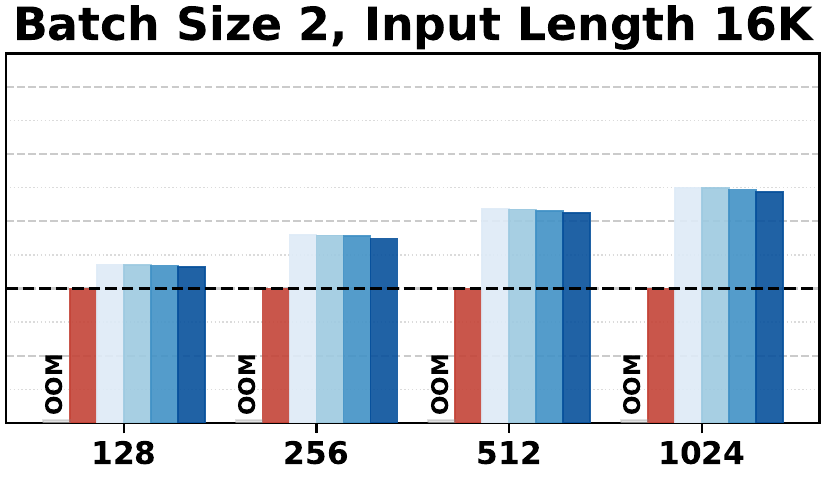}

    \includegraphics[width=0.274\linewidth]{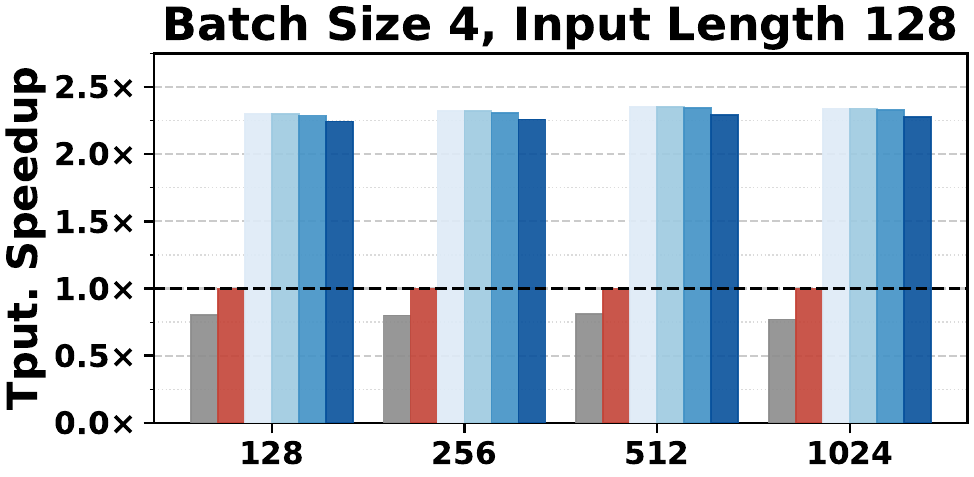}
    \hfill
    \includegraphics[width=0.234\linewidth]{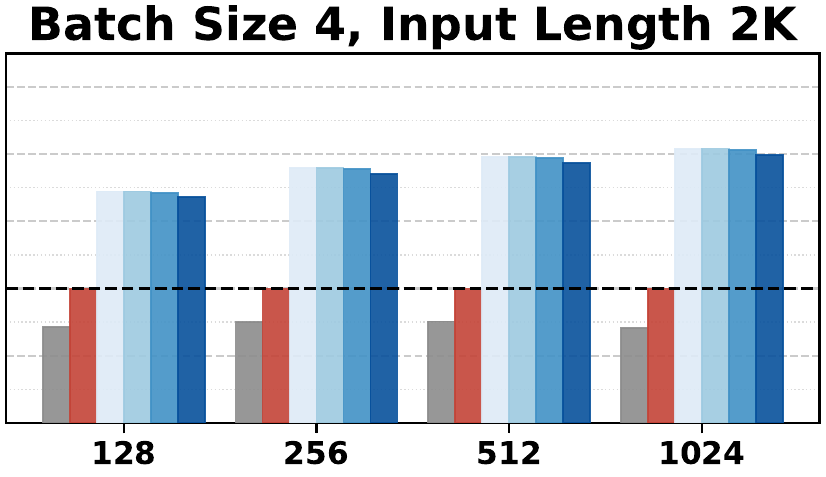}
    \hfill
    \includegraphics[width=0.234\linewidth]{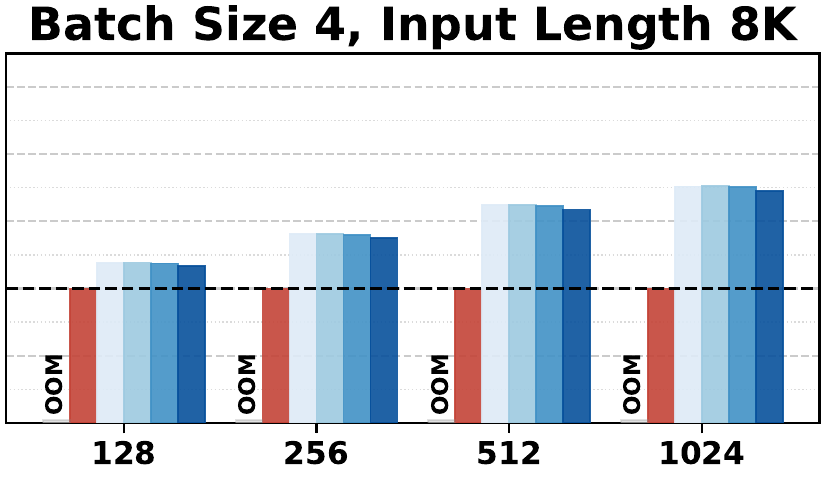}
    \hfill
    \includegraphics[width=0.234\linewidth]{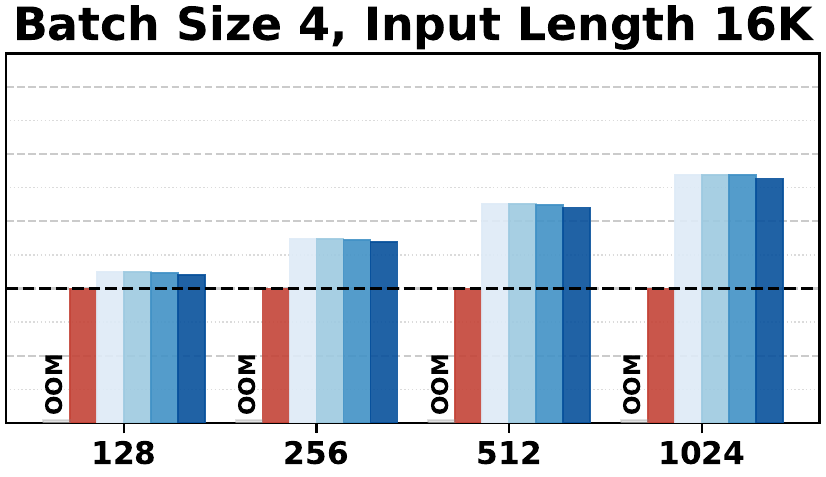}

    \includegraphics[width=0.274\linewidth]{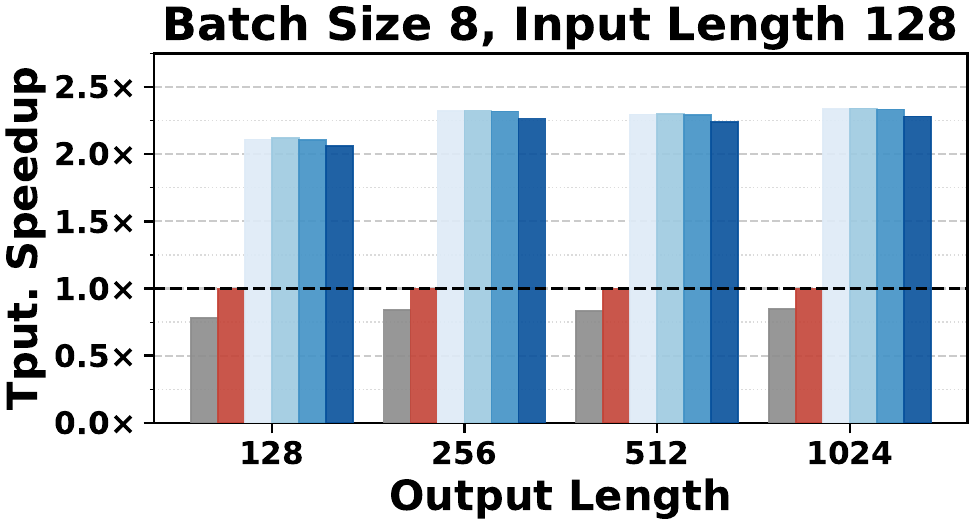}
    \hfill
    \includegraphics[width=0.234\linewidth]{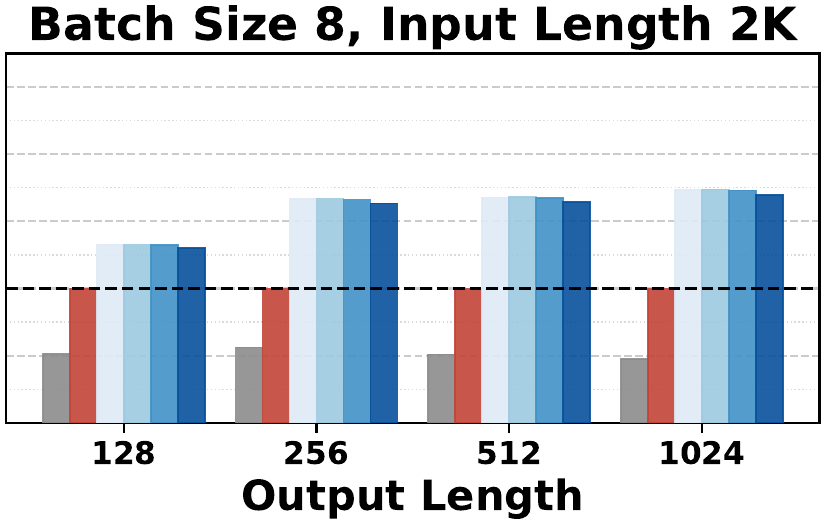}
    \hfill
    \includegraphics[width=0.234\linewidth]{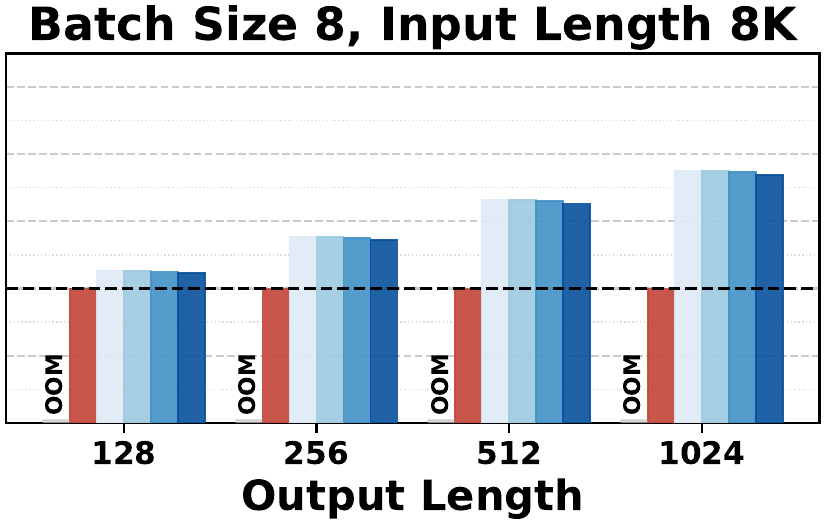}
    \hfill
    \includegraphics[width=0.234\linewidth]{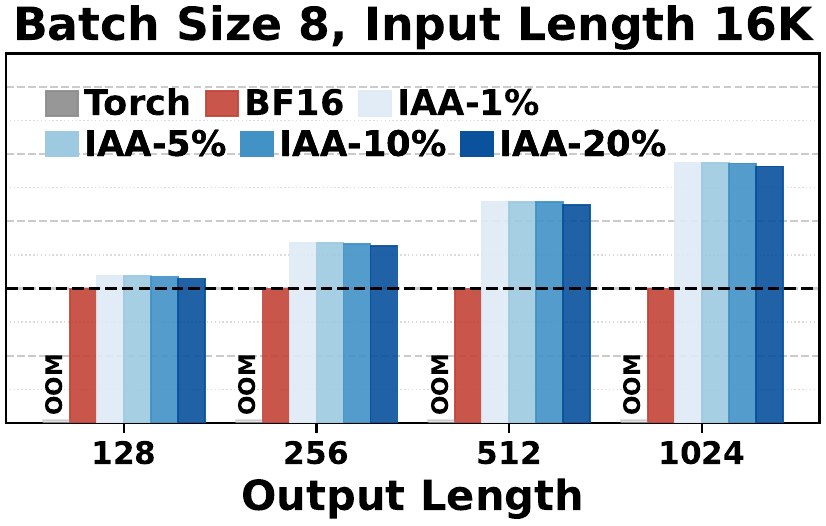}

    \caption{
    Generation-throughput speedup relative to BF16 on Llama-3.2-3B for \textbf{input-activation-aware (IAA)} \modelname variants.
    Rows vary batch size over 1, 2, 4, and 8; columns vary prompt length over 128, 2K, 8K, and 16K tokens.
    Each panel varies output length and compares IAA retention ratios; OOM denotes an out-of-memory run.
    }
    \label{fig:generation_throughput_iaa}
    
\end{figure*}

\begin{figure*}[t]
    \centering

    \includegraphics[width=0.274\linewidth]{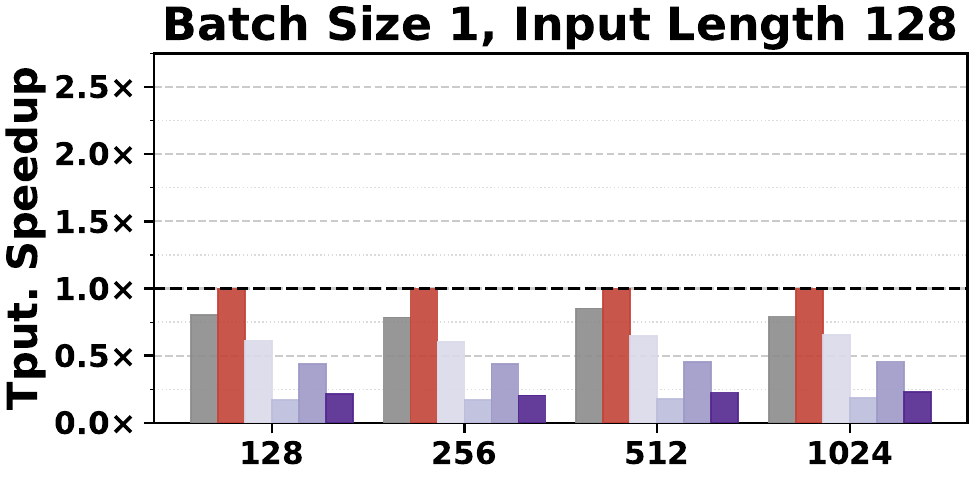}
    \hfill
    \includegraphics[width=0.234\linewidth]{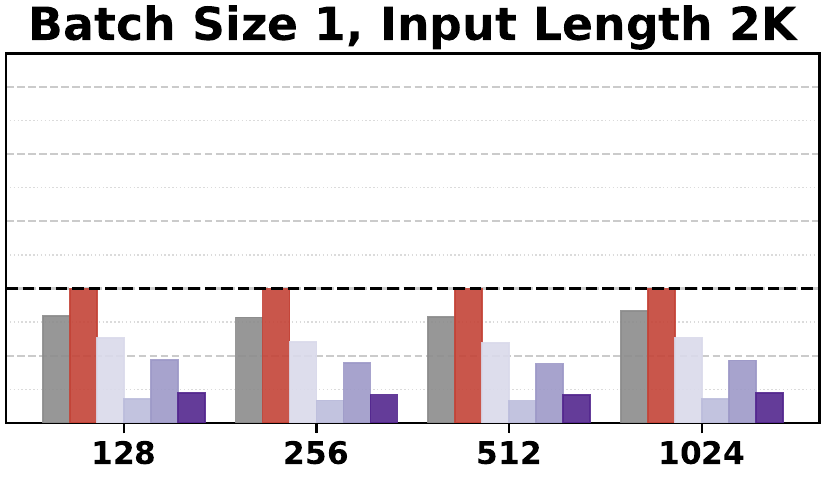}
    \hfill
    \includegraphics[width=0.234\linewidth]{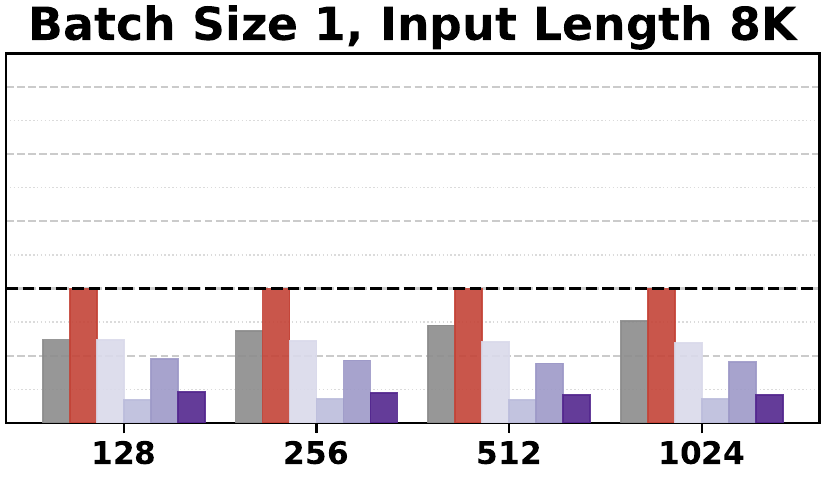}
    \hfill
    \includegraphics[width=0.234\linewidth]{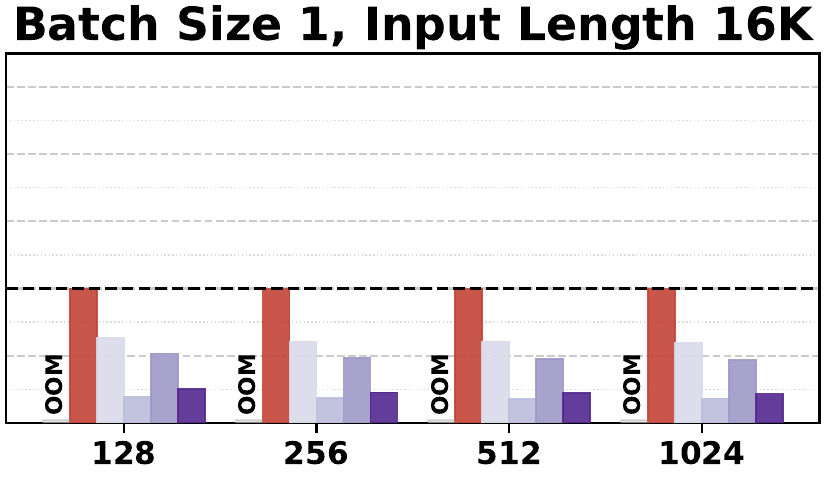}

    \includegraphics[width=0.274\linewidth]{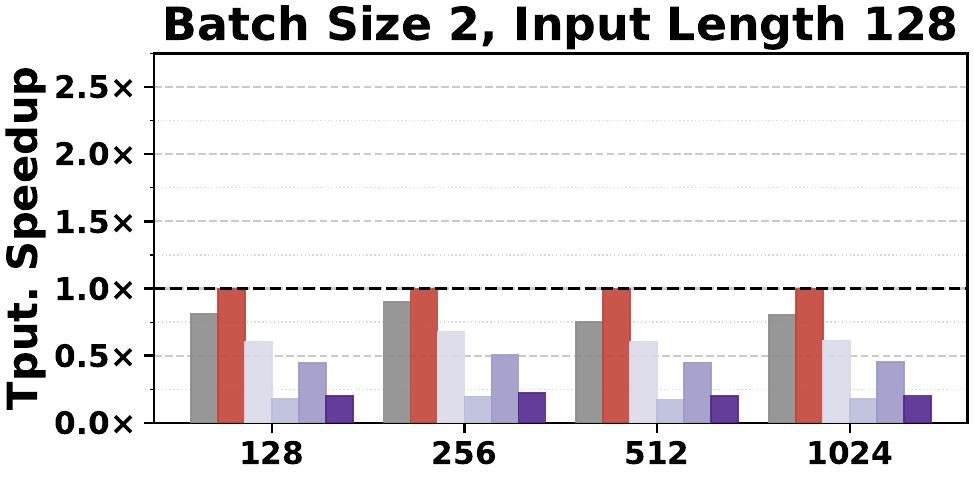}
    \hfill
    \includegraphics[width=0.234\linewidth]{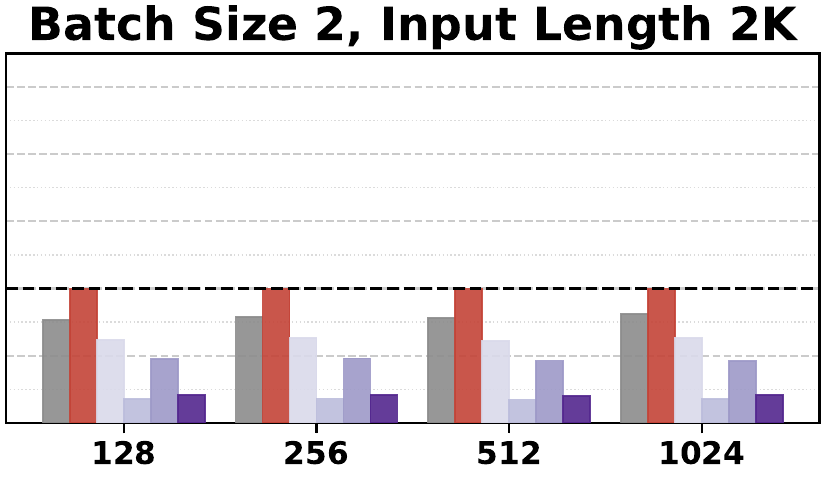}
    \hfill
    \includegraphics[width=0.234\linewidth]{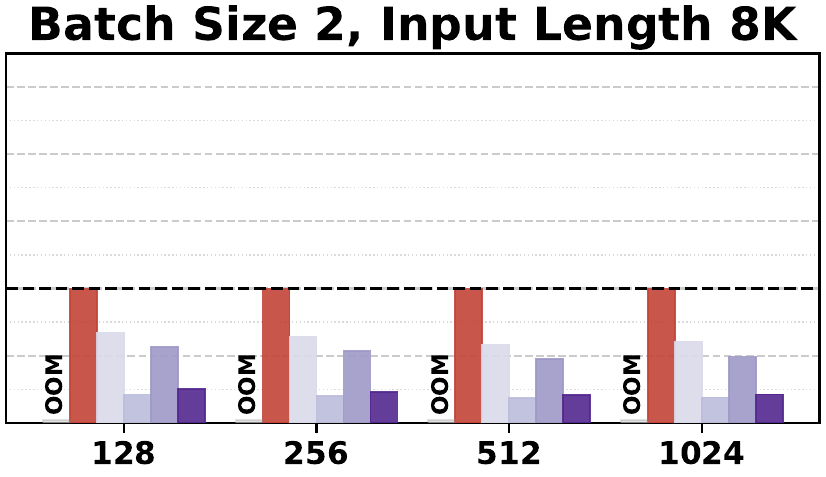}
    \hfill
    \includegraphics[width=0.234\linewidth]{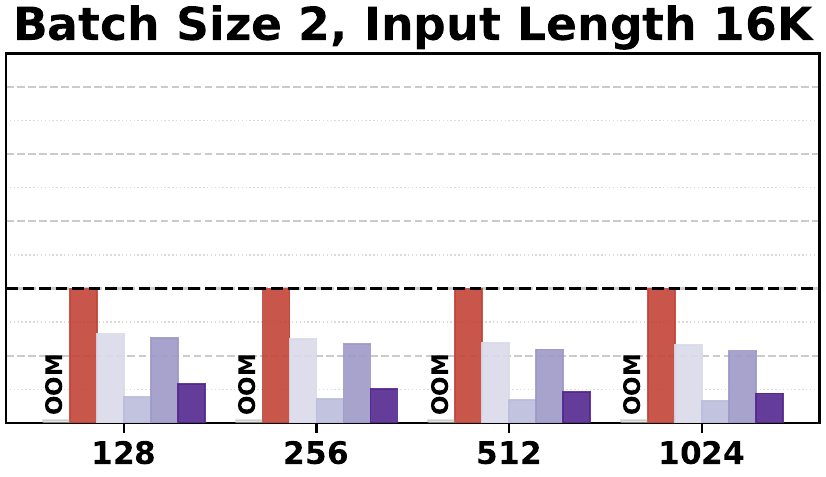}

    \includegraphics[width=0.274\linewidth]{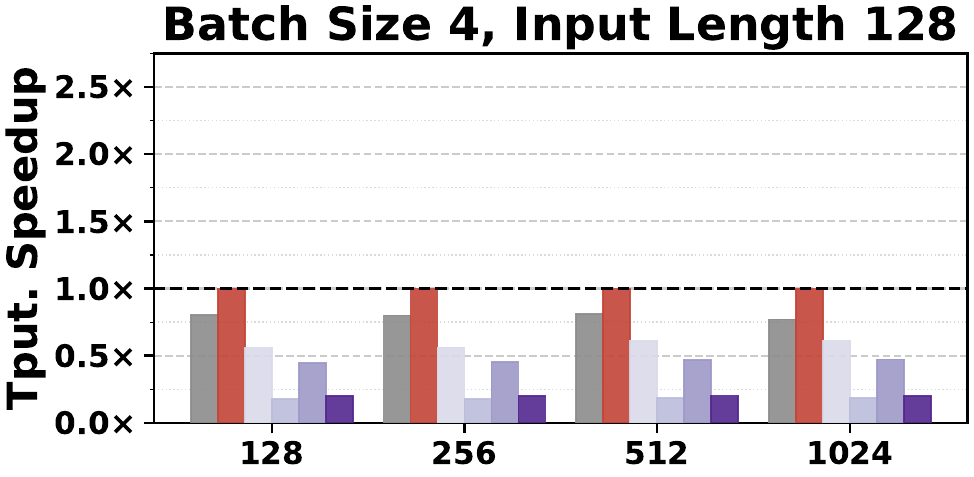}
    \hfill
    \includegraphics[width=0.234\linewidth]{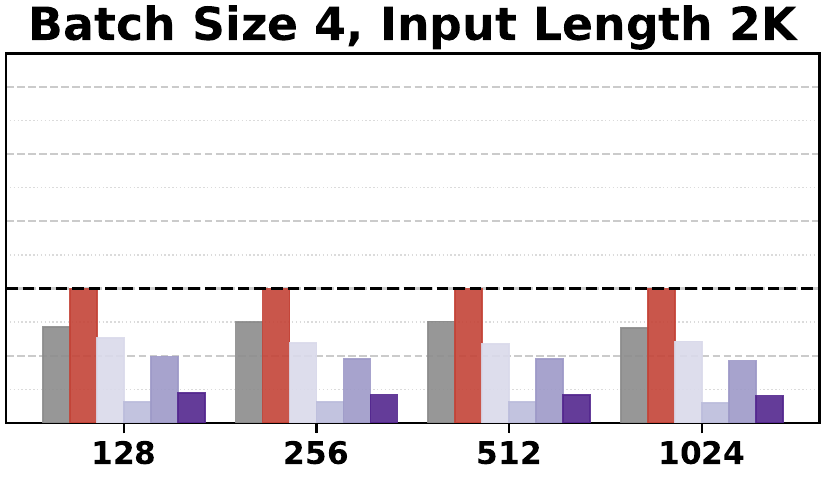}
    \hfill
    \includegraphics[width=0.234\linewidth]{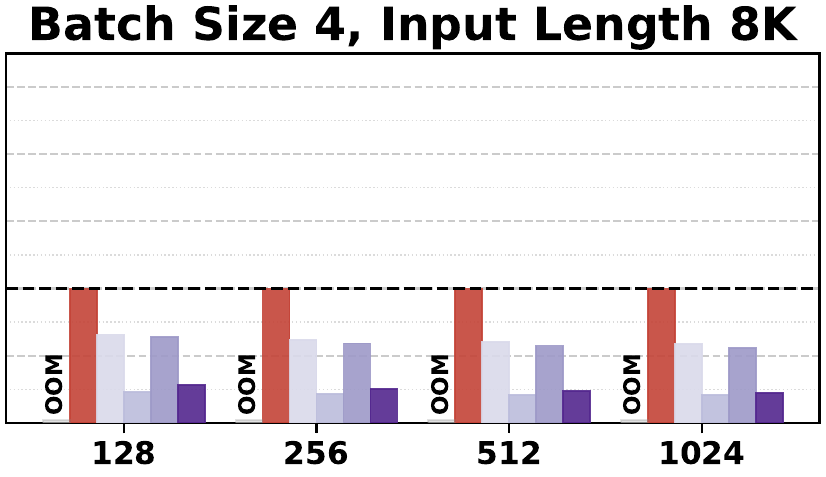}
    \hfill
    \includegraphics[width=0.234\linewidth]{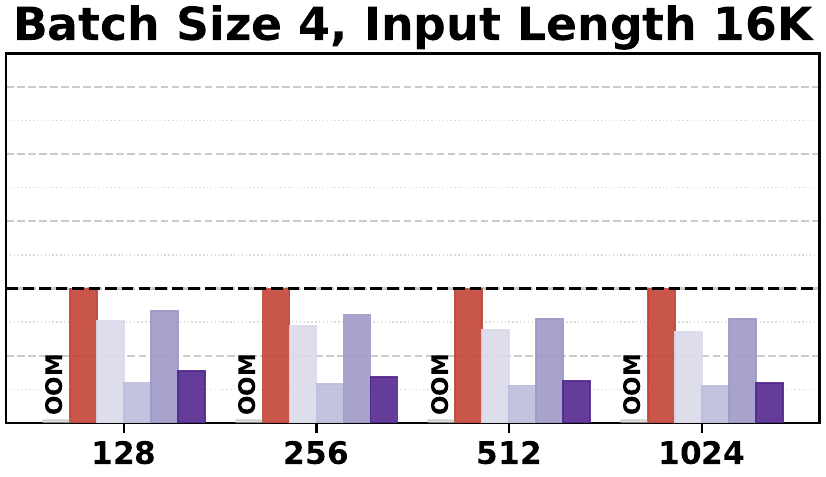}

    \includegraphics[width=0.274\linewidth]{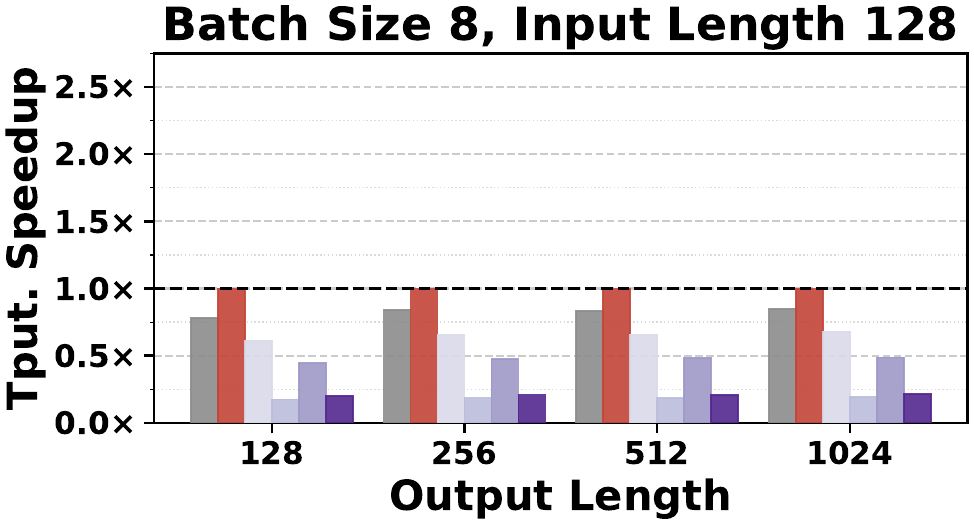}
    \hfill
    \includegraphics[width=0.234\linewidth]{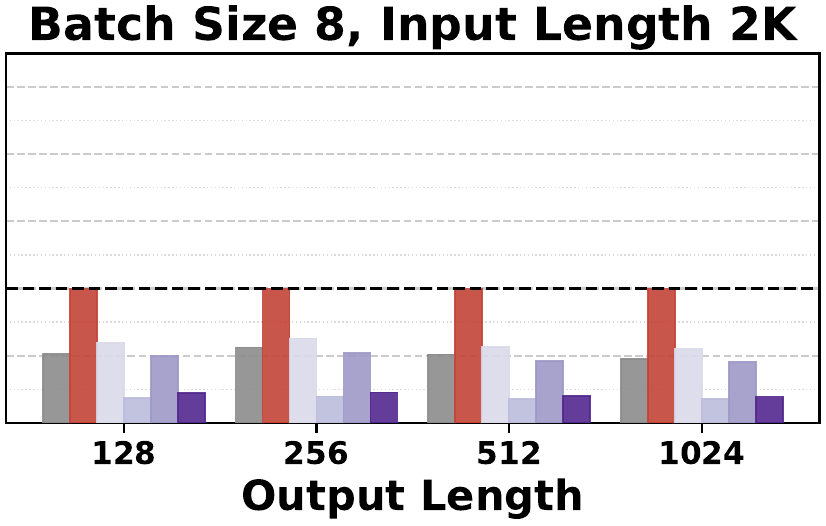}
    \hfill
    \includegraphics[width=0.234\linewidth]{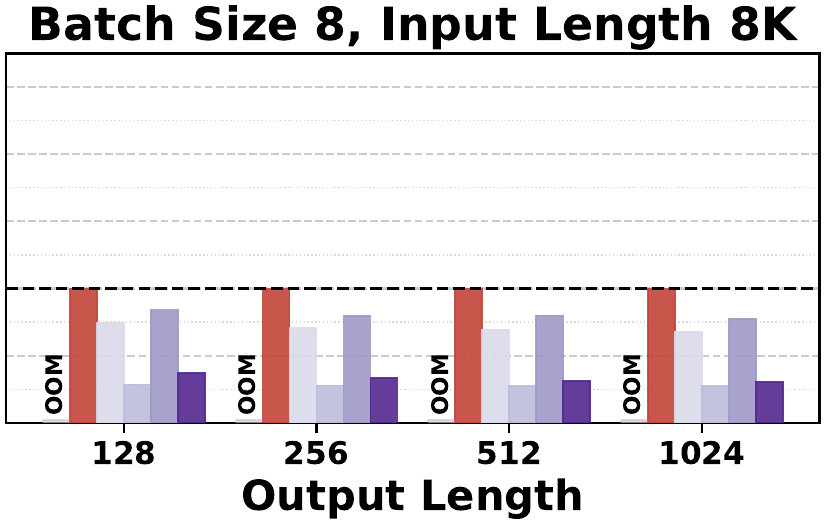}
    \hfill
    \includegraphics[width=0.234\linewidth]{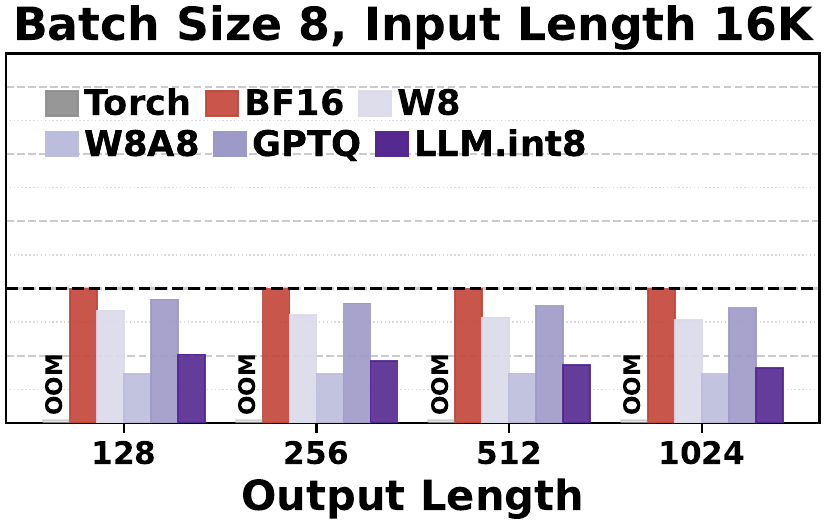}

    \caption{
    Generation-throughput speedup relative to BF16 on Llama-3.2-3B for \textbf{baseline inference methods}.
    Rows vary batch size over 1, 2, 4, and 8; columns vary prompt length over 128, 2K, 8K, and 16K tokens.
    Each panel varies output length and compares the baseline methods; OOM denotes an out-of-memory run.
    }
    \label{fig:generation_throughput_baselines}
    
\end{figure*}

\begin{figure*}[t]
    \centering

    \includegraphics[width=0.274\linewidth]{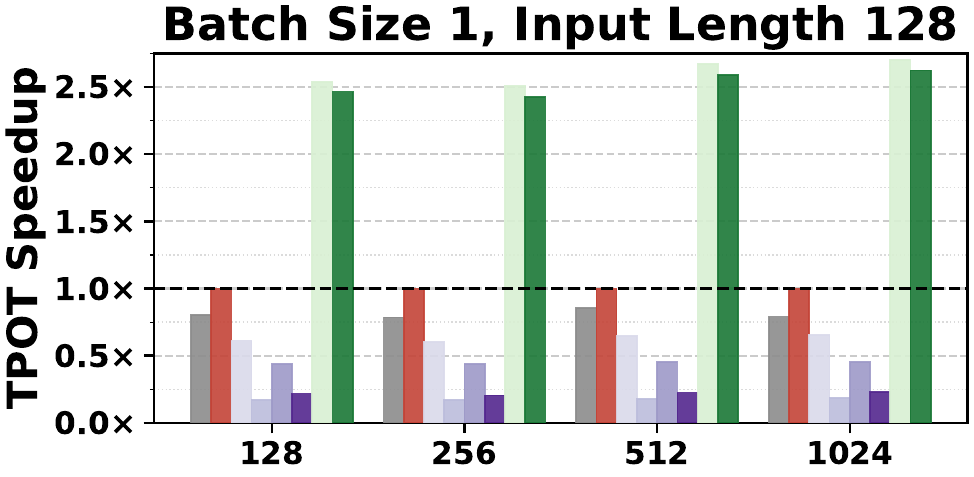}
    \hfill
    \includegraphics[width=0.234\linewidth]{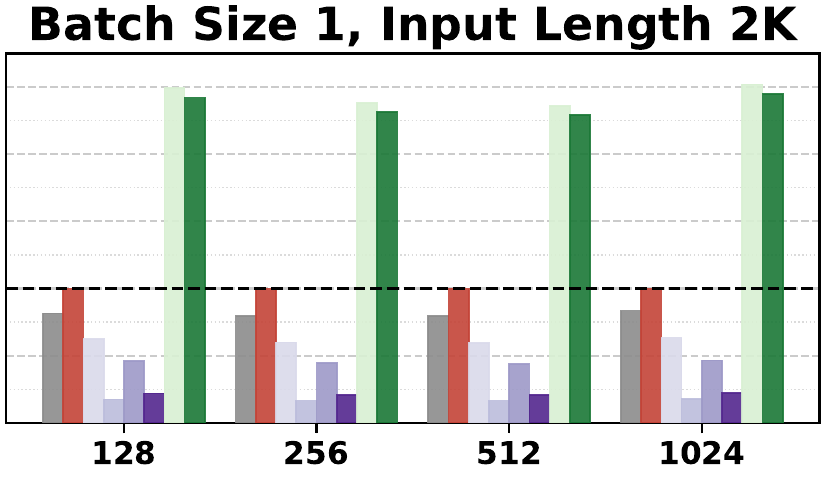}
    \hfill
    \includegraphics[width=0.234\linewidth]{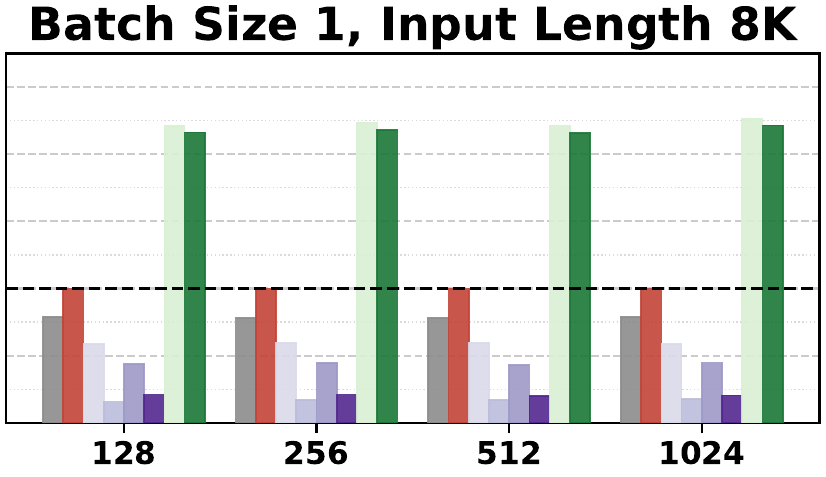}
    \hfill
    \includegraphics[width=0.234\linewidth]{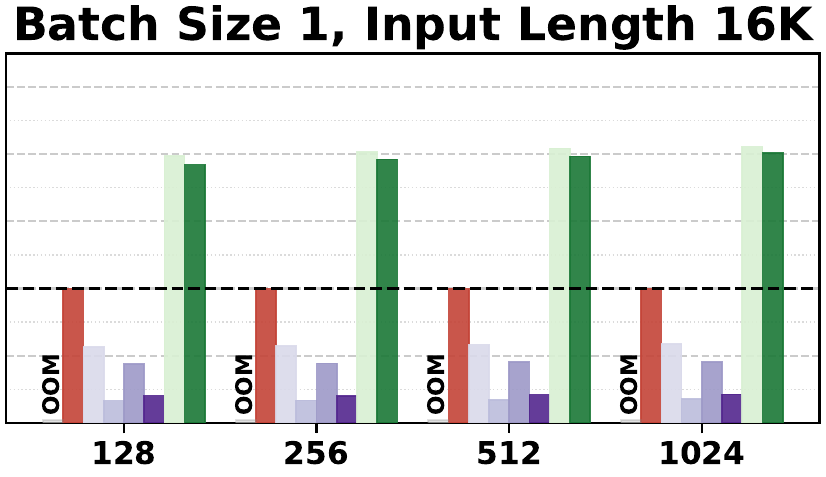}

    \includegraphics[width=0.274\linewidth]{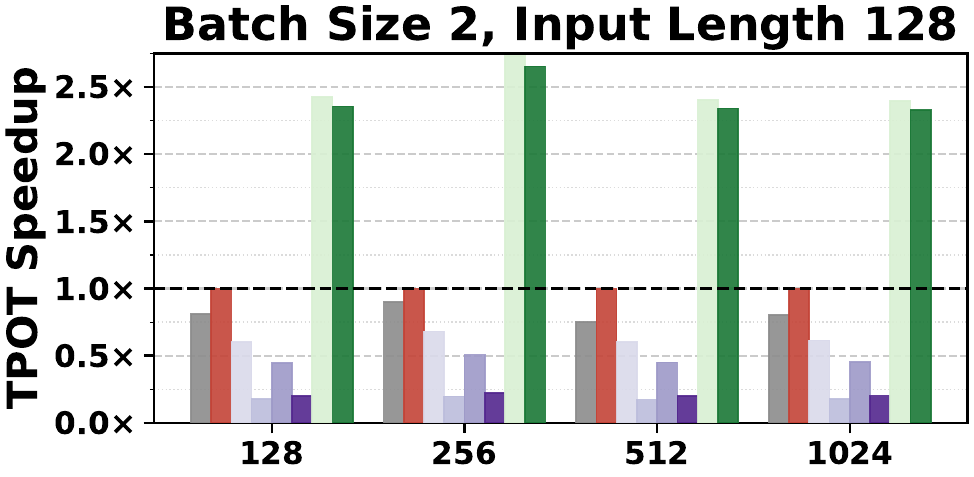}
    \hfill
    \includegraphics[width=0.234\linewidth]{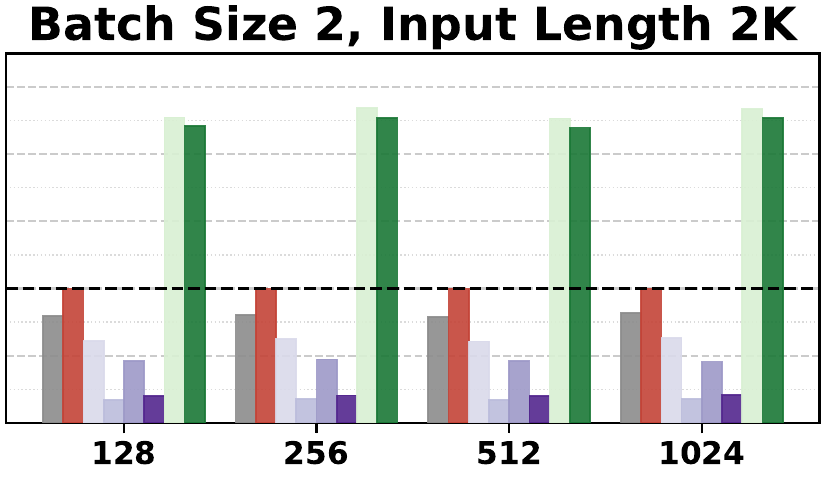}
    \hfill
    \includegraphics[width=0.234\linewidth]{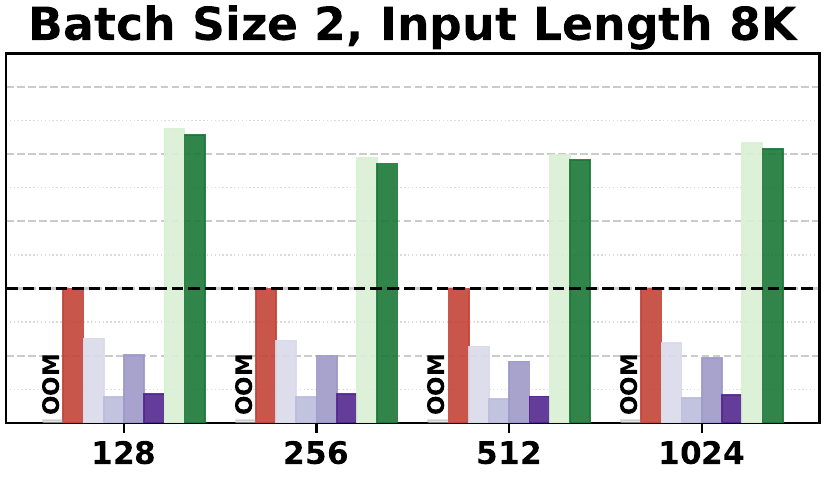}
    \hfill
    \includegraphics[width=0.234\linewidth]{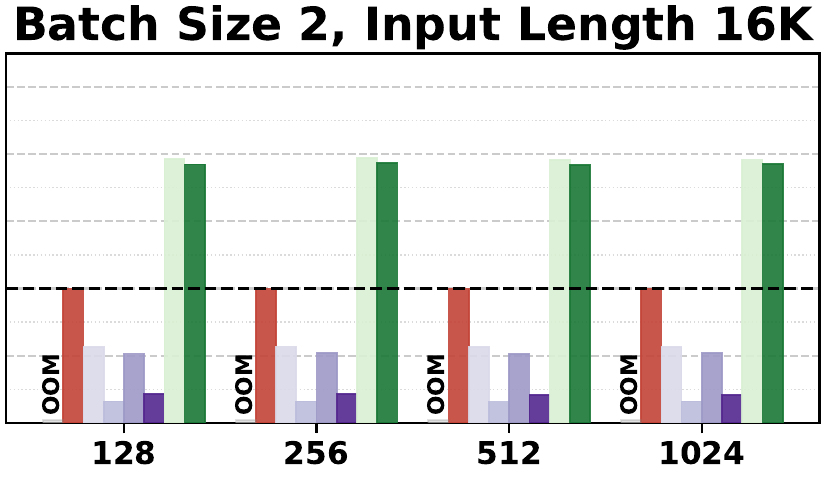}

    \includegraphics[width=0.274\linewidth]{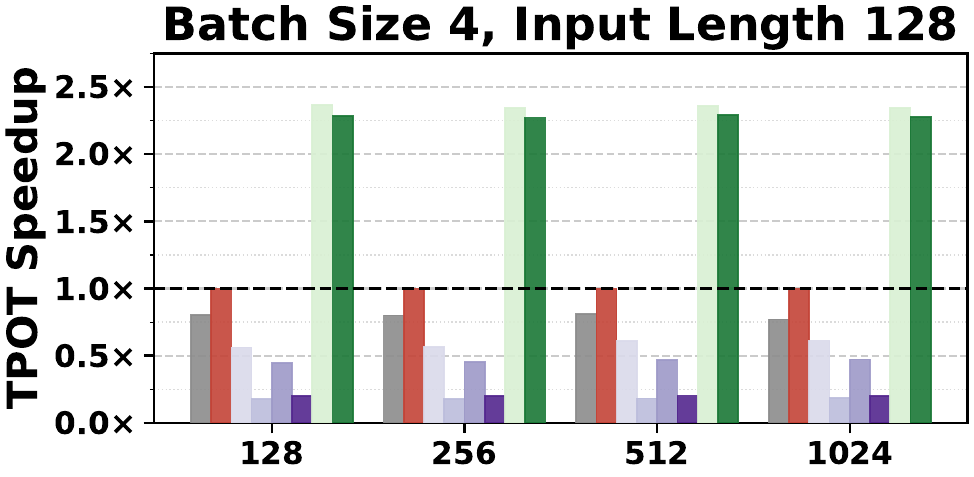}
    \hfill
    \includegraphics[width=0.234\linewidth]{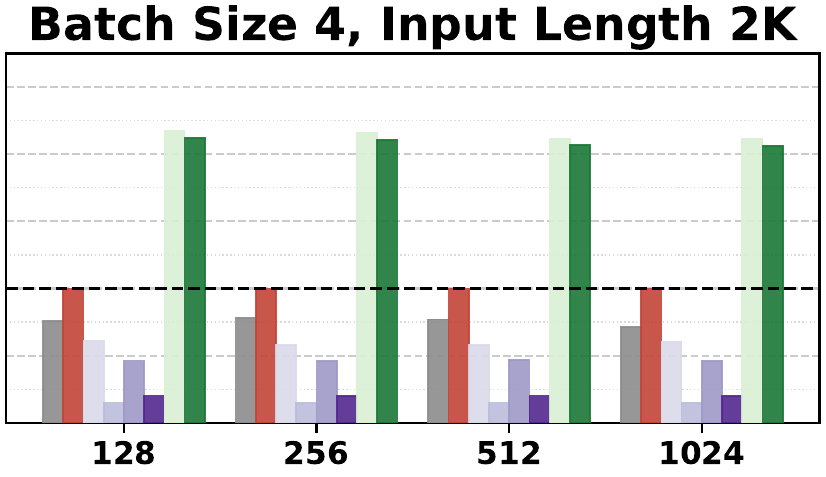}
    \hfill
    \includegraphics[width=0.234\linewidth]{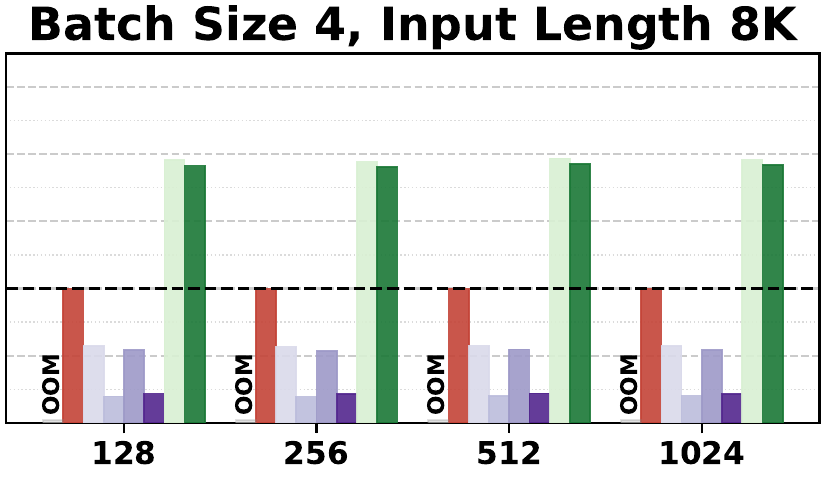}
    \hfill
    \includegraphics[width=0.234\linewidth]{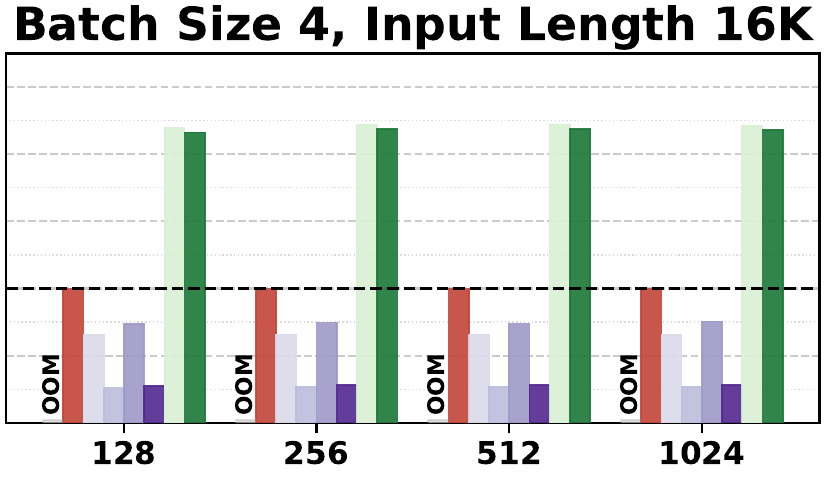}

    \includegraphics[width=0.274\linewidth]{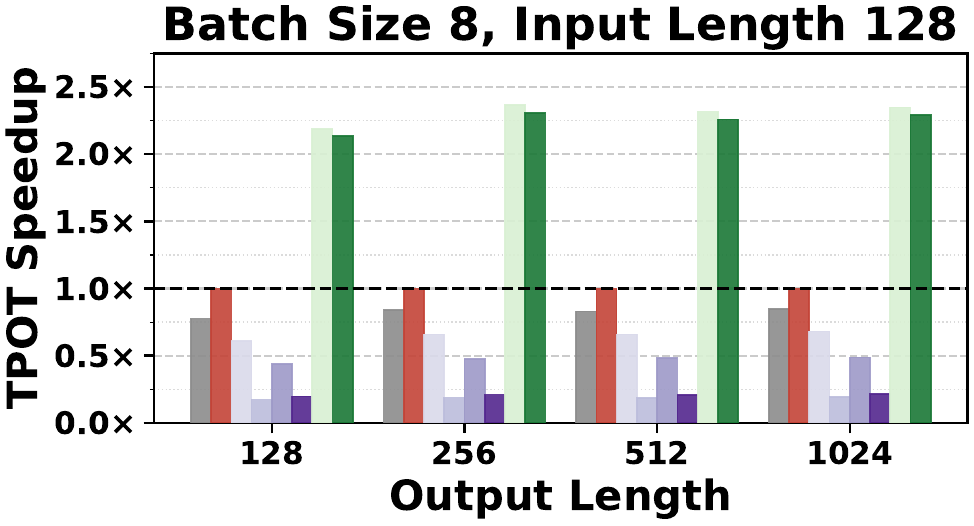}
    \hfill
    \includegraphics[width=0.234\linewidth]{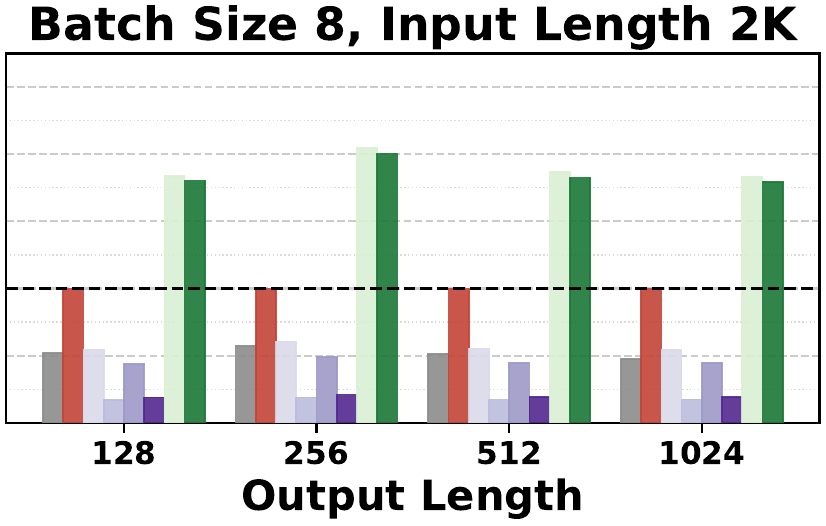}
    \hfill
    \includegraphics[width=0.234\linewidth]{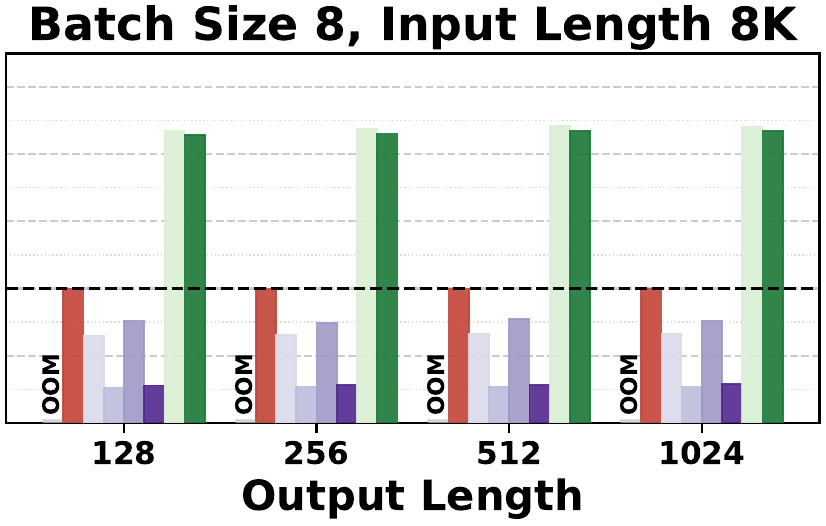}
    \hfill
    \includegraphics[width=0.234\linewidth]{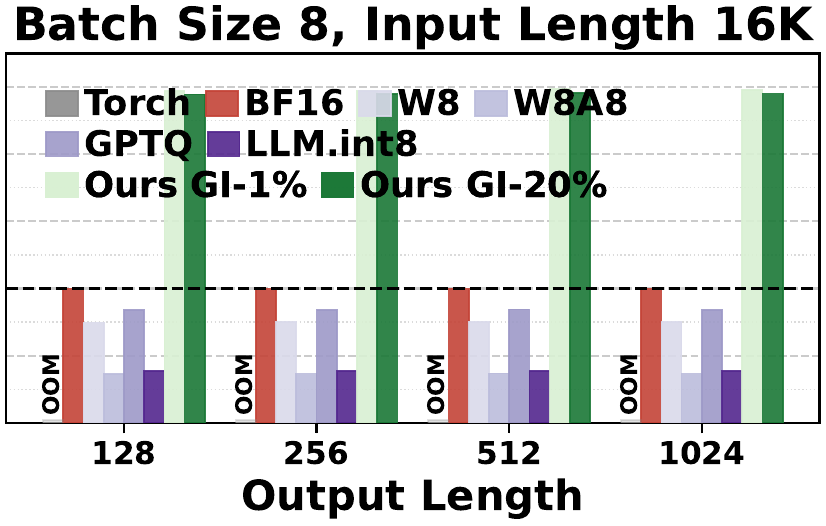}

    \caption{
    Time-per-output-token (TPOT) comparison on Llama-3.2-3B.
    Rows vary batch size over 1, 2, 4, and 8; columns vary prompt length over 128, 2K, 8K, and 16K tokens.
    Each panel compares inference methods across output lengths of 128, 256, 512, and 1024 tokens; OOM denotes an out-of-memory run.
    }
    \label{fig:tpot_main}
    
\end{figure*}

\begin{figure*}[t]
    \centering

    \includegraphics[width=0.274\linewidth]{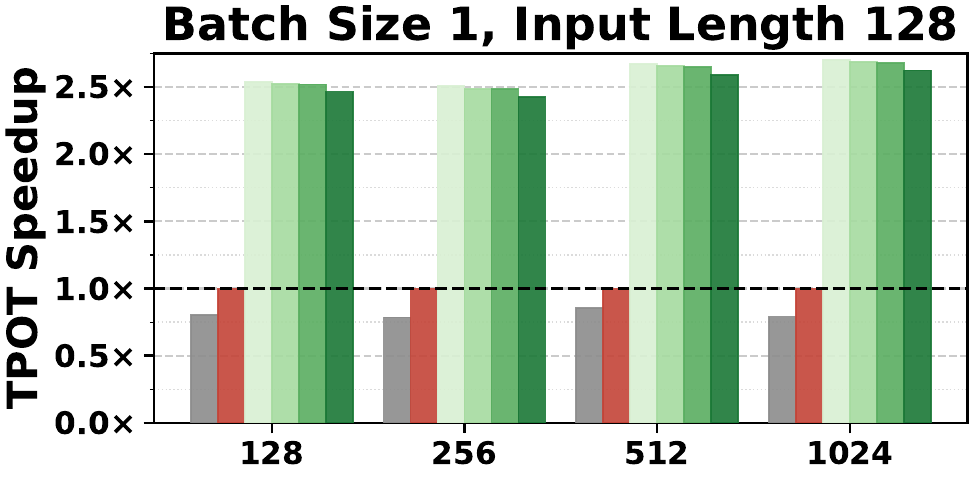}
    \hfill
    \includegraphics[width=0.234\linewidth]{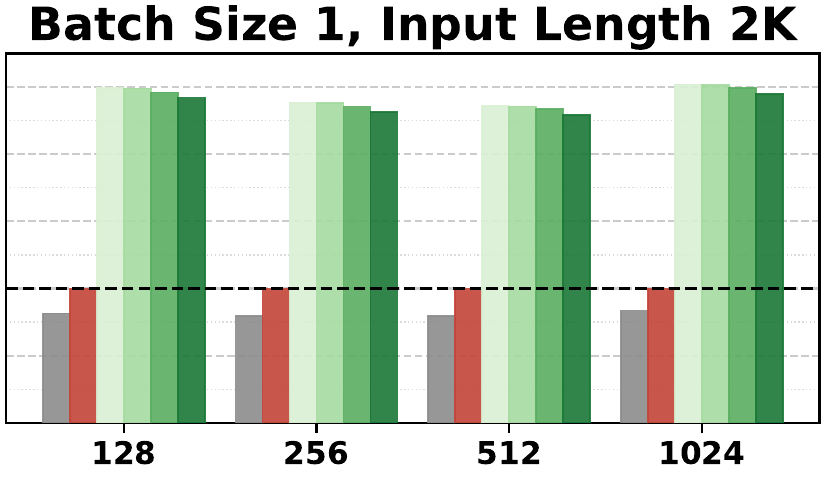}
    \hfill
    \includegraphics[width=0.234\linewidth]{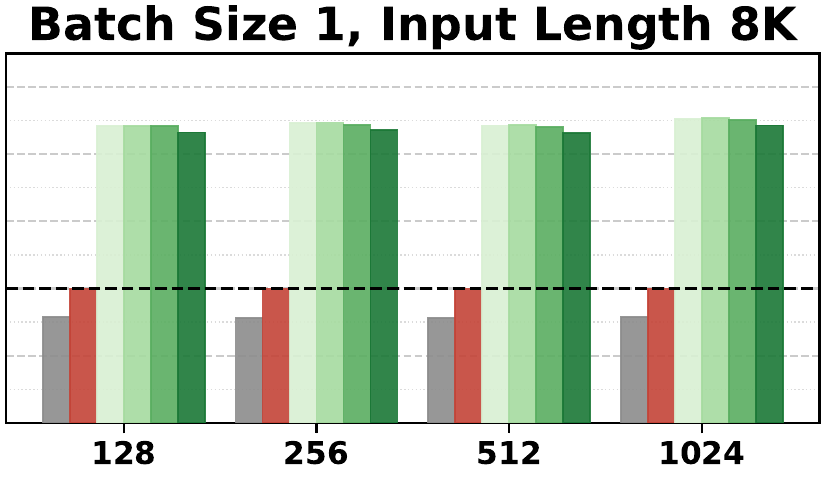}
    \hfill
    \includegraphics[width=0.234\linewidth]{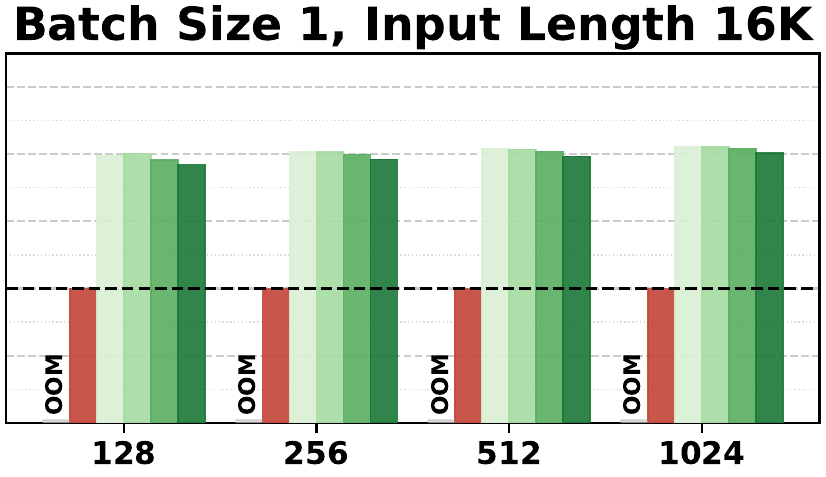}

    \includegraphics[width=0.274\linewidth]{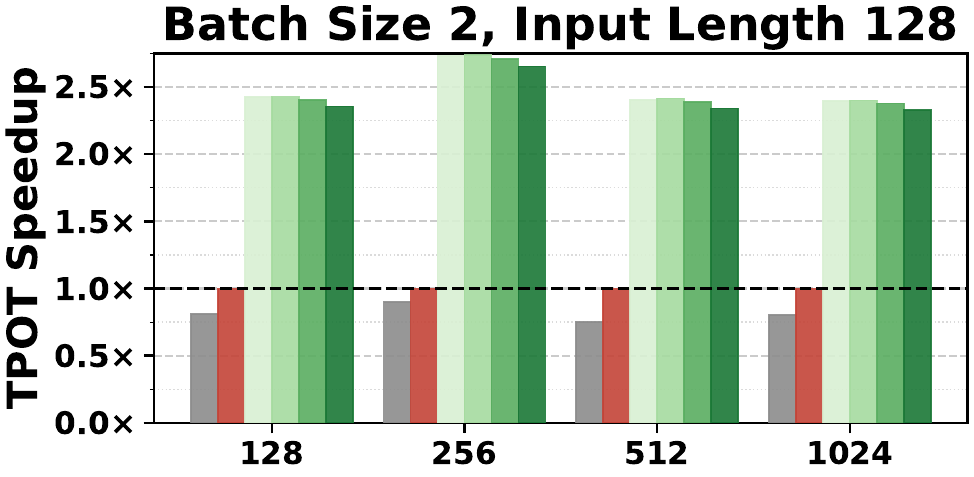}
    \hfill
    \includegraphics[width=0.234\linewidth]{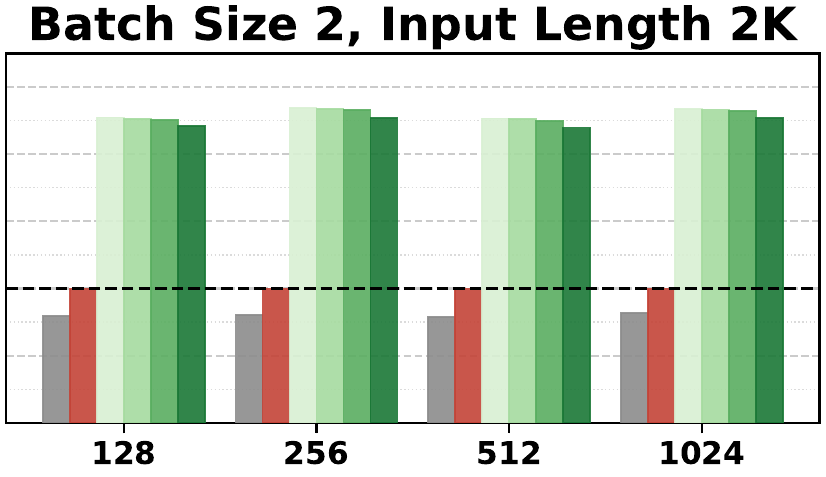}
    \hfill
    \includegraphics[width=0.234\linewidth]{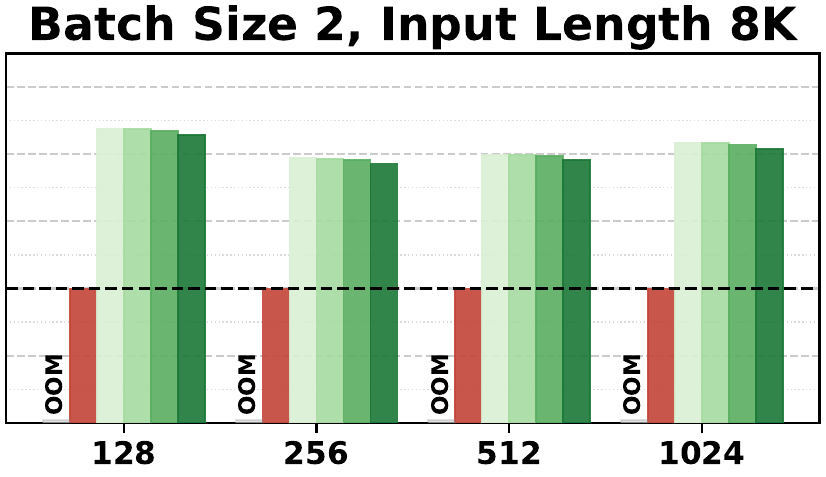}
    \hfill
    \includegraphics[width=0.234\linewidth]{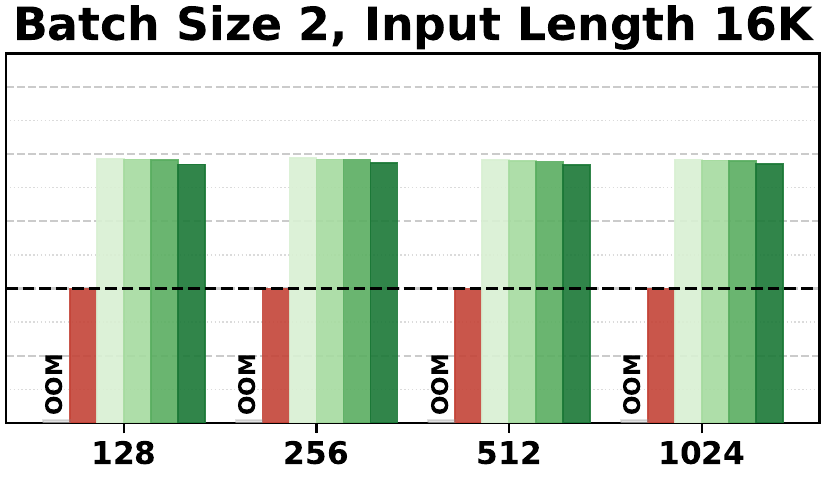}

    \includegraphics[width=0.274\linewidth]{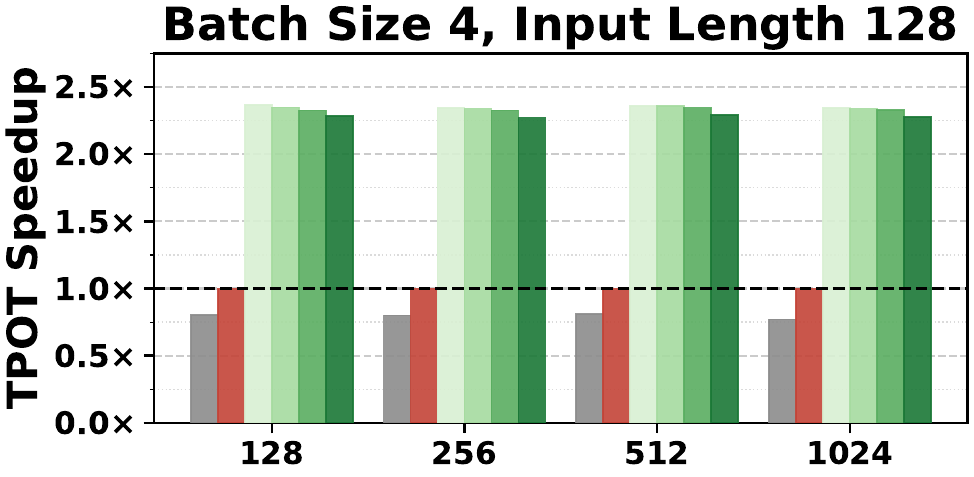}
    \hfill
    \includegraphics[width=0.234\linewidth]{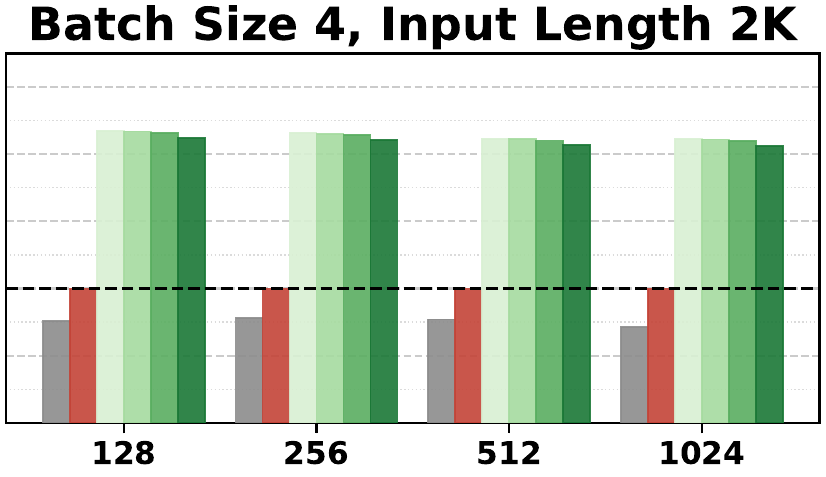}
    \hfill
    \includegraphics[width=0.234\linewidth]{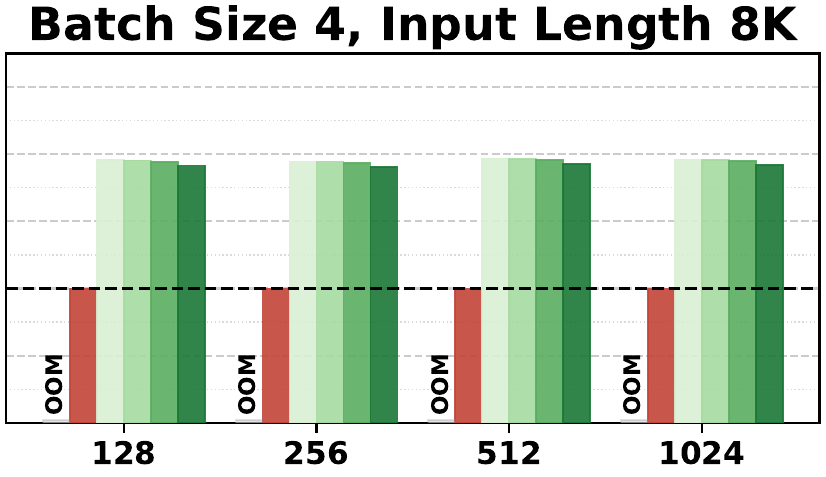}
    \hfill
    \includegraphics[width=0.234\linewidth]{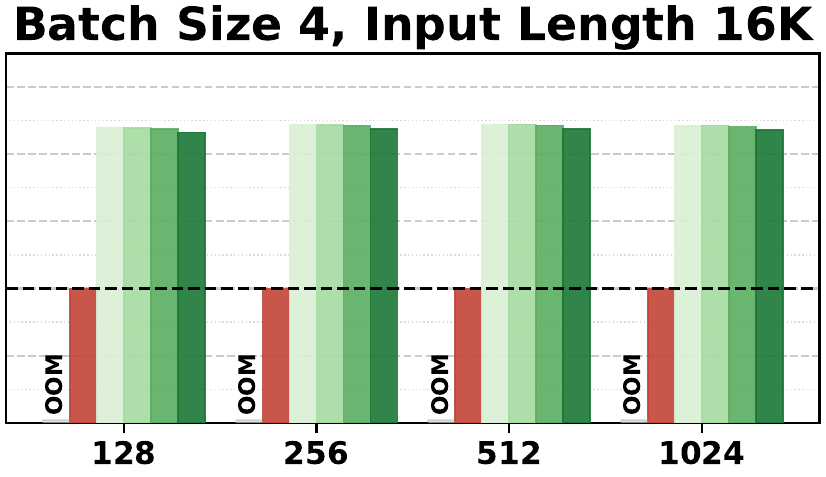}

    \includegraphics[width=0.274\linewidth]{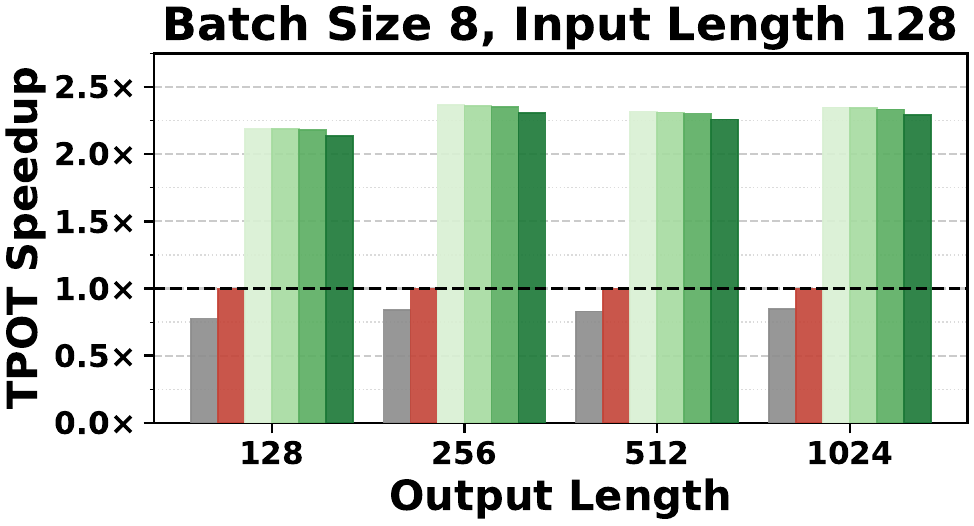}
    \hfill
    \includegraphics[width=0.234\linewidth]{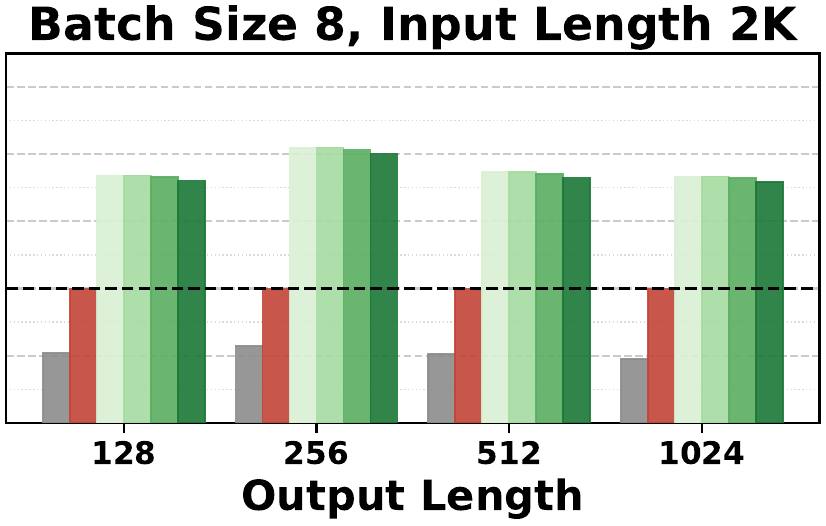}
    \hfill
    \includegraphics[width=0.234\linewidth]{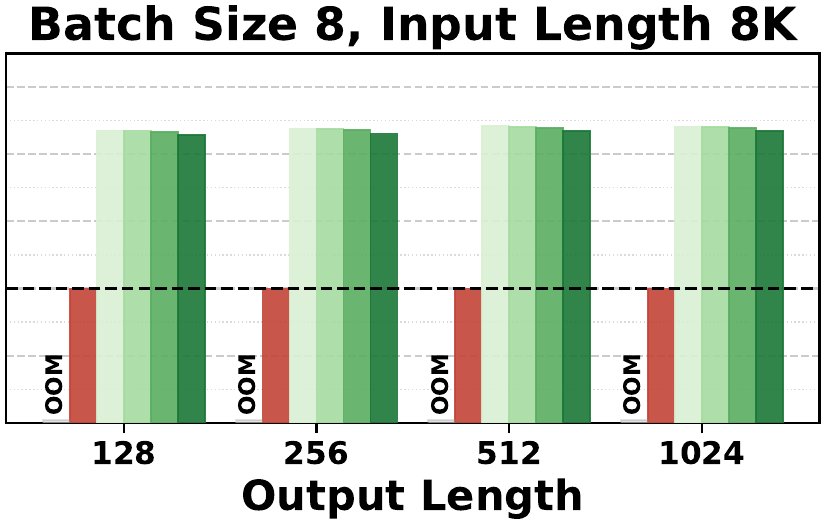}
    \hfill
    \includegraphics[width=0.234\linewidth]{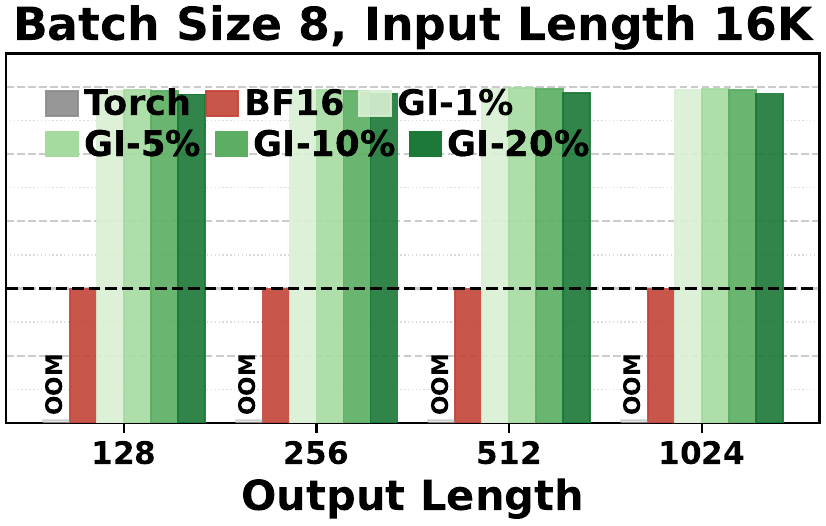}

    \caption{
    Time-per-output-token (TPOT) comparison for \textbf{gradient-informed (GI)} \modelname variants on Llama-3.2-3B.
    Rows vary batch size over 1, 2, 4, and 8; columns vary prompt length over 128, 2K, 8K, and 16K tokens.
    Each panel compares GI retention configurations across output lengths; OOM denotes an out-of-memory run.
    }
    \label{fig:tpot_gi}
    
\end{figure*}

\begin{figure*}[t]
    \centering

    \includegraphics[width=0.274\linewidth]{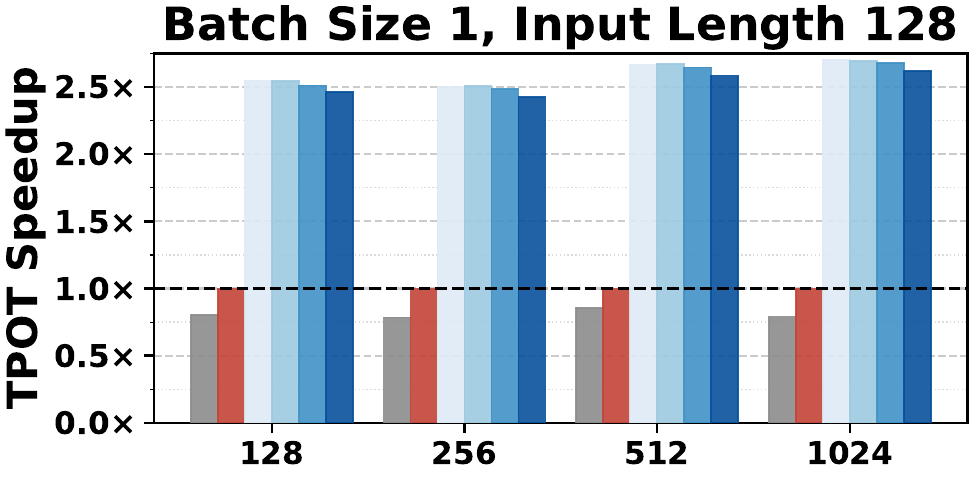}
    \hfill
    \includegraphics[width=0.234\linewidth]{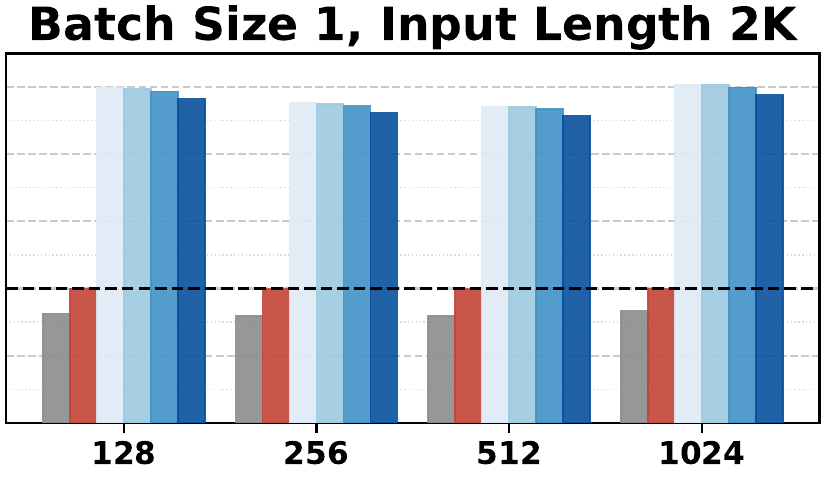}
    \hfill
    \includegraphics[width=0.234\linewidth]{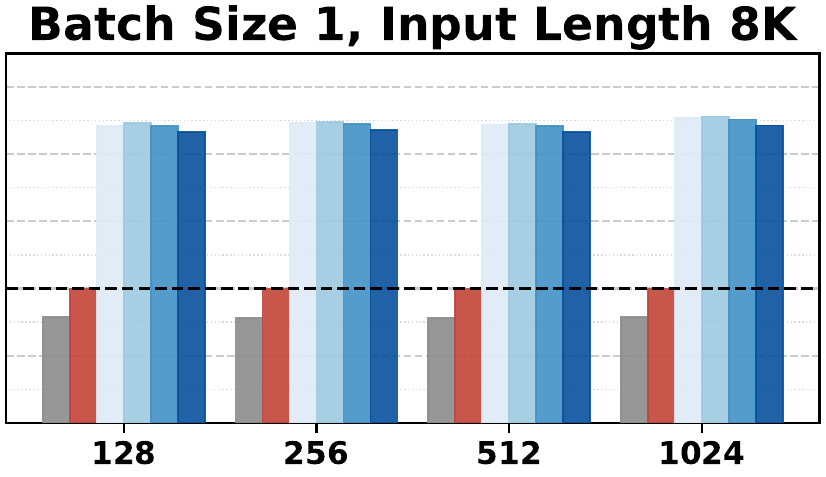}
    \hfill
    \includegraphics[width=0.234\linewidth]{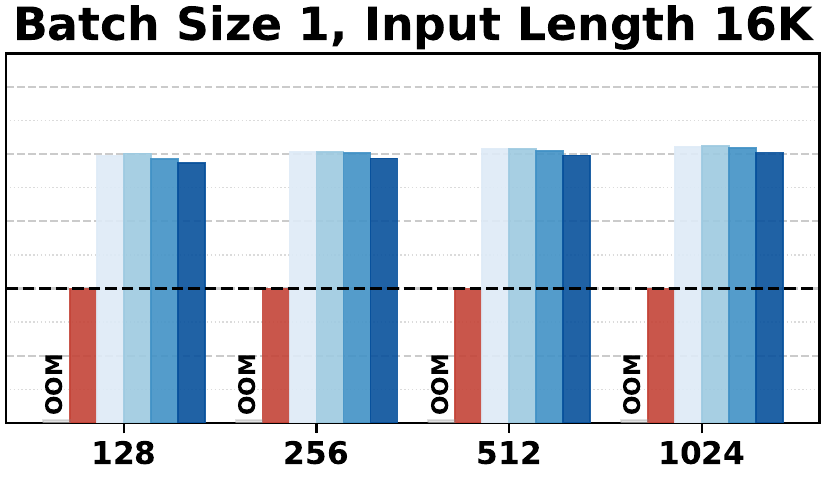}

    \includegraphics[width=0.274\linewidth]{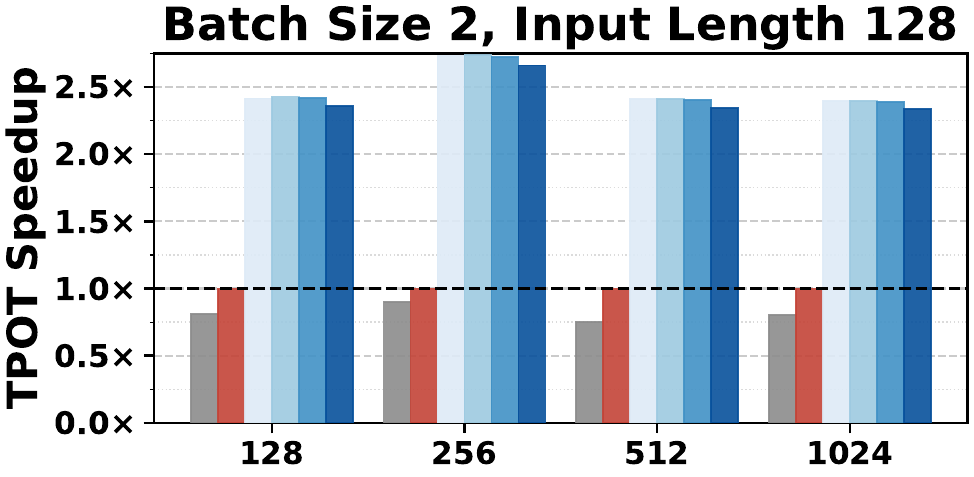}
    \hfill
    \includegraphics[width=0.234\linewidth]{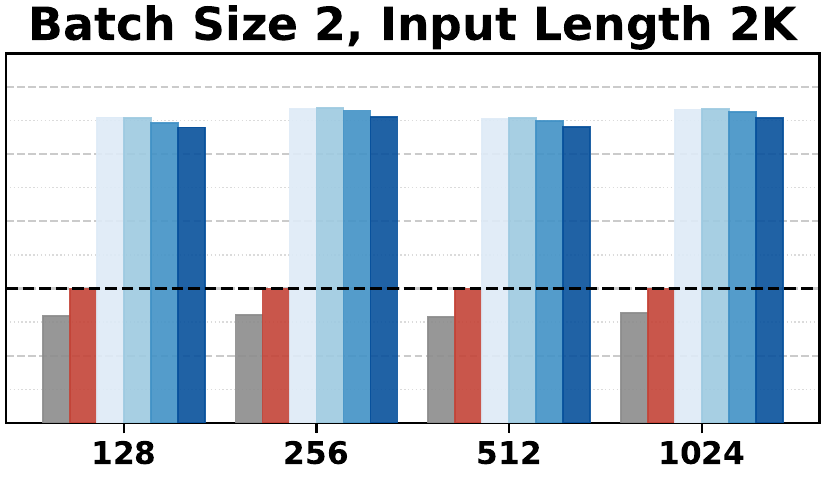}
    \hfill
    \includegraphics[width=0.234\linewidth]{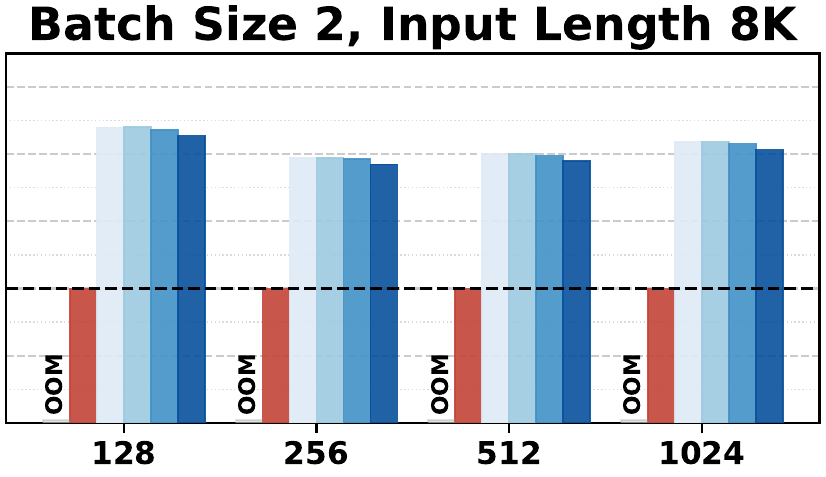}
    \hfill
    \includegraphics[width=0.234\linewidth]{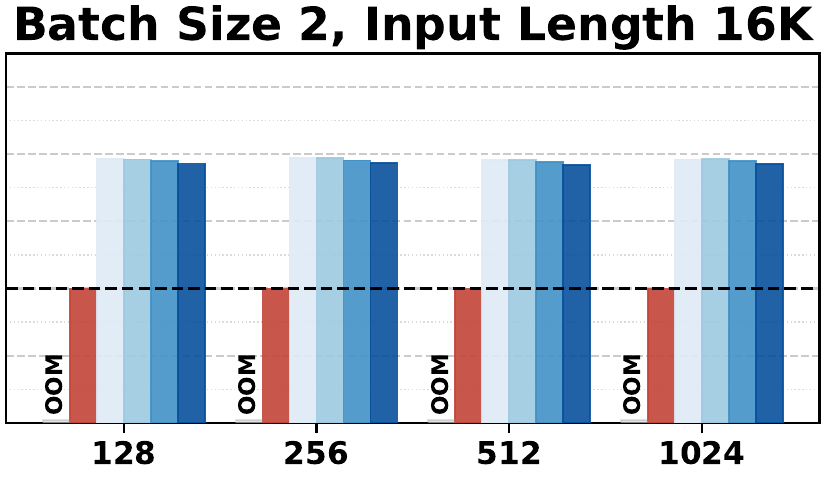}

    \includegraphics[width=0.274\linewidth]{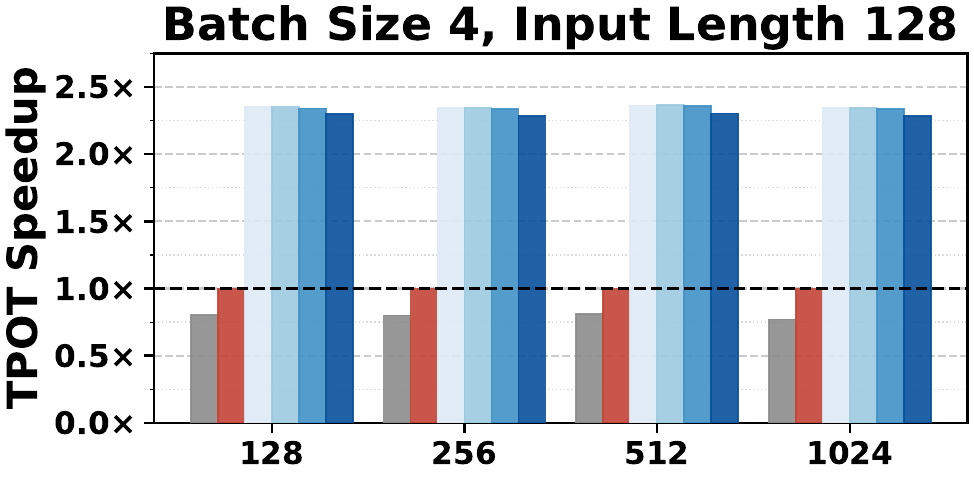}
    \hfill
    \includegraphics[width=0.234\linewidth]{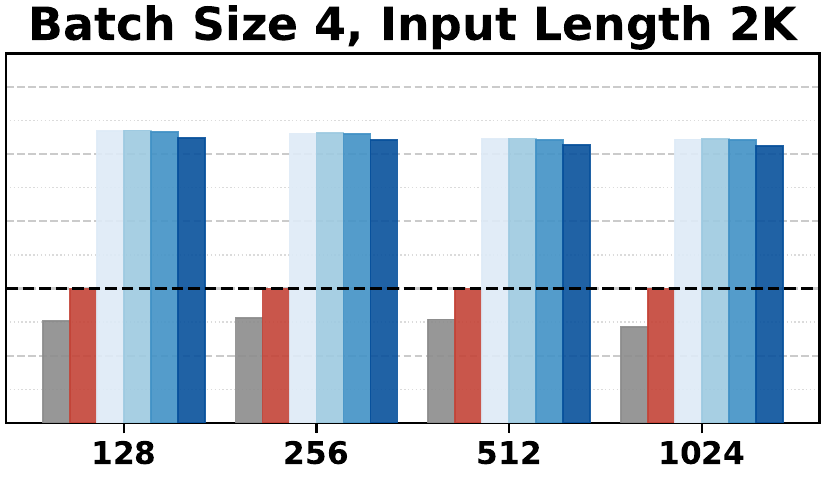}
    \hfill
    \includegraphics[width=0.234\linewidth]{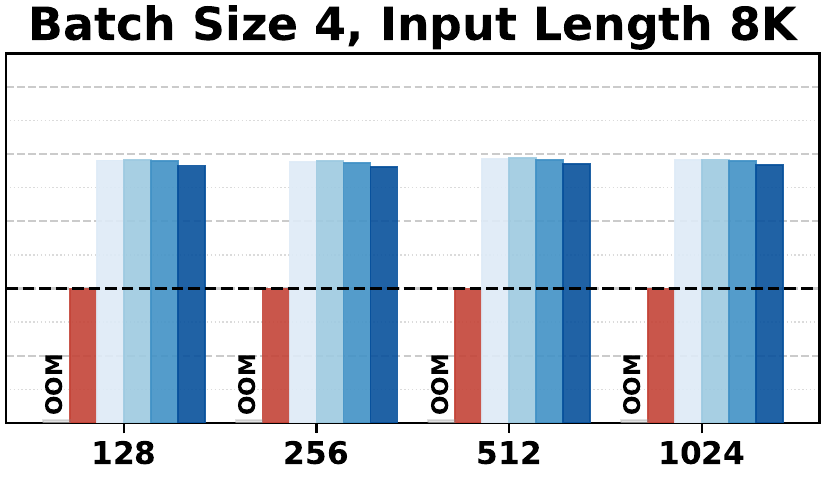}
    \hfill
    \includegraphics[width=0.234\linewidth]{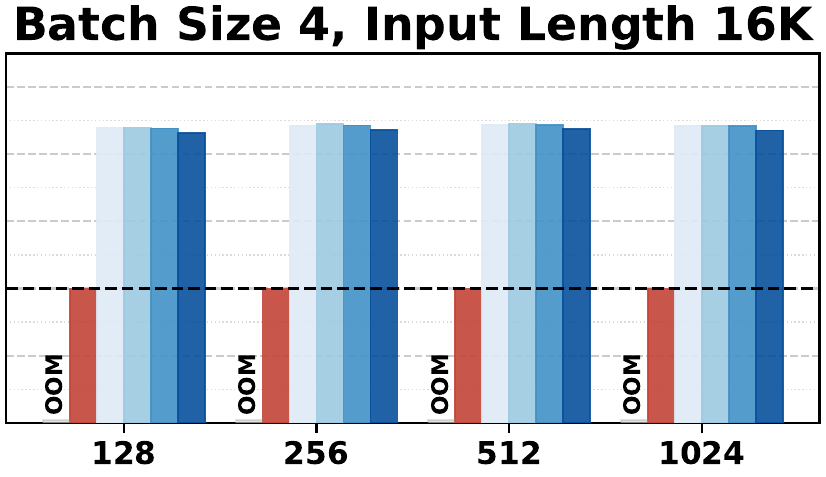}

    \includegraphics[width=0.274\linewidth]{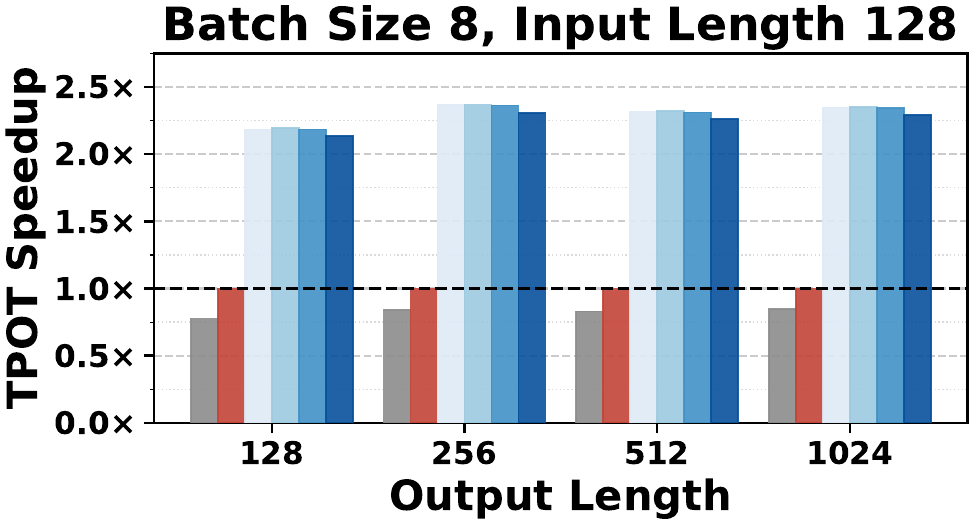}
    \hfill
    \includegraphics[width=0.234\linewidth]{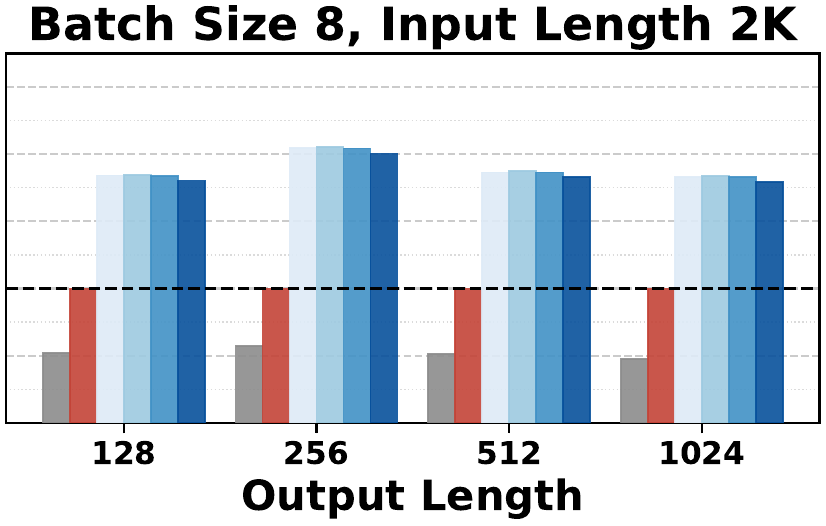}
    \hfill
    \includegraphics[width=0.234\linewidth]{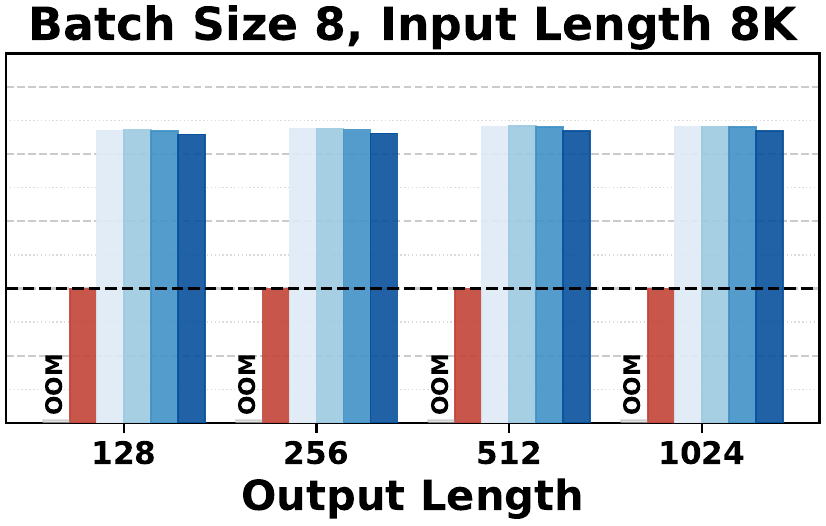}
    \hfill
    \includegraphics[width=0.234\linewidth]{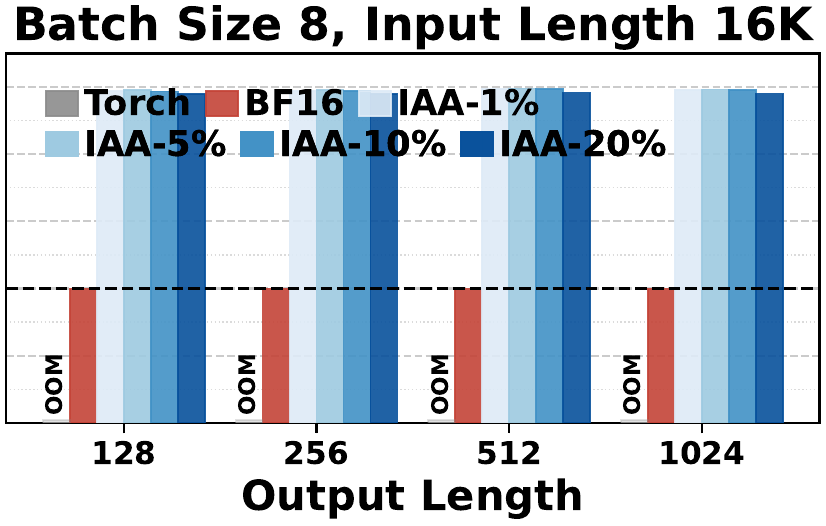}

    \caption{
    Time-per-output-token (TPOT) comparison for \textbf{input-activation-aware (IAA)} \modelname variants on Llama-3.2-3B.
    Rows vary batch size over 1, 2, 4, and 8; columns vary prompt length over 128, 2K, 8K, and 16K tokens.
    Each panel compares IAA retention ratios across output lengths; OOM denotes an out-of-memory run.
    }
    \label{fig:tpot_iaa}
    
\end{figure*}

\begin{figure*}[t]
    \centering

    \includegraphics[width=0.274\linewidth]{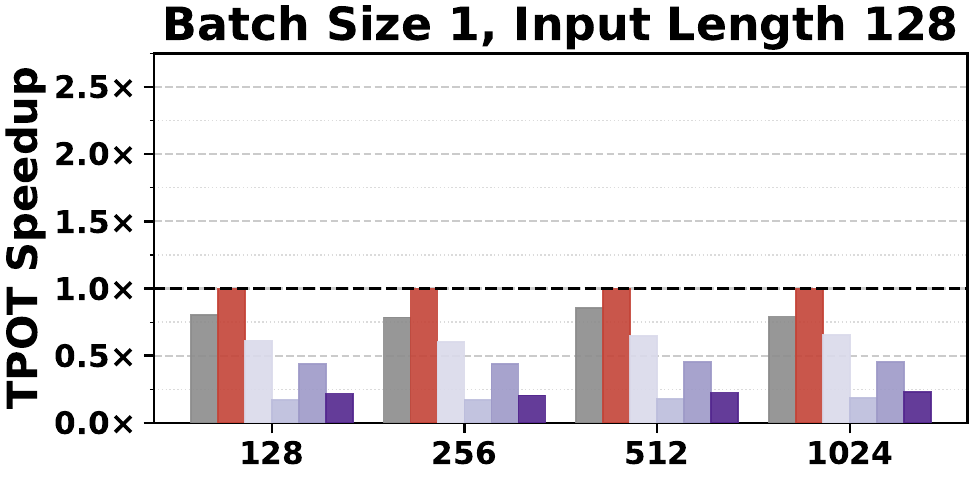}
    \hfill
    \includegraphics[width=0.234\linewidth]{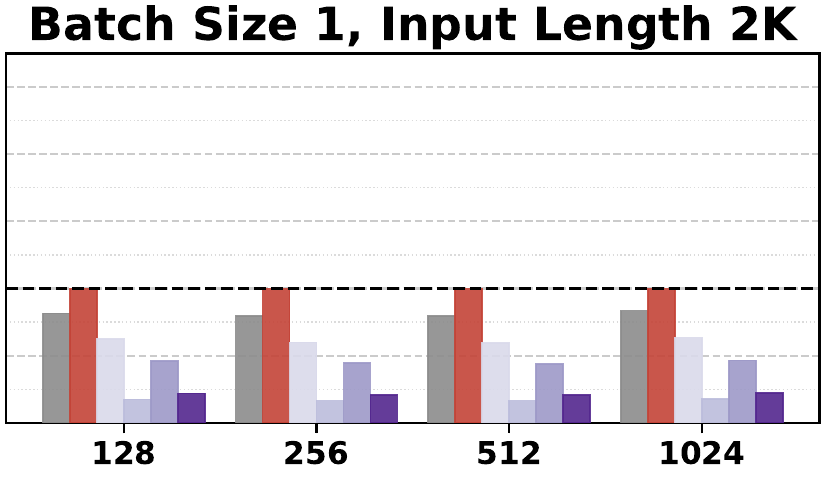}
    \hfill
    \includegraphics[width=0.234\linewidth]{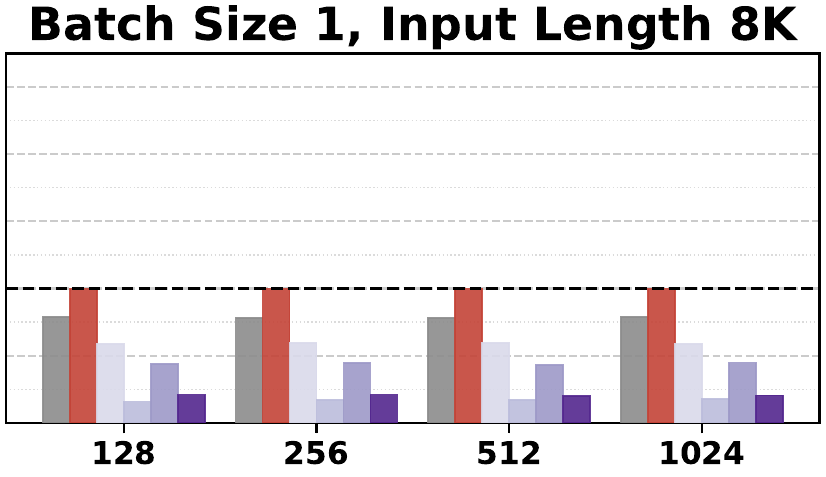}
    \hfill
    \includegraphics[width=0.234\linewidth]{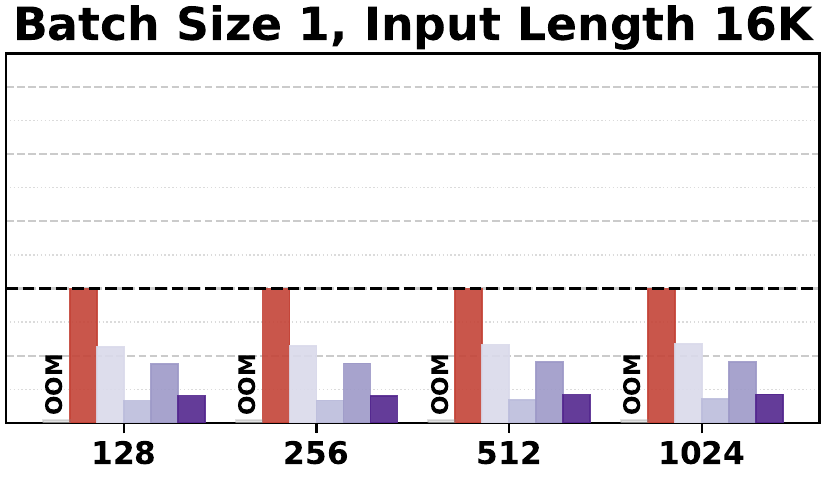}

    \includegraphics[width=0.274\linewidth]{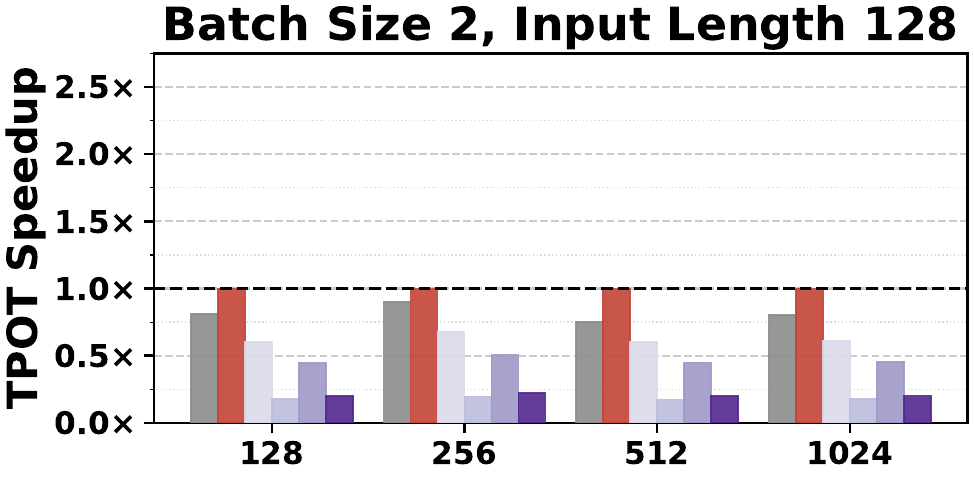}
    \hfill
    \includegraphics[width=0.234\linewidth]{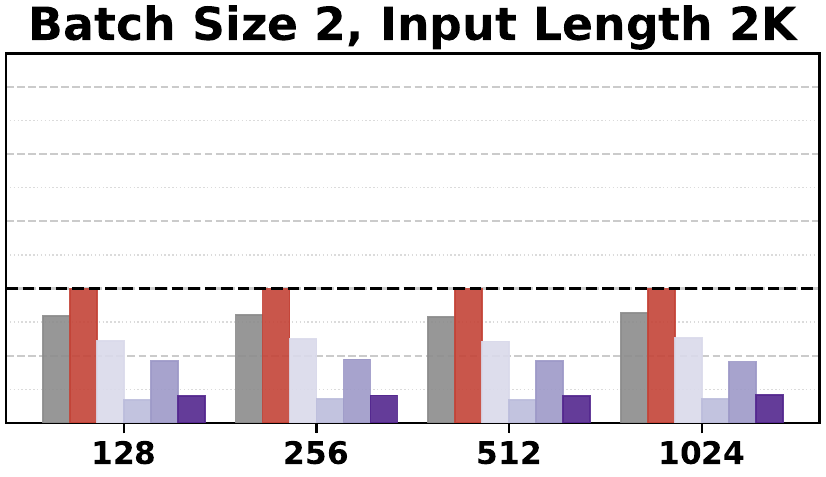}
    \hfill
    \includegraphics[width=0.234\linewidth]{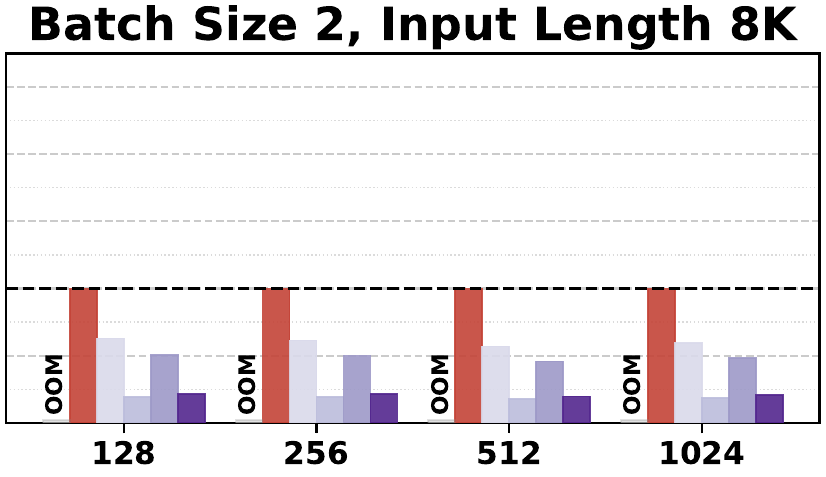}
    \hfill
    \includegraphics[width=0.234\linewidth]{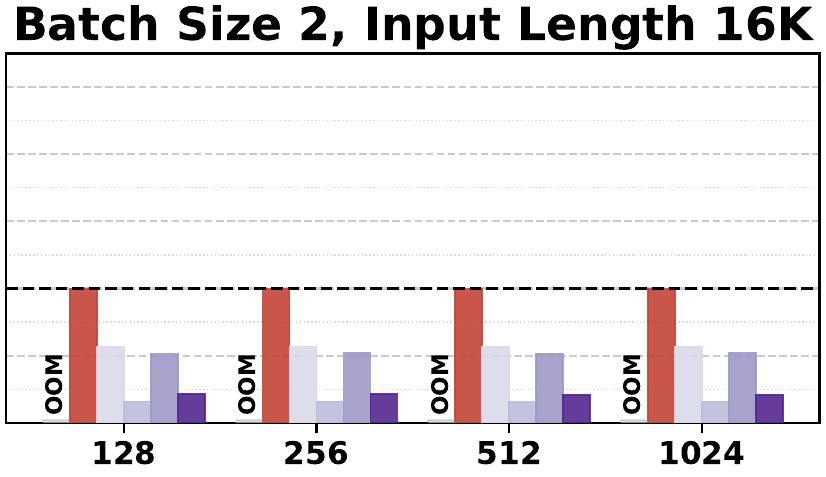}

    \includegraphics[width=0.274\linewidth]{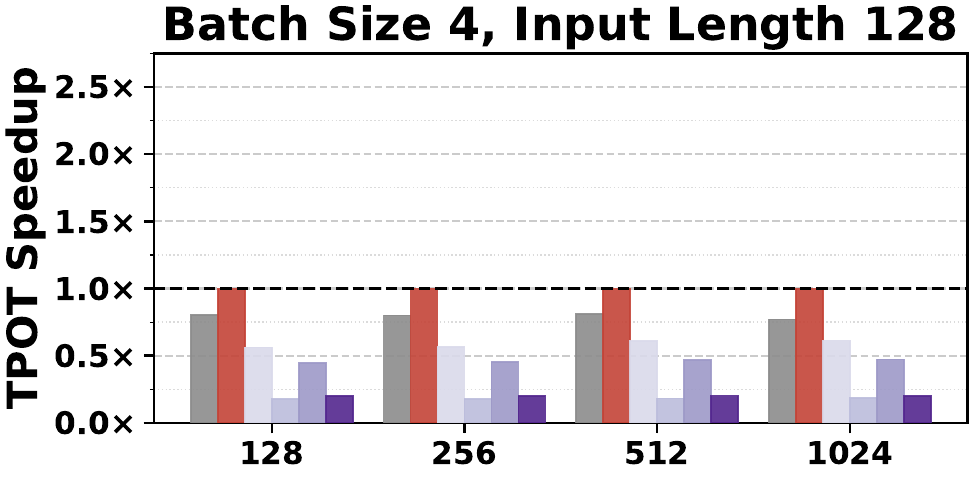}
    \hfill
    \includegraphics[width=0.234\linewidth]{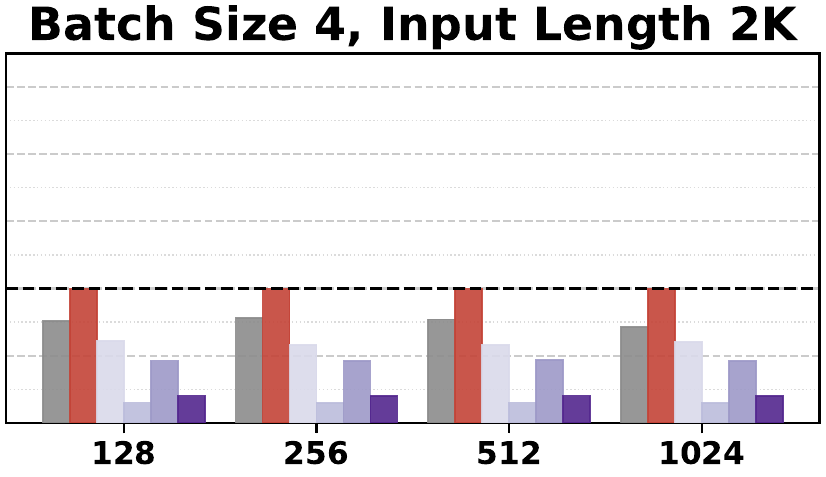}
    \hfill
    \includegraphics[width=0.234\linewidth]{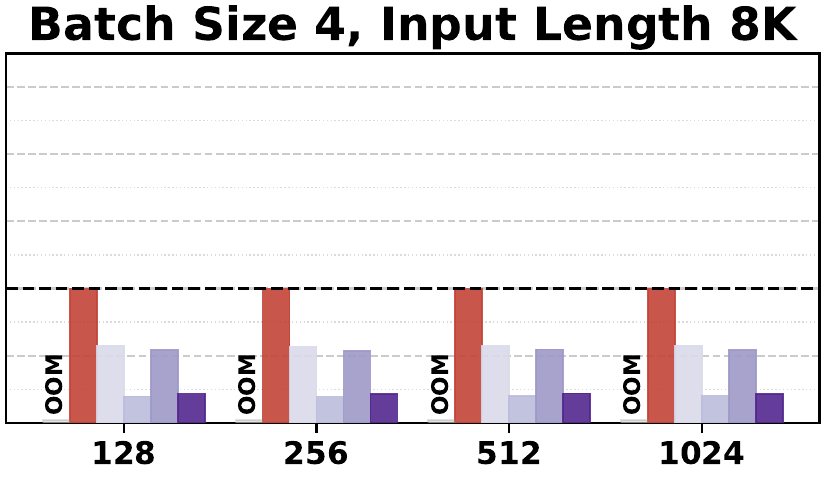}
    \hfill
    \includegraphics[width=0.234\linewidth]{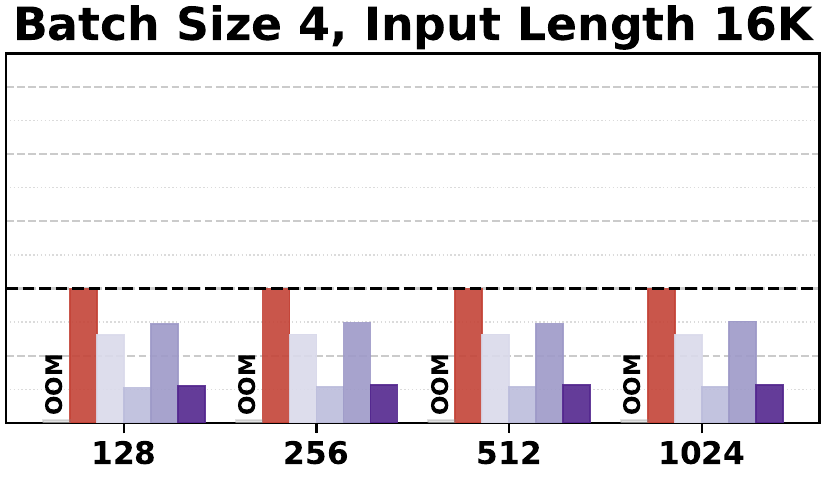}

    \includegraphics[width=0.274\linewidth]{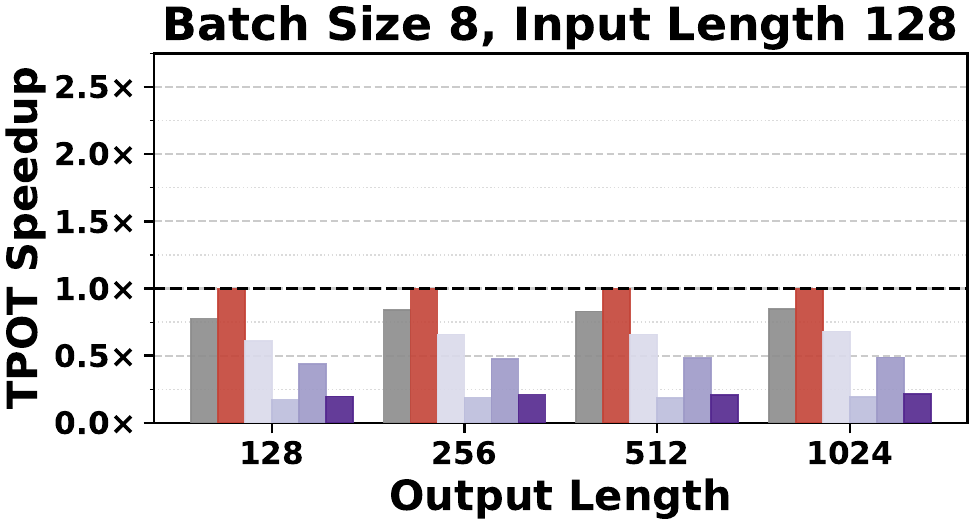}
    \hfill
    \includegraphics[width=0.234\linewidth]{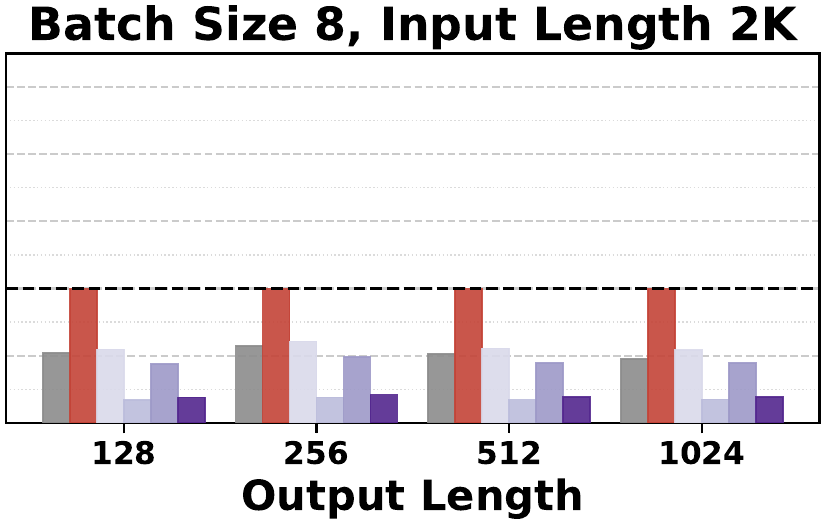}
    \hfill
    \includegraphics[width=0.234\linewidth]{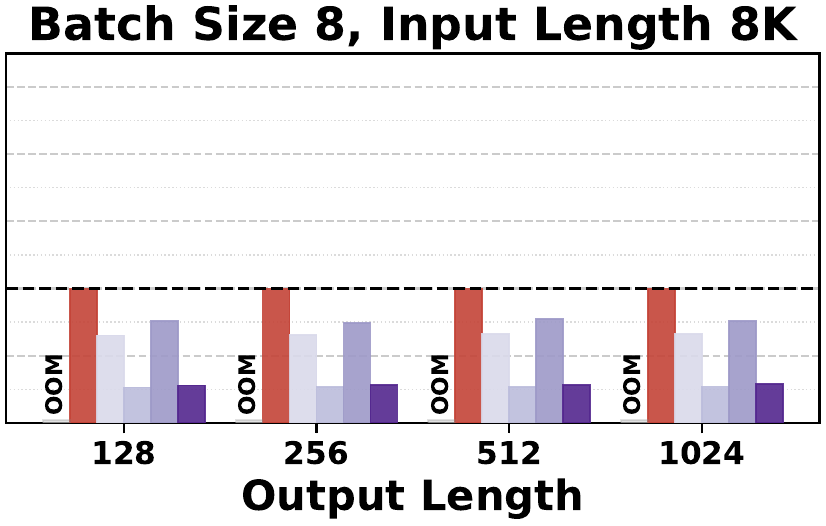}
    \hfill
    \includegraphics[width=0.234\linewidth]{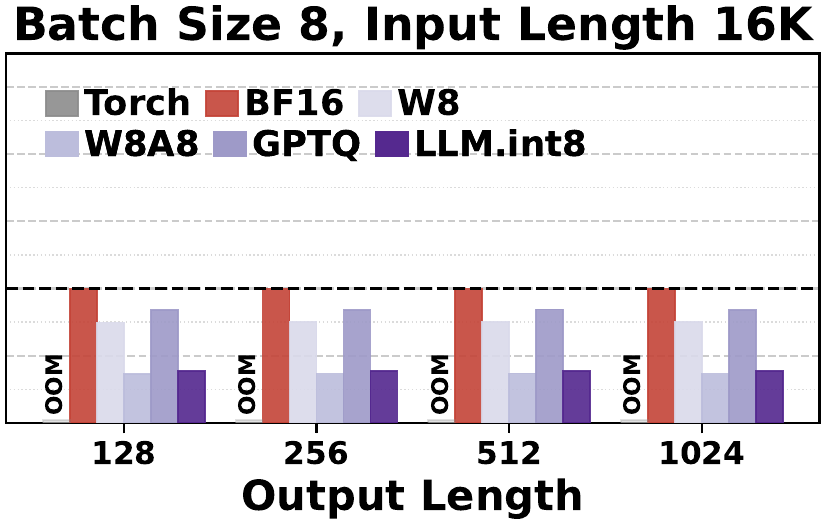}

    \caption{
    Time-per-output-token (TPOT) comparison for \textbf{baseline inference methods} on Llama-3.2-3B.
    Rows vary batch size over 1, 2, 4, and 8; columns vary prompt length over 128, 2K, 8K, and 16K tokens.
    Each panel compares the baseline configurations across output lengths; OOM denotes an out-of-memory run.
    }
    \label{fig:tpot_baselines}
    
\end{figure*}

\begin{figure*}[t]
    \centering

    \includegraphics[width=0.274\linewidth]{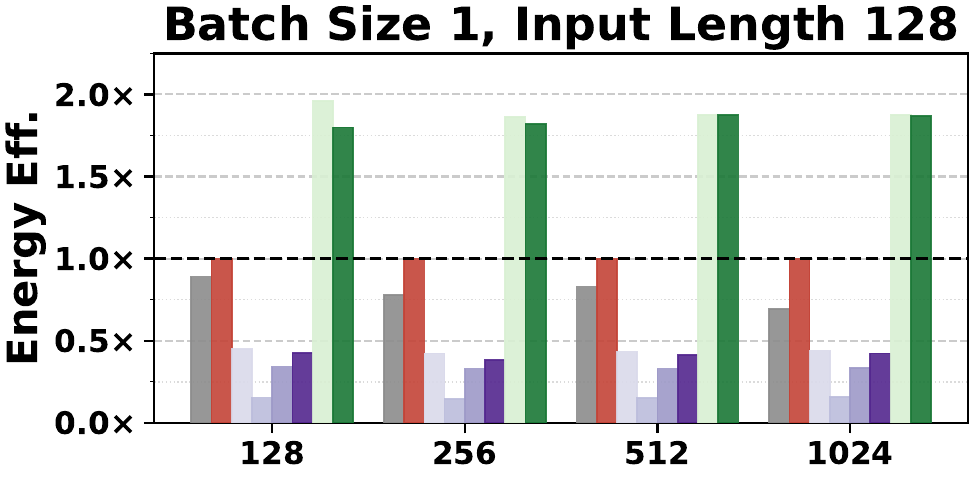}
    \hfill
    \includegraphics[width=0.234\linewidth]{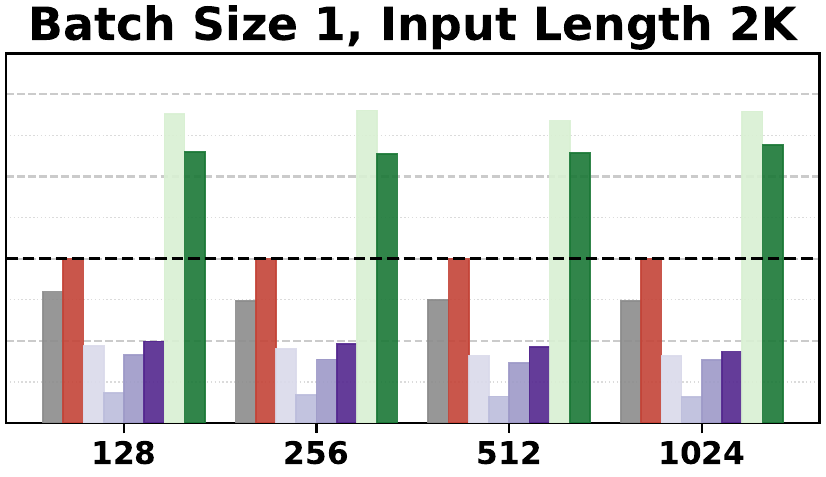}
    \hfill
    \includegraphics[width=0.234\linewidth]{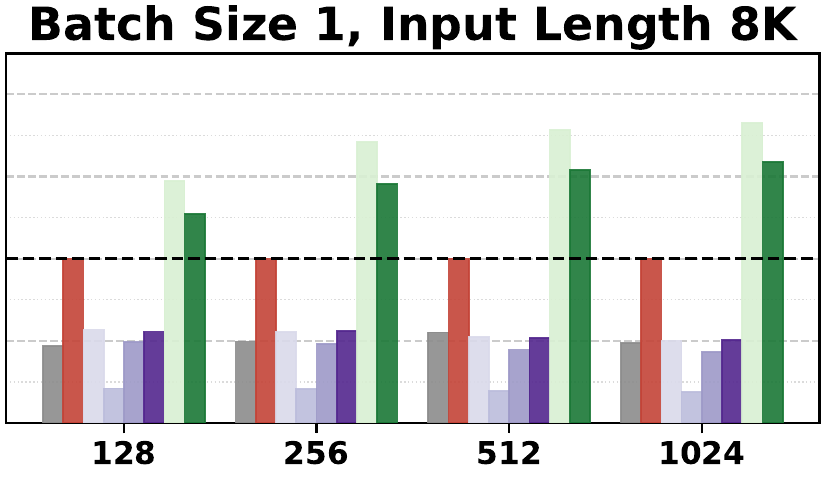}
    \hfill
    \includegraphics[width=0.234\linewidth]{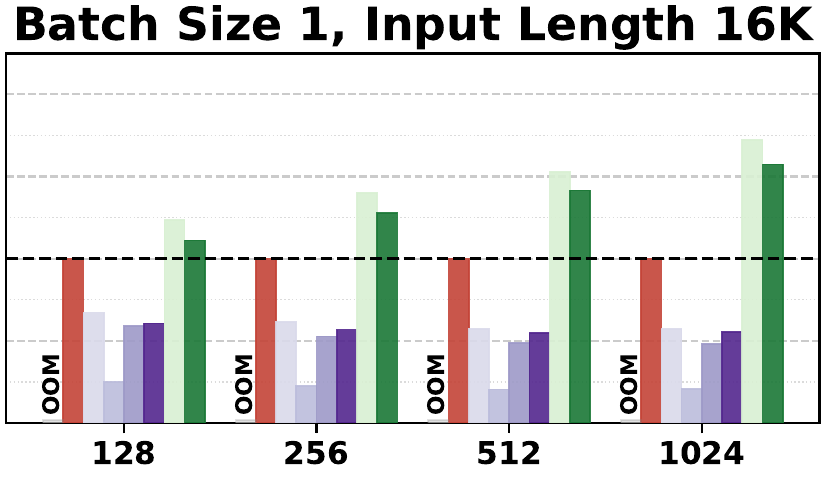}

    \includegraphics[width=0.274\linewidth]{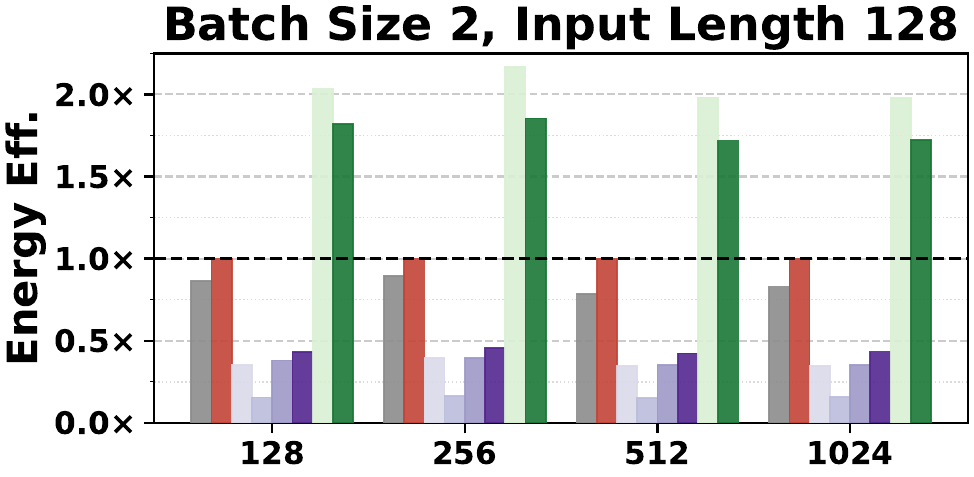}
    \hfill
    \includegraphics[width=0.234\linewidth]{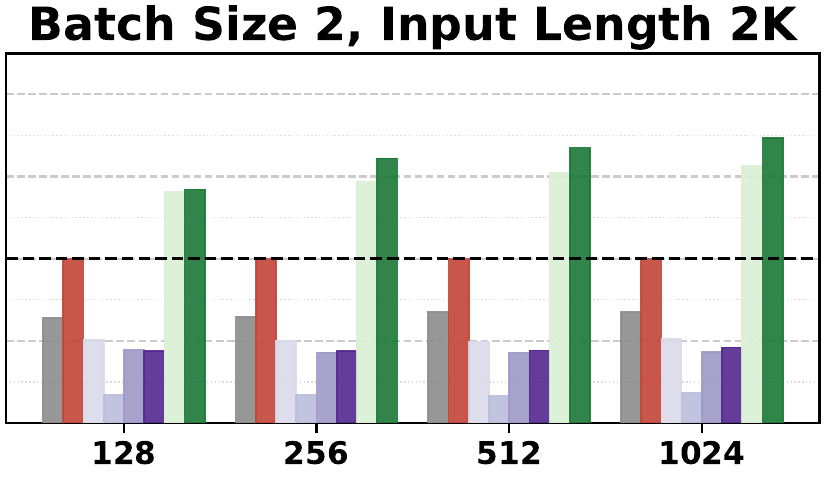}
    \hfill
    \includegraphics[width=0.234\linewidth]{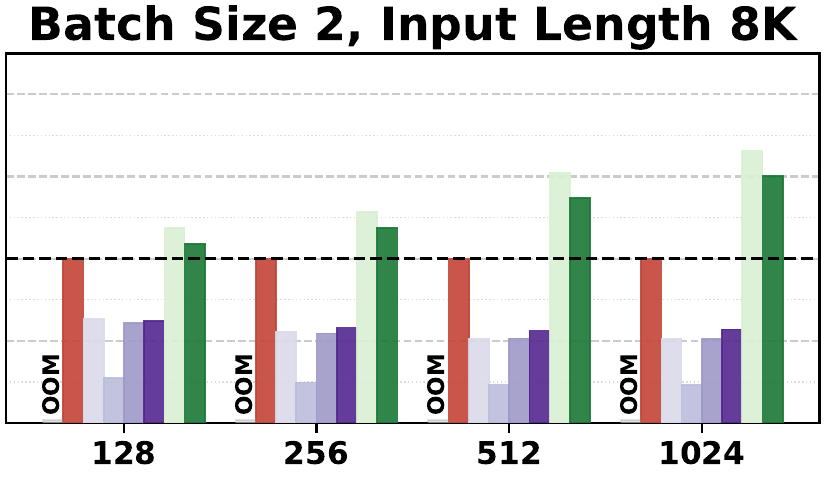}
    \hfill
    \includegraphics[width=0.234\linewidth]{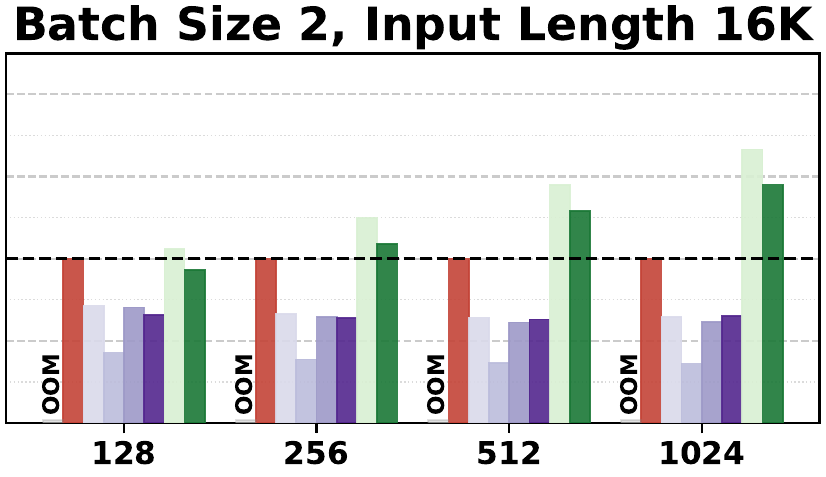}

    \includegraphics[width=0.274\linewidth]{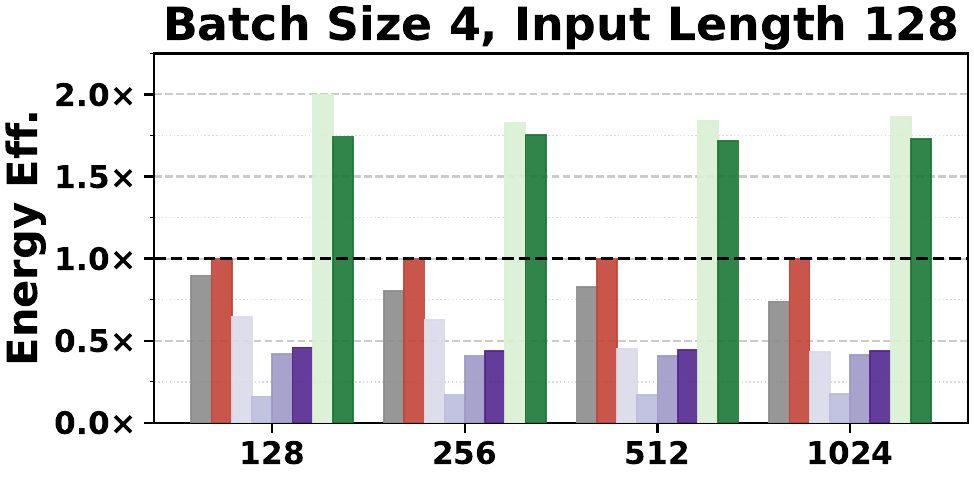}
    \hfill
    \includegraphics[width=0.234\linewidth]{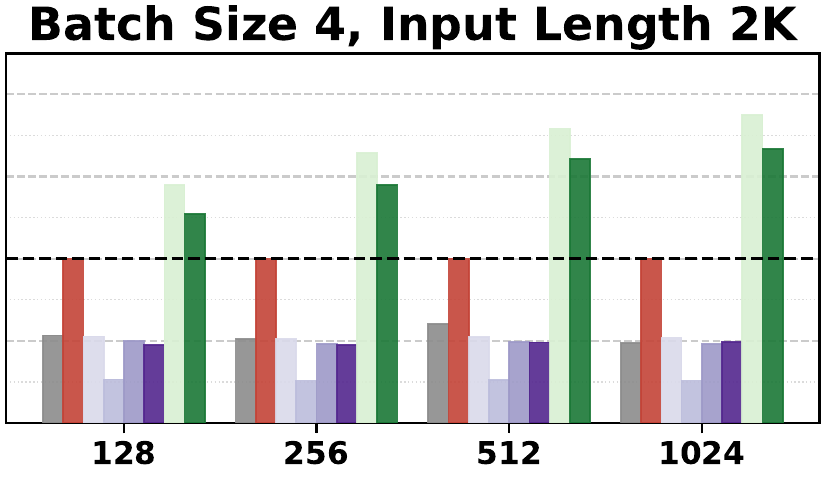}
    \hfill
    \includegraphics[width=0.234\linewidth]{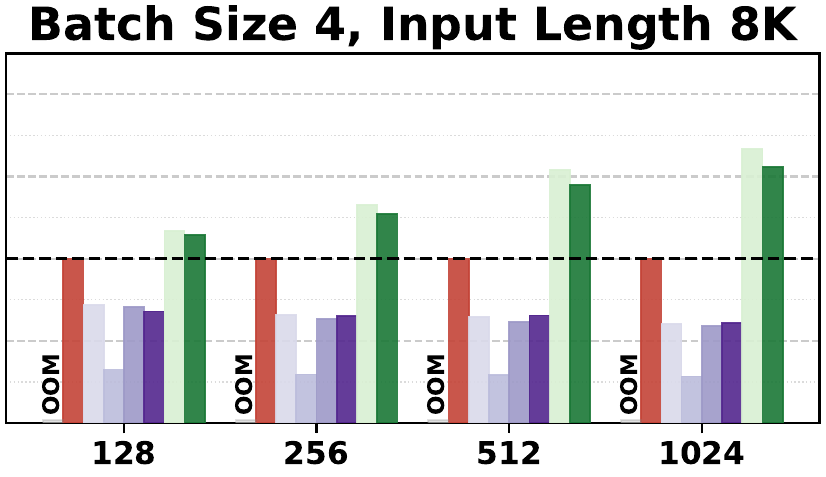}
    \hfill
    \includegraphics[width=0.234\linewidth]{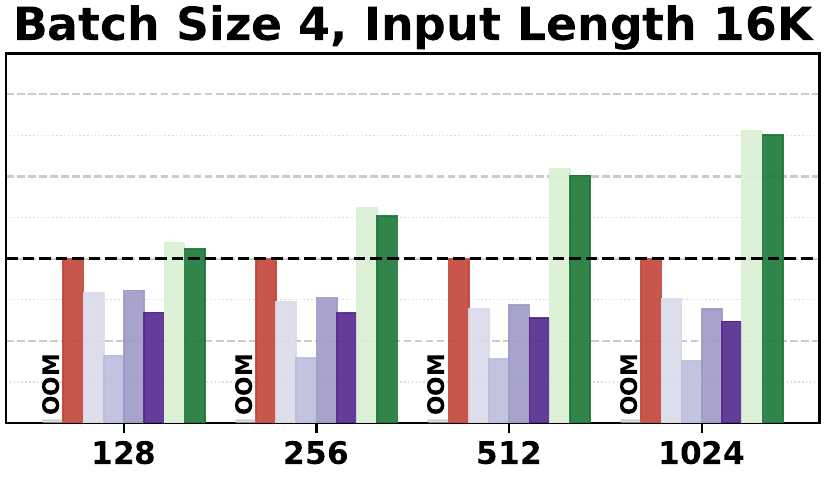}

    \includegraphics[width=0.274\linewidth]{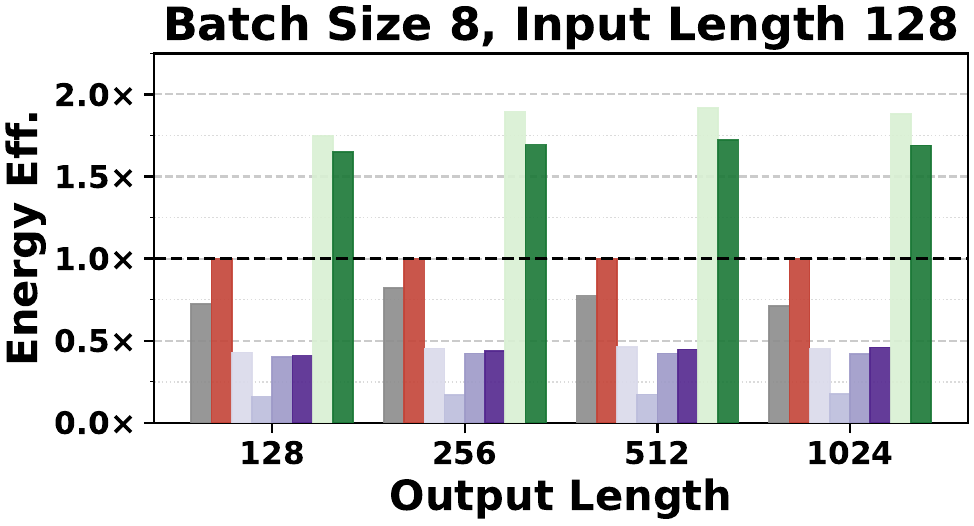}
    \hfill
    \includegraphics[width=0.234\linewidth]{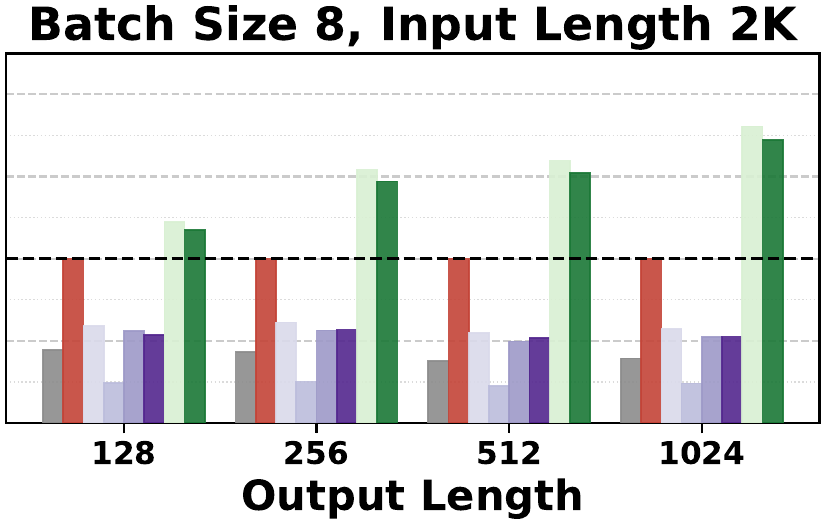}
    \hfill
    \includegraphics[width=0.234\linewidth]{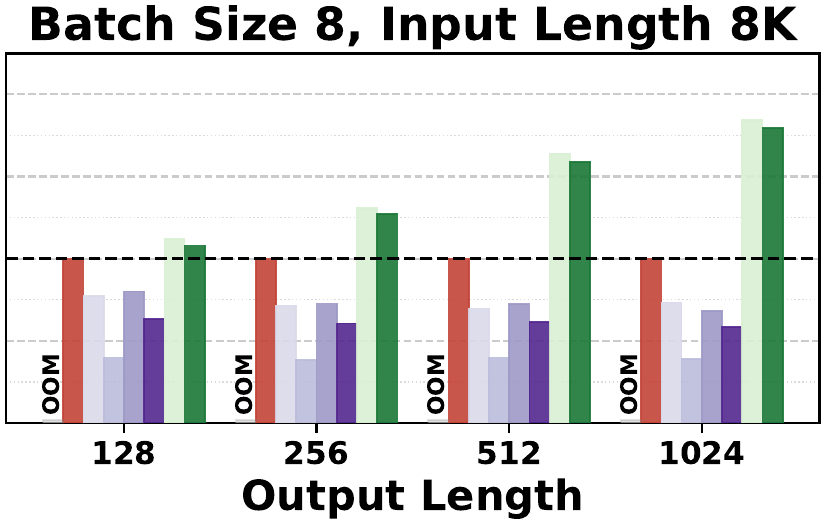}
    \hfill
    \includegraphics[width=0.234\linewidth]{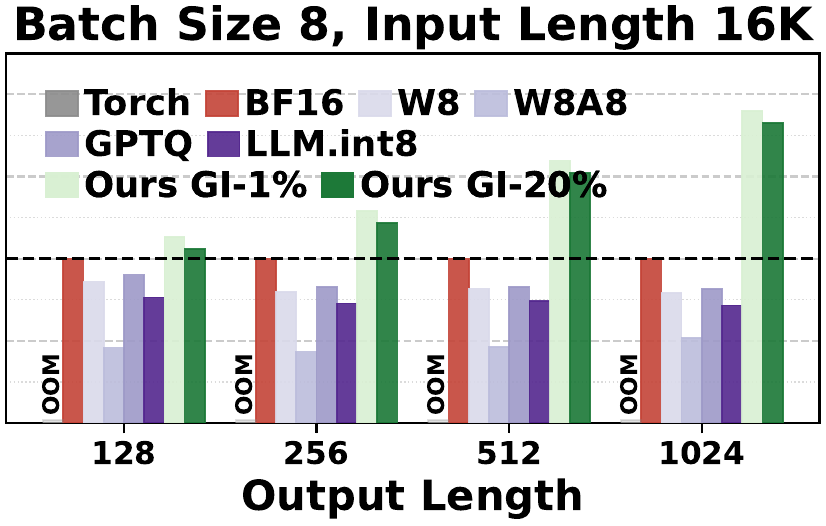}

    \caption{
    Generation-energy comparison on Llama-3.2-3B.
    Rows vary batch size over 1, 2, 4, and 8; columns vary prompt length over 128, 2K, 8K, and 16K tokens.
    Each panel compares inference methods across output lengths of 128, 256, 512, and 1024 tokens; OOM denotes an out-of-memory run.
    }
    \label{fig:energy_main}
    
\end{figure*}

\begin{figure*}[t]
    \centering

    \includegraphics[width=0.274\linewidth]{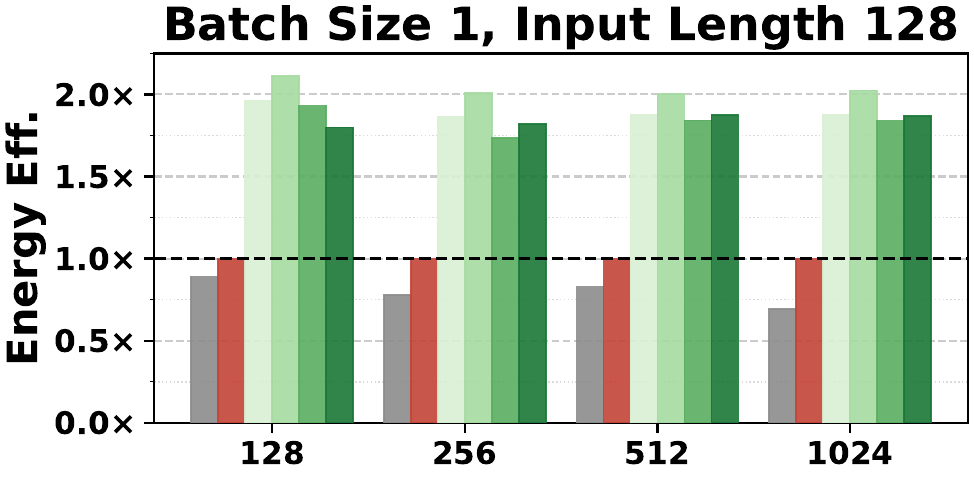}
    \hfill
    \includegraphics[width=0.234\linewidth]{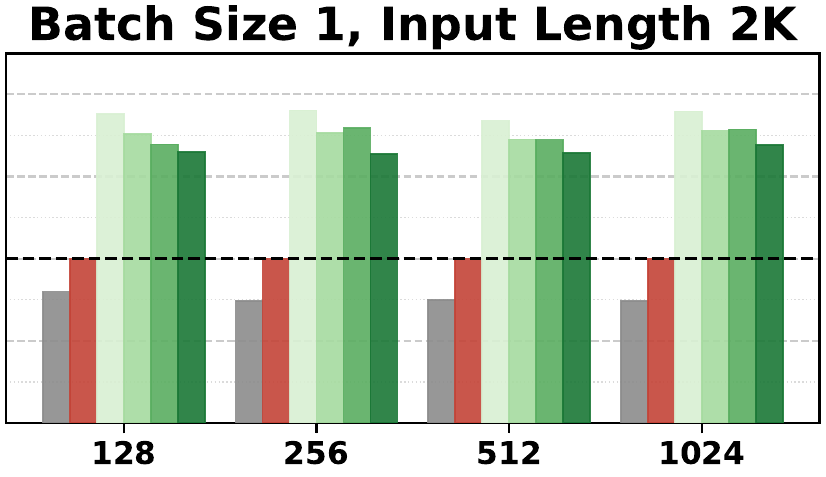}
    \hfill
    \includegraphics[width=0.234\linewidth]{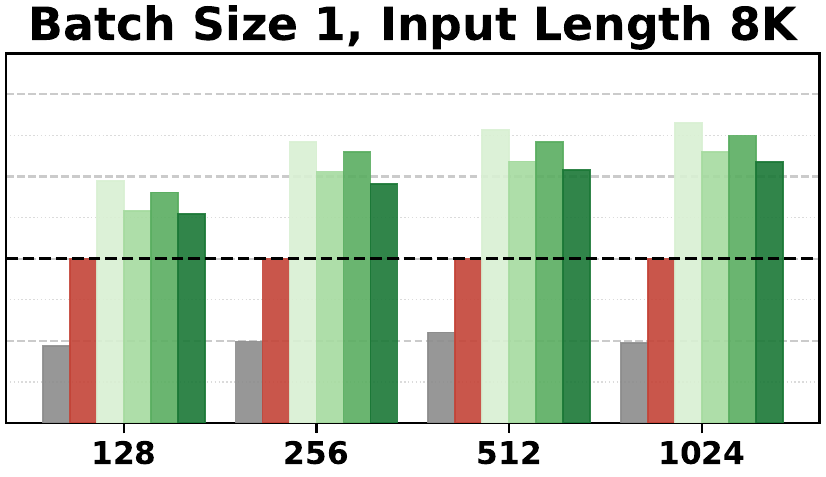}
    \hfill
    \includegraphics[width=0.234\linewidth]{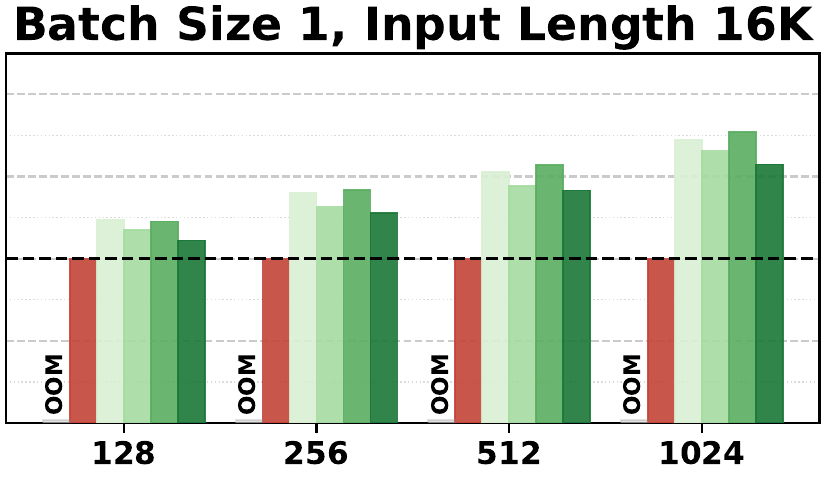}

    \includegraphics[width=0.274\linewidth]{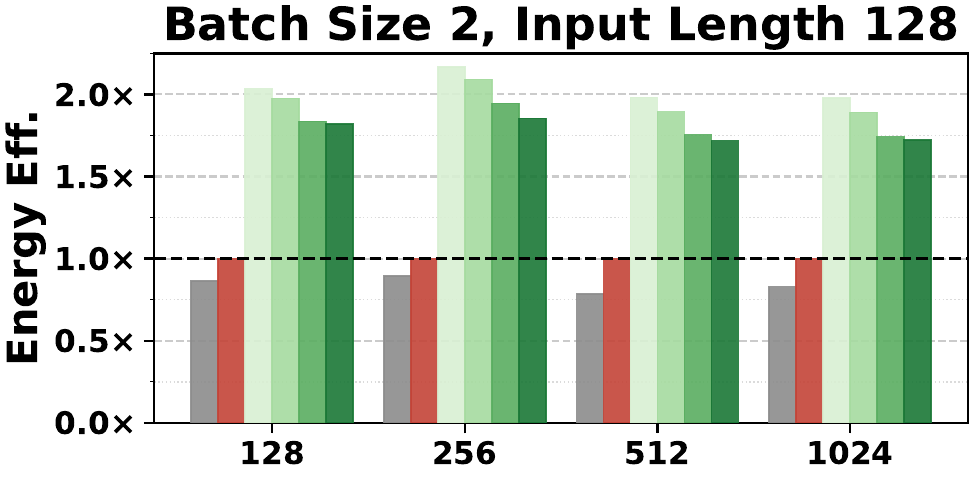}
    \hfill
    \includegraphics[width=0.234\linewidth]{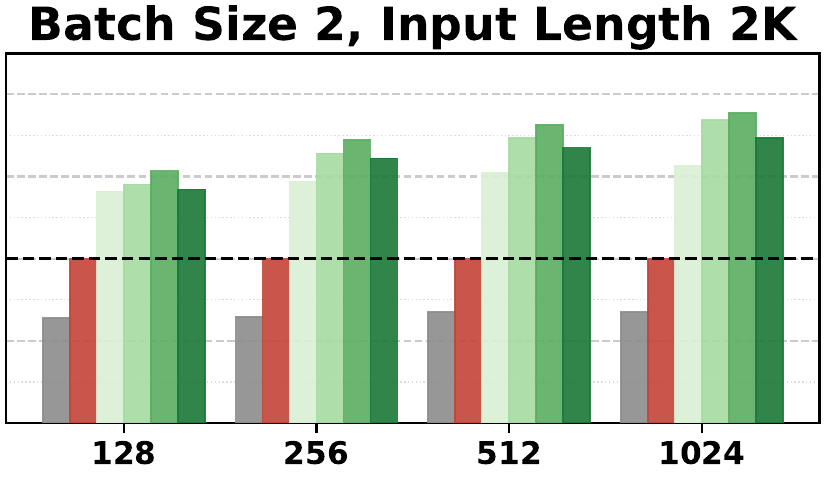}
    \hfill
    \includegraphics[width=0.234\linewidth]{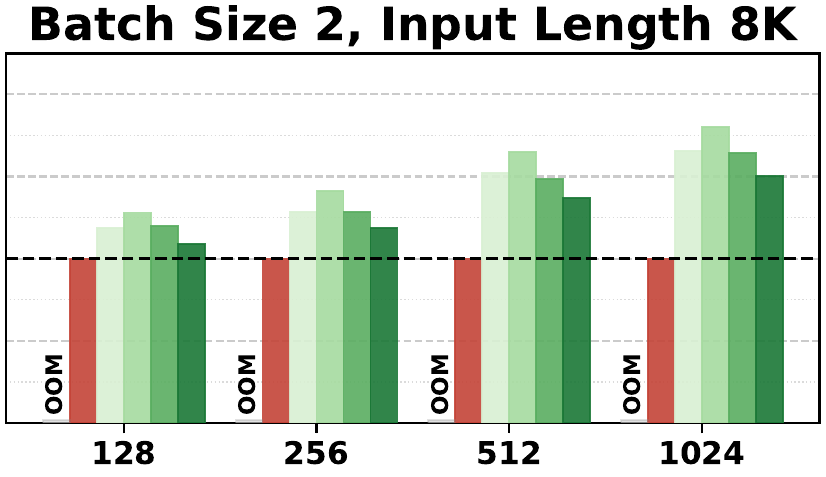}
    \hfill
    \includegraphics[width=0.234\linewidth]{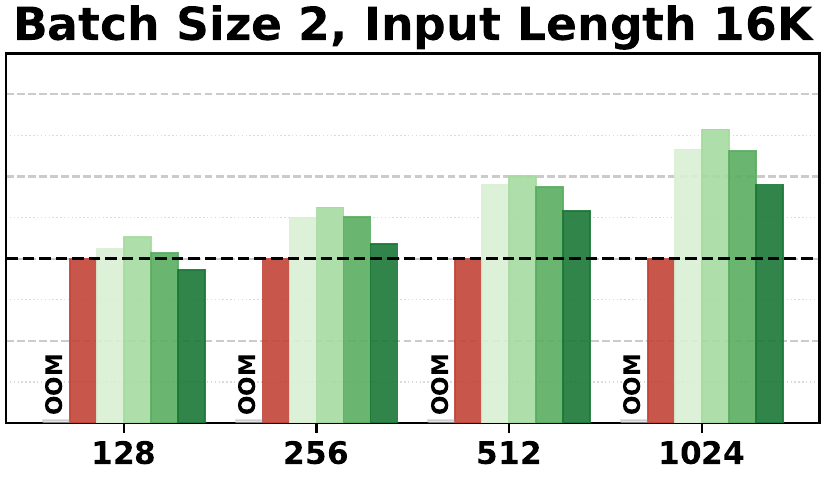}

    \includegraphics[width=0.274\linewidth]{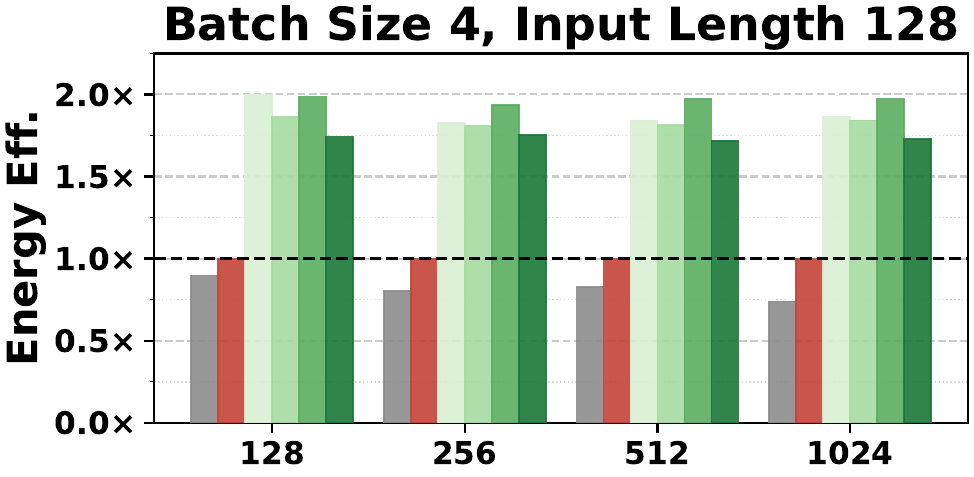}
    \hfill
    \includegraphics[width=0.234\linewidth]{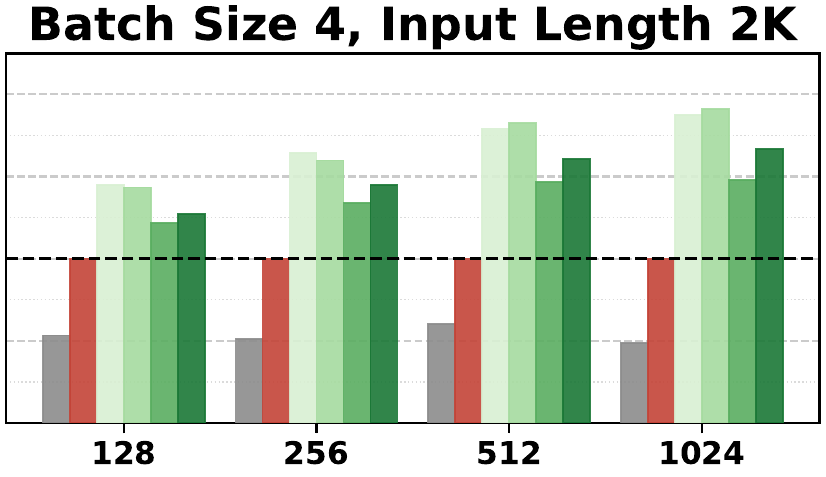}
    \hfill
    \includegraphics[width=0.234\linewidth]{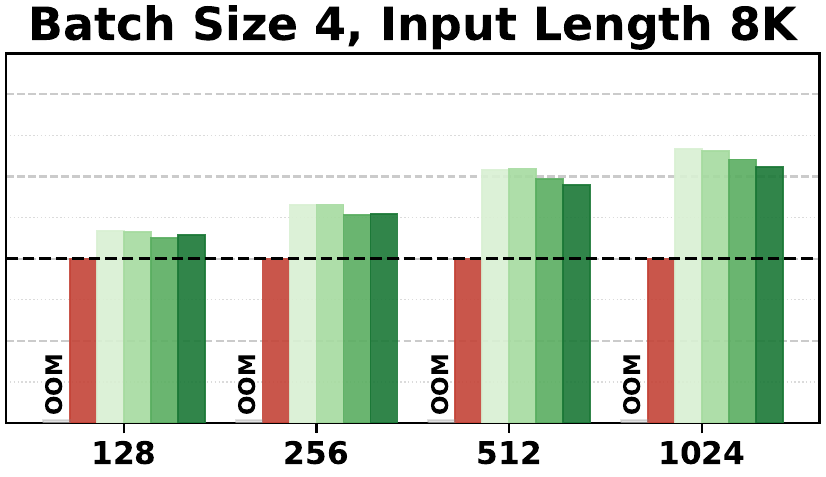}
    \hfill
    \includegraphics[width=0.234\linewidth]{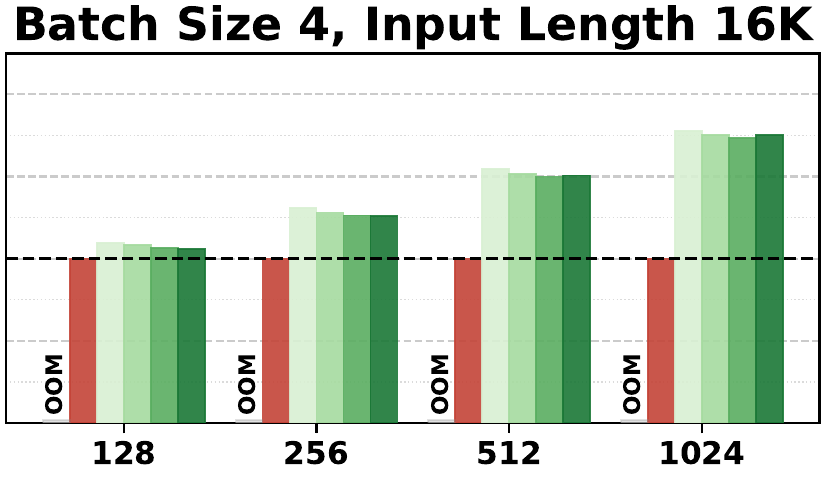}

    \includegraphics[width=0.274\linewidth]{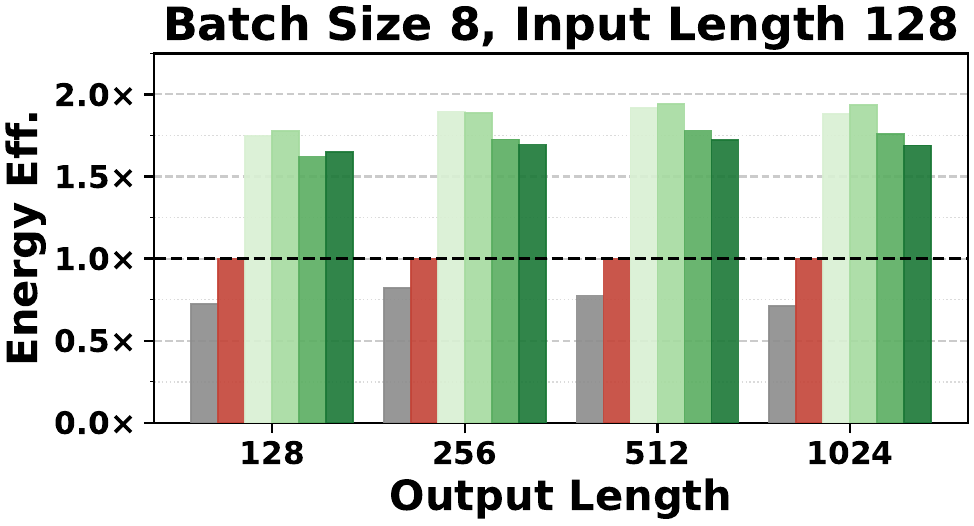}
    \hfill
    \includegraphics[width=0.234\linewidth]{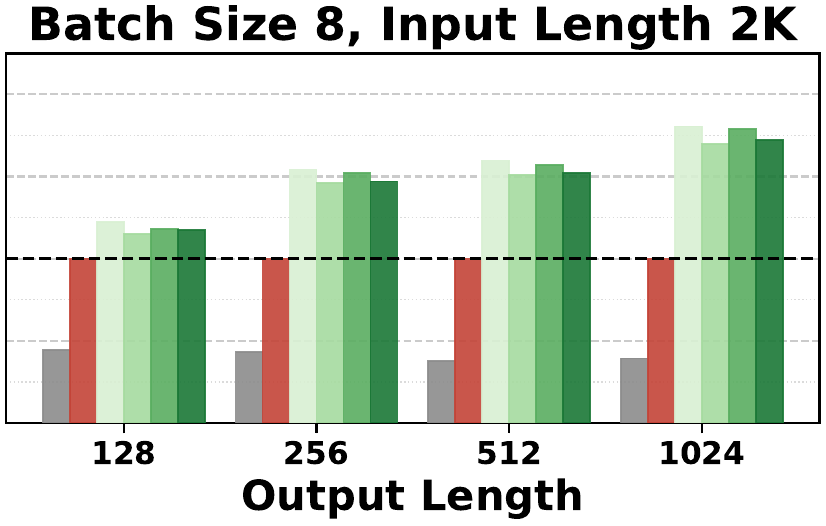}
    \hfill
    \includegraphics[width=0.234\linewidth]{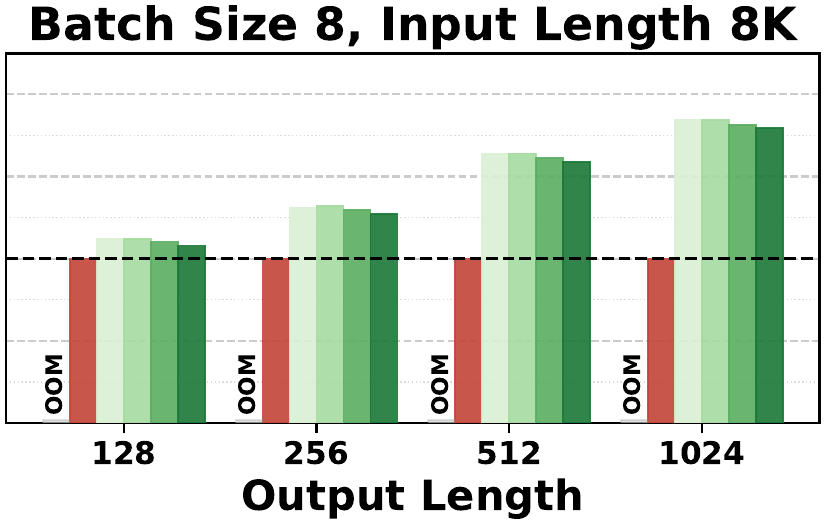}
    \hfill
    \includegraphics[width=0.234\linewidth]{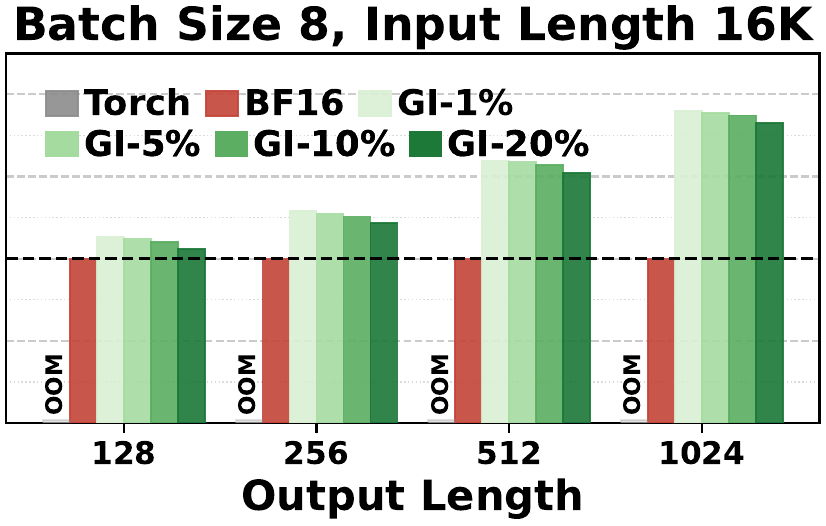}

    \caption{
    Generation-energy comparison for \textbf{gradient-informed (GI)} \modelname variants on Llama-3.2-3B.
    Rows vary batch size over 1, 2, 4, and 8; columns vary prompt length over 128, 2K, 8K, and 16K tokens.
    Each panel compares GI retention configurations across output lengths; OOM denotes an out-of-memory run.
    }
    \label{fig:energy_gi}
    
\end{figure*}

\begin{figure*}[t]
    \centering

    \includegraphics[width=0.274\linewidth]{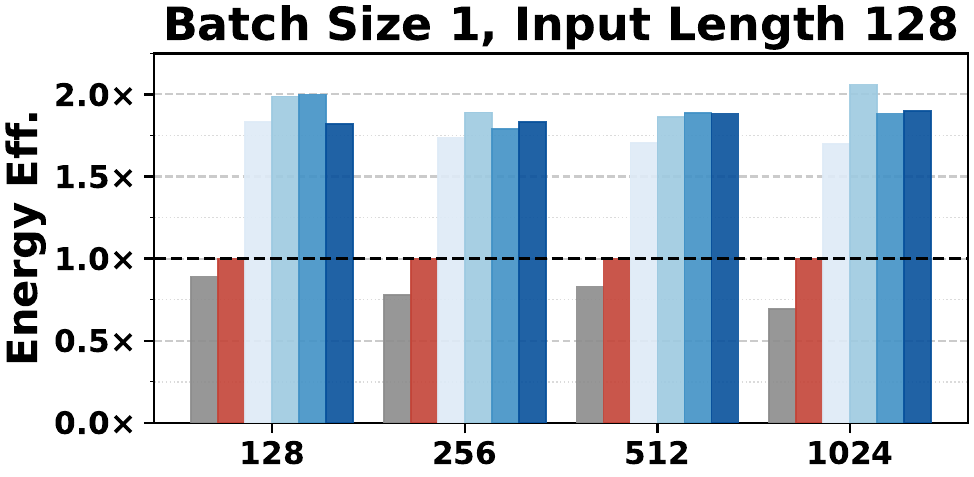}
    \hfill
    \includegraphics[width=0.234\linewidth]{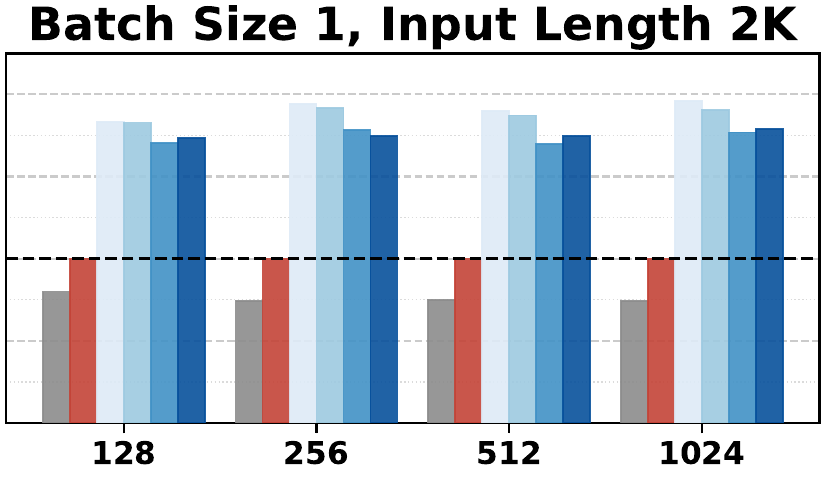}
    \hfill
    \includegraphics[width=0.234\linewidth]{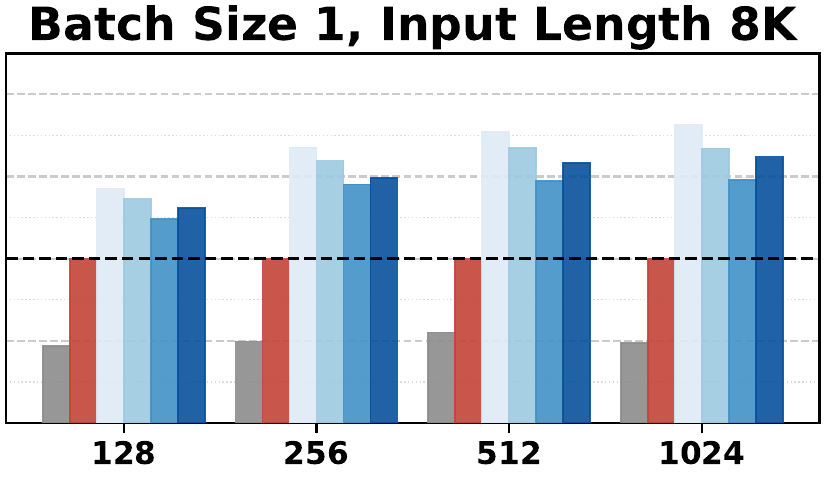}
    \hfill
    \includegraphics[width=0.234\linewidth]{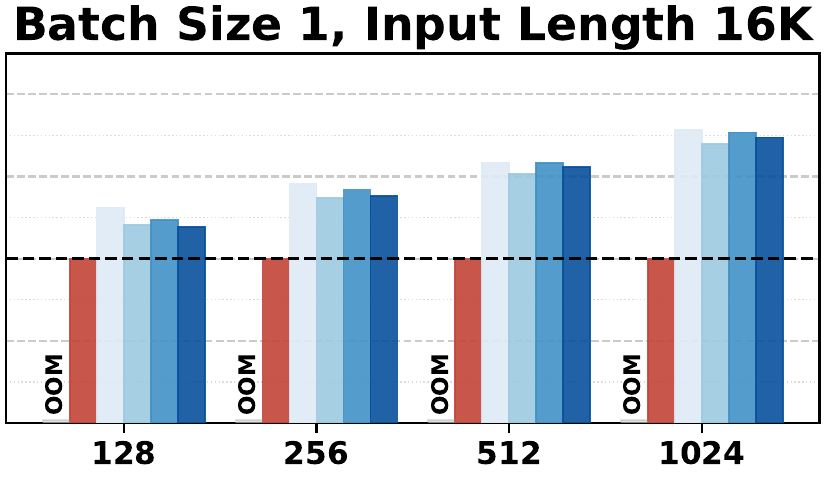}

    \includegraphics[width=0.274\linewidth]{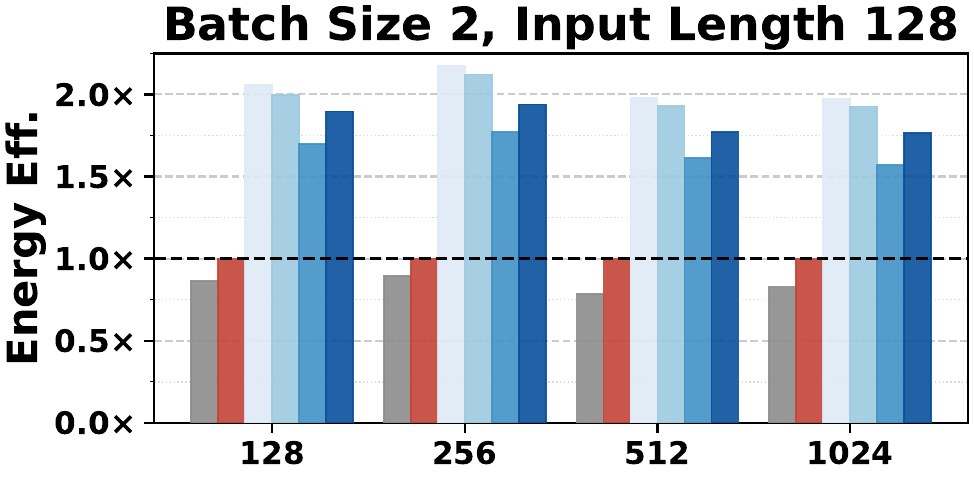}
    \hfill
    \includegraphics[width=0.234\linewidth]{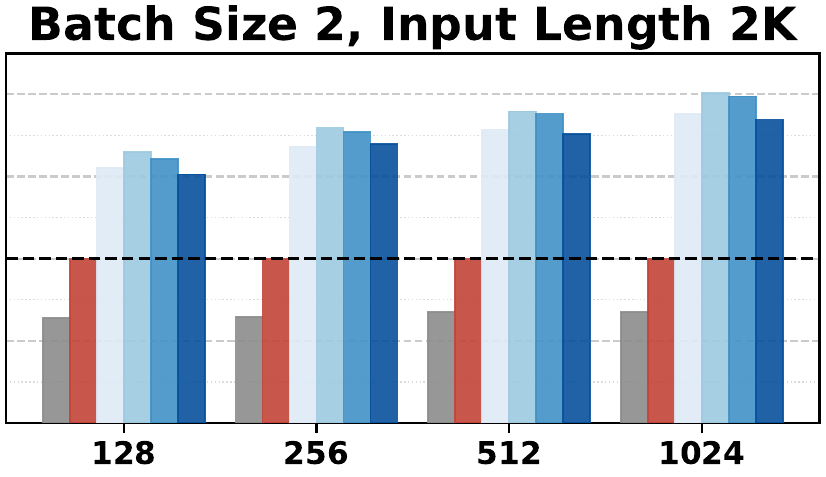}
    \hfill
    \includegraphics[width=0.234\linewidth]{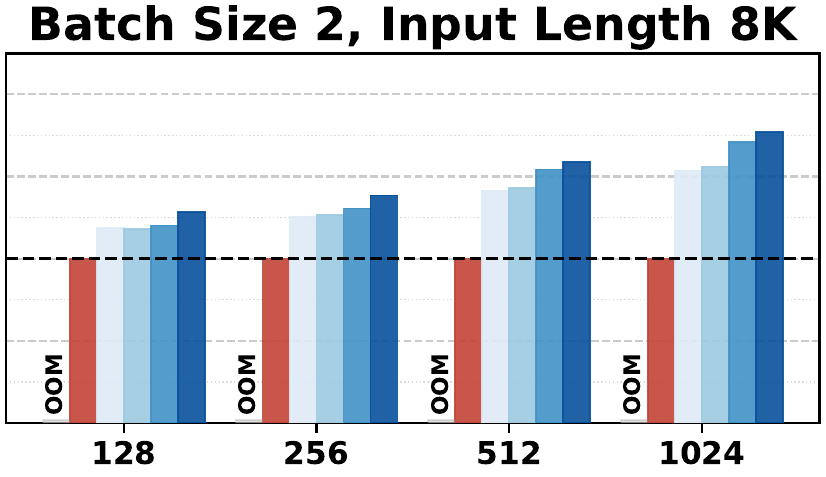}
    \hfill
    \includegraphics[width=0.234\linewidth]{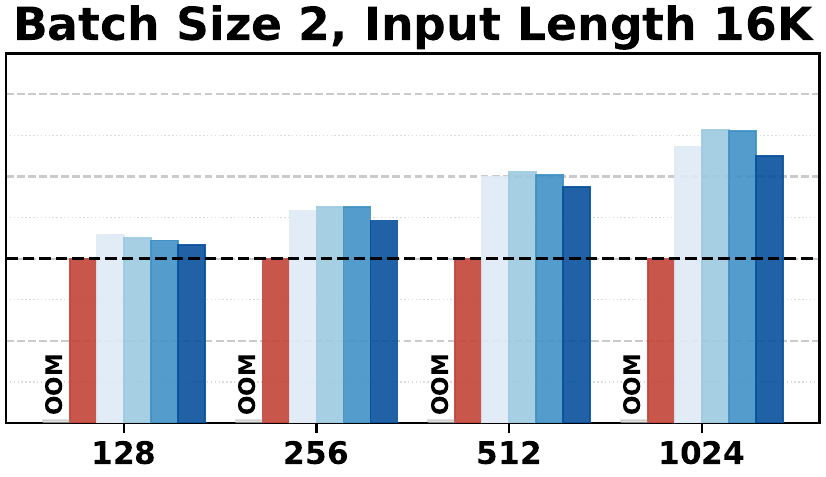}

    \includegraphics[width=0.274\linewidth]{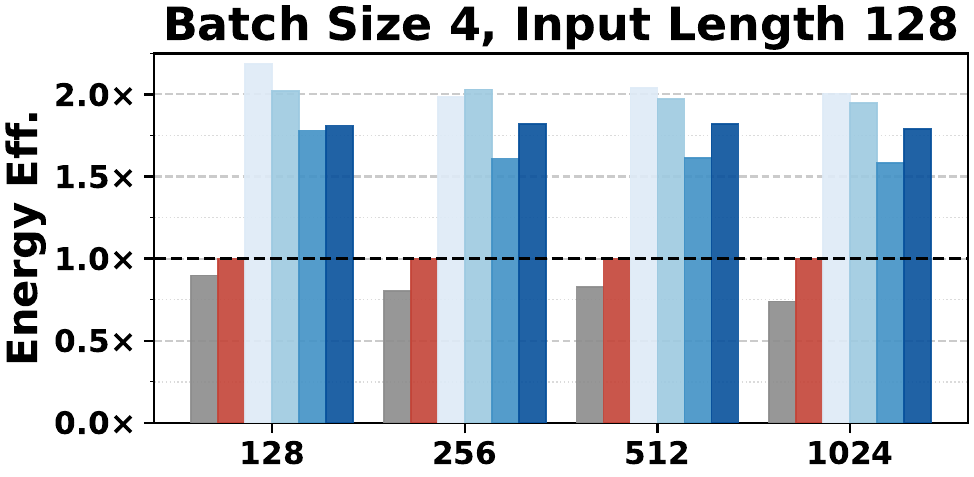}
    \hfill
    \includegraphics[width=0.234\linewidth]{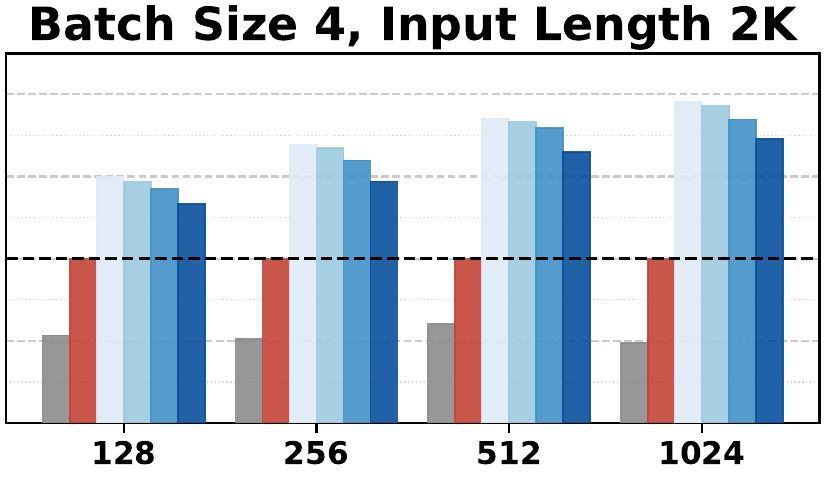}
    \hfill
    \includegraphics[width=0.234\linewidth]{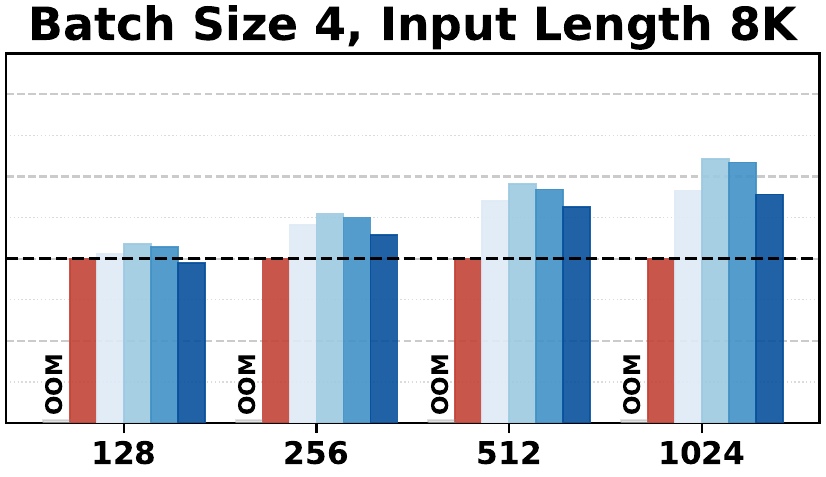}
    \hfill
    \includegraphics[width=0.234\linewidth]{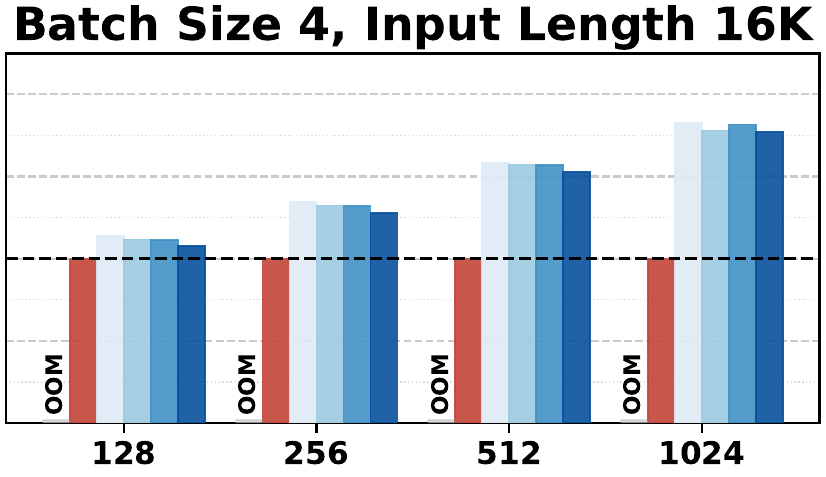}

    \includegraphics[width=0.274\linewidth]{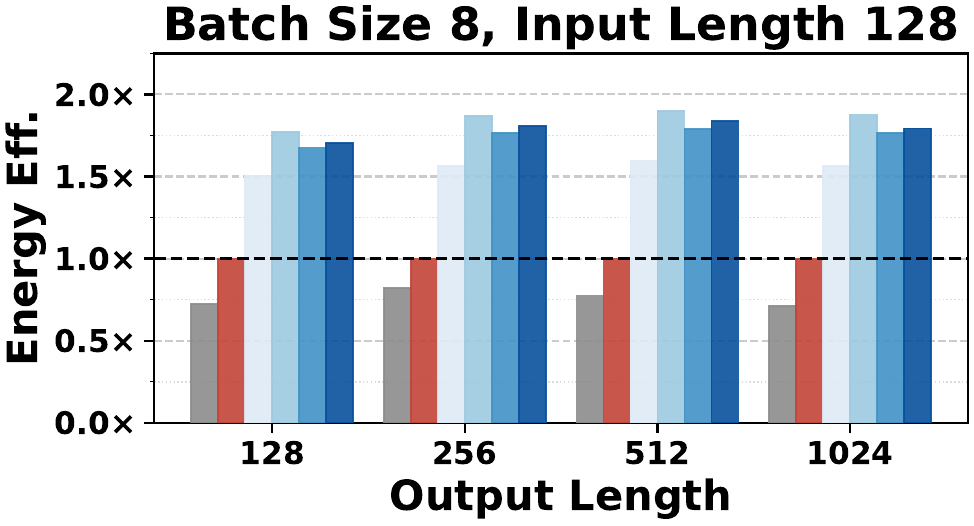}
    \hfill
    \includegraphics[width=0.234\linewidth]{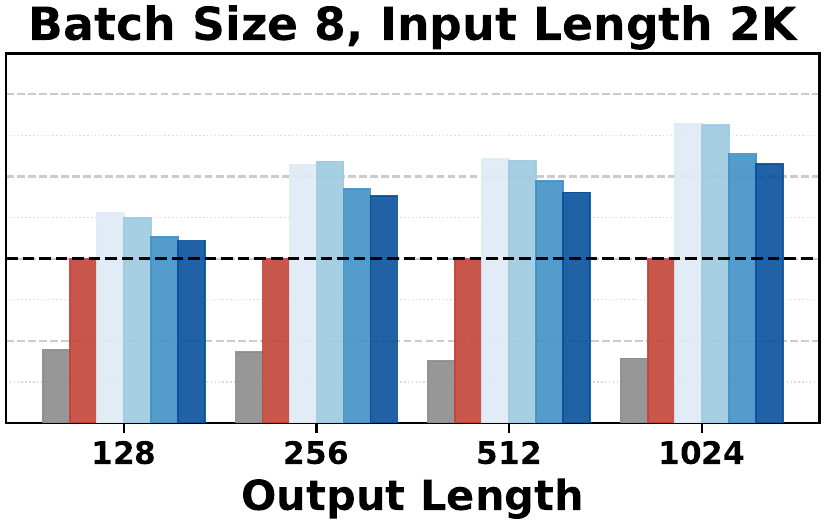}
    \hfill
    \includegraphics[width=0.234\linewidth]{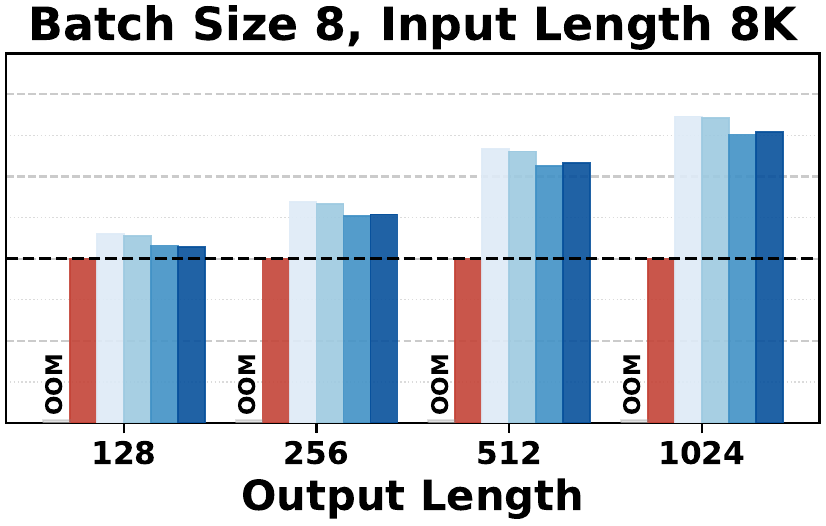}
    \hfill
    \includegraphics[width=0.234\linewidth]{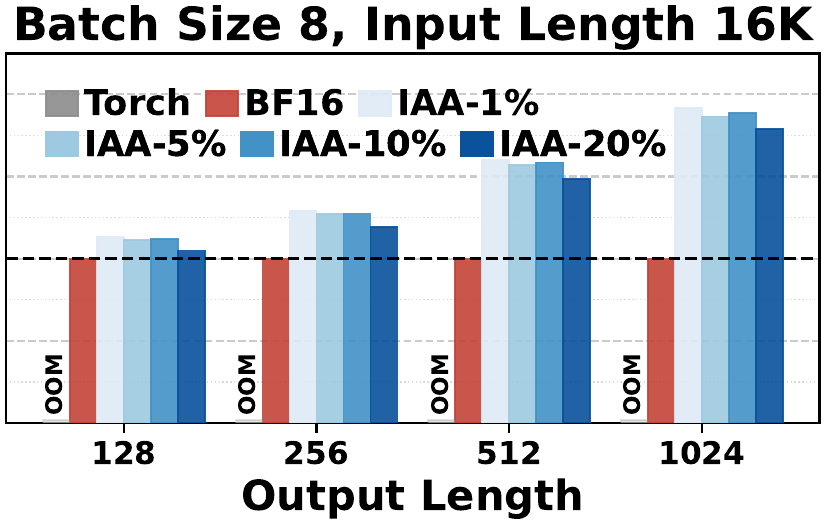}

    \caption{
    Generation-energy comparison for \textbf{input-activation-aware (IAA)} \modelname variants on Llama-3.2-3B.
    Rows vary batch size over 1, 2, 4, and 8; columns vary prompt length over 128, 2K, 8K, and 16K tokens.
    Each panel compares IAA retention configurations across output lengths; OOM denotes an out-of-memory run.
    }
    \label{fig:energy_iaa}
    
\end{figure*}

\begin{figure*}[t]
    \centering

    \includegraphics[width=0.274\linewidth]{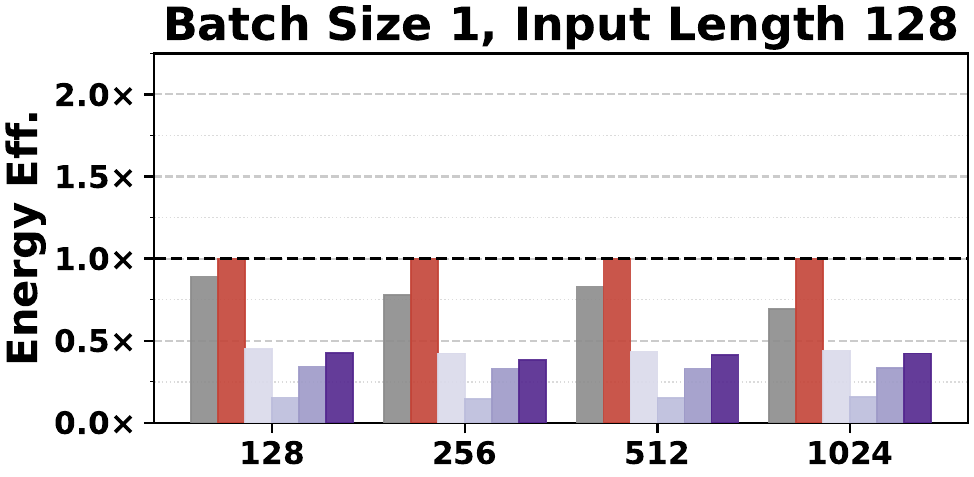}
    \hfill
    \includegraphics[width=0.234\linewidth]{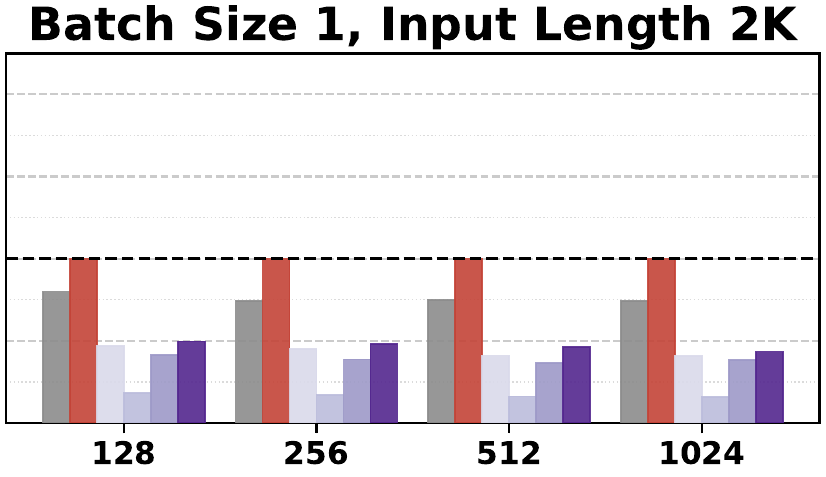}
    \hfill
    \includegraphics[width=0.234\linewidth]{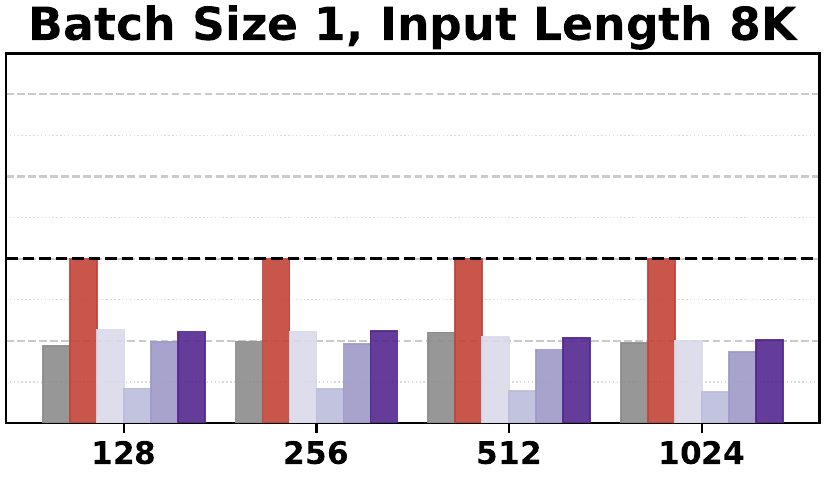}
    \hfill
    \includegraphics[width=0.234\linewidth]{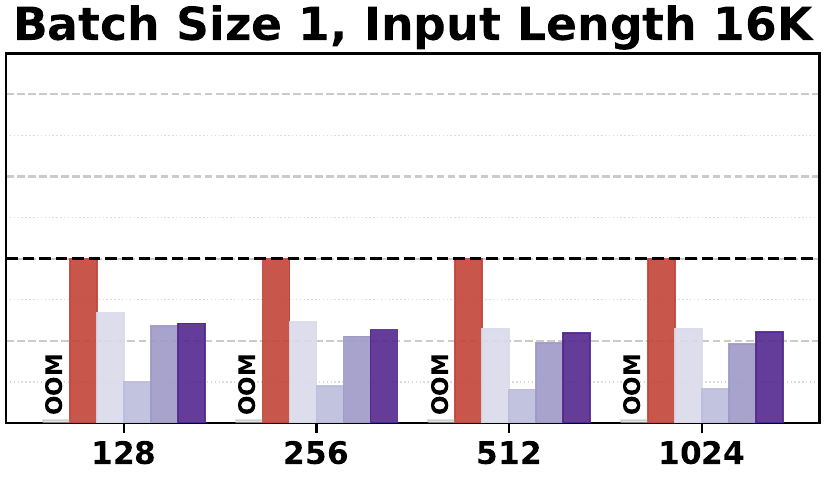}

    \includegraphics[width=0.274\linewidth]{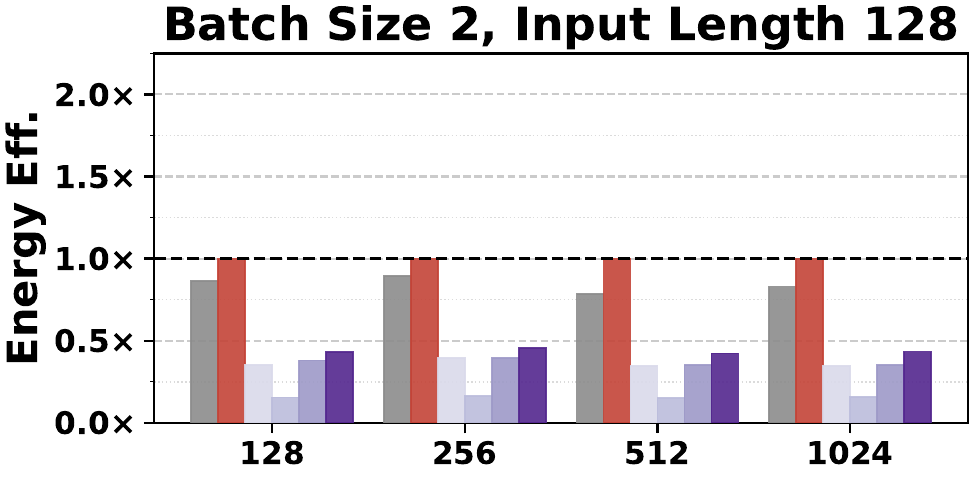}
    \hfill
    \includegraphics[width=0.234\linewidth]{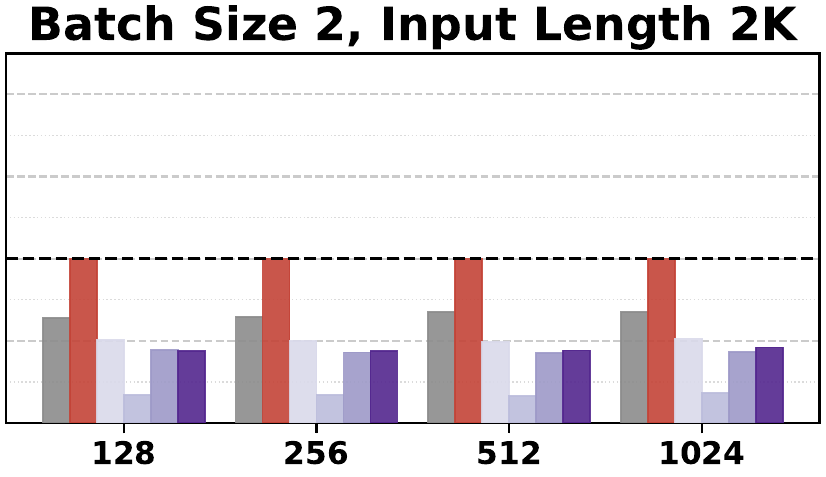}
    \hfill
    \includegraphics[width=0.234\linewidth]{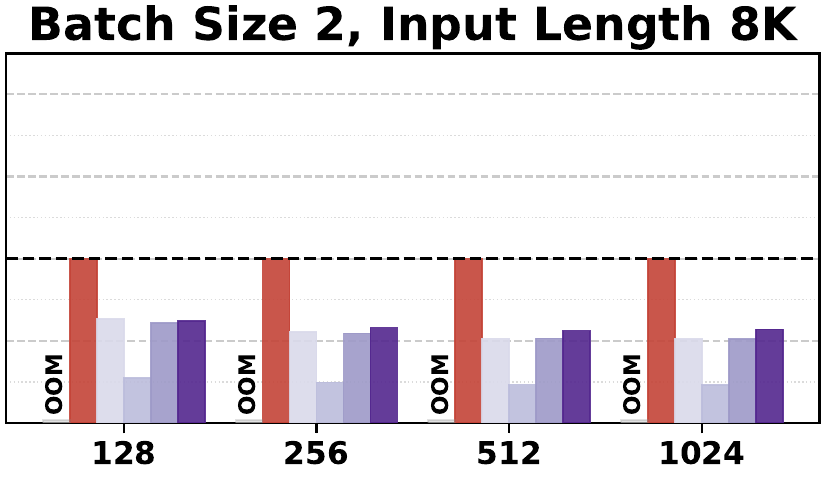}
    \hfill
    \includegraphics[width=0.234\linewidth]{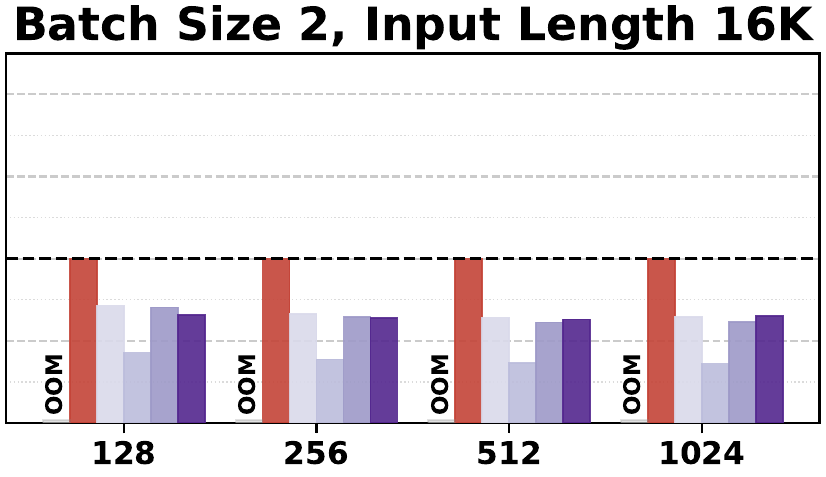}

    \includegraphics[width=0.274\linewidth]{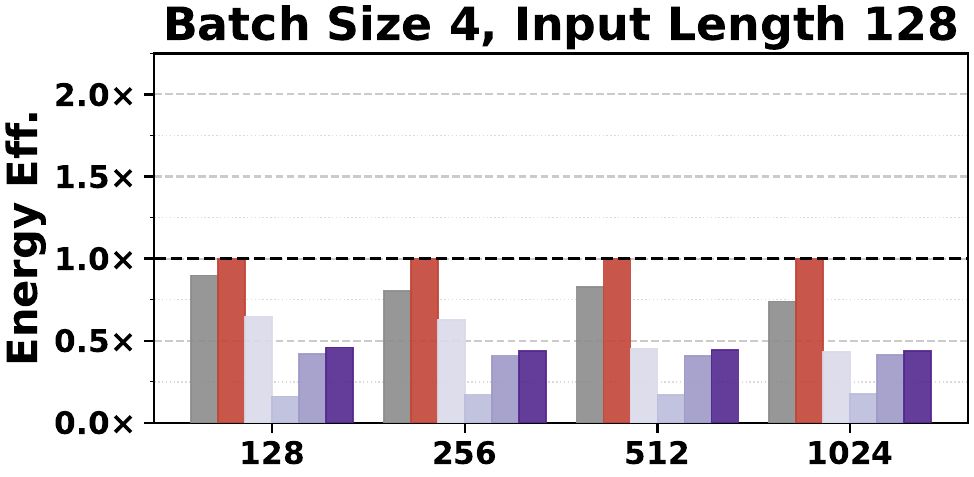}
    \hfill
    \includegraphics[width=0.234\linewidth]{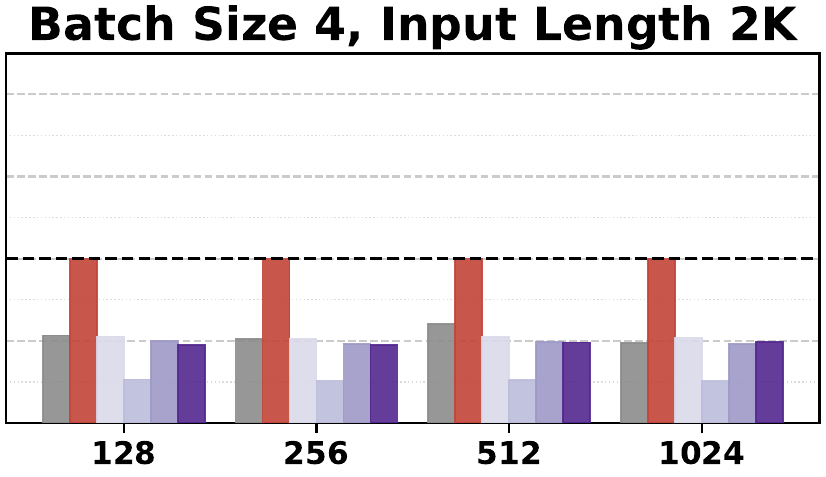}
    \hfill
    \includegraphics[width=0.234\linewidth]{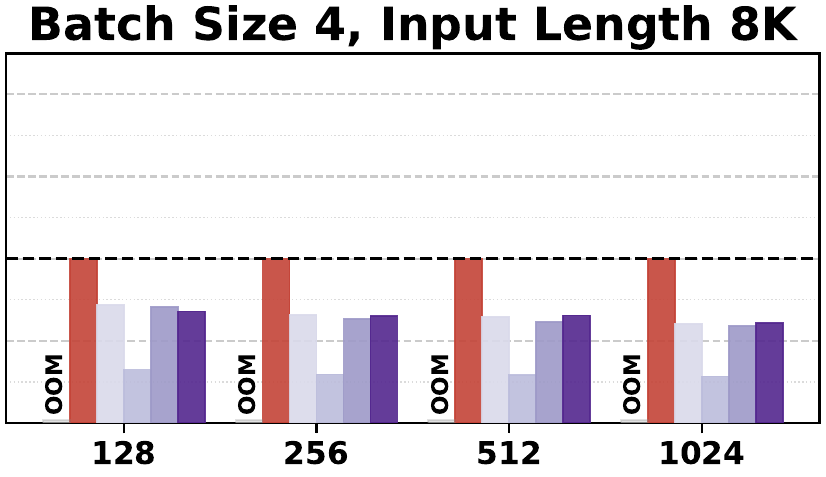}
    \hfill
    \includegraphics[width=0.234\linewidth]{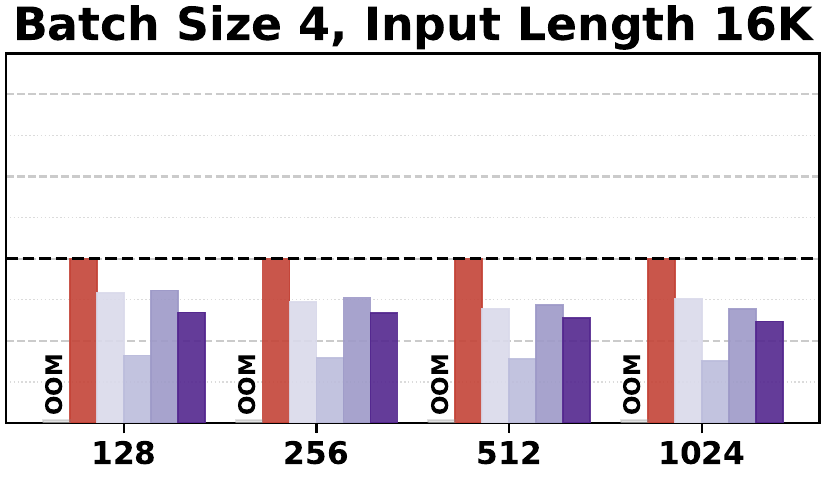}

    \includegraphics[width=0.274\linewidth]{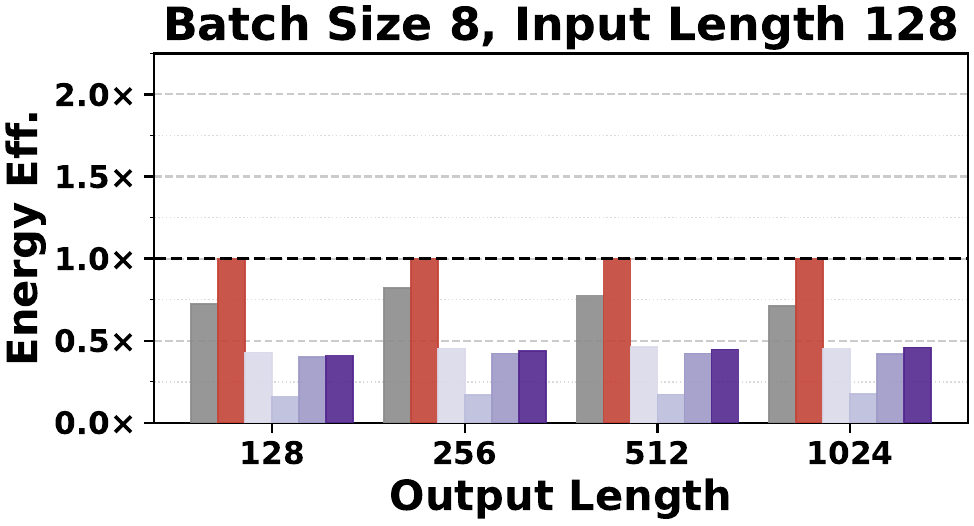}
    \hfill
    \includegraphics[width=0.234\linewidth]{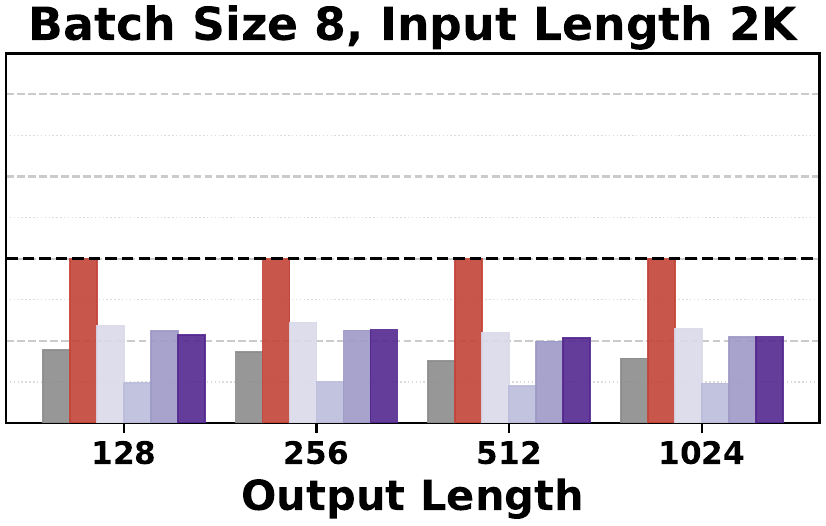}
    \hfill
    \includegraphics[width=0.234\linewidth]{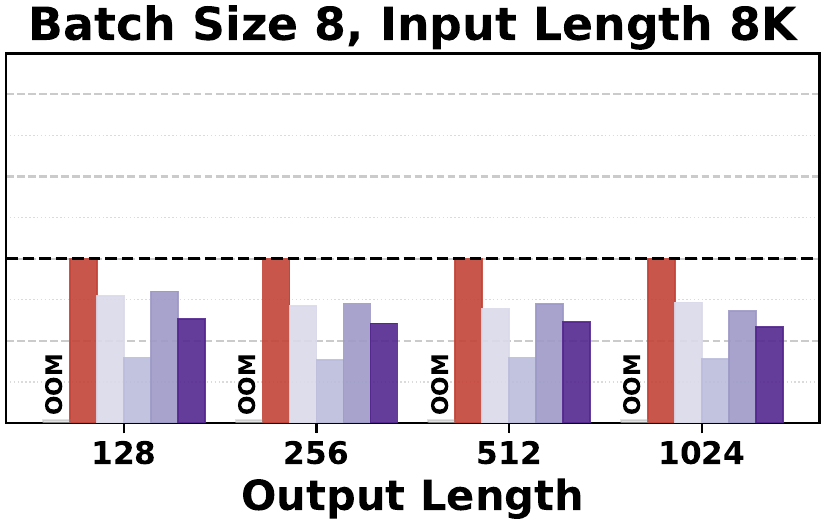}
    \hfill
    \includegraphics[width=0.234\linewidth]{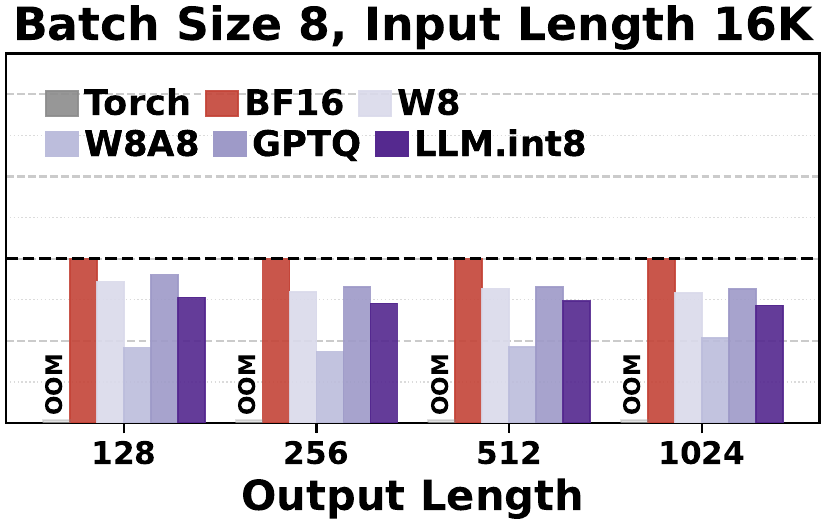}

    \caption{
    Generation-energy comparison for \textbf{baseline inference methods} on Llama-3.2-3B.
    Rows vary batch size over 1, 2, 4, and 8; columns vary prompt length over 128, 2K, 8K, and 16K tokens.
    Each panel compares the baseline configurations across output lengths; OOM denotes an out-of-memory run.
    }
    \label{fig:energy_baselines}
    
\end{figure*}

\begin{table}[!t]
\centering
\footnotesize
\setlength{\tabcolsep}{3pt}
\renewcommand{\arraystretch}{0.95}
\caption{Prefill peak VRAM (GB) across prompt lengths for Llama-3.2-3B, Qwen2.5-7B, and Qwen2.5-14B. GI denotes gradient-informed tile retention; subscripts give the target 16-bit tile-retention percentage.}
\label{tab:prefill_peak_vram_all_models}
\begin{adjustbox}{width=\columnwidth}
\begin{tabular}{@{}c l c c c c@{}}
\toprule
\textbf{Model} & \textbf{Len} & \textbf{GI$_1$} & \textbf{GI$_5$} & \textbf{GI$_{10}$} & \textbf{GI$_{20}$} \\
\midrule
\multirow{5}{*}{\rotatebox{90}{\resizebox{45pt}{!}{\textbf{Llama-3.2-3B}}}} & 1K & 3.68 & 3.91 & 3.96 & 4.31 \\
 & 4K & 4.64 & 4.82 & 4.87 & 5.09 \\
 & 8K & 5.26 & 5.44 & 5.50 & 5.71 \\
 & 16K & 6.89 & 7.07 & 7.12 & 7.34 \\
 & 32K & 10.64 & 10.82 & 10.87 & 11.09 \\
\midrule
\multirow{5}{*}{\rotatebox{90}{\resizebox{45pt}{!}{\textbf{Qwen2.5-7B}}}} & 1K & 8.56 & 8.82 & 9.26 & 9.72 \\
 & 4K & 9.68 & 9.94 & 10.38 & 10.84 \\
 & 8K & 10.12 & 10.37 & 10.82 & 11.27 \\
 & 16K & 12.43 & 12.69 & 13.13 & 13.58 \\
 & 32K & 15.90 & 16.15 & 16.60 & 17.05 \\
\midrule
\multirow{5}{*}{\rotatebox{90}{\resizebox{45pt}{!}{\textbf{Qwen2.5-14B}}}} & 1K & 16.12 & 16.74 & 17.35 & 18.29 \\
 & 4K & 17.65 & 18.27 & 18.88 & 19.82 \\
 & 8K & 18.49 & 19.32 & 19.72 & 20.66 \\
 & 16K & 21.65 & 22.48 & 22.88 & 23.82 \\
 & 32K & 27.96 & 27.95 & 29.19 & 30.13 \\
\bottomrule
\end{tabular}
\end{adjustbox}
\end{table}

\begin{table*}[t]
\centering
\small
\setlength{\tabcolsep}{4pt}
\renewcommand{\arraystretch}{0.55}

\caption{
GPU memory comparison on the Llama-3.2-3B model.
We compare the original model, representative PTQ baselines, and tile-level mixed-precision configurations across retained floating-point tile ratios.
GI and IAA denote Gradient-informed and Input-activation-aware tile-retention policies, with subscripts indicating the retained floating-point tile ratio in percent.
Values are in GB. Prefill peak VRAM is measured at a prompt length of 32K.
}
\label{tab:memory_llama3b_full}

\begin{adjustbox}{width=\textwidth}
\begin{tabular}{l c @{\vsep} c c c c c c c c @{\vsep} c c @{\vsep} c c}
\toprule
\multirow{2}{*}{\textbf{Metric}} &
\multirow{2}{*}{\textbf{Original}} &
\multicolumn{8}{c}{\textbf{Ours}} &
\multicolumn{2}{c}{\textbf{Uniform}} &
\multirow{2}{*}{\textbf{GPTQ}} &
\multirow{2}{*}{\textbf{LLM.int8}} \\
\cmidrule(lr){3-10}\cmidrule(lr){11-12}
& &
\textbf{GI$_1$} & \textbf{IAA$_1$} &
\textbf{GI$_5$} & \textbf{IAA$_5$} &
\textbf{GI$_{10}$} & \textbf{IAA$_{10}$} &
\textbf{GI$_{20}$} & \textbf{IAA$_{20}$} &
\textbf{W8} & \textbf{W8A8} &
& \\
\midrule
Storage
& 5.98
& 3.39 & 3.39
& 3.50 & 3.50
& 3.63 & 3.63
& 3.89 & 3.89
& 3.36 & 3.36
& 3.42 & 3.36 \\

Post-load VRAM
& 5.99
& 3.51 & 3.51
& 3.64 & 3.64
& 3.69 & 3.69
& 4.02 & 4.02
& 5.99 & 5.99
& 3.55 & 3.46 \\

Peak VRAM
& 20.15
& 16.14 & 16.14
& 16.32 & 16.32
& 16.38 & 16.38
& 16.59 & 16.59
& 16.75 & 21.00
& 17.59 & 19.15 \\

Prefill Peak VRAM
& 12.64
& 10.64 & 10.64
& 10.82 & 10.82
& 10.87 & 10.87
& 11.09 & 11.09
& 10.24 & 12.24
& 10.09 & 11.40 \\

\bottomrule
\end{tabular}
\end{adjustbox}

\end{table*}

\begin{table*}[t]
\centering
\small
\setlength{\tabcolsep}{4pt}
\renewcommand{\arraystretch}{0.55}

\caption{
GPU memory comparison on the Qwen2.5-7B model.
We compare the original model, representative PTQ baselines, and tile-level mixed-precision configurations across retained floating-point tile ratios.
GI and IAA denote Gradient-informed and Input-activation-aware tile-retention policies, with subscripts indicating the retained floating-point tile ratio in percent.
Values are in GB. Prefill peak VRAM is measured at a prompt length of 32K.
}
\label{tab:memory_qwen25_7b}

\begin{adjustbox}{width=\textwidth}
\begin{tabular}{l c @{\vsep} c c c c c c c c @{\vsep} c c @{\vsep} c c}
\toprule
\multirow{2}{*}{\textbf{Metric}} &
\multirow{2}{*}{\textbf{Original}} &
\multicolumn{8}{c}{\textbf{Ours}} &
\multicolumn{2}{c}{\textbf{Uniform}} &
\multirow{2}{*}{\textbf{GPTQ}} &
\multirow{2}{*}{\textbf{LLM.int8}} \\
\cmidrule(lr){3-10}\cmidrule(lr){11-12}
& &
\textbf{GI$_1$} & \textbf{IAA$_1$} &
\textbf{GI$_5$} & \textbf{IAA$_5$} &
\textbf{GI$_{10}$} & \textbf{IAA$_{10}$} &
\textbf{GI$_{20}$} & \textbf{IAA$_{20}$} &
\textbf{W8} & \textbf{W8A8} &
& \\
\midrule
Storage
& 14.19
& 8.18 & 8.18
& 8.42 & 8.42
& 8.73 & 8.73
& 9.33 & 9.33
& 8.11 & 8.11
& 8.25 & 8.11 \\

Post-load VRAM
& 14.19
& 8.25 & 8.25
& 8.52 & 8.52
& 8.97 & 8.97
& 9.40 & 9.40
& 13.20 & 13.20
& 8.35 & 8.44 \\

Peak VRAM
& 28.27
& 22.83 & 22.83
& 23.09 & 23.09
& 23.54 & 23.54
& 23.99 & 23.99
& 24.82 & 38.70
& 22.34 & 26.86 \\

Prefill Peak VRAM
& 21.33
& 15.90 & 15.90
& 16.15 & 16.15
& 16.60 & 16.60
& 17.05 & 17.05
& 17.88 & 24.82
& 15.40 & 17.61 \\

\bottomrule
\end{tabular}
\end{adjustbox}

\end{table*}

\begin{table*}[t]
\centering
\small
\setlength{\tabcolsep}{4pt}
\renewcommand{\arraystretch}{0.55}

\caption{
GPU memory comparison on the Qwen2.5-14B model.
We compare the original model, representative PTQ baselines, and tile-level mixed-precision configurations across retained floating-point tile ratios.
GI and IAA denote Gradient-informed and Input-activation-aware tile-retention policies, with subscripts indicating the retained floating-point tile ratio in percent.
Values are in GB. Prefill peak VRAM is measured at a prompt length of 32K.
}
\label{tab:memory_qwen25_14b}

\begin{adjustbox}{width=\textwidth}
\begin{tabular}{l c @{\vsep} c c c c c c c c @{\vsep} c c @{\vsep} c c}
\toprule
\multirow{2}{*}{\textbf{Metric}} &
\multirow{2}{*}{\textbf{Original}} &
\multicolumn{8}{c}{\textbf{Ours}} &
\multicolumn{2}{c}{\textbf{Uniform}} &
\multirow{2}{*}{\textbf{GPTQ}} &
\multirow{2}{*}{\textbf{LLM.int8}} \\
\cmidrule(lr){3-10}\cmidrule(lr){11-12}
& &
\textbf{GI$_1$} & \textbf{IAA$_1$} &
\textbf{GI$_5$} & \textbf{IAA$_5$} &
\textbf{GI$_{10}$} & \textbf{IAA$_{10}$} &
\textbf{GI$_{20}$} & \textbf{IAA$_{20}$} &
\textbf{W8} & \textbf{W8A8} &
& \\
\midrule
Storage
& 27.51
& 15.35 & 15.35
& 15.85 & 15.85
& 16.46 & 16.46
& 17.69 & 17.69
& 15.21 & 15.21
& 15.50 & 15.21 \\

Post-load VRAM
& 27.51
& 15.67 & 15.67
& 16.26 & 16.26
& 16.90 & 16.90
& 17.85 & 17.85
& 27.52 & 27.52
& 15.75 & 15.95 \\

Peak VRAM
& 40.10
& 27.96 & 27.99
& 27.95 & 27.95
& 29.19 & 29.19
& 30.13 & 30.13
& 38.47 & 34.28
& 27.68 & 28.92 \\

Prefill Peak VRAM
& 40.10
& 27.96 & 27.99
& 27.95 & 27.95
& 29.19 & 29.19
& 30.13 & 30.13
& 29.22 & 34.28
& 27.68 & 28.92 \\

\bottomrule
\end{tabular}
\end{adjustbox}

\end{table*}

\begin{table*}[t]
\centering
\scriptsize
\setlength{\tabcolsep}{2.0pt}
\renewcommand{\arraystretch}{0.75}

\caption{
General benchmark results for \textbf{Llama-3.2-3B} across language modeling and downstream reasoning tasks.
GI and IAA denote Gradient-informed and Input-activation-aware tile-retention policies, with subscripts indicating the retained floating-point tile ratio in percent.
Arrows next to benchmark names indicate the preferred metric direction: $\downarrow$ means lower is better and $\uparrow$ means higher is better.
Accuracy values are reported as percentages, while WikiText-2 is reported with word perplexity.
Bold values indicate the best result in each row.
Cell colors are relative to the row-wise Original model under the corresponding metric direction: green for better than or equal to Original, blue for worse than Original, and gray for the Original column.
}
\label{tab:general_llama3b_full}

\begin{adjustbox}{width=\textwidth}
\begin{tabular}{l c c @{\vsep} c c c c c c c c @{\vsep} c c @{\vsep} c c c}
\toprule
\multirow{2}{*}{\textbf{Benchmark}} &
\multirow{2}{*}{\textbf{Metric}} &
\multirow{2}{*}{\textbf{Original}} &
\multicolumn{8}{c}{\textbf{Ours}} &
\multicolumn{2}{c}{\textbf{Uniform}} &
\multirow{2}{*}{\rotatebox{25}{\textbf{GPTQ}}} &
\multirow{2}{*}{\rotatebox{25}{\textbf{SmoothQ}}} &
\multirow{2}{*}{\rotatebox{25}{\textbf{LLM.int8}}} \\
\cmidrule(lr){4-11}\cmidrule(lr){12-13}
& & &
\textbf{GI$_1$} & \textbf{IAA$_1$} &
\textbf{GI$_5$} & \textbf{IAA$_5$} &
\textbf{GI$_{10}$} & \textbf{IAA$_{10}$} &
\textbf{GI$_{20}$} & \textbf{IAA$_{20}$} &
\textbf{W8} & \textbf{W8A8} &
& & \\
\midrule

WikiText-2 $\downarrow$
& PPL \setbase{12.487} & \cellcolor{heatneutral}\base
& \pcell{12.518} & \pcell{12.537}
& \pcell{12.473} & \pcell{12.531}
& \pcell{12.456} & \pcell{12.494}
& {\bfseries\pcell{12.403}} & \pcell{12.438}
& \pcell{12.453} & \pcell{12.630}
& \pcell{12.515} & \pcell{12.609}
& \pcell{12.639} \\

WinoGrande $\uparrow$ & Acc. \setbase{71.43} & \cellcolor{heatneutral}\base & \gcell{71.38} & \gcell{70.98} & {\bfseries\gcell{72.01}} & \gcell{70.67} & \gcell{70.98} & \gcell{71.30} & \gcell{71.46} & \gcell{71.30} & \gcell{70.88} & \gcell{70.64} & \gcell{71.43} & \gcell{71.59} & \gcell{71.11} \\

PIQA $\uparrow$
& Acc. \setbase{77.31} & \cellcolor{heatneutral}\base
& \gcell{76.61} & \gcell{77.09}
& \gcell{77.04} & \gcell{76.61}
& \gcell{76.93} & \gcell{76.61}
& {\bfseries\gcell{77.42}} & \gcell{77.15}
& \gcell{76.61} & \gcell{76.55}
& \gcell{76.82} & \gcell{76.55}
& \gcell{76.82} \\

LAMBADA $\uparrow$
& Acc. \setbase{64.23} & \cellcolor{heatneutral}\base
& \gcell{65.62} & \gcell{65.07}
& \gcell{65.49} & \gcell{65.58}
& \gcell{65.64} & \gcell{65.66}
& {\bfseries\gcell{65.78}} & \gcell{65.49}
& \gcell{64.41} & \gcell{63.96}
& \gcell{64.43} & \gcell{63.98}
& \gcell{64.31} \\

ARC-Challenge $\uparrow$
& Acc. \setbase{52.56} & \cellcolor{heatneutral}\base
& \gcell{52.82} & \gcell{52.30}
& \gcell{52.82} & \gcell{52.47}
& \gcell{53.56} & \gcell{52.71}
& {\bfseries\gcell{53.69}} & \gcell{52.88}
& \gcell{52.30} & \gcell{51.79}
& \gcell{52.73} & \gcell{51.71}
& \gcell{51.88} \\

HellaSwag $\uparrow$
& Acc. \setbase{73.65} & \cellcolor{heatneutral}\textbf{\base}
& \gcell{73.25} & \gcell{73.09}
& \gcell{73.22} & \gcell{72.72}
& \gcell{73.10} & \gcell{72.50}
& \gcell{72.95} & \gcell{72.50}
& \gcell{73.59} & \gcell{73.53}
& \gcell{73.55} & \gcell{73.47}
& \gcell{73.48} \\

MMLU $\uparrow$
& Acc. \setbase{60.47} & \cellcolor{heatneutral}\base
& \gcell{60.64} & \gcell{60.78}
& \gcell{60.90} & \gcell{60.74}
& \gcell{60.89} & \gcell{60.71}
& {\bfseries\gcell{61.10}} & \gcell{60.54}
& \gcell{60.51} & \gcell{60.80}
& \gcell{60.46} & \gcell{60.48}
& \gcell{60.89} \\

\bottomrule
\end{tabular}
\end{adjustbox}
\end{table*}

\begin{table*}[t]

\centering
\scriptsize
\setlength{\tabcolsep}{2.0pt}
\renewcommand{\arraystretch}{0.75}

\caption{
General benchmark results for \textbf{Qwen2.5-7B} across language modeling and downstream reasoning tasks.
GI and IAA denote Gradient-informed and Input-activation-aware tile-retention policies, with subscripts indicating the retained floating-point tile ratio in percent.
Arrows next to benchmark names indicate the preferred metric direction: $\downarrow$ means lower is better and $\uparrow$ means higher is better.
Accuracy values are reported as percentages, while WikiText-2 is reported with word perplexity.
Bold values indicate the best result in each row.
Cell colors are relative to the row-wise Original model under the corresponding metric direction: green for better than or equal to Original, blue for worse than Original, and gray for the Original column.
}
\label{tab:general_qwen7b_full}

\begin{adjustbox}{width=\textwidth}
\begin{tabular}{l c c @{\vsep} c c c c c c c c @{\vsep} c c @{\vsep} c c c}
\toprule
\multirow{2}{*}{\textbf{Benchmark}} &
\multirow{2}{*}{\textbf{Metric}} &
\multirow{2}{*}{\textbf{Original}} &
\multicolumn{8}{c}{\textbf{Ours}} &
\multicolumn{2}{c}{\textbf{Uniform}} &
\multirow{2}{*}{\rotatebox{25}{\textbf{GPTQ}}} &
\multirow{2}{*}{\rotatebox{25}{\textbf{SmoothQ}}} &
\multirow{2}{*}{\rotatebox{25}{\textbf{LLM.int8}}} \\
\cmidrule(lr){4-11}\cmidrule(lr){12-13}
& & &
\textbf{GI$_1$} & \textbf{IAA$_1$} &
\textbf{GI$_5$} & \textbf{IAA$_5$} &
\textbf{GI$_{10}$} & \textbf{IAA$_{10}$} &
\textbf{GI$_{20}$} & \textbf{IAA$_{20}$} &
\textbf{W8} & \textbf{W8A8} &
& & \\
\midrule

WikiText-2 $\downarrow$
& PPL \setbase{9.407} & \cellcolor{heatneutral}\base
& \pcell{9.542} & \pcell{9.576}
& \pcell{9.466} & \pcell{9.519}
& \pcell{9.434} & \pcell{9.502}
& {\bfseries\pcell{9.353}} & \pcell{9.408}
& \pcell{9.420} & \pcell{9.719}
& \pcell{9.413} & \pcell{9.753}
& \pcell{9.464} \\

WinoGrande $\uparrow$
& Acc. \setbase{75.53} & \cellcolor{heatneutral}\base
& \gcell{75.40} & \gcell{75.51}
& \gcell{75.17} & \gcell{75.14}
& \gcell{75.24} & \gcell{74.80}
& \gcell{75.32} & {\bfseries\gcell{75.80}}
& \gcell{75.30} & \gcell{74.59}
& \gcell{75.22} & \gcell{74.66}
& \gcell{75.37} \\

PIQA $\uparrow$
& Acc. \setbase{80.58} & \cellcolor{heatneutral}\base
& \gcell{80.44} & \gcell{80.32}
& \gcell{80.33} & \gcell{80.88}
& \gcell{80.55} & \gcell{81.01}
& {\bfseries\gcell{81.15}} & \gcell{80.66}
& \gcell{80.58} & \gcell{79.05}
& \gcell{80.30} & \gcell{78.84}
& \gcell{80.20} \\

LAMBADA $\uparrow$
& Acc. \setbase{69.49} & \cellcolor{heatneutral}\base
& \gcell{69.14} & \gcell{68.99}
& \gcell{69.37} & \gcell{69.11}
& \gcell{69.56} & \gcell{69.07}
& {\bfseries\gcell{69.89}} & \gcell{69.22}
& \gcell{69.47} & \gcell{67.30}
& \gcell{69.38} & \gcell{64.41}
& \gcell{69.71} \\

ARC-Challenge $\uparrow$
& Acc. \setbase{67.06} & \cellcolor{heatneutral}\base
& \gcell{66.90} & \gcell{66.47}
& \gcell{67.33} & \gcell{66.81}
& \gcell{67.33} & {\bfseries\gcell{67.75}}
& \gcell{67.58} & \gcell{67.58}
& \gcell{66.89} & \gcell{67.15}
& \gcell{66.89} & \gcell{65.96}
& \gcell{67.06} \\

HellaSwag $\uparrow$
& Acc. \setbase{81.39} & \cellcolor{heatneutral}\base
& \gcell{81.10} & \gcell{81.24}
& \gcell{81.49} & \gcell{81.24}
& \gcell{81.35} & \gcell{81.30}
& \gcell{81.15} & {\bfseries\gcell{81.54}}
& \gcell{81.26} & \gcell{80.85}
& \gcell{81.40} & \gcell{80.38}
& \gcell{81.33} \\

MMLU $\uparrow$
& Acc. \setbase{74.23} & \cellcolor{heatneutral}\base
& \gcell{75.37} & \gcell{75.09}
& \gcell{75.26} & \gcell{75.19}
& {\bfseries\gcell{75.55}} & \gcell{75.22}
& \gcell{75.39} & \gcell{75.51}
& \gcell{75.22} & \gcell{74.73}
& \gcell{74.23} & \gcell{73.56}
& \gcell{75.29} \\

\bottomrule
\end{tabular}
\end{adjustbox}

\end{table*}

\begin{table*}[t]

\centering
\scriptsize
\setlength{\tabcolsep}{2.0pt}
\renewcommand{\arraystretch}{0.75}

\caption{
General benchmark results for \textbf{Qwen2.5-14B} across language modeling and downstream reasoning tasks.
GI and IAA denote Gradient-informed and Input-activation-aware tile-retention policies, with subscripts indicating the retained floating-point tile ratio in percent.
Arrows next to benchmark names indicate the preferred metric direction: $\downarrow$ means lower is better and $\uparrow$ means higher is better.
Accuracy values are reported as percentages, while WikiText-2 is reported with word perplexity.
Bold values indicate the best result in each row.
Cell colors are relative to the row-wise Original model under the corresponding metric direction: green for better than or equal to Original, blue for worse than Original, and gray for the Original column.
}
\label{tab:general_qwen14b_full}

\begin{adjustbox}{width=\textwidth}
\begin{tabular}{l c c @{\vsep} c c c c c c c c @{\vsep} c c @{\vsep} c c c}
\toprule
\multirow{2}{*}{\textbf{Benchmark}} &
\multirow{2}{*}{\textbf{Metric}} &
\multirow{2}{*}{\textbf{Original}} &
\multicolumn{8}{c}{\textbf{Ours}} &
\multicolumn{2}{c}{\textbf{Uniform}} &
\multirow{2}{*}{\rotatebox{25}{\textbf{GPTQ}}} &
\multirow{2}{*}{\rotatebox{25}{\textbf{SmoothQ}}} &
\multirow{2}{*}{\rotatebox{25}{\textbf{LLM.int8}}} \\
\cmidrule(lr){4-11}\cmidrule(lr){12-13}
& & &
\textbf{GI$_1$} & \textbf{IAA$_1$} &
\textbf{GI$_5$} & \textbf{IAA$_5$} &
\textbf{GI$_{10}$} & \textbf{IAA$_{10}$} &
\textbf{GI$_{20}$} & \textbf{IAA$_{20}$} &
\textbf{W8} & \textbf{W8A8} &
& & \\
\midrule

WikiText-2 $\downarrow$
& PPL \setbase{6.661} & \cellcolor{heatneutral}\base
& \pcell{6.776} & \pcell{6.774}
& \pcell{6.725} & \pcell{6.740}
& \pcell{6.685} & \pcell{6.695}
& {\bfseries\pcell{6.611}} & \pcell{6.637}
& \pcell{6.667} & \pcell{7.721}
& \pcell{6.661} & \pcell{6.928}
& \pcell{6.751} \\

WinoGrande $\uparrow$
& Acc. \setbase{79.16} & \cellcolor{heatneutral}\base
& \gcell{79.28} & \gcell{79.80}
& \gcell{80.45} & \gcell{79.73}
& \gcell{79.71} & \gcell{79.80}
& \gcell{80.42} & {\bfseries\gcell{80.67}}
& \gcell{79.48} & \gcell{80.58}
& \gcell{79.48} & \gcell{79.48}
& \gcell{79.16} \\

PIQA $\uparrow$
& Acc. \setbase{81.77} & \cellcolor{heatneutral}\base
& \gcell{81.83} & \gcell{81.54}
& \gcell{81.59} & \gcell{81.31}
& \gcell{81.48} & \gcell{81.35}
& {\bfseries\gcell{81.97}} & \gcell{81.68}
& \gcell{81.83} & \gcell{81.83}
& \gcell{81.66} & \gcell{81.34}
& \gcell{81.56} \\

LAMBADA $\uparrow$
& Acc. \setbase{73.10} & \cellcolor{heatneutral}\textbf{\base}
& \gcell{72.76} & \gcell{72.75}
& \gcell{72.93} & \gcell{72.79}
& \gcell{72.97} & \gcell{72.94}
& \gcell{73.06} & \gcell{72.79}
& \gcell{72.46} & \gcell{72.07}
& \gcell{72.95} & \gcell{72.21}
& \gcell{72.99} \\

ARC-Challenge $\uparrow$
& Acc. \setbase{72.70} & \cellcolor{heatneutral}\textbf{\base}
& \gcell{71.16} & \gcell{71.42}
& \gcell{71.59} & \gcell{71.33}
& \gcell{71.76} & \gcell{71.50}
& \gcell{71.25} & \gcell{70.99}
& \gcell{72.35} & \gcell{71.67}
& \gcell{72.61} & \gcell{71.16}
& \gcell{71.93} \\

HellaSwag $\uparrow$
& Acc. \setbase{85.07} & \cellcolor{heatneutral}\base
& \gcell{84.94} & \gcell{84.78}
& \gcell{85.05} & \gcell{84.42}
& \gcell{86.38} & \gcell{85.19}
& {\bfseries\gcell{86.49}} & \gcell{86.06}
& \gcell{85.28} & \gcell{85.06}
& \gcell{85.03} & \gcell{84.80}
& \gcell{85.24} \\

MMLU $\uparrow$
& Acc. \setbase{79.77} & \cellcolor{heatneutral}\base
& \gcell{80.39} & \gcell{80.67}
& \gcell{80.57} & \gcell{80.76}
& {\bfseries\gcell{80.92}} & \gcell{80.66}
& \gcell{80.81} & \gcell{80.81}
& \gcell{80.56} & \gcell{80.46}
& \gcell{80.40} & \gcell{80.42}
& \gcell{80.51} \\

\bottomrule
\end{tabular}
\end{adjustbox}
\end{table*}

\begin{figure*}[t]
    \centering

    \begin{subfigure}[b]{0.49\linewidth}
        \centering
        \includegraphics[width=\linewidth]{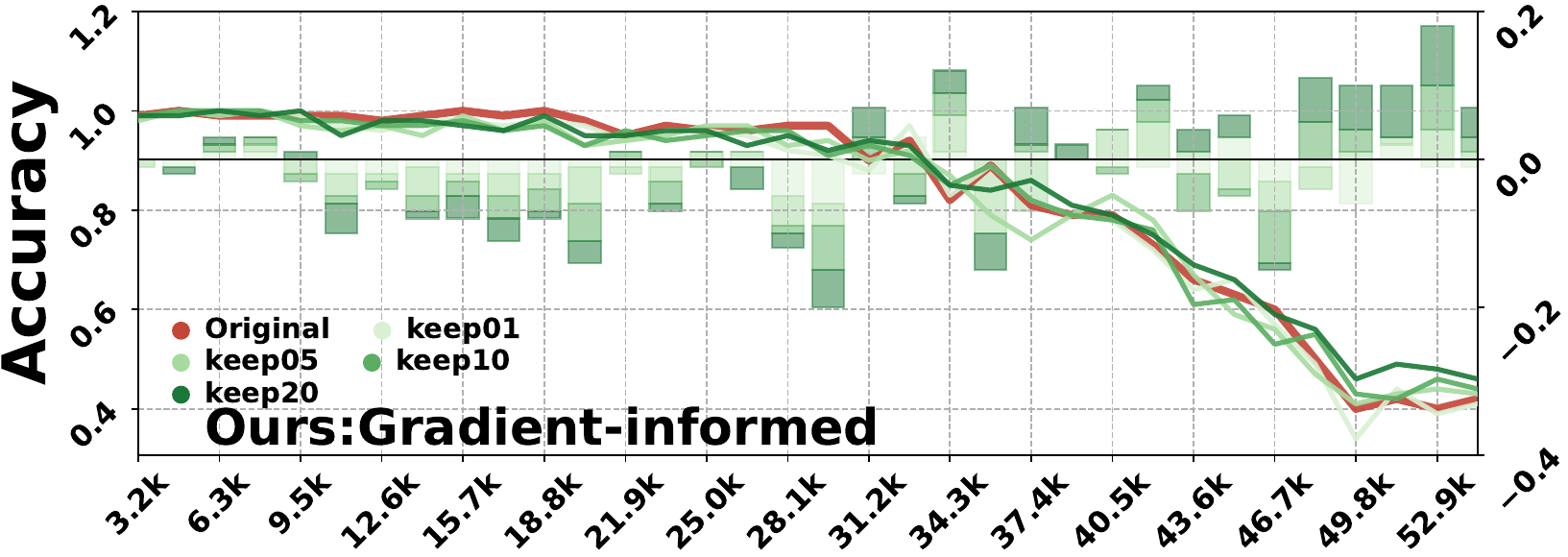}
    \end{subfigure}
    \hfill
    \begin{subfigure}[b]{0.495\linewidth}
        \centering
        \includegraphics[width=\linewidth]{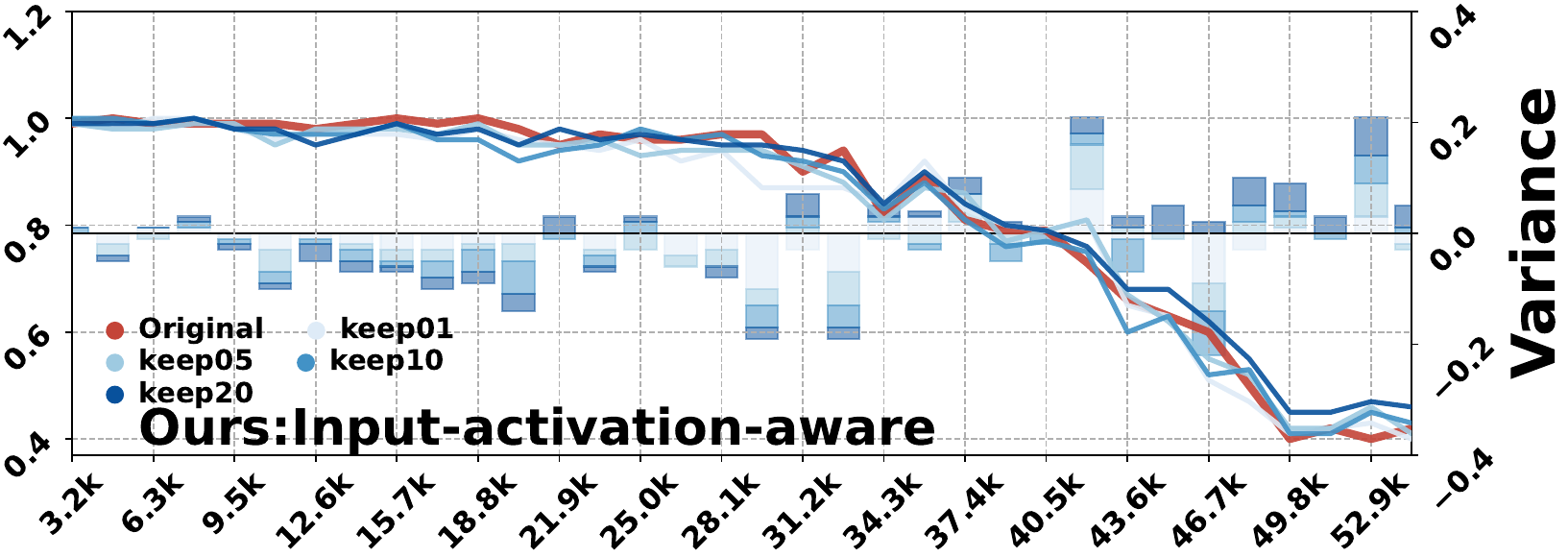}
    \end{subfigure}

    \begin{subfigure}[b]{0.495\linewidth}
        \centering
        \includegraphics[width=\linewidth]{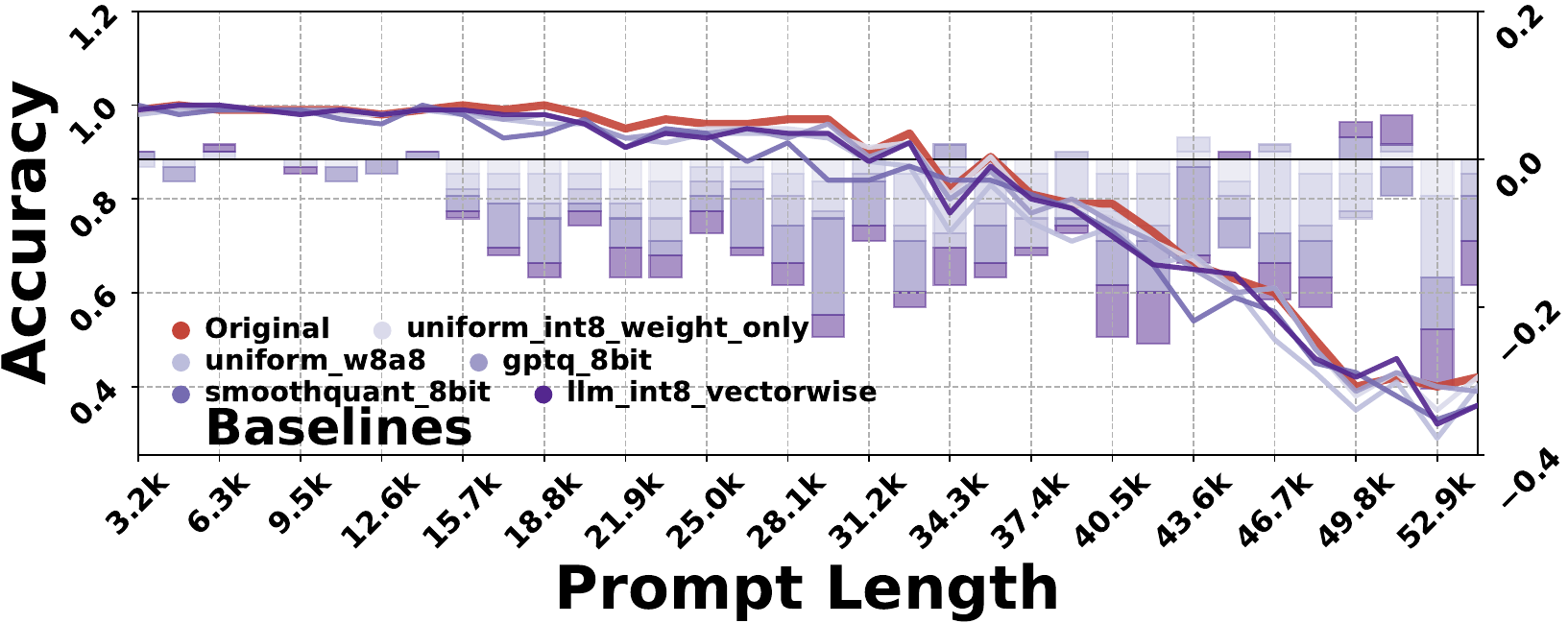}
    \end{subfigure}
    \hfill
    \begin{subfigure}[b]{0.49\linewidth}
        \centering
        \includegraphics[width=\linewidth]{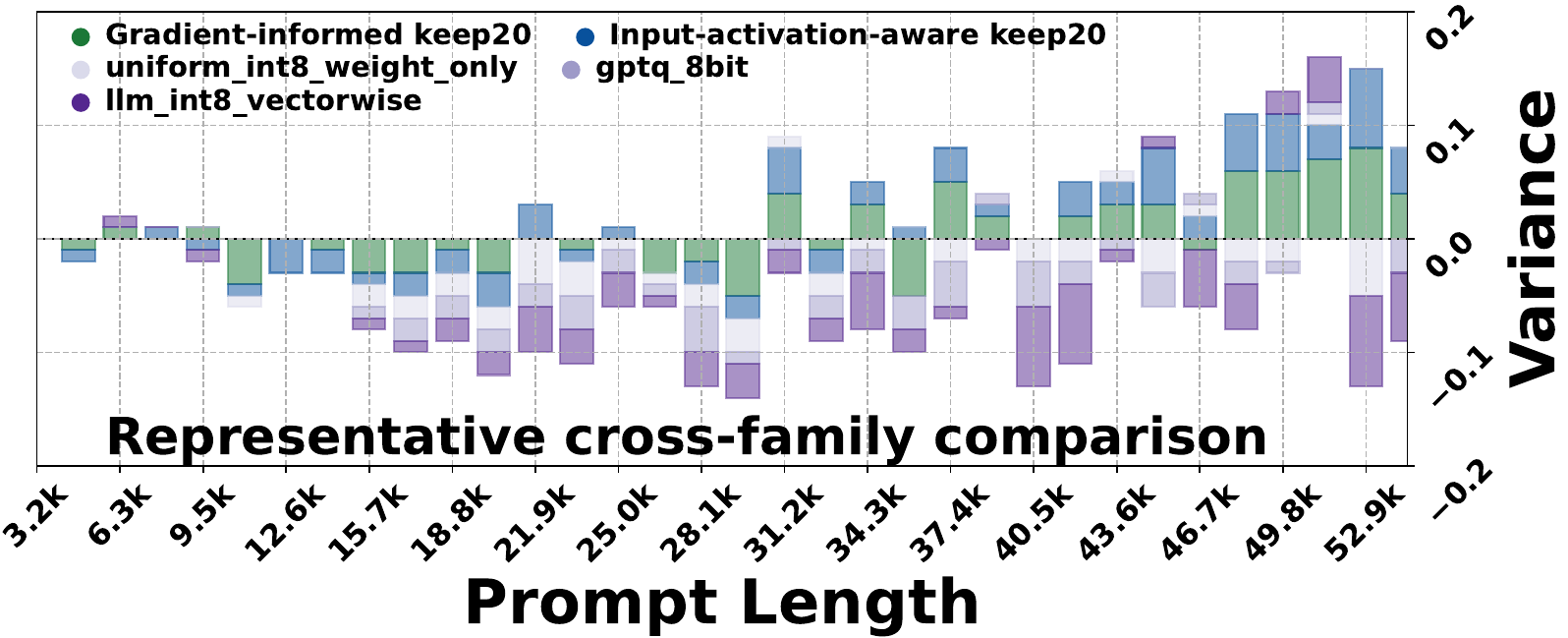}
    \end{subfigure}
    
    \caption{
    Line-level retrieval for \textbf{Qwen2.5-7B} across prompt lengths from 3.2K to 52.9K tokens.
    \textbf{Line plots (left y-axis)} report exact-match accuracy, while \textbf{bar plots (right y-axis)} show accuracy differences relative to the original baseline.
    Top-left: \modelname with gradient-informed maps, where keep01, keep05, keep10, and keep20 retain 1\%, 5\%, 10\%, and 20\% of tiles in the deployment floating-point dtype.
    Top-right: \modelname with input-activation-aware maps under the same keep ratios.
    Bottom-left: baseline PTQ methods, including uniform INT8 weight-only quantization, uniform W8A8 quantization, GPTQ, SmoothQuant, and LLM.int8.
    Bottom-right: representative cross-family comparison between the strongest \modelname configurations and strongest baseline methods.
    }
    
    \label{fig:longeval_qwen7b}
\end{figure*}

\begin{figure*}[t]
    \centering

    \begin{subfigure}[b]{0.49\linewidth}
        \centering
        \includegraphics[width=\linewidth]{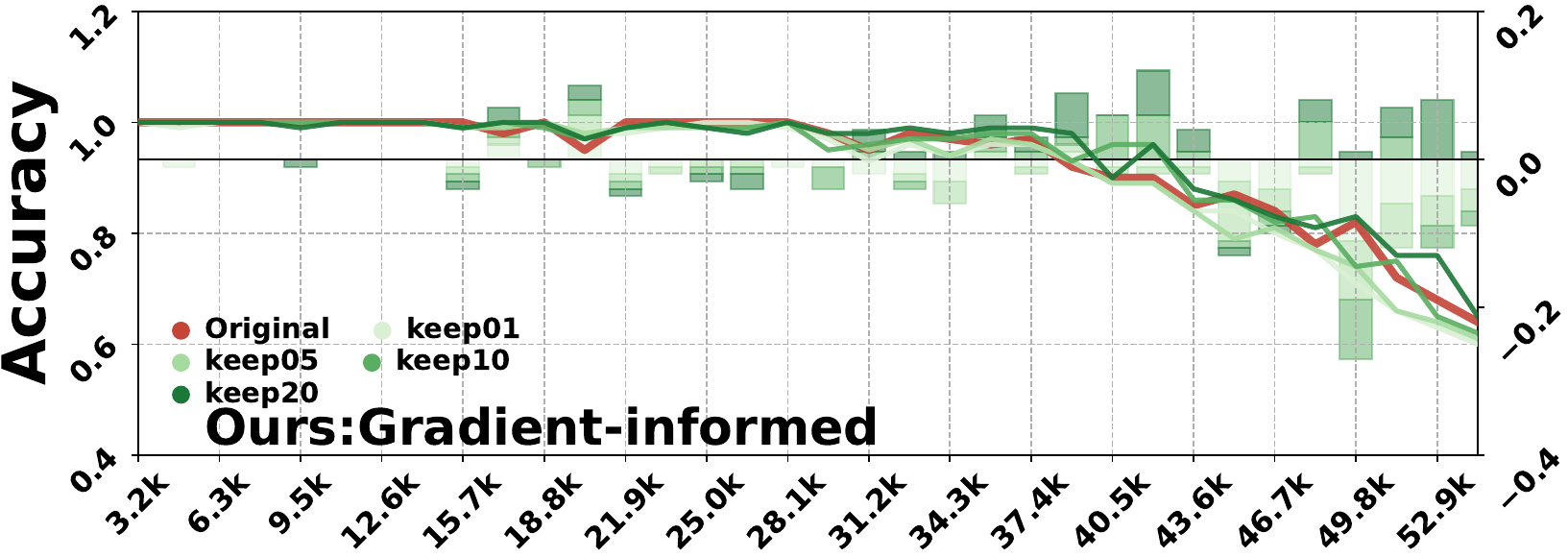}
    \end{subfigure}
    \hfill
    \begin{subfigure}[b]{0.495\linewidth}
        \centering
        \includegraphics[width=\linewidth]{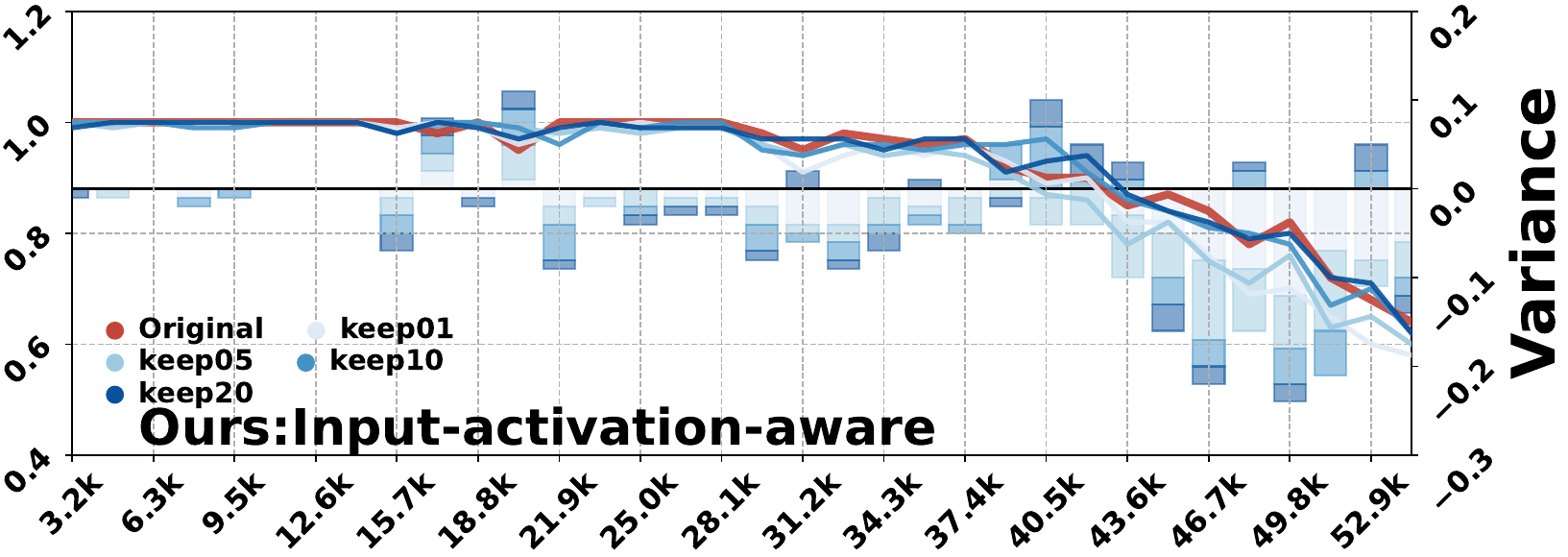}
    \end{subfigure}

    \begin{subfigure}[b]{0.495\linewidth}
        \centering
        \includegraphics[width=\linewidth]{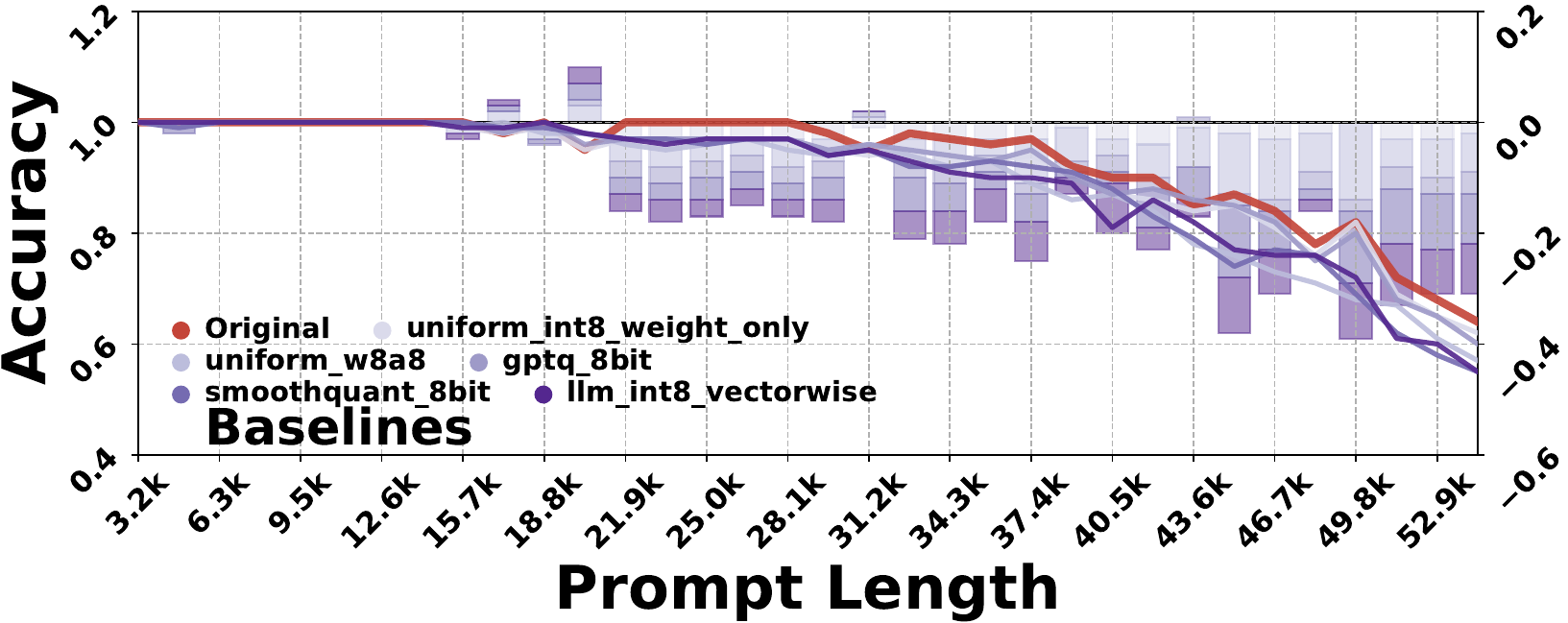}
    \end{subfigure}
    \hfill
    \begin{subfigure}[b]{0.49\linewidth}
        \centering
        \includegraphics[width=\linewidth]{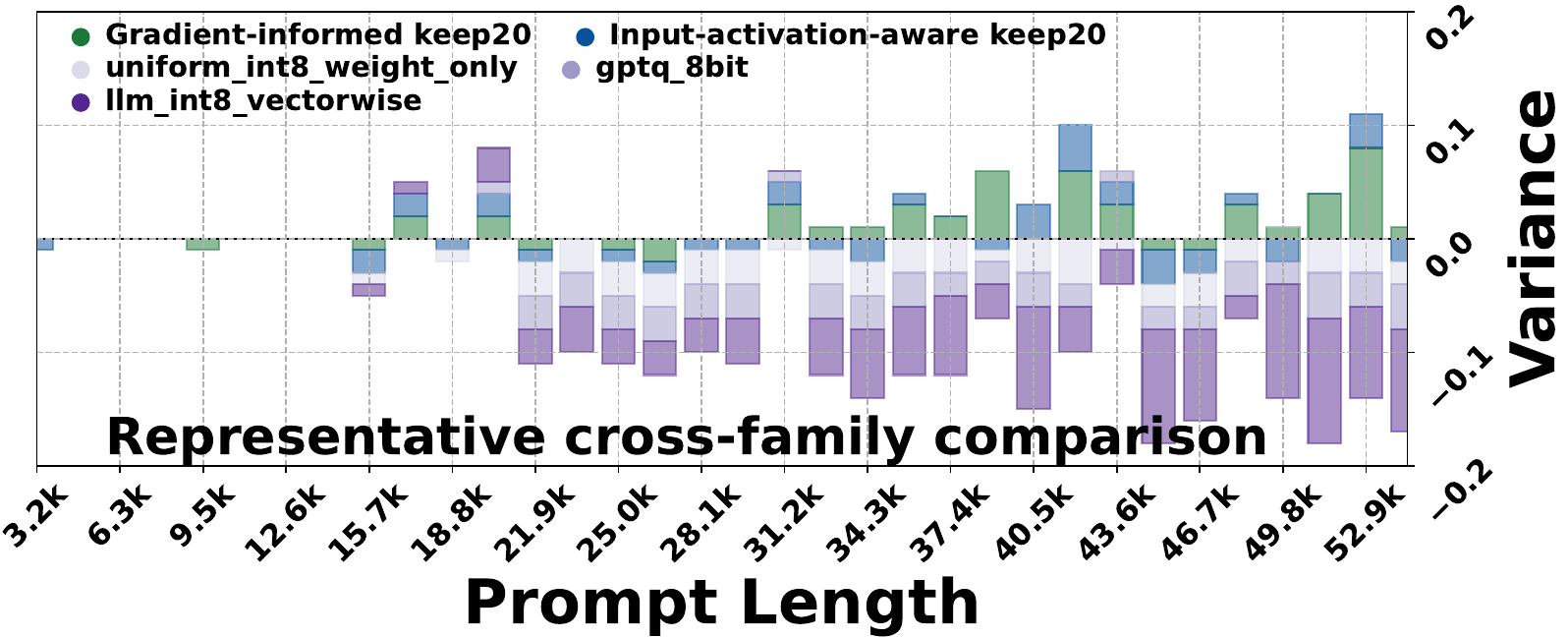}
    \end{subfigure}

    \caption{
    Line-level retrieval accuracy for \textbf{Qwen2.5-14B} on LongEval across prompt lengths from 3.2K to 52.9K tokens. Line plots (left y-axis) report exact-match retrieval accuracy, while bar plots (right y-axis) show accuracy differences relative to the Original floating-point model.
    }
    
    \label{fig:longeval_qwen14b}
\end{figure*}

\begin{table*}[t]
\centering
\scriptsize
\setlength{\tabcolsep}{1.8pt}
\renewcommand{\arraystretch}{0.9}
\caption{
Full long-context question answering results for \textbf{Llama-3.2-3B} on LV-Eval across 9 datasets and context lengths from 16K to 256K.
GI and IAA denote Gradient-informed and Input-activation-aware tile-retention policies, with subscripts indicating the retained floating-point tile ratio in percent.
We follow the LV-Eval metric protocol: keyword-recall F1 by default, blacklist-based F1 for \textit{cmrc\_mixup}, and blacklist-based ROUGE-L for \textit{dureader\_mixup}.
Cell colors are relative to the row-wise Original reference: green and blue indicate higher and lower scores, respectively, while gray marks Original and values within the neutral comparison range.
}
\label{tab:lveval_llama3b_full}

\begin{adjustbox}{width=\textwidth}
\begin{tabular}{c c c @{\vsep} c c c c c c c c @{\vsep} c c @{\vsep} c c @{\vsep} c}
\toprule
\multirow{2}{*}{\textbf{Dataset}} &
\multirow{2}{*}{\textbf{Len}} &
\multirow{2}{*}{\textbf{Original}} &
\multicolumn{8}{c}{\textbf{Ours}} &
\multicolumn{2}{c}{\textbf{Uniform}} &
\multirow{2}{*}{\rotatebox{25}{\textbf{GPTQ}}} &
\multirow{2}{*}{\rotatebox{25}{\textbf{SmoothQ}}} &
\multirow{2}{*}{\rotatebox{25}{\textbf{LLM.int8}}} \\
\cmidrule(lr){4-11}\cmidrule(lr){12-13}
& & &
\textbf{GI$_{1}$} & \textbf{IAA$_{1}$} &
\textbf{GI$_{5}$} & \textbf{IAA$_{5}$} &
\textbf{GI$_{10}$} & \textbf{IAA$_{10}$} &
\textbf{GI$_{20}$} & \textbf{IAA$_{20}$} &
\textbf{W8} & \textbf{W8A8} &
& & \\
\midrule

\multirow{5}{*}{\ds{1}}
& 16k \setbase{21.41} & \cellcolor{heatneutral}\base
& \tcell{23.82} & {\bfseries\tcell{24.21}}
& \tcell{24.01} & \tcell{23.82}
& \tcell{23.92} & \tcell{23.76}
& \tcell{24.06} & \tcell{23.87}
& \tcell{21.15} & \tcell{21.39}
& \tcell{21.53} & \tcell{20.15}
& \tcell{20} \\
& 32k \setbase{16.86} & \cellcolor{heatneutral}\base
& \tcell{18.16} & \tcell{18.23}
& \tcell{18.34} & \tcell{18.19}
& \tcell{18.26} & \tcell{18.13}
& {\bfseries\tcell{18.38}} & \tcell{18.23}
& \tcell{17.95} & \tcell{17.05}
& \tcell{17.39} & \tcell{16.06}
& \tcell{16.31} \\
& 64k \setbase{7.5} & \cellcolor{heatneutral}\base
& \tcell{8.48} & \tcell{8}
& \tcell{8.58} & \tcell{8.03}
& \tcell{8.54} & \tcell{8.02}
& {\bfseries\tcell{8.59}} & \tcell{8.04}
& \tcell{7.46} & \tcell{6.93}
& \tcell{7.58} & \tcell{7.24}
& \tcell{6.47} \\
& 128k \setbase{3.08} & \cellcolor{heatneutral}\base
& \tcell{4.11} & \tcell{3.63}
& {\bfseries\tcell{4.16}} & \tcell{3.64}
& \tcell{4.14} & \tcell{3.64}
& {\bfseries\tcell{4.16}} & \tcell{3.65}
& \tcell{3.23} & \tcell{3.99}
& \tcell{3.16} & \tcell{3.57}
& \tcell{3.58} \\
& 256k \setbase{2.37} & \cellcolor{heatneutral}\base
& \tcell{2.68} & \tcell{2.52}
& \tcell{2.72} & \tcell{2.53}
& \tcell{2.71} & \tcell{2.53}
& {\bfseries\tcell{2.73}} & \tcell{2.54}
& \tcell{2.43} & \tcell{2.61}
& \tcell{2.38} & \tcell{2.43}
& \tcell{2.47} \\
\midrule

\multirow{5}{*}{\ds{2}}
& 16k \setbase{16.46} & \cellcolor{heatneutral}\base
& \tcell{16.45} & \tcell{17.37}
& \tcell{16.6} & \tcell{16.46}
& \tcell{16.54} & \tcell{16.42}
& \tcell{16.64} & \tcell{16.5}
& \tcell{16.54} & \tcell{17.33}
& \tcell{16.07} & \tcell{17.47}
& {\bfseries\tcell{17.88}} \\
& 32k \setbase{17.1} & {\bfseries\cellcolor{heatneutral}\base}
& \tcell{16.48} & \tcell{16.88}
& \tcell{16.35} & \tcell{17.02}
& \tcell{16.29} & \tcell{16.17}
& \tcell{16.39} & \tcell{17.03}
& \tcell{16.6} & \tcell{16.88}
& \tcell{15.83} & \tcell{16.64}
& \tcell{16.37} \\
& 64k \setbase{15.52} & \cellcolor{heatneutral}\base
& \tcell{15.4} & \tcell{14.72}
& \tcell{15.72} & \tcell{15.46}
& \tcell{15.64} & \tcell{15.43}
& {\bfseries\tcell{15.76}} & \tcell{15.47}
& \tcell{14.55} & \tcell{14.97}
& \tcell{15.04} & \tcell{14.77}
& \tcell{14.84} \\
& 128k \setbase{11.1} & \cellcolor{heatneutral}\base
& \tcell{11} & \tcell{11.4}
& \tcell{11.27} & \tcell{11.13}
& \tcell{11.2} & \tcell{11.08}
& \tcell{11.3} & \tcell{11.15}
& {\bfseries\tcell{11.6}} & \tcell{11.04}
& \tcell{11.15} & \tcell{11.26}
& \tcell{10.92} \\
& 256k \setbase{7.43} & \cellcolor{heatneutral}\base
& \tcell{6.66} & \tcell{7.07}
& \tcell{7.56} & \tcell{7.41}
& \tcell{7.51} & \tcell{7.4}
& {\bfseries\tcell{7.58}} & \tcell{7.42}
& \tcell{6.87} & \tcell{6.6}
& \tcell{6.88} & \tcell{6.61}
& \tcell{6.6} \\
\midrule

\multirow{5}{*}{\ds{3}}
& 16k \setbase{24.49} & \cellcolor{heatneutral}\base
& \tcell{25.29} & \tcell{24.22}
& \tcell{25.51} & \tcell{24.36}
& \tcell{25.41} & \tcell{24.33}
& {\bfseries\tcell{25.56}} & \tcell{24.37}
& \tcell{24.33} & \tcell{20.71}
& \tcell{24.74} & \tcell{23.15}
& \tcell{23.52} \\
& 32k \setbase{19.23} & \cellcolor{heatneutral}\base
& \tcell{18.43} & \tcell{17.28}
& \tcell{19.43} & \tcell{19.14}
& \tcell{19.35} & \tcell{19.11}
& \tcell{19.48} & \tcell{19.15}
& \tcell{17.82} & \tcell{19.12}
& \tcell{18.35} & {\bfseries\tcell{20.74}}
& \tcell{17.7} \\
& 64k \setbase{15.86} & \cellcolor{heatneutral}\base
& \tcell{13.61} & {\bfseries\tcell{17.02}}
& \tcell{16.06} & \tcell{15.92}
& \tcell{15.98} & \tcell{15.86}
& \tcell{16.1} & \tcell{15.96}
& \tcell{15.89} & \tcell{14.24}
& \tcell{15.08} & \tcell{15.73}
& \tcell{15.82} \\
& 128k \setbase{16.15} & \cellcolor{heatneutral}\base
& \tcell{13.41} & {\bfseries\tcell{16.78}}
& \tcell{15.82} & \tcell{15.68}
& \tcell{15.74} & \tcell{15.62}
& \tcell{15.86} & \tcell{15.72}
& \tcell{14.21} & \tcell{14.03}
& \tcell{14.85} & \tcell{16.12}
& \tcell{15.58} \\
& 256k \setbase{10.44} & \cellcolor{heatneutral}\base
& \tcell{9.46} & {\bfseries\tcell{11.12}}
& \tcell{10.62} & \tcell{10.49}
& \tcell{10.55} & \tcell{10.44}
& \tcell{10.65} & \tcell{10.52}
& \tcell{10.31} & \tcell{9.93}
& \tcell{10.65} & \tcell{9.57}
& \tcell{10.66} \\
\midrule

\multirow{5}{*}{\ds{4}}
& 16k \setbase{11.23} & \cellcolor{heatneutral}\base
& \tcell{14.08} & \tcell{14.05}
& \tcell{14.18} & \tcell{14.08}
& \tcell{14.13} & \tcell{14.05}
& {\bfseries\tcell{14.21}} & \tcell{14.09}
& \tcell{11.17} & \tcell{11.99}
& \tcell{10.76} & \tcell{11.74}
& \tcell{12.39} \\
& 32k \setbase{11.39} & \cellcolor{heatneutral}\base
& \tcell{12.68} & \tcell{10.17}
& \tcell{12.8} & \tcell{11.33}
& \tcell{12.75} & \tcell{11.32}
& {\bfseries\tcell{12.83}} & \tcell{11.34}
& \tcell{10.84} & \tcell{11.6}
& \tcell{10.72} & \tcell{11.17}
& \tcell{12.17} \\
& 64k \setbase{3.45} & \cellcolor{heatneutral}\base
& \tcell{5.22} & {\bfseries\tcell{5.3}}
& \tcell{5.26} & \tcell{5.23}
& \tcell{5.25} & \tcell{5.22}
& \tcell{5.27} & \tcell{5.24}
& \tcell{4.29} & \tcell{4.43}
& \tcell{4.04} & \tcell{3.93}
& \tcell{4.36} \\
& 128k \setbase{3.43} & \cellcolor{heatneutral}\base
& \tcell{3.48} & {\bfseries\tcell{3.97}}
& \tcell{3.53} & \tcell{3.49}
& \tcell{3.51} & \tcell{3.47}
& \tcell{3.54} & \tcell{3.5}
& \tcell{3.07} & \tcell{3.75}
& \tcell{3.06} & \tcell{3.45}
& \tcell{3.37} \\
& 256k \setbase{2.37} & \cellcolor{heatneutral}\base
& \tcell{2.5} & {\bfseries\tcell{2.92}}
& \tcell{2.54} & \tcell{2.51}
& \tcell{2.53} & \tcell{2.5}
& \tcell{2.55} & \tcell{2.52}
& \tcell{2.37} & \tcell{2.1}
& \tcell{2.5} & \tcell{2.71}
& \tcell{2.25} \\
\midrule

\multirow{5}{*}{\ds{5}}
& 16k \setbase{22.39} & \cellcolor{heatneutral}\base
& \tcell{24.12} & \tcell{24.04}
& \tcell{24.32} & \tcell{24.1}
& \tcell{24.23} & \tcell{24.06}
& {\bfseries\tcell{24.37}} & \tcell{24.11}
& \tcell{22.98} & \tcell{24.01}
& \tcell{22.37} & \tcell{23.01}
& \tcell{22.68} \\
& 32k \setbase{23.02} & \cellcolor{heatneutral}\base
& \tcell{24.32} & {\bfseries\tcell{24.63}}
& \tcell{24.56} & \tcell{24.36}
& \tcell{24.46} & \tcell{24.29}
& \tcell{24.61} & \tcell{24.41}
& \tcell{23.61} & \tcell{23.6}
& \tcell{23.12} & \tcell{23.51}
& \tcell{22.39} \\
& 64k \setbase{15.49} & \cellcolor{heatneutral}\base
& \tcell{15.6} & \tcell{15.67}
& \tcell{15.8} & \tcell{15.66}
& \tcell{15.72} & \tcell{15.6}
& \tcell{15.84} & \tcell{15.7}
& \tcell{16.68} & \tcell{16.49}
& \tcell{15.54} & {\bfseries\tcell{17.08}}
& \tcell{15.24} \\
& 128k \setbase{14.98} & \cellcolor{heatneutral}\base
& \tcell{13.99} & \tcell{13.75}
& \tcell{15.21} & \tcell{14.93}
& \tcell{15.12} & \tcell{14.9}
& \tcell{15.24} & \tcell{14.95}
& \tcell{15.06} & \tcell{14.53}
& \tcell{15.2} & \tcell{14.29}
& {\bfseries\tcell{15.47}} \\
& 256k \setbase{10.47} & \cellcolor{heatneutral}\base
& \tcell{10.34} & {\bfseries\tcell{11.95}}
& \tcell{10.65} & \tcell{10.52}
& \tcell{10.58} & \tcell{10.47}
& \tcell{10.68} & \tcell{10.55}
& \tcell{10.62} & \tcell{11.58}
& \tcell{10.81} & \tcell{11.46}
& \tcell{11.09} \\
\midrule

\multirow{5}{*}{\ds{6}}
& 16k \setbase{17.88} & \cellcolor{heatneutral}\base
& \tcell{19.77} & \tcell{19.65}
& \tcell{19.93} & \tcell{19.7}
& \tcell{19.86} & \tcell{19.68}
& {\bfseries\tcell{19.97}} & \tcell{19.71}
& \tcell{18.61} & \tcell{19.21}
& \tcell{18.62} & \tcell{18.85}
& \tcell{18.26} \\
& 32k \setbase{15.77} & \cellcolor{heatneutral}\base
& \tcell{16.13} & \tcell{15.77}
& \tcell{16.29} & \tcell{15.82}
& \tcell{16.22} & \tcell{15.8}
& \tcell{16.33} & \tcell{15.83}
& \tcell{15.91} & \tcell{16.13}
& \tcell{15.3} & {\bfseries\tcell{16.39}}
& \tcell{16.01} \\
& 64k \setbase{11.77} & \cellcolor{heatneutral}\base
& \tcell{12.48} & \tcell{12.01}
& \tcell{12.63} & \tcell{12.06}
& \tcell{12.57} & \tcell{12.04}
& {\bfseries\tcell{12.66}} & \tcell{12.07}
& \tcell{12.45} & \tcell{11.52}
& \tcell{11.63} & \tcell{12.39}
& \tcell{11.97} \\
& 128k \setbase{7.52} & \cellcolor{heatneutral}\base
& \tcell{7.91} & \tcell{7.84}
& \tcell{8.02} & \tcell{7.88}
& \tcell{7.98} & \tcell{7.86}
& {\bfseries\tcell{8.04}} & \tcell{7.89}
& \tcell{7.55} & \tcell{7.91}
& \tcell{7.31} & \tcell{6.89}
& \tcell{7.12} \\
& 256k \setbase{6.13} & \cellcolor{heatneutral}\base
& \tcell{5.61} & \tcell{6.83}
& \tcell{6.24} & \tcell{6.16}
& \tcell{6.2} & \tcell{6.13}
& \tcell{6.25} & \tcell{6.17}
& \tcell{6.18} & \tcell{6.84}
& \tcell{6.25} & {\bfseries\tcell{7.03}}
& \tcell{6.24} \\
\midrule

\multirow{5}{*}{\ds{7}}
& 16k \setbase{52.61} & \cellcolor{heatneutral}\base
& \tcell{51.86} & {\bfseries\tcell{53.73}}
& \tcell{52.83} & \tcell{52.61}
& \tcell{52.73} & \tcell{52.55}
& \tcell{52.89} & \tcell{52.67}
& \tcell{52.64} & \tcell{51.64}
& \tcell{52.84} & \tcell{53.19}
& \tcell{52.69} \\
& 32k \setbase{37.75} & \cellcolor{heatneutral}\base
& \tcell{37.72} & \tcell{37.97}
& \tcell{38.01} & \tcell{37.79}
& \tcell{37.9} & \tcell{37.71}
& {\bfseries\tcell{38.07}} & \tcell{37.85}
& \tcell{36.52} & \tcell{36.64}
& \tcell{36.98} & \tcell{35.57}
& \tcell{38.03} \\
& 64k \setbase{24.31} & \cellcolor{heatneutral}\base
& \tcell{26.22} & \tcell{23.7}
& \tcell{26.53} & \tcell{24.21}
& \tcell{26.41} & \tcell{24.18}
& {\bfseries\tcell{26.59}} & \tcell{24.24}
& \tcell{24.64} & \tcell{24.14}
& \tcell{24.69} & \tcell{26.24}
& \tcell{25.66} \\
& 128k \setbase{14.61} & \cellcolor{heatneutral}\base
& \tcell{16.16} & {\bfseries\tcell{16.97}}
& \tcell{16.38} & \tcell{16.2}
& \tcell{16.3} & \tcell{16.14}
& \tcell{16.42} & \tcell{16.23}
& \tcell{14.83} & \tcell{15.7}
& \tcell{15.2} & \tcell{16.69}
& \tcell{14.92} \\
& 256k \setbase{10.93} & \cellcolor{heatneutral}\base
& \tcell{10.93} & \tcell{10.74}
& \tcell{11.12} & \tcell{10.9}
& \tcell{11.05} & \tcell{10.88}
& \tcell{11.15} & \tcell{10.92}
& \tcell{10.56} & \tcell{10.65}
& \tcell{10.87} & \tcell{10.83}
& {\bfseries\tcell{11.24}} \\
\midrule

\multirow{5}{*}{\ds{8}}
& 16k \setbase{33.81} & \cellcolor{heatneutral}\base
& \tcell{34.28} & {\bfseries\tcell{36.01}}
& \tcell{34.5} & \tcell{34.28}
& \tcell{34.4} & \tcell{34.22}
& \tcell{34.56} & \tcell{34.34}
& \tcell{33.98} & \tcell{34.1}
& \tcell{32.69} & \tcell{35.75}
& \tcell{34.64} \\
& 32k \setbase{25.95} & \cellcolor{heatneutral}\base
& \tcell{27.51} & \tcell{25.08}
& \tcell{27.77} & \tcell{25.83}
& \tcell{27.66} & \tcell{25.8}
& {\bfseries\tcell{27.83}} & \tcell{25.85}
& \tcell{25.14} & \tcell{26.03}
& \tcell{25.84} & \tcell{26.5}
& \tcell{25.74} \\
& 64k \setbase{21.24} & \cellcolor{heatneutral}\base
& \tcell{20.22} & \tcell{21.27}
& \tcell{21.51} & \tcell{21.32}
& \tcell{21.4} & \tcell{21.24}
& \tcell{21.56} & \tcell{21.38}
& \tcell{21.39} & \tcell{20.01}
& {\bfseries\tcell{21.58}} & \tcell{20.89}
& \tcell{21.42} \\
& 128k \setbase{12.21} & \cellcolor{heatneutral}\base
& \tcell{12.65} & \tcell{14.08}
& \tcell{12.84} & \tcell{12.68}
& \tcell{12.76} & \tcell{12.63}
& \tcell{12.86} & \tcell{12.71}
& {\bfseries\tcell{14.13}} & \tcell{12.95}
& \tcell{13.04} & \tcell{13.5}
& \tcell{12.98} \\
& 256k \setbase{7.65} & \cellcolor{heatneutral}\base
& \tcell{7.97} & \tcell{7.21}
& \tcell{8.1} & \tcell{7.63}
& \tcell{8.05} & \tcell{7.62}
& \tcell{8.12} & \tcell{7.64}
& \tcell{7.99} & \tcell{8.01}
& {\bfseries\tcell{8.24}} & \tcell{8.04}
& \tcell{7.73} \\
\midrule

\multirow{5}{*}{\ds{9}}
& 16k \setbase{19.05} & \cellcolor{heatneutral}\base
& \tcell{23.38} & \tcell{17.99}
& \tcell{23.55} & \tcell{18.95}
& \tcell{23.47} & \tcell{18.93}
& {\bfseries\tcell{23.59}} & \tcell{18.96}
& \tcell{20.91} & \tcell{19.41}
& \tcell{18.84} & \tcell{20.2}
& \tcell{22.97} \\
& 32k \setbase{13.48} & \cellcolor{heatneutral}\base
& \tcell{15.95} & \tcell{15.37}
& \tcell{16.09} & \tcell{15.41}
& \tcell{16.03} & \tcell{15.4}
& \tcell{16.12} & \tcell{15.42}
& \tcell{14.77} & \tcell{15.62}
& \tcell{14.19} & {\bfseries\tcell{18.02}}
& \tcell{14.35} \\
& 64k \setbase{13.71} & \cellcolor{heatneutral}\base
& \tcell{13.85} & \tcell{11.61}
& \tcell{14.03} & \tcell{13.65}
& \tcell{13.95} & \tcell{13.63}
& \tcell{14.06} & \tcell{13.67}
& \tcell{13.51} & \tcell{13.72}
& \tcell{13.95} & {\bfseries\tcell{14.49}}
& \tcell{12.57} \\
& 128k \setbase{6.6} & \cellcolor{heatneutral}\base
& \tcell{8.44} & \tcell{6.79}
& \tcell{8.54} & \tcell{6.82}
& \tcell{8.5} & \tcell{6.81}
& {\bfseries\tcell{8.56}} & \tcell{6.83}
& \tcell{6.93} & \tcell{6.94}
& \tcell{6.96} & \tcell{6.98}
& \tcell{7.31} \\
& 256k \setbase{4.78} & \cellcolor{heatneutral}\base
& \tcell{5.35} & \tcell{4.84}
& \tcell{5.43} & \tcell{4.87}
& \tcell{5.4} & \tcell{4.86}
& \tcell{5.45} & \tcell{4.87}
& {\bfseries\tcell{5.73}} & \tcell{4.96}
& \tcell{5.37} & \tcell{5.51}
& \tcell{5.48} \\
\bottomrule
\end{tabular}
\end{adjustbox}
\end{table*}

\begin{table*}[t]
\centering
\scriptsize
\setlength{\tabcolsep}{1.8pt}
\renewcommand{\arraystretch}{0.9}
\caption{
Full long-context question answering results for \textbf{Qwen2.5-7B} on LV-Eval across 9 datasets and context lengths from 16K to 256K.
GI and IAA denote Gradient-informed and Input-activation-aware tile-retention policies, with subscripts indicating the retained floating-point tile ratio in percent.
We follow the LV-Eval metric protocol: keyword-recall F1 by default, blacklist-based F1 for \textit{cmrc\_mixup}, and blacklist-based ROUGE-L for \textit{dureader\_mixup}.
Cell colors are relative to the row-wise Original reference: green and blue indicate higher and lower scores, respectively, while gray marks Original and values within the neutral comparison range.
}
\label{tab:lveval_qwen7b_full}

\begin{adjustbox}{width=\textwidth}
\begin{tabular}{c c c @{\vsep} c c c c c c c c @{\vsep} c c @{\vsep} c c @{\vsep} c}
\toprule
\multirow{2}{*}{\textbf{Dataset}} &
\multirow{2}{*}{\textbf{Len}} &
\multirow{2}{*}{\textbf{Original}} &
\multicolumn{8}{c}{\textbf{Ours}} &
\multicolumn{2}{c}{\textbf{Uniform}} &
\multirow{2}{*}{\rotatebox{25}{\textbf{GPTQ}}} &
\multirow{2}{*}{\rotatebox{25}{\textbf{SmoothQ}}} &
\multirow{2}{*}{\rotatebox{25}{\textbf{LLM.int8}}} \\
\cmidrule(lr){4-11}\cmidrule(lr){12-13}
& & &
\textbf{GI$_{1}$} & \textbf{IAA$_{1}$} &
\textbf{GI$_{5}$} & \textbf{IAA$_{5}$} &
\textbf{GI$_{10}$} & \textbf{IAA$_{10}$} &
\textbf{GI$_{20}$} & \textbf{IAA$_{20}$} &
\textbf{W8} & \textbf{W8A8} &
& & \\
\midrule

\multirow{5}{*}{\ds{1}}
& 16k \setbase{24.6} & \cellcolor{heatneutral}\base
& \tcell{22.38} & \tcell{22.18}
& \tcell{23.4} & \tcell{23.33}
& \tcell{23.85} & \tcell{23.81}
& \tcell{24.63} & {\bfseries\tcell{24.66}}
& \tcell{23.99} & \tcell{24.26}
& \tcell{23.33} & \tcell{22.63}
& \tcell{23.35} \\
& 32k \setbase{23.19} & \cellcolor{heatneutral}\base
& \tcell{19.17} & \tcell{20.18}
& \tcell{21.66} & \tcell{21.6}
& \tcell{22.32} & \tcell{22.35}
& \tcell{23.46} & {\bfseries\tcell{23.49}}
& \tcell{22.51} & \tcell{22.16}
& \tcell{21.91} & \tcell{21.01}
& \tcell{23.41} \\
& 64k \setbase{11.83} & \cellcolor{heatneutral}\base
& \tcell{11.13} & \tcell{10.06}
& \tcell{11.88} & \tcell{11.85}
& \tcell{12.22} & \tcell{12.19}
& {\bfseries\tcell{12.81}} & \tcell{12.8}
& \tcell{12.63} & \tcell{11.77}
& \tcell{12.76} & \tcell{11.79}
& \tcell{12.12} \\
& 128k \setbase{7.41} & \cellcolor{heatneutral}\base
& \tcell{6.5} & \tcell{5.69}
& \tcell{6.97} & \tcell{6.95}
& \tcell{7.18} & \tcell{7.16}
& \tcell{7.54} & {\bfseries\tcell{7.56}}
& \tcell{7.51} & \tcell{6.87}
& \tcell{7.33} & \tcell{6.12}
& \tcell{7.46} \\
& 256k \setbase{4.01} & \cellcolor{heatneutral}\base
& \tcell{3.72} & \tcell{3.84}
& \tcell{4.17} & \tcell{4.16}
& \tcell{4.31} & \tcell{4.32}
& \tcell{4.56} & {\bfseries\tcell{4.57}}
& \tcell{4.36} & \tcell{4.13}
& \tcell{4.32} & \tcell{3.98}
& \tcell{4.54} \\
\midrule

\multirow{5}{*}{\ds{2}}
& 16k \setbase{20.89} & \cellcolor{heatneutral}\base
& \tcell{17.05} & \tcell{17.82}
& \tcell{19.22} & \tcell{19.16}
& \tcell{19.84} & \tcell{19.87}
& {\bfseries\tcell{20.94}} & \tcell{20.92}
& \tcell{20.13} & \tcell{20.16}
& \tcell{20.22} & \tcell{20.24}
& \tcell{19.85} \\
& 32k \setbase{19.06} & \cellcolor{heatneutral}\base
& \tcell{16.79} & \tcell{17.47}
& \tcell{18.82} & \tcell{18.76}
& \tcell{19.26} & \tcell{19.21}
& {\bfseries\tcell{20.05}} & \tcell{20.03}
& \tcell{19.37} & \tcell{19.99}
& \tcell{18.35} & \tcell{18.39}
& \tcell{18.89} \\
& 64k \setbase{17.57} & \cellcolor{heatneutral}\base
& \tcell{16.38} & \tcell{18.01}
& \tcell{18.28} & \tcell{18.27}
& \tcell{18.41} & \tcell{18.41}
& \tcell{18.61} & {\bfseries\tcell{18.63}}
& \tcell{16.9} & \tcell{16.52}
& \tcell{17.15} & \tcell{18.55}
& \tcell{16.4} \\
& 128k \setbase{14.66} & \cellcolor{heatneutral}\base
& \tcell{12.95} & \tcell{13.17}
& \tcell{14.15} & \tcell{14.11}
& \tcell{14.58} & \tcell{14.61}
& {\bfseries\tcell{15.36}} & \tcell{15.33}
& \tcell{15.03} & \tcell{14.35}
& \tcell{15.27} & \tcell{13.26}
& \tcell{14.48} \\
& 256k \setbase{8.69} & \cellcolor{heatneutral}\base
& \tcell{8.0} & \tcell{8.31}
& \tcell{8.55} & \tcell{8.53}
& \tcell{8.66} & \tcell{8.66}
& {\bfseries\tcell{8.85}} & \tcell{8.84}
& \tcell{8.79} & \tcell{8.6}
& \tcell{8.58} & \tcell{8.27}
& \tcell{8.74} \\
\midrule

\multirow{5}{*}{\ds{3}}
& 16k \setbase{5.06} & \cellcolor{heatneutral}\base
& \tcell{4.53} & \tcell{4.57}
& \tcell{4.09} & \tcell{4.99}
& \tcell{5.22} & \tcell{5.23}
& {\bfseries\tcell{5.57}} & \tcell{5.56}
& \tcell{4.93} & \tcell{5.47}
& \tcell{5} & \tcell{5.56}
& \tcell{5.39} \\
& 32k \setbase{3.4} & \cellcolor{heatneutral}\base
& \tcell{2.72} & \tcell{2.4}
& \tcell{2.82} & \tcell{2.87}
& \tcell{3.25} & \tcell{3.15}
& {\bfseries\tcell{3.53}} & \tcell{3.51}
& \tcell{3.22} & \tcell{3}
& \tcell{3.3} & \tcell{3.52}
& \tcell{3.22} \\
& 64k \setbase{2.61} & \cellcolor{heatneutral}\base
& \tcell{2.47} & \tcell{2.44}
& \tcell{2.36} & \tcell{2.8}
& \tcell{3} & \tcell{3}
& {\bfseries\tcell{3.29}} & \tcell{3.28}
& \tcell{2.81} & \tcell{2.93}
& \tcell{2.71} & \tcell{2.69}
& \tcell{3.28} \\
& 128k \setbase{2.21} & \cellcolor{heatneutral}\base
& \tcell{1.88} & \tcell{1.97}
& \tcell{2.12} & \tcell{2.12}
& \tcell{2.19} & \tcell{2.2}
& {\bfseries\tcell{2.31}} & \tcell{2.3}
& \tcell{2.17} & \tcell{2.02}
& \tcell{2.17} & \tcell{2.09}
& \tcell{2.3} \\
& 256k \setbase{1.42} & \cellcolor{heatneutral}\base
& \tcell{1.52} & \tcell{1.28}
& {\bfseries\tcell{1.53}} & \tcell{1.52}
& {\bfseries\tcell{1.53}} & \tcell{1.52}
& {\bfseries\tcell{1.53}} & \tcell{1.51}
& \tcell{1.36} & \tcell{1.3}
& \tcell{1.41} & \tcell{1.47}
& \tcell{1.42} \\
\midrule

\multirow{5}{*}{\ds{4}}
& 16k \setbase{18.89} & \cellcolor{heatneutral}\base
& \tcell{18.86} & \tcell{18.04}
& \tcell{19.49} & \tcell{19.44}
& \tcell{19.77} & \tcell{19.72}
& {\bfseries\tcell{20.27}} & \tcell{20.25}
& \tcell{19.57} & \tcell{20.2}
& \tcell{19.77} & \tcell{20.22}
& \tcell{18.68} \\
& 32k \setbase{20.54} & {\bfseries\cellcolor{heatneutral}\base}
& \tcell{18.41} & \tcell{17.77}
& \tcell{20.08} & \tcell{20.02}
& \tcell{19.47} & \tcell{19.42}
& \tcell{19.97} & \tcell{19.95}
& \tcell{19.28} & \tcell{19.9}
& \tcell{19.47} & \tcell{20.13}
& \tcell{18.4} \\
& 64k \setbase{13.8} & \cellcolor{heatneutral}\base
& \tcell{13.05} & \tcell{11.54}
& \tcell{13.7} & \tcell{13.66}
& \tcell{13.98} & \tcell{13.95}
& \tcell{14.48} & {\bfseries\tcell{14.49}}
& \tcell{14.43} & \tcell{13.3}
& \tcell{14.15} & \tcell{13.73}
& \tcell{13.68} \\
& 128k \setbase{5.39} & \cellcolor{heatneutral}\base
& \tcell{5.07} & \tcell{5.4}
& \tcell{5.71} & \tcell{5.69}
& \tcell{5.85} & \tcell{5.85}
& {\bfseries\tcell{6.09}} & \tcell{6.08}
& \tcell{5.24} & \tcell{6.06}
& \tcell{5.45} & \tcell{4.86}
& \tcell{5.5} \\
& 256k \setbase{2.97} & \cellcolor{heatneutral}\base
& \tcell{3.47} & \tcell{2.74}
& \tcell{3.48} & \tcell{3.47}
& \tcell{3.48} & \tcell{3.48}
& {\bfseries\tcell{3.49}} & {\bfseries\tcell{3.49}}
& \tcell{2.91} & \tcell{3.46}
& \tcell{3.32} & \tcell{3.24}
& \tcell{2.9} \\
\midrule

\multirow{5}{*}{\ds{5}}
& 16k \setbase{14.67} & \cellcolor{heatneutral}\base
& \tcell{13.74} & \tcell{13.82}
& \tcell{14.35} & \tcell{14.32}
& \tcell{14.59} & \tcell{14.6}
& {\bfseries\tcell{15.01}} & \tcell{14.99}
& \tcell{14.97} & \tcell{13.92}
& \tcell{14.41} & \tcell{14.79}
& \tcell{14.26} \\
& 32k \setbase{13.21} & \cellcolor{heatneutral}\base
& \tcell{13.13} & \tcell{12.24}
& \tcell{13.38} & \tcell{13.34}
& \tcell{13.49} & \tcell{13.46}
& {\bfseries\tcell{13.68}} & \tcell{13.67}
& \tcell{12.77} & \tcell{13.21}
& \tcell{13.19} & \tcell{12.98}
& \tcell{13.64} \\
& 64k \setbase{10.69} & \cellcolor{heatneutral}\base
& \tcell{10.17} & \tcell{10.05}
& \tcell{10.59} & \tcell{10.56}
& \tcell{10.78} & \tcell{10.75}
& \tcell{11.1} & {\bfseries\tcell{11.11}}
& \tcell{11.06} & \tcell{10.96}
& \tcell{10.84} & \tcell{10.66}
& \tcell{10.9} \\
& 128k \setbase{10.52} & \cellcolor{heatneutral}\base
& \tcell{9.51} & \tcell{9.79}
& \tcell{10.14} & \tcell{10.12}
& \tcell{10.3} & \tcell{10.31}
& {\bfseries\tcell{10.59}} & \tcell{10.56}
& \tcell{10.22} & \tcell{10.12}
& \tcell{10.1} & \tcell{10.15}
& \tcell{10.15} \\
& 256k \setbase{9.87} & \cellcolor{heatneutral}\base
& \tcell{8.63} & \tcell{8.75}
& \tcell{9.28} & \tcell{9.25}
& \tcell{9.52} & \tcell{9.53}
& {\bfseries\tcell{9.94}} & \tcell{9.92}
& \tcell{9.41} & \tcell{9.44}
& \tcell{9.44} & \tcell{8.41}
& \tcell{9.47} \\
\midrule

\multirow{5}{*}{\ds{6}}
& 16k \setbase{10.41} & \cellcolor{heatneutral}\base
& \tcell{10.13} & \tcell{9.37}
& \tcell{10.55} & \tcell{10.52}
& \tcell{10.74} & \tcell{10.71}
& \tcell{11.05} & {\bfseries\tcell{11.07}}
& \tcell{11} & \tcell{11.04}
& \tcell{10.83} & \tcell{10.39}
& \tcell{10.84} \\
& 32k \setbase{9.2} & \cellcolor{heatneutral}\base
& \tcell{8.34} & \tcell{7.98}
& \tcell{9.09} & \tcell{9.06}
& \tcell{9.42} & \tcell{9.39}
& \tcell{9.99} & {\bfseries\tcell{10}}
& \tcell{9.32} & \tcell{8.98}
& \tcell{9.51} & \tcell{9.97}
& \tcell{9.7} \\
& 64k \setbase{7.15} & \cellcolor{heatneutral}\base
& \tcell{6.97} & \tcell{6.6}
& \tcell{7.3} & \tcell{7.28}
& \tcell{7.44} & \tcell{7.43}
& {\bfseries\tcell{7.7}} & \tcell{7.69}
& \tcell{7.21} & \tcell{7.47}
& \tcell{7.36} & \tcell{7.33}
& \tcell{7.67} \\
& 128k \setbase{5.96} & \cellcolor{heatneutral}\base
& \tcell{5.46} & \tcell{5.38}
& \tcell{5.72} & \tcell{5.7}
& \tcell{5.84} & \tcell{5.82}
& \tcell{6.04} & {\bfseries\tcell{6.05}}
& \tcell{5.72} & \tcell{5.75}
& \tcell{6.01} & \tcell{6}
& \tcell{5.78} \\
& 256k \setbase{5.59} & \cellcolor{heatneutral}\base
& \tcell{5.35} & \tcell{5.27}
& \tcell{5.66} & \tcell{5.64}
& \tcell{5.8} & \tcell{5.78}
& \tcell{6.03} & {\bfseries\tcell{6.04}}
& \tcell{5.52} & \tcell{5.71}
& \tcell{5.83} & \tcell{6}
& \tcell{5.77} \\
\midrule

\multirow{5}{*}{\ds{7}}
& 16k \setbase{26.3} & \cellcolor{heatneutral}\base
& \tcell{21.84} & \tcell{22.36}
& \tcell{24.61} & \tcell{24.54}
& \tcell{25.6} & \tcell{25.65}
& {\bfseries\tcell{27.37}} & \tcell{27.34}
& \tcell{25.81} & \tcell{25.19}
& \tcell{27.3} & \tcell{25.47}
& \tcell{25.67} \\
& 32k \setbase{19.84} & \cellcolor{heatneutral}\base
& \tcell{17.87} & \tcell{17.81}
& \tcell{19.05} & \tcell{18.99}
& \tcell{19.57} & \tcell{19.58}
& {\bfseries\tcell{20.5}} & \tcell{20.47}
& \tcell{19.84} & \tcell{19.31}
& \tcell{19.72} & \tcell{20.43}
& \tcell{19.35} \\
& 64k \setbase{13.23} & \cellcolor{heatneutral}\base
& \tcell{12.21} & \tcell{12.36}
& \tcell{12.88} & \tcell{12.85}
& \tcell{13.12} & \tcell{13.13}
& \tcell{13.51} & {\bfseries\tcell{13.53}}
& \tcell{13.44} & \tcell{12.53}
& \tcell{13.46} & \tcell{13.47}
& \tcell{12.29} \\
& 128k \setbase{9.55} & \cellcolor{heatneutral}\base
& \tcell{8.85} & \tcell{8.61}
& \tcell{9.55} & \tcell{9.53}
& \tcell{9.86} & \tcell{9.83}
& {\bfseries\tcell{10.42}} & \tcell{10.4}
& \tcell{9.58} & \tcell{10.06}
& \tcell{9.82} & \tcell{10.36}
& \tcell{9.4} \\
& 256k \setbase{7.14} & \cellcolor{heatneutral}\base
& \tcell{7.49} & \tcell{6.78}
& \tcell{7.6} & \tcell{7.58}
& \tcell{7.65} & \tcell{7.63}
& {\bfseries\tcell{7.74}} & \tcell{7.73}
& \tcell{7.49} & \tcell{7.69}
& \tcell{7.11} & \tcell{7.2}
& \tcell{7.06} \\
\midrule

\multirow{5}{*}{\ds{8}}
& 16k \setbase{18.05} & \cellcolor{heatneutral}\base
& \tcell{16.62} & \tcell{16.97}
& \tcell{17.51} & \tcell{17.47}
& \tcell{17.75} & \tcell{17.76}
& \tcell{18.16} & {\bfseries\tcell{18.18}}
& \tcell{17.49} & \tcell{16.5}
& \tcell{18.13} & \tcell{18.01}
& \tcell{17.96} \\
& 32k \setbase{14.24} & \cellcolor{heatneutral}\base
& \tcell{12.57} & \tcell{12.9}
& \tcell{14.3} & \tcell{14.26}
& \tcell{14.93} & \tcell{14.96}
& \tcell{16.01} & {\bfseries\tcell{16.03}}
& \tcell{14.49} & \tcell{14.42}
& \tcell{13.72} & \tcell{15.98}
& \tcell{14.43} \\
& 64k \setbase{12.1} & \cellcolor{heatneutral}\base
& \tcell{11.69} & \tcell{11.77}
& \tcell{11.97} & \tcell{11.95}
& \tcell{12.05} & \tcell{12.06}
& {\bfseries\tcell{12.22}} & \tcell{12.2}
& \tcell{11.93} & \tcell{11.64}
& \tcell{11.45} & \tcell{11.63}
& \tcell{12.16} \\
& 128k \setbase{9.86} & \cellcolor{heatneutral}\base
& \tcell{9.66} & \tcell{9.87}
& \tcell{10.4} & \tcell{10.37}
& \tcell{10.63} & \tcell{10.64}
& \tcell{11.03} & {\bfseries\tcell{11.05}}
& \tcell{9.62} & \tcell{10.87}
& \tcell{9.97} & \tcell{10.99}
& \tcell{9.48} \\
& 256k \setbase{8.52} & \cellcolor{heatneutral}\base
& \tcell{8.6} & \tcell{8.74}
& \tcell{9.18} & \tcell{9.16}
& \tcell{9.38} & \tcell{9.39}
& \tcell{9.72} & {\bfseries\tcell{9.73}}
& \tcell{9.67} & \tcell{9.04}
& \tcell{8.56} & \tcell{9.67}
& \tcell{9.01} \\
\midrule

\multirow{5}{*}{\ds{9}}
& 16k \setbase{14.57} & \cellcolor{heatneutral}\base
& \tcell{15.29} & \tcell{14.91}
& \tcell{15.3} & \tcell{15.26}
& \tcell{15.31} & \tcell{15.28}
& {\bfseries\tcell{15.33}} & \tcell{15.31}
& \tcell{14.69} & \tcell{14.65}
& \tcell{15.11} & \tcell{13.65}
& \tcell{14} \\
& 32k \setbase{13.51} & \cellcolor{heatneutral}\base
& \tcell{13.88} & \tcell{13.69}
& \tcell{14.09} & \tcell{14.05}
& \tcell{14.18} & \tcell{14.15}
& {\bfseries\tcell{14.35}} & \tcell{14.33}
& \tcell{13.11} & \tcell{14.16}
& \tcell{13.77} & \tcell{12.87}
& \tcell{14.3} \\
& 64k \setbase{9.98} & \cellcolor{heatneutral}\base
& \tcell{9.94} & \tcell{10.56}
& \tcell{10.8} & \tcell{10.79}
& \tcell{10.91} & \tcell{10.92}
& \tcell{11.09} & {\bfseries\tcell{11.11}}
& \tcell{10.94} & \tcell{9.88}
& \tcell{10.82} & \tcell{11.06}
& \tcell{10.74} \\
& 128k \setbase{7.15} & \cellcolor{heatneutral}\base
& \tcell{7.17} & \tcell{7.61}
& \tcell{7.63} & \tcell{7.63}
& \tcell{7.63} & \tcell{7.63}
& {\bfseries\tcell{7.65}} & \tcell{7.64}
& \tcell{6.47} & \tcell{6.92}
& \tcell{6.87} & \tcell{7.26}
& \tcell{7.58} \\
& 256k \setbase{5.21} & \cellcolor{heatneutral}\base
& \tcell{4.46} & \tcell{4.99}
& \tcell{5.34} & \tcell{5.32}
& \tcell{5.49} & \tcell{5.5}
& {\bfseries\tcell{5.77}} & \tcell{5.76}
& \tcell{4.9} & \tcell{4.11}
& \tcell{4.49} & \tcell{5.73}
& \tcell{5.03} \\
\bottomrule
\end{tabular}
\end{adjustbox}
\end{table*}

\begin{table*}[t]
\centering
\scriptsize
\setlength{\tabcolsep}{1.8pt}
\renewcommand{\arraystretch}{0.9}
\caption{
Full long-context question answering results for \textbf{Qwen2.5-14B} on LV-Eval across 9 datasets and context lengths from 16K to 256K.
GI and IAA denote Gradient-informed and Input-activation-aware tile-retention policies, with subscripts indicating the retained floating-point tile ratio in percent.
We follow the LV-Eval metric protocol: keyword-recall F1 by default, blacklist-based F1 for \textit{cmrc\_mixup}, and blacklist-based ROUGE-L for \textit{dureader\_mixup}.
Cell colors are relative to the row-wise Original reference: green and blue indicate higher and lower scores, respectively, while gray marks Original and values within the neutral comparison range.
}
\label{tab:lveval_qwen14b_full}

\begin{adjustbox}{width=\textwidth}
\begin{tabular}{c c c @{\vsep} c c c c c c c c @{\vsep} c c @{\vsep} c c @{\vsep} c}
\toprule
\multirow{2}{*}{\textbf{Dataset}} &
\multirow{2}{*}{\textbf{Len}} &
\multirow{2}{*}{\textbf{Original}} &
\multicolumn{8}{c}{\textbf{Ours}} &
\multicolumn{2}{c}{\textbf{Uniform}} &
\multirow{2}{*}{\rotatebox{25}{\textbf{GPTQ}}} &
\multirow{2}{*}{\rotatebox{25}{\textbf{SmoothQ}}} &
\multirow{2}{*}{\rotatebox{25}{\textbf{LLM.int8}}} \\
\cmidrule(lr){4-11}\cmidrule(lr){12-13}
& & &
\textbf{GI$_{1}$} & \textbf{IAA$_{1}$} &
\textbf{GI$_{5}$} & \textbf{IAA$_{5}$} &
\textbf{GI$_{10}$} & \textbf{IAA$_{10}$} &
\textbf{GI$_{20}$} & \textbf{IAA$_{20}$} &
\textbf{W8} & \textbf{W8A8} &
& & \\
\midrule

\multirow{5}{*}{\ds{1}}
& 16k \setbase{27.62} & \cellcolor{heatneutral}\base
& \tcell{26.61} & \tcell{25.96}
& \tcell{28.26} & \tcell{28.18}
& \tcell{29.25} & \tcell{29.17}
& {\bfseries\tcell{30.28}} & \tcell{30.2}
& \tcell{28.11} & \tcell{28.77}
& \tcell{27.79} & \tcell{26.69}
& \tcell{30.12} \\
& 32k \setbase{26.84} & \cellcolor{heatneutral}\base
& \tcell{25.51} & \tcell{24.79}
& \tcell{26.57} & \tcell{26.49}
& \tcell{27.21} & \tcell{27.13}
& {\bfseries\tcell{27.87}} & \tcell{27.79}
& \tcell{27.19} & \tcell{27.68}
& \tcell{26.75} & \tcell{25.59}
& \tcell{25.84} \\
& 64k \setbase{12.82} & \cellcolor{heatneutral}\base
& \tcell{12.01} & \tcell{13.23}
& \tcell{14.3} & \tcell{14.26}
& \tcell{14.95} & \tcell{14.9}
& {\bfseries\tcell{15.61}} & \tcell{15.57}
& \tcell{12.19} & \tcell{15.5}
& \tcell{11.48} & \tcell{12.57}
& \tcell{12.8} \\
& 128k \setbase{8.86} & \cellcolor{heatneutral}\base
& \tcell{7.25} & \tcell{7.63}
& \tcell{8.24} & \tcell{8.2}
& \tcell{8.6} & \tcell{8.57}
& {\bfseries\tcell{8.98}} & \tcell{8.94}
& \tcell{8.7} & \tcell{8.88}
& \tcell{8.63} & \tcell{8.1}
& \tcell{8.57} \\
& 256k \setbase{4.41} & \cellcolor{heatneutral}\base
& \tcell{4.95} & \tcell{4.51}
& \tcell{5.19} & \tcell{5.17}
& \tcell{5.33} & \tcell{5.31}
& {\bfseries\tcell{5.48}} & \tcell{5.46}
& \tcell{4.61} & \tcell{5.42}
& \tcell{4.67} & \tcell{4.88}
& \tcell{4.82} \\
\midrule

\multirow{5}{*}{\ds{2}}
& 16k \setbase{20.59} & \cellcolor{heatneutral}\base
& \tcell{19.4} & \tcell{19.22}
& \tcell{19.99} & \tcell{19.93}
& \tcell{20.35} & \tcell{20.29}
& {\bfseries\tcell{20.72}} & \tcell{20.66}
& \tcell{20.45} & \tcell{19.13}
& \tcell{20.37} & \tcell{20.2}
& \tcell{19.91} \\
& 32k \setbase{20.2} & \cellcolor{heatneutral}\base
& \tcell{17.47} & \tcell{19.22}
& \tcell{20.15} & \tcell{20.09}
& \tcell{20.71} & \tcell{20.65}
& {\bfseries\tcell{21.29}} & \tcell{21.23}
& \tcell{20.26} & \tcell{19.02}
& \tcell{20.06} & \tcell{20.06}
& \tcell{20.36} \\
& 64k \setbase{19.09} & \cellcolor{heatneutral}\base
& \tcell{17.13} & \tcell{17.24}
& \tcell{18.25} & \tcell{18.19}
& \tcell{18.85} & \tcell{18.79}
& {\bfseries\tcell{19.48}} & \tcell{19.42}
& \tcell{18.98} & \tcell{18.27}
& \tcell{19.31} & \tcell{18.39}
& \tcell{18.13} \\
& 128k \setbase{15.17} & \cellcolor{heatneutral}\base
& \tcell{14.59} & \tcell{14.39}
& \tcell{15.34} & \tcell{15.27}
& \tcell{15.79} & \tcell{15.72}
& {\bfseries\tcell{16.26}} & \tcell{16.19}
& \tcell{15.44} & \tcell{14.83}
& \tcell{15.39} & \tcell{15.98}
& \tcell{16.08} \\
& 256k \setbase{8.26} & \cellcolor{heatneutral}\base
& \tcell{10.48} & \tcell{9.51}
& \tcell{10.53} & \tcell{10.49}
& \tcell{10.56} & \tcell{10.52}
& {\bfseries\tcell{10.59}} & \tcell{10.55}
& \tcell{8.66} & \tcell{8.76}
& \tcell{8.23} & \tcell{8.21}
& \tcell{8.72} \\
\midrule

\multirow{5}{*}{\ds{3}}
& 16k \setbase{4.21} & \cellcolor{heatneutral}\base
& \tcell{4.03} & \tcell{3.54}
& \tcell{4.21} & \tcell{4.2}
& \tcell{4.27} & \tcell{4.23}
& {\bfseries\tcell{4.37}} & \tcell{4.35}
& \tcell{4.23} & \tcell{4.33}
& \tcell{4.34} & \tcell{4.24}
& \tcell{4.09} \\
& 32k \setbase{2.76} & \cellcolor{heatneutral}\base
& \tcell{2.53} & \tcell{2.9}
& \tcell{3.35} & \tcell{3.22}
& \tcell{3.36} & \tcell{3.22}
& {\bfseries\tcell{3.37}} & \tcell{3.36}
& \tcell{2.83} & \tcell{3.18}
& \tcell{2.78} & \tcell{3.33}
& \tcell{3.07} \\
& 64k \setbase{2.54} & \cellcolor{heatneutral}\base
& \tcell{2.3} & \tcell{2.19}
& \tcell{2.41} & \tcell{2.39}
& \tcell{2.52} & \tcell{2.39}
& {\bfseries\tcell{2.6}} & \tcell{2.59}
& \tcell{2.48} & \tcell{2.58}
& \tcell{2.5} & \tcell{2.48}
& \tcell{2.45} \\
& 128k \setbase{2.2} & \cellcolor{heatneutral}\base
& \tcell{2.36} & \tcell{2.01}
& \tcell{2.12} & \tcell{2.12}
& \tcell{2.38} & \tcell{2.3}
& {\bfseries\tcell{2.39}} & \tcell{2.38}
& \tcell{2.3} & \tcell{2.04}
& \tcell{2.24} & \tcell{2.18}
& \tcell{2.32} \\
& 256k \setbase{1.62} & \cellcolor{heatneutral}\base
& \tcell{1.94} & \tcell{1.75}
& \tcell{1.78} & \tcell{1.78}
& \tcell{1.96} & \tcell{1.92}
& {\bfseries\tcell{1.97}} & \tcell{1.96}
& \tcell{1.68} & \tcell{1.85}
& \tcell{1.69} & \tcell{1.76}
& \tcell{1.59} \\
\midrule

\multirow{5}{*}{\ds{4}}
& 16k \setbase{21.18} & \cellcolor{heatneutral}\base
& \tcell{22.6} & \tcell{22.63}
& \tcell{23.21} & \tcell{23.15}
& \tcell{23.56} & \tcell{23.5}
& {\bfseries\tcell{23.93}} & \tcell{23.86}
& \tcell{22.07} & \tcell{23.79}
& \tcell{21.07} & \tcell{23.16}
& \tcell{22.73} \\
& 32k \setbase{24.51} & \cellcolor{heatneutral}\base
& \tcell{22.32} & \tcell{21.52}
& \tcell{23.39} & \tcell{23.31}
& \tcell{24.03} & \tcell{23.95}
& {\bfseries\tcell{24.7}} & \tcell{24.62}
& \tcell{21.74} & \tcell{23.42}
& \tcell{20.75} & \tcell{22.81}
& \tcell{22.75} \\
& 64k \setbase{13.2} & \cellcolor{heatneutral}\base
& \tcell{13.53} & \tcell{14.03}
& \tcell{14.42} & \tcell{14.37}
& \tcell{14.65} & \tcell{14.6}
& {\bfseries\tcell{14.89}} & \tcell{14.84}
& \tcell{13.28} & \tcell{14.49}
& \tcell{12.77} & \tcell{14.77}
& \tcell{14} \\
& 128k \setbase{6.11} & \cellcolor{heatneutral}\base
& \tcell{7.1} & \tcell{6.65}
& \tcell{7.13} & \tcell{7.11}
& \tcell{7.15} & \tcell{7.12}
& {\bfseries\tcell{7.17}} & \tcell{7.14}
& \tcell{6.57} & \tcell{6.94}
& \tcell{6.25} & \tcell{6.21}
& \tcell{6.62} \\
& 256k \setbase{4.62} & \cellcolor{heatneutral}\base
& \tcell{4.89} & \tcell{4.56}
& \tcell{5.07} & \tcell{5.05}
& \tcell{5.17} & \tcell{5.15}
& {\bfseries\tcell{5.28}} & \tcell{5.26}
& \tcell{4.67} & \tcell{4.62}
& \tcell{4.62} & \tcell{5.22}
& \tcell{4.65} \\
\midrule

\multirow{5}{*}{\ds{5}}
& 16k \setbase{13.99} & \cellcolor{heatneutral}\base
& \tcell{14.05} & \tcell{13.52}
& \tcell{14.21} & \tcell{14.17}
& \tcell{14.31} & \tcell{14.26}
& {\bfseries\tcell{14.41}} & \tcell{14.36}
& \tcell{14.32} & \tcell{14.2}
& \tcell{13.69} & \tcell{13.69}
& \tcell{14.29} \\
& 32k \setbase{14.18} & \cellcolor{heatneutral}\base
& \tcell{12.9} & \tcell{13.32}
& \tcell{13.76} & \tcell{13.71}
& \tcell{14.02} & \tcell{13.97}
& {\bfseries\tcell{14.29}} & \tcell{14.24}
& \tcell{13.72} & \tcell{13.62}
& \tcell{13.59} & \tcell{13.99}
& \tcell{13.28} \\
& 64k \setbase{10.26} & \cellcolor{heatneutral}\base
& \tcell{10.26} & \tcell{10.85}
& \tcell{10.96} & \tcell{10.93}
& \tcell{11.02} & \tcell{10.99}
& {\bfseries\tcell{11.09}} & \tcell{11.06}
& \tcell{10.27} & \tcell{10.14}
& \tcell{10.63} & \tcell{10.36}
& \tcell{11} \\
& 128k \setbase{10.89} & \cellcolor{heatneutral}\base
& \tcell{10.2} & \tcell{10.03}
& \tcell{10.59} & \tcell{10.54}
& \tcell{10.82} & \tcell{10.77}
& {\bfseries\tcell{11.06}} & \tcell{11.01}
& \tcell{10.12} & \tcell{9.99}
& \tcell{10.56} & \tcell{10.29}
& \tcell{10.36} \\
& 256k \setbase{8.56} & \cellcolor{heatneutral}\base
& \tcell{8.72} & \tcell{8.85}
& \tcell{8.9} & \tcell{8.88}
& \tcell{8.93} & \tcell{8.9}
& {\bfseries\tcell{8.96}} & \tcell{8.93}
& \tcell{8.71} & \tcell{8.75}
& \tcell{8.83} & \tcell{8.33}
& \tcell{8.19} \\
\midrule

\multirow{5}{*}{\ds{6}}
& 16k \setbase{11.56} & \cellcolor{heatneutral}\base
& \tcell{10.44} & \tcell{10.36}
& \tcell{10.98} & \tcell{10.94}
& \tcell{11.3} & \tcell{11.26}
& {\bfseries\tcell{11.63}} & \tcell{11.6}
& \tcell{11.12} & \tcell{10.03}
& \tcell{10.86} & \tcell{9.9}
& \tcell{10.55} \\
& 32k \setbase{9.84} & \cellcolor{heatneutral}\base
& \tcell{9.35} & \tcell{9.34}
& \tcell{9.6} & \tcell{9.57}
& \tcell{9.76} & \tcell{9.73}
& {\bfseries\tcell{9.91}} & \tcell{9.88}
& \tcell{9.73} & \tcell{8.91}
& \tcell{9.39} & \tcell{8.74}
& \tcell{9.07} \\
& 64k \setbase{7.53} & \cellcolor{heatneutral}\base
& \tcell{7.52} & \tcell{7.72}
& \tcell{7.97} & \tcell{7.95}
& \tcell{8.12} & \tcell{8.1}
& {\bfseries\tcell{8.28}} & \tcell{8.25}
& \tcell{8.21} & \tcell{7.94}
& \tcell{7.81} & \tcell{8.17}
& \tcell{7.73} \\
& 128k \setbase{6.29} & \cellcolor{heatneutral}\base
& \tcell{5.67} & \tcell{5.83}
& \tcell{6.07} & \tcell{6.04}
& \tcell{6.21} & \tcell{6.19}
& {\bfseries\tcell{6.36}} & \tcell{6.34}
& \tcell{6.19} & \tcell{5.69}
& \tcell{6} & \tcell{5.95}
& \tcell{6.06} \\
& 256k \setbase{4.94} & \cellcolor{heatneutral}\base
& \tcell{4.91} & \tcell{4.95}
& \tcell{5.13} & \tcell{5.11}
& \tcell{5.24} & \tcell{5.22}
& {\bfseries\tcell{5.36}} & \tcell{5.33}
& \tcell{4.88} & \tcell{5.11}
& \tcell{5.02} & \tcell{5.29}
& \tcell{5.07} \\
\midrule

\multirow{5}{*}{\ds{7}}
& 16k \setbase{28.85} & \cellcolor{heatneutral}\base
& \tcell{23.05} & \tcell{22.8}
& \tcell{22.01} & \tcell{21.99}
& \tcell{27.34} & \tcell{22.68}
& {\bfseries\tcell{29.01}} & \tcell{28.93}
& \tcell{27.39} & \tcell{25.34}
& \tcell{28.62} & \tcell{25.94}
& \tcell{24.8} \\
& 32k \setbase{24.0} & \cellcolor{heatneutral}\base
& \tcell{17.82} & \tcell{18.25}
& \tcell{21.14} & \tcell{21.06}
& \tcell{22.87} & \tcell{22.79}
& {\bfseries\tcell{24.66}} & \tcell{24.59}
& \tcell{23.58} & \tcell{22.82}
& \tcell{24.48} & \tcell{22}
& \tcell{22.3} \\
& 64k \setbase{14.74} & \cellcolor{heatneutral}\base
& \tcell{12.59} & \tcell{13.49}
& \tcell{14.25} & \tcell{14.2}
& \tcell{14.7} & \tcell{14.65}
& {\bfseries\tcell{15.17}} & \tcell{15.12}
& \tcell{15.04} & \tcell{14.99}
& \tcell{14.42} & \tcell{13.79}
& \tcell{14.32} \\
& 128k \setbase{10.93} & \cellcolor{heatneutral}\base
& \tcell{8.94} & \tcell{8.69}
& \tcell{10.16} & \tcell{10.11}
& \tcell{10.9} & \tcell{10.85}
& {\bfseries\tcell{11.66}} & \tcell{11.61}
& \tcell{11} & \tcell{11.53}
& \tcell{10.8} & \tcell{10.08}
& \tcell{9.79} \\
& 256k \setbase{8.03} & \cellcolor{heatneutral}\base
& \tcell{7.5} & \tcell{7.58}
& \tcell{7.95} & \tcell{7.92}
& \tcell{8.17} & \tcell{8.14}
& {\bfseries\tcell{8.41}} & \tcell{8.37}
& \tcell{7.65} & \tcell{8.26}
& \tcell{8.3} & \tcell{7.19}
& \tcell{7.35} \\
\midrule

\multirow{5}{*}{\ds{8}}
& 16k \setbase{16.43} & \cellcolor{heatneutral}\base
& \tcell{17.3} & \tcell{16.74}
& \tcell{17.35} & \tcell{17.29}
& \tcell{17.38} & \tcell{17.32}
& {\bfseries\tcell{17.41}} & \tcell{17.35}
& \tcell{16.1} & \tcell{16.55}
& \tcell{16.4} & \tcell{16.83}
& \tcell{17.1} \\
& 32k \setbase{14.92} & \cellcolor{heatneutral}\base
& \tcell{12.76} & \tcell{14.39}
& \tcell{14.68} & \tcell{14.63}
& \tcell{14.85} & \tcell{14.81}
& {\bfseries\tcell{15.03}} & \tcell{14.99}
& \tcell{14.16} & \tcell{14.35}
& \tcell{14.73} & \tcell{13.38}
& \tcell{14.7} \\
& 64k \setbase{11.95} & \cellcolor{heatneutral}\base
& \tcell{11.4} & \tcell{11.98}
& \tcell{12.05} & \tcell{12.02}
& \tcell{12.08} & \tcell{12.05}
& {\bfseries\tcell{12.13}} & \tcell{12.09}
& \tcell{11.73} & \tcell{12.02}
& \tcell{11.72} & \tcell{11.45}
& \tcell{11.76} \\
& 128k \setbase{9.75} & \cellcolor{heatneutral}\base
& \tcell{9.33} & \tcell{9.39}
& \tcell{9.74} & \tcell{9.69}
& \tcell{9.95} & \tcell{9.9}
& {\bfseries\tcell{10.16}} & \tcell{10.12}
& \tcell{10.05} & \tcell{9.72}
& \tcell{9.72} & \tcell{9.74}
& \tcell{9.93} \\
& 256k \setbase{7.27} & \cellcolor{heatneutral}\base
& \tcell{7.73} & \tcell{7.86}
& \tcell{8.16} & \tcell{8.13}
& \tcell{8.34} & \tcell{8.31}
& {\bfseries\tcell{8.53}} & \tcell{8.49}
& \tcell{7.66} & \tcell{8.43}
& \tcell{7.89} & \tcell{8.07}
& \tcell{7.71} \\
\midrule

\multirow{5}{*}{\ds{9}}
& 16k \setbase{16.9} & \cellcolor{heatneutral}\base
& \tcell{15.65} & \tcell{18.54}
& \tcell{19.23} & \tcell{19.17}
& \tcell{19.64} & \tcell{19.59}
& {\bfseries\tcell{20.07}} & \tcell{20.01}
& \tcell{16.41} & \tcell{15.9}
& \tcell{16} & \tcell{19.96}
& \tcell{17.9} \\
& 32k \setbase{16.3} & \cellcolor{heatneutral}\base
& \tcell{14.85} & \tcell{15.0}
& \tcell{15.96} & \tcell{15.91}
& \tcell{16.54} & \tcell{16.49}
& {\bfseries\tcell{17.14}} & \tcell{17.09}
& \tcell{15.73} & \tcell{13.55}
& \tcell{15.34} & \tcell{17.02}
& \tcell{16.72} \\
& 64k \setbase{14.27} & \cellcolor{heatneutral}\base
& \tcell{14.07} & \tcell{13.59}
& \tcell{14.63} & \tcell{14.58}
& \tcell{14.97} & \tcell{14.92}
& {\bfseries\tcell{15.32}} & \tcell{15.27}
& \tcell{14.2} & \tcell{13.35}
& \tcell{14.25} & \tcell{14.36}
& \tcell{14.56} \\
& 128k \setbase{7.54} & \cellcolor{heatneutral}\base
& \tcell{7.59} & \tcell{6.83}
& \tcell{8.51} & \tcell{8.48}
& \tcell{9.06} & \tcell{9.03}
& {\bfseries\tcell{9.64}} & \tcell{9.6}
& \tcell{8.67} & \tcell{9.13}
& \tcell{8.57} & \tcell{9.55}
& \tcell{8.51} \\
& 256k \setbase{5.56} & \cellcolor{heatneutral}\base
& \tcell{5.49} & \tcell{6.35}
& \tcell{6.38} & \tcell{6.37}
& \tcell{6.4} & \tcell{6.38}
& {\bfseries\tcell{6.42}} & \tcell{6.4}
& \tcell{5.55} & \tcell{6.13}
& \tcell{5.51} & \tcell{5.36}
& \tcell{5.17} \\
\bottomrule
\end{tabular}
\end{adjustbox}
\end{table*}

\section{Additional Experiments}
\label{appendix:exp}

This appendix extends the main-text evaluation in the same order: inference efficiency, general tasks, and long-context tasks. We first detail the generation protocol and examine throughput, decoding latency, energy, and GPU memory. We then report complete quality results across model scales, retention ratios, and context lengths.

\subsection{Experimental Setup}
\label{appendix:exp_setup}

\noindent \textbf{Generation workloads.}
End-to-end efficiency is evaluated on Llama-3.2-3B-Instruct using an NVIDIA A100 GPU with 40\,GB memory. Each model instance runs on one GPU, with four independent processes per node assigned separate GPUs and corresponding CPU/NUMA resources. We vary batch size $B\in\{1,2,4,8\}$, prompt length $P\in\{128,2048,8192,16384\}$, and generated length $O\in\{128,256,512,1024\}$, producing 64 request configurations per method. All methods receive identical synthetic token sequences constructed from a repeated text pattern. Greedy decoding generates exactly $O$ tokens for each request, even if an end-of-sequence token is produced.

\noindent \textbf{Execution configurations.}
The BF16 reference uses the original model with scaled dot-product attention (SDPA). Torch FP16 uses Hugging Face generation with eager attention. Uniform 8-bit weight-only quantization uses torchao's \texttt{Int8WeightOnlyConfig} through Hugging Face \texttt{TorchAoConfig}, while uniform 8-bit weight-and-activation quantization uses \texttt{Int8DynamicActivationInt8WeightConfig} with cuBLAS INT8 GEMM. GPTQ uses GPTQModel with 8-bit weights, group size 128, and the torch backend. LLM.int8 uses bitsandbytes with an outlier threshold of 6.0. For \modelname, both GI and IAA are evaluated at 1\%, 5\%, 10\%, and 20\% retention. A pure 8-bit configuration using the same \modelname execution implementation provides the zero-retention control.

\noindent \textbf{Measurement protocol.}
Each configuration uses two warm-up runs followed by five measured runs. We report median timing measurements and maximum observed GPU memory, retaining out-of-memory (OOM) and execution-failure statuses. Time to first token (TTFT) measures the interval from request submission to the first generated token, while time per output token (TPOT) averages the generation time after the first token. Output throughput is $BO/T_{\mathrm{total}}$, where $T_{\mathrm{total}}$ is the complete request duration. Energy is measured through NVIDIA Management Library (NVML) accumulated GPU energy and normalized to joules per 1,000 generated tokens, including idle power during the measured request. Throughput, latency, and energy comparisons use these definitions.

\subsection{Inference Efficiency}
\label{appendix:exp_efficiency}

The main text establishes the generation-throughput advantage of \modelname across request shapes. Here, we examine how the two calibration routes and their retention budgets affect throughput, per-token latency, and energy. These measurements use the same 64 request configurations, allowing each metric to be compared under identical generated workloads.

\noindent \textbf{Generation throughput.}
Figures~\ref{fig:generation_throughput_gi}, \ref{fig:generation_throughput_iaa}, and~\ref{fig:generation_throughput_baselines} expand the main throughput comparison into separate GI, IAA, and baseline groups. The GI and IAA panels include all four retention ratios, showing how additional 16-bit capacity affects complete-model generation. The baseline panels provide the corresponding executable-method comparisons across batch sizes, prompt lengths, and output lengths.

The throughput profiles show that increasing the retained fraction produces a modest change in execution efficiency. With batch size 1, a 128-token prompt, and 512 generated tokens, GI$_1$ and GI$_{20}$ achieve 154.98 and 150.17 tokens/s, respectively. At batch size 8 with 16K prompt tokens and 1024 generated tokens, they achieve 306.20 and 301.93 tokens/s. IAA exhibits closely matched throughput at the corresponding retention ratios. These results show that the common tile-native execution structure accommodates different calibration maps and retained capacities with efficient generation.

\noindent \textbf{Time per output token.}
Complete-request throughput includes both prompt processing and repeated decoding. We therefore examine time per output token (TPOT) after the first generated token. Figures~\ref{fig:tpot_main}, \ref{fig:tpot_gi}, \ref{fig:tpot_iaa}, and~\ref{fig:tpot_baselines} report TPOT speedup relative to BF16, defined as $\mathrm{TPOT}_{\mathrm{BF16}}/\mathrm{TPOT}_{\mathrm{method}}$ at matched request shapes. Higher values indicate lower decoding latency. Both GI$_1$ and GI$_{20}$ exceed the BF16 reference throughout the displayed grid, showing that the generation gains extend to repeated decoding.

\noindent \textbf{Generation energy.}
We further measure GPU energy per 1,000 generated tokens over the complete request, including prompt processing and decoding. Figures~\ref{fig:energy_main}, \ref{fig:energy_gi}, \ref{fig:energy_iaa}, and~\ref{fig:energy_baselines} report energy efficiency relative to BF16, defined as $E_{\mathrm{BF16}}/E_{\mathrm{method}}$ for the same request shape. Higher values therefore indicate lower energy consumption per generated token. Across the displayed grid, GI$_1$ achieves higher throughput and lower measured energy consumption than BF16.

\subsubsection{GPU Memory Results}
\label{appendix:exp_memory}

We further examine how the retained-tile budget controls the memory footprint of reconstructed models across three model scales. These measurements cover checkpoint storage, post-load VRAM, peak VRAM, and prefill peak VRAM under the model-scale evaluation protocol. The end-to-end generation measurements follow the separate workload protocol in Appendix~\ref{appendix:exp_setup}.

Table~\ref{tab:prefill_peak_vram_all_models} reports prefill peak VRAM across prompt lengths from 1K to 32K. Tables~\ref{tab:memory_llama3b_full}, \ref{tab:memory_qwen25_7b}, and \ref{tab:memory_qwen25_14b} provide the complete memory profiles for Llama-3.2-3B, Qwen2.5-7B, and Qwen2.5-14B.

The results show that the retention ratio provides direct control over the persistent memory footprint. On Llama-3.2-3B, increasing retention from 1\% to 20\% changes post-load memory from 3.51 to 4.02\,GB, compared with 5.99\,GB for Original. The corresponding values are 8.25 to 9.40\,GB on Qwen2.5-7B and 15.67 to 17.85\,GB on Qwen2.5-14B. GI and IAA have identical or closely matched memory usage at the same retention ratio because the number and format of retained weights remain the same, while their selected locations differ.

Prefill measurements further show how the compact representation carries into longer inputs. At 32K prompt tokens, GI$_{20}$ uses 11.09\,GB on Llama-3.2-3B, 17.05\,GB on Qwen2.5-7B, and 30.13\,GB on Qwen2.5-14B, compared with 12.64, 21.33, and 40.10\,GB for their respective original models. The retention ratio controls persistent weight storage, while longer prompts additionally require activation and attention memory. Together with the quality evaluations, these measurements support selecting the retained capacity and calibration route jointly for deployment memory and accuracy.

\subsection{General Tasks}
\label{appendix:exp_general}

After evaluating execution efficiency, we extend the general-task quality analysis to three model scales and the complete set of retention ratios. Tables~\ref{tab:general_llama3b_full}, \ref{tab:general_qwen7b_full}, and~\ref{tab:general_qwen14b_full} report WikiText-2 perplexity and six downstream benchmarks for Llama-3.2-3B, Qwen2.5-7B, and Qwen2.5-14B. Both GI and IAA are evaluated with 1\%, 5\%, 10\%, and 20\% of target weight tiles retained at 16-bit, extending the main text's 1\% and 20\% configurations.

Across the three models, the full retention sweep shows that locally preserving a limited number of sensitive tiles maintains strong general-task performance. On Qwen2.5-7B, GI$_{20}$ achieves 9.353 WikiText-2 perplexity, compared with 9.407 for Original. On Qwen2.5-14B, the corresponding values are 6.611 and 6.661. The improvements extend to selected downstream tasks: GI$_{10}$ reaches 75.55\% MMLU accuracy on Qwen2.5-7B versus 74.23\% for Original, while GI$_{20}$ reaches 86.49\% HellaSwag accuracy on Qwen2.5-14B versus 85.07\% for Original. These results extend the quality preservation observed on Llama-3.2-3B to larger models using the same tile allocation rule.

The two calibration routes also provide different quality profiles at the same retained capacity. On Qwen2.5-14B at 20\% retention, IAA reaches 80.67\% WinoGrande accuracy, compared with 80.42\% for GI and 79.16\% for Original. On HellaSwag, GI reaches 86.49\%, compared with 86.06\% for IAA and 85.07\% for Original. These task-dependent rankings show how calibration affects quality with retained capacity and mixed-precision execution held fixed.

\subsection{Long Context Tasks}\label{appendix:exp_long}

We further examine whether calibration-derived local precision choices preserve retrieval and question-answering capabilities as context length increases. Following the main text, LongEval evaluates evidence localization across input lengths, while LV-Eval extends the analysis to question answering over multiple documents and datasets.

\noindent \textbf{Long-context retrieval.}
The main text evaluates Llama-3.2-3B on LongEval from 2.7K to 43.4K tokens. Figures~\ref{fig:longeval_qwen7b} and~\ref{fig:longeval_qwen14b} extend the same retrieval analysis to Qwen2.5-7B and Qwen2.5-14B, covering input lengths from 3.2K to 52.9K tokens. Each figure compares GI and IAA retention sweeps with the quantization baselines, using exact-match accuracy to measure successful line retrieval.

The Qwen results extend the retrieval evaluation to larger models and prompts up to 52.9K tokens. Within each model, the retention sweep examines how high-precision capacity affects line retrieval, while the matched-budget GI and IAA comparison evaluates the two calibration routes. Each resulting map is reused throughout the length sweep, connecting offline calibration to retrieval under changing input lengths.

\noindent \textbf{Long-context question answering.}
LV-Eval extends the retrieval analysis to answering questions from long input contexts. Tables~\ref{tab:lveval_llama3b_full}, \ref{tab:lveval_qwen7b_full}, and~\ref{tab:lveval_qwen14b_full} report complete results for Llama-3.2-3B, Qwen2.5-7B, and Qwen2.5-14B at five context-length levels: 16K, 32K, 64K, 128K, and 256K. Each model includes all four retention ratios for IAA and GI.

The nine datasets follow the numbering in Table~\ref{tab:lveval_llama3b}: \ds{1} \textit{cmrc\_mixup}, \ds{2} \textit{dureader\_mixup}, \ds{3} \textit{hotpotwikiqa\_mixup}, \ds{4} \textit{lic\_mixup}, \ds{5} \textit{loogle\_CR\_mixup}, \ds{6} \textit{loogle\_MIR\_mixup}, \ds{7} \textit{loogle\_SD\_mixup}, \ds{8} \textit{multifieldqa\_en\_mixup}, and \ds{9} \textit{multifieldqa\_zh\_mixup}. We follow the LV-Eval protocol, reporting keyword-recall-based F1 by default, blacklist-based F1 for \textit{cmrc\_mixup}, and blacklist-based ROUGE-L for \textit{dureader\_mixup}.

The 128K and 256K LV-Eval levels extend the main-text question-answering evaluation to longer contexts. On Qwen2.5-14B, GI$_{20}$ scores 11.66 on LooGLE-SD at 128K, compared with 10.93 for Original. On Qwen2.5-7B, it scores 9.72 on English MultiFieldQA at 256K, compared with 8.52 for Original. Each model reuses its calibrated maps across these length levels, showing that the selected local regions remain useful for answering questions in longer contexts.

Across the full LV-Eval evaluation, the retention ratio sets the available 16-bit capacity, and the calibration route determines its allocation and the associated quantization parameters. The 20\% configurations frequently achieve strong results across datasets and lengths, while lower-retention variants provide additional operating points with smaller weight representations. Each model uses its own calibration-derived maps, which are reused across that model's retrieval and question-answering tasks and input lengths.

\begin{table*}[t]
\centering
\footnotesize
\setlength{\tabcolsep}{3.0pt}
\renewcommand{\arraystretch}{0.9}
\setlength{\aboverulesep}{0.25ex}
\setlength{\belowrulesep}{0.35ex}
\caption{
Matched-budget tile-selection controls on \textbf{Qwen2.5-7B} with $64\times64$ tiles. Matched-control PPL uses the fixed subset of 62 WikiText-2 test documents.
The target-linear payload is shared by all selection rules within each budget.
Random PPL is reported as mean $\pm$ sample standard deviation over five draws.
Bold marks the best displayed value in each row.
}
\label{tab:selection_controls}
\begin{tabular}{@{}>{\centering\arraybackslash}p{\dimexpr0.1\textwidth-\tabcolsep\relax} >{\centering\arraybackslash}p{\dimexpr0.13\textwidth-2\tabcolsep\relax} >{\raggedright\arraybackslash}p{\dimexpr0.38\textwidth-2\tabcolsep\relax} >{\centering\arraybackslash}p{\dimexpr0.13\textwidth-2\tabcolsep\relax} >{\centering\arraybackslash}p{\dimexpr0.13\textwidth-2\tabcolsep\relax} >{\centering\arraybackslash}p{\dimexpr0.13\textwidth-\tabcolsep\relax}@{}}
\toprule
\multirow{2}{*}{\shortstack{\textbf{Retention}\\\textbf{(\%)}}} &
\multirow{2}{*}{\shortstack{\textbf{Target-linear}\\\textbf{(GiB)}}} &
\multirow{2}{*}{\textbf{Metric}} &
\multicolumn{2}{c}{\textbf{Controls}} &
\textbf{Ours} \\
\cmidrule(lr){4-5}\cmidrule(lr){6-6}
& & & \textbf{Random} & \textbf{Magnitude} & \textbf{IAA} \\
\midrule

\multirow{2}{*}{5} & \multirow{2}{*}{6.39}
& Retained saliency mass (\%) $\uparrow$ 
& 5.00
& 5.34
& {\bfseries7.05} \\

& & Matched-control PPL $\downarrow$ 
& $9.92\pm0.03$
& 9.93
& {\bfseries9.85} \\

\midrule

\multirow{2}{*}{20} & \multirow{2}{*}{7.30}
& Retained saliency mass (\%) $\uparrow$ 
& 20.00
& 20.76
& {\bfseries23.90} \\

& & Matched-control PPL $\downarrow$ 
& $9.82\pm0.03$
& 9.78
& {\bfseries9.7650} \\

\bottomrule
\end{tabular}
\end{table*}

\begin{table*}[t]
\centering
\footnotesize
\setlength{\tabcolsep}{2.2pt}
\renewcommand{\arraystretch}{0.9}
\setlength{\aboverulesep}{0.25ex}
\setlength{\belowrulesep}{0.35ex}
\caption{
Allocation-geometry controls on \textbf{Qwen2.5-7B} at a matched 20\% high-precision retention target.
Total payload denotes the stored representation measured in this control.
Retained saliency mass is reported as a fraction, and PPL uses the fixed 62-document subset.
Bold marks the best displayed value in each row.
}
\label{tab:geometry_quality}
\begin{tabular}{@{}>{\raggedright\arraybackslash}p{\dimexpr0.34\textwidth-\tabcolsep\relax} >{\centering\arraybackslash}p{\dimexpr0.1\textwidth-2\tabcolsep\relax} >{\centering\arraybackslash}p{\dimexpr0.12\textwidth-2\tabcolsep\relax} >{\centering\arraybackslash}p{\dimexpr0.11\textwidth-2\tabcolsep\relax} >{\centering\arraybackslash}p{\dimexpr0.11\textwidth-2\tabcolsep\relax} >{\centering\arraybackslash}p{\dimexpr0.11\textwidth-2\tabcolsep\relax} >{\centering\arraybackslash}p{\dimexpr0.11\textwidth-\tabcolsep\relax}@{}}
\toprule
\multirow{2}{*}{\textbf{Metric}} &
\multirow{2}{*}{\textbf{Unit}} &
\multicolumn{4}{c}{\textbf{Allocation controls}} &
\textbf{Ours} \\
\cmidrule(lr){3-6}\cmidrule(lr){7-7}
& &
\textbf{Full column} &
\textbf{$1\times64$} &
\textbf{$64\times1$} &
\textbf{Full row} &
\textbf{$64\times64$} \\
\midrule

Total payload $\downarrow$ & GiB 
& 7.2995
& 7.3744
& 7.3744
& 7.2997
& {\bfseries7.2984} \\

Metadata $\downarrow$ & MiB 
& \textbf{0.43}
& 77.21
& 77.21
& 0.53
& 0.61 \\

Retained saliency mass $\uparrow$ & Fraction 
& 0.31
& 0.26
& {\bfseries0.32}
& 0.23
& 0.24 \\

Matched-control PPL $\downarrow$ & PPL 
& 9.72
& 9.87
& {\bfseries9.71}
& 9.98
& 9.7650 \\

\bottomrule
\end{tabular}
\end{table*}

\begin{table}[!t]
\centering
\footnotesize
\setlength{\tabcolsep}{2.2pt}
\renewcommand{\arraystretch}{0.95}
\caption{
Geometry-specific execution on \textbf{Qwen2.5-7B} with IAA scores and 20\% 16-bit retention.
Native tiles are $64\times64$. FP-intersected tiles contain retained weights; partially mixed tiles contain both precisions.
B1 and B4 denote batch sizes 1 and 4; TPOT is time per output token.
Bold marks the minimum in each row.
}
\label{tab:geometry_execution}
\begin{adjustbox}{max width=\columnwidth,center}
\begin{tabular}{l c @{\vsep} c c @{\vsep} c}
\toprule
\multirow{2}{*}{\textbf{Metric}} &
\multirow{2}{*}{\textbf{Unit}} &
\multicolumn{2}{c}{\textbf{One-dimensional paths}} &
\textbf{Ours} \\
\cmidrule(lr){3-4}\cmidrule(lr){5-5}
& &
\shortstack{\textbf{Full}\\\textbf{column}} &
\shortstack{\textbf{$64\times1$}\\\textbf{compact strip}} &
\shortstack{\textbf{$64\times64$}\\\textbf{tile}} \\
\midrule

Metadata $\downarrow$ & MiB 
& \textbf{0.436}
& 77.208
& 0.611 \\

\shortstack[l]{FP-intersected\\native tiles} $\downarrow$ & \% 
& 99.50
& 99.53
& {\bfseries20.01} \\

\shortstack[l]{Partially mixed\\native tiles} $\downarrow$ & Count
& 1,581,552
& 1,583,860
& \textbf{0} \\

\shortstack[l]{Prefill\\B1/8K} $\downarrow$ & ms 
& 1548.57
& 4183.85
& {\bfseries1211.46} \\

\shortstack[l]{Decode B1\\TPOT} $\downarrow$ & \shortstack{ms/\\token} 
& 57.12
& 58.67
& {\bfseries40.08} \\

\shortstack[l]{Decode B4\\TPOT} $\downarrow$ & \shortstack{ms/\\token} 
& 58.06
& 62.93
& {\bfseries39.73} \\

\shortstack[l]{Post-load\\memory} $\downarrow$ & GiB 
& 10.574
& 11.674
& {\bfseries9.377} \\

\bottomrule
\end{tabular}
\end{adjustbox}
\end{table}

\begin{table}[!t]
\centering
\footnotesize
\setlength{\tabcolsep}{2.2pt}
\renewcommand{\arraystretch}{0.95}
\caption{
Exact-budget tile-size sensitivity on \textbf{Qwen2.5-7B}.
Each configuration retains 66,977,792 high-precision weight elements.
Map time measures decision-map construction.
PPL is the matched-control value, MMLU accuracy is a percentage, and LongEval accuracy is a fraction at 52.9K tokens.
Bold marks the best displayed value in each row.
}
\label{tab:tile_size_control}
\begin{adjustbox}{max width=\columnwidth,center}
\begin{tabular}{l c @{\vsep} c c c}
\toprule
\multirow{2}{*}{\textbf{Metric}} &
\multirow{2}{*}{\textbf{Unit}} &
\multicolumn{3}{c}{\textbf{Weight tile size}} \\
\cmidrule(lr){3-5}
& &
\textbf{$32\times32$} &
\textbf{$64\times64$} &
\textbf{$128\times128$} \\
\midrule

Target-linear payload $\downarrow$ & GiB 
& 6.1457
& \textbf{6.1454}
& {\bfseries6.1454} \\

Metadata $\downarrow$ & Bytes
& 246,848
& 35,840
& \textbf{11,312} \\

Map-construction time $\downarrow$ & s 
& 5.48
& 3.16
& {\bfseries2.16} \\

Matched-control PPL $\downarrow$ & PPL 
& 9.9368
& \textbf{9.9319}
& 9.9322 \\

MMLU $\uparrow$ & \% 
& 73.89
& \textbf{74.06}
& 73.80 \\

LongEval at 52.9K $\uparrow$ & Accuracy 
& 0.40
& \textbf{0.50}
& 0.30 \\

\bottomrule
\end{tabular}
\end{adjustbox}
\end{table}

\begin{table}[!t]
\centering
\footnotesize
\setlength{\tabcolsep}{2.2pt}
\renewcommand{\arraystretch}{0.95}
\caption{
Offline calibration time and peak GPU memory for GI and IAA at three sample counts. Speedup is the ratio of GI time to IAA time at the same sample count. Bold marks the lower-cost route in each pair.
}
\label{tab:calibration_cost}
\begin{adjustbox}{max width=\columnwidth,center}
\begin{tabular}{c @{\vsep} c c c @{\vsep} c c}
\toprule
\multirow{2}{*}{\textbf{Samples}} &
\multicolumn{2}{c}{\textbf{Time (s)} $\downarrow$} &
\multirow{2}{*}{\shortstack{\textbf{IAA}\\\textbf{speedup} ($\times$)}} &
\multicolumn{2}{c}{\textbf{Peak memory (GiB)} $\downarrow$} \\
\cmidrule(lr){2-3}\cmidrule(lr){5-6}
& \textbf{GI} & \textbf{IAA} & & \textbf{GI} & \textbf{IAA} \\
\midrule

128 
& 112.4
& {\bfseries36.8}
& 3.05 
& 25.41
& {\bfseries9.24} \\

256 
& 218.3
& {\bfseries77.0}
& 2.83 
& 25.41
& {\bfseries9.24} \\

512 
& 407.1
& {\bfseries132.6}
& 3.07 
& 25.41
& {\bfseries9.24} \\

\bottomrule
\end{tabular}
\end{adjustbox}
\end{table}

\begin{table}[!t]
\centering
\footnotesize
\setlength{\tabcolsep}{2.2pt}
\renewcommand{\arraystretch}{0.95}
\caption{
Retained-set stability across three independent calibration draws at 1\% and 20\% retention. Values report Jaccard similarity between selected tile sets. Bold marks the larger value in each pair, including ties.
}
\label{tab:retained_set_stability}
\begin{adjustbox}{max width=\columnwidth,center}
\begin{tabular}{l @{\vsep} c c @{\vsep} c c}
\toprule
\multirow{2}{*}{\textbf{Statistic}} &
\multicolumn{2}{c}{\textbf{1\% retention}} &
\multicolumn{2}{c}{\textbf{20\% retention}} \\
\cmidrule(lr){2-3}\cmidrule(lr){4-5}
& \textbf{GI} & \textbf{IAA} & \textbf{GI} & \textbf{IAA} \\
\midrule

Pairwise Jaccard mean $\uparrow$ 
& 0.98
& {\bfseries0.99} 
& \textbf{0.99}
& {\bfseries0.99} \\

Pairwise Jaccard minimum $\uparrow$ 
& 0.97
& {\bfseries0.99} 
& \textbf{0.99}
& {\bfseries0.99} \\

Per-layer Jaccard median $\uparrow$ 
& 0.97
& {\bfseries0.99} 
& \textbf{0.99}
& {\bfseries0.99} \\

Per-layer Jaccard minimum $\uparrow$ 
& 0.93
& {\bfseries0.97} 
& 0.97
& {\bfseries0.99} \\

\bottomrule
\end{tabular}
\end{adjustbox}
\end{table}

\begin{table*}[t]
\centering
\begin{minipage}[t]{\columnwidth}
\centering
\footnotesize
\setlength{\tabcolsep}{2.2pt}
\renewcommand{\arraystretch}{0.95}
\caption{
Overlap between GI and IAA retained sets across models and retention ratios.
Jaccard is reported as a fraction, shared/retained as a percentage, and Jaccard/chance as the reported ratio to random-map overlap.
These columns describe agreement between selection routes across models.
}
\label{tab:gi_iaa_overlap}
\begin{adjustbox}{max width=\columnwidth,center}
\begin{tabular}{l c @{\vsep} c c c}
\toprule
\textbf{Model} &
\shortstack{\textbf{Retention}\\\textbf{(\%)}} &
\textbf{Jaccard} &
\shortstack{\textbf{Jaccard /}\\\textbf{chance} ($\times$)} &
\shortstack{\textbf{Shared /}\\\textbf{retained (\%)}} \\
\midrule

\multirow{4}{*}{Llama-3.2-3B}
& 1  & 0.04 & 8.60 & 8.29 \\
& 5  & 0.09 & 3.45 & 16.25 \\
& 10 & 0.13 & 2.52 & 23.39 \\
& 20 & 0.22 & 1.94 & 35.44 \\

\midrule

\multirow{4}{*}{Qwen2.5-7B}
& 1  & 0.15 & 29.19 & 25.58 \\
& 5  & 0.29 & 11.20 & 44.63 \\
& 10 & 0.36 & 6.87 & 53.11 \\
& 20 & 0.42 & 3.78 & 59.14 \\

\midrule

\multirow{4}{*}{Qwen2.5-14B}
& 1  & 0.13 & 25.67 & 22.85 \\
& 5  & 0.22 & 8.77 & 36.71 \\
& 10 & 0.29 & 5.47 & 44.71 \\
& 20 & 0.37 & 3.30 & 53.62 \\

\bottomrule
\end{tabular}
\end{adjustbox}
\end{minipage}\hfill
\begin{minipage}[t]{\columnwidth}
\centering
\footnotesize
\setlength{\tabcolsep}{2.2pt}
\renewcommand{\arraystretch}{0.95}
\caption{
Projection-level saliency coverage and GI/IAA retained-set overlap on \textbf{Qwen2.5-7B} at 20\% retention. All values are percentages.
The GI and IAA columns report retained saliency mass, while the final two columns describe agreement between their retained sets.
These statistics characterize selected regions by projection family and scoring route.
}
\label{tab:projection_map_analysis}
\begin{adjustbox}{max width=\columnwidth,center}
\begin{tabular}{l @{\vsep} c c @{\vsep} c c}
\toprule
\multirow{2}{*}{\textbf{Projection}} &
\multicolumn{2}{c}{\shortstack{\textbf{Retained saliency}\\\textbf{mass (\%)}}} &
\multirow{2}{*}{\shortstack{\textbf{Jaccard}\\\textbf{(\%)}}} &
\multirow{2}{*}{\shortstack{\textbf{Shared /}\\\textbf{retained (\%)}}} \\
\cmidrule(lr){2-3}
& \textbf{GI} & \textbf{IAA} & & \\
\midrule

\texttt{q\_proj}
& 32.37 & 25.33 & 23.89 & 38.03 \\

\texttt{k\_proj}
& 28.62 & 26.13 & 31.12 & 45.67 \\

\texttt{v\_proj}
& 27.86 & 24.07 & 22.30 & 32.66 \\

\texttt{o\_proj}
& 29.35 & 29.70 & 58.51 & 71.97 \\

\texttt{up\_proj}
& 22.75 & 21.94 & 39.12 & 55.83 \\

\texttt{gate\_proj}
& 24.03 & 25.59 & 49.74 & 66.13 \\

\texttt{down\_proj}
& 22.58 & 22.09 & 41.40 & 58.11 \\

\bottomrule
\end{tabular}
\end{adjustbox}
\end{minipage}
\end{table*}

\section{Calibration and Allocation Analysis}
\label{appendix:allocation_analysis}

We examine how retained locations, allocation geometry, and calibration choices contribute to \modelname. The first three studies isolate tile selection, allocation geometry, and tile size under matched high-precision budgets on Qwen2.5-7B. We then evaluate calibration cost, the reproducibility of retained sets, and the differences between GI and IAA maps. For the selection, geometry, and tile-size controls, \emph{matched-control perplexity (PPL)} is evaluated on a fixed subset of 62 WikiText-2 test documents using a common evaluation wrapper. Full-benchmark results are in Appendix~\ref{appendix:exp_general}.

\subsection{Matched-Budget Tile Selection}
\label{appendix:selection_controls}

To isolate the value of retained locations, we fix the $64\times64$ tile grid and compare IAA, magnitude-based selection, and random selection. Within each budget, all three rules retain the same number of high-precision weights: 326,516,736 at 5\% and 1,305,608,192 at 20\%. Random selection uses five independent draws. Table~\ref{tab:selection_controls} reports the resulting matched-control PPL and retained saliency mass, the percentage of calibration saliency covered by the retained weights.

At 5\% retention, IAA achieves 9.85 PPL, compared with 9.93 for magnitude selection and 9.92 for random selection, while covering 7.05\% of the calibration saliency. At 20\%, IAA further achieves 9.7650 PPL, compared with 9.78 and 9.82 for the two controls. Since all methods retain the same high-precision capacity within each comparison, these results show that activation-aware scores effectively prioritize useful local weight regions. We next examine how the geometry of these regions affects their representation and execution.

\subsection{Allocation Geometry and Execution}
\label{appendix:geometry_controls}

Selection scores determine priority, while the allocation unit determines which weights share a precision decision. We therefore compare $1\times64$ and $64\times1$ strips, full rows, full columns, and $64\times64$ tiles at a matched 20\% retention target. Table~\ref{tab:geometry_quality} reports the payload, metadata, retained saliency mass, and matched-control PPL for these alternatives.

At 20\% retention, $64\times64$ tiles achieve 9.7650 matched-control PPL, close to 9.72 for full columns and 9.71 for $64\times1$ strips. Their metadata occupies 0.61\,MiB, compared with 77.21\,MiB for the compact strips. Thus, 2D tiles preserve similar language-modeling quality in this control with substantially less metadata than fine-grained strips. The execution measurements below test how these representations affect prefill and decoding.

We further compare executable full-column, compact-strip, and tile implementations using the same IAA entry scores and 20\% retention target. These geometry measurements use an A100-40GB GPU, with 10 warm-up requests followed by 30 timed requests in each of three fresh processes. Table~\ref{tab:geometry_execution} reports prefill latency, decode TPOT, and post-load memory for this dedicated control.

Relative to full-column allocation, the tile implementation reduces 8K prefill latency by 21.8\%, B1 decode TPOT by 29.8\%, and B4 decode TPOT by 31.6\%, with approximately 1.20\,GiB less post-load memory. Together with the quality comparison, these measurements support the hardware-aligned tile as an effective allocation and execution unit among those evaluated.

\subsection{Tile-Size Sensitivity}
\label{appendix:tile_size}

Within the 2D design, tile size controls the resolution of precision placement and the amount of selection metadata. We compare $32\times32$, $64\times64$, and $128\times128$ tiles while retaining exactly 66,977,792 high-precision weight elements in every configuration, corresponding to approximately 1.026\% retention. Table~\ref{tab:tile_size_control} therefore evaluates size changes at equal retained capacity.

The $64\times64$ configuration achieves the highest MMLU accuracy and LongEval accuracy among the three evaluated sizes, reaching 74.06\% and 0.50, respectively. Compared with $32\times32$ tiles, it requires 85.5\% less metadata and 42.3\% less map-construction time. These results support the selected tile dimensions for local precision allocation and compact metadata construction.

\enlargethispage{-75pt}
\subsection{Calibration Cost}
\label{appendix:calibration_cost}

The two scoring routes share the selection and execution pipeline, while collecting different calibration signals. To quantify their offline resource requirements, we compare wall-clock time and peak memory at matched sample counts of 128, 256, and 512 (Table~\ref{tab:calibration_cost}). These cost measurements use the stated sample counts, while the full quality evaluation uses maps calibrated with 8,192 examples.

IAA is 2.83--3.07$\times$ faster across these sample counts, and its peak memory is 9.24\,GiB versus 25.41\,GiB for GI, a reduction of 63.6\%. This difference follows the information collected by each route: IAA uses forward statistics, while GI additionally computes gradients of the prediction loss. The shared pipeline therefore supports lower-cost calibration alongside loss-sensitive selection.

\enlargethispage{-80pt}
\subsection{Retained-Set Stability}
\label{appendix:retained_stability}

We examine whether changing the calibration records changes which tiles receive high precision. Calibration is repeated across three independent draws for each route at 1\% and 20\% retention. For retained sets $\mathcal{R}_1$ and $\mathcal{R}_2$ from two draws, Jaccard similarity is $J(\mathcal{R}_1,\mathcal{R}_2)=|\mathcal{R}_1\cap\mathcal{R}_2|/|\mathcal{R}_1\cup\mathcal{R}_2|$. Table~\ref{tab:retained_set_stability} reports agreement between full-model retained sets and within individual projections, measuring the repeatability of local precision selection.

The mean pairwise Jaccard is 0.98 for GI at 1\% retention and 0.99 for the other three configurations. Layer-level minima range from 0.93 to 0.99, showing that the agreement extends to individual layers. Both routes therefore produce reproducible retained locations under the evaluated sampling changes. We next compare GI with IAA to distinguish this within-route stability from differences between the two scoring signals.

\subsection{GI/IAA Retained-Set Overlap}
\label{appendix:map_overlap}

GI and IAA use the same capacity rule but evaluate importance through different signals. Table~\ref{tab:gi_iaa_overlap} compares their retained sets across three models and four retention ratios. Jaccard measures intersection relative to union, while shared/retained measures the fraction of either equally sized retained set selected by both routes. The chance-normalized column compares Jaccard with the expected overlap of random maps at the same retention ratio.

Agreement exceeds the reported chance reference by 1.94--29.19$\times$. At 20\% retention, the two routes share 35.44\% of retained tiles on Llama-3.2-3B, 59.14\% on Qwen2.5-7B, and 53.62\% on Qwen2.5-14B. The maps therefore contain shared selections alongside locations preferred by each scoring signal. These overlaps describe the difference in retained locations, while the quality and calibration-cost experiments evaluate the practical value of each route.

To locate differences between the two retained maps, Table~\ref{tab:projection_map_analysis} groups the 20\% retention results on Qwen2.5-7B by projection family across 28 decoder layers. Both routes retain 318,752 tiles, with 188,508 shared selections, corresponding to 41.98\% Jaccard similarity and 59.14\% shared/retained overlap. Retained saliency mass measures the fraction of importance scores covered by selected regions in each projection family.

Agreement is highest in the attention-output and gate projections, while the query and value projections show less overlap. The retained budget is identical for the two routes within each projection, so these differences concern the locations receiving high precision. Projection-level comparisons therefore identify where the two calibration routes make the most different local selections.

\end{document}